\documentclass[letterpaper]{article} 
\usepackage{aaai2026}  
\usepackage{times}  
\usepackage{helvet}  
\usepackage{courier}  
\usepackage[hyphens]{url}  
\usepackage{graphicx} 
\usepackage{natbib}  
\usepackage{caption} 
\usepackage{amsthm} 

\usepackage{algorithm}
\usepackage{algorithmic}

\usepackage{newfloat}
\usepackage{listings}
\DeclareCaptionStyle{ruled}{labelfont=normalfont,labelsep=colon,strut=off} 
\floatstyle{ruled}
\newfloat{listing}{tb}{lst}{}
\floatname{listing}{Listing}
\usepackage[utf8]{inputenc} 
\usepackage[T1]{fontenc}    
\usepackage{url}            
\usepackage{booktabs}       
\usepackage{amsfonts}       
\usepackage{amsmath}        
\usepackage{nicefrac}       
\usepackage{microtype}      
\usepackage{graphicx}
\usepackage{amssymb}
\usepackage{pifont}
\usepackage{subcaption}

\usepackage[table]{xcolor}
\usepackage{multirow} 

\newtheorem{theorem}{Theorem}
\usepackage{graphicx}

\newcommand{\x}{\ensuremath{\boldsymbol{x}}}
\newcommand{\xw}{\ensuremath{\boldsymbol{x}^w}}
\newcommand{\xl}{\ensuremath{\boldsymbol{x}^l}}

\newcommand{\vc}{\ensuremath{\boldsymbol{c}}}
\newcommand{\I}{\ensuremath{\boldsymbol{I}}}
\newcommand{\bbE}{\ensuremath{\mathbb{E}}}
\newcommand{\bbD}{\ensuremath{\mathbb{D}}}
\newcommand{\kl}{\ensuremath{\bbD_{\text{KL}}}}

\newcommand{\calvar}[1]{\ensuremath{\mathcal{#1}}}
\newcommand{\calD}{\calvar{D}}
\newcommand{\calN}{\calvar{N}}

\newcommand{\noise}{\ensuremath{\boldsymbol{\epsilon}}}

\newcommand{\pref}{p_{\text{ref}}}
\newcommand{\pinit}{p_{\text{init}}}

\title{Rethinking Direct Preference Optimization in
Diffusion Models}
\author {
    Junyong Kang\equalcontrib\textsuperscript{\rm 1},
    Seohyun Lim\equalcontrib\textsuperscript{\rm 1},
    Kyungjune Baek\textsuperscript{\rm 2}, 
     Hyunjung Shim\dag\textsuperscript{\rm 1}
}
\affiliations{
    \textsuperscript{\rm 1}Korea Advanced Institute of Science and Technology\\
    \textsuperscript{\rm 2}Sejong University\\

{\{jykang,seohyunlim,kateshim\}@kaist.ac.kr}\\
{\{kyungjune.baek\}@sejong.ac.kr}
}

\usepackage{bibentry}

\begin{document}

\maketitle


\begin{abstract}
    Aligning text-to-image (T2I) diffusion models with human preferences has emerged as a critical research challenge. While Direct Preference Optimization (DPO) has established a foundation for preference learning in large language models (LLMs), its extension to diffusion models remains limited in alignment performance. 
    In this work, we propose an enhanced version of Diffusion-DPO by introducing a stable reference model update strategy.
    This strategy facilitates the exploration of better alignment solutions while maintaining training stability. 
    Moreover, we design a timestep-aware optimization strategy that further boosts performance by addressing preference learning imbalance across timesteps.
    Through the synergistic combination of our exploration and timestep-aware optimization, our method significantly improves the alignment performance of Diffusion-DPO on human preference evaluation benchmarks, achieving state-of-the-art results. The code is available at the Github: \url{https://github.com/kaist-cvml/RethinkingDPO_Diffusion_Models}.
\end{abstract}

\section{Introduction}
Diffusion models~\cite{ddpm, song2019generative, song2021score} have emerged as a powerful generative framework, achieving remarkable success in text-to-image (T2I) generation~\cite{sdxl, saharia2022photo}. 
By leveraging large-scale image-text pairs during training, these models can synthesize high-fidelity images conditioned on natural language prompts.
However, due to the uncurated and noisy nature of web-scale datasets, their outputs often misalign with human aesthetic and semantic preferences.

To address these challenges, the field of aligning with human feedback has emerged as a crucial research direction. Inspired by advances in aligning language models with human feedback~\cite{ouyang2022human, rafailov2023direct}, recent efforts have extended alignment techniques to the vision domain.  
These methods can be broadly categorized into two prominent approaches: reward model-based fine-tuning ~\cite{blacktraining,fan2023dpok,xu2024imagereward,clarkdirectly,prabhudesai2023aligning} and Direct Preference Optimization (DPO)~\cite{wallace2024diffusion,lialigning,yang2024dense,zhu2025dspo}.

\begin{figure*}[t!]
    \centering
    \begin{subfigure}[b]{0.24\textwidth}
        \centering
        \includegraphics[width=\linewidth]{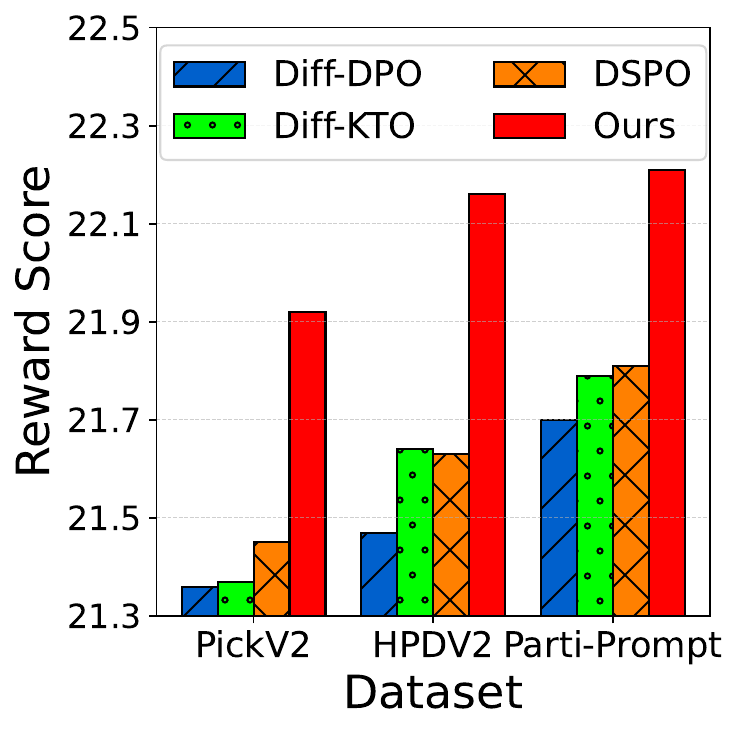}
        \caption{Performance comparison}
    \end{subfigure}
    \begin{subfigure}[b]{0.4\textwidth}
        \centering
        \includegraphics[width=\linewidth]{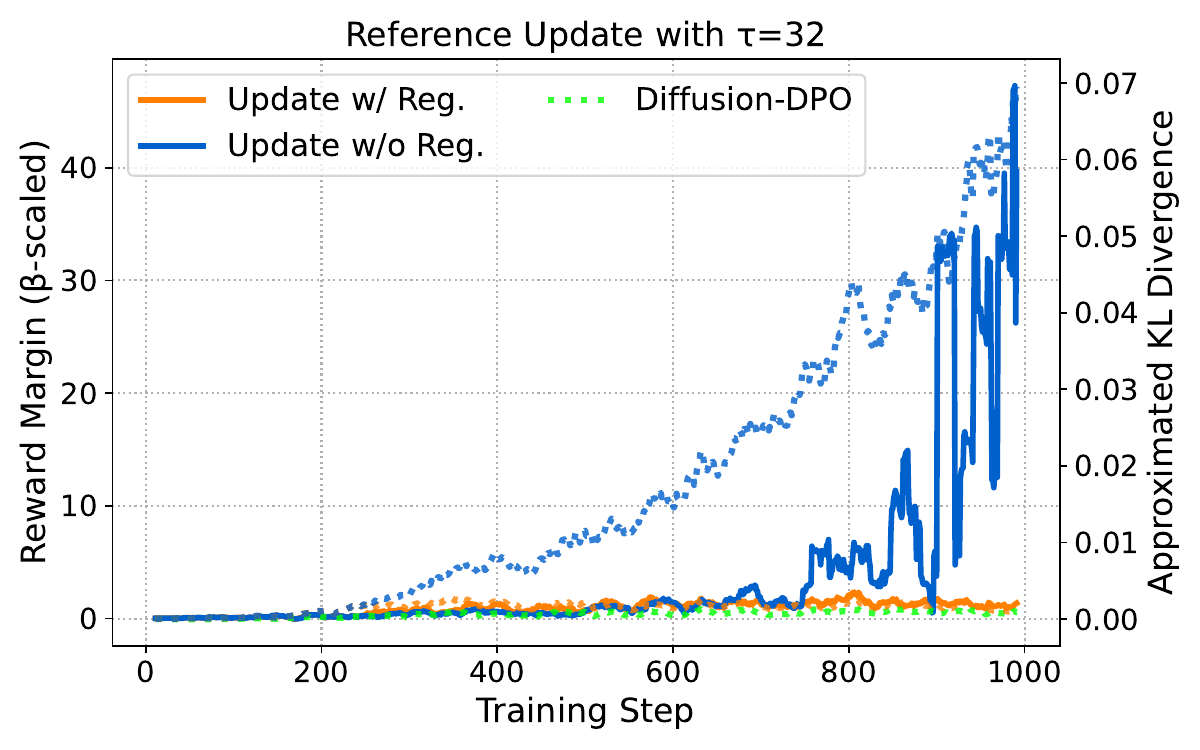}
        \caption{Divergence and reward margin}
    \end{subfigure}
    \begin{subfigure}[b]{0.3\textwidth}
        \centering
        \includegraphics[width=\linewidth]{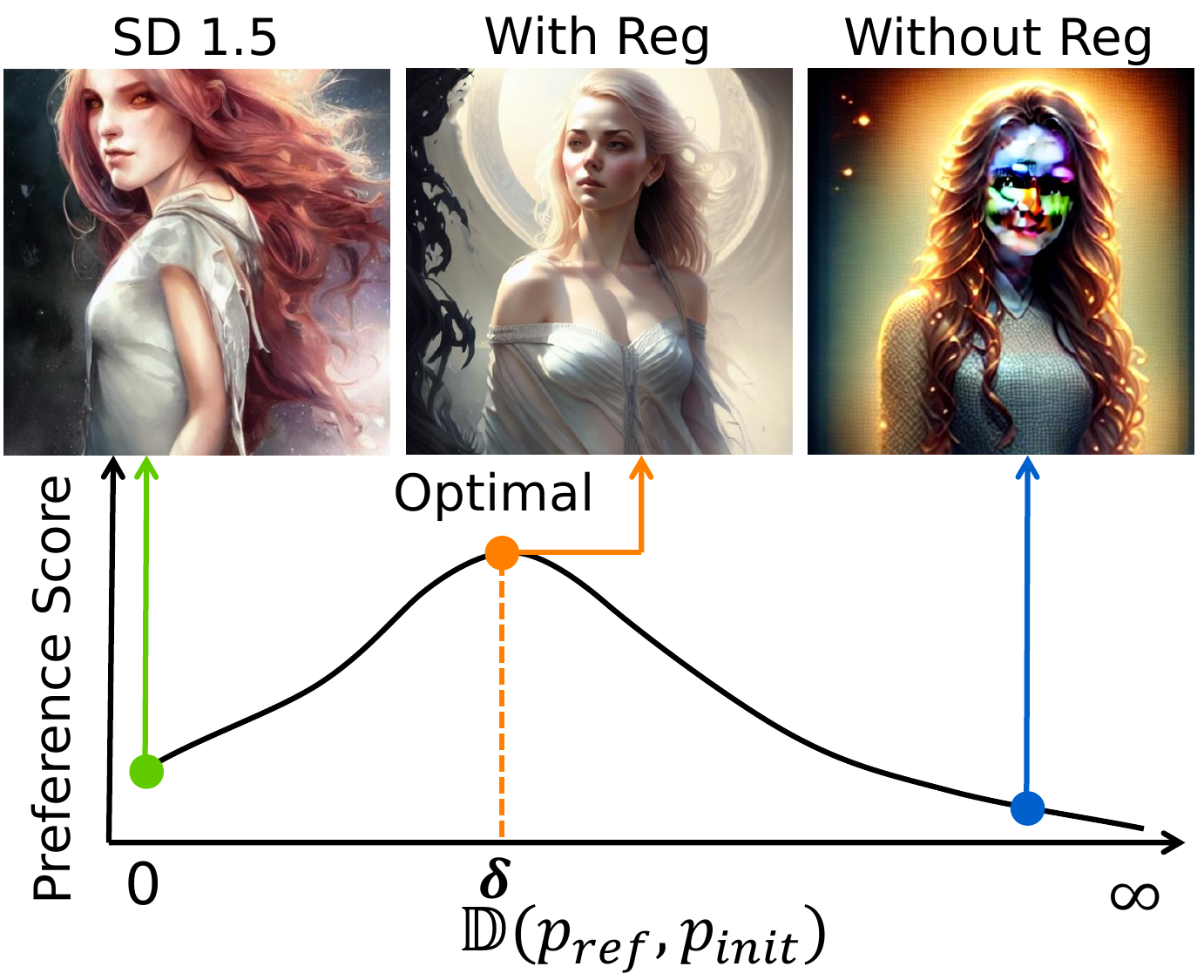}  
        \caption{Illustration of our exploration method}
    \end{subfigure}

    \caption{(a) Alignment performance of Diffusion-DPO, baselines, and our proposed method on SD1.5 with PickScore reward. Our method significantly improves the alignment performance over Diffusion-DPO.  
    (b) (solid lines) Implicit reward margin under the reference update strategy, with and without our regularization. (dotted lines) Approximated KL divergence between the training model and the pre-trained model (Diffusion-DPO), and between the reference model and the pre-trained model (ours).
    (c) Relationship between the divergence from the pre-trained model and the preference score. The illustration shows that controlled divergence enables effective exploration while excessive deviation results in a decline in preference score.
    }
    \label{fig:overview}
\end{figure*}

Reward model-based approaches typically rely on large vision-language models, such as PickScore~\cite{kirstain2023pick} and ImageReward~\cite{xu2024imagereward}. 
They are known to suffer from unstable training and reward overoptimization problems~\cite{hu2025better,kim2025testtime}.
In contrast, DPO~\cite{rafailov2023direct} offers a more stable alternative by directly optimizing the human preference data without the use of an explicit reward model. 
Extensions of DPO to diffusion models, such as Diffusion-DPO~\cite{wallace2024diffusion} and D3PO~\cite{yang2024using}, have shown early promise in the image generation domain. 
However, their alignment performance remain suboptimal compared to recent state-of-the-art methods~\cite{ethayarajh2024kto,zhu2025dspo}, as shown in Figure~\ref{fig:overview}(a). 

In this work, we identify a key limitation in current DPO adaptations in diffusion as constrained model exploration. 
Na\'ive Diffusion-DPO has relatively small divergence from the pre-trained model, suggesting limited traversal in model space (Figure~\ref{fig:overview}(b)).
This motivates our key hypothesis: encouraging greater exploration can help the model discover improved alignment solutions.

To this end, we adopt a reference update framework to promote exploratory behavior.
We find that updating the reference model forces the model to quickly change its prediction, leading to more exploration.
However, unrestricted reference updates lead to a model divergence problem, where the model loses its generative prior and degrades image quality.

Based on the observation that model error grows as the reference model diverges from the pre-trained model, we introduce a regularization algorithm to constrain the deviation of the reference model.
This adaptive strategy restricts excessive updates to the reference model when the deviation becomes large, preserving the generative prior while enabling meaningful exploration.
Despite its simplicity, this method offers important insights into the training stability of DPO for diffusion models and significantly improves alignment performance.

In addition, we observe that the impact of preference optimization with our exploration is imbalanced across diffusion timesteps, showing the need for emphasizing the learning of early timesteps. 
As several prior works~\cite{balaji2022ediff, wang2024first} demonstrated that diffusion models synthesize semantic structures during early timesteps, we aim to prioritize preference learning in early timesteps.
To accomplish this, we propose a timestep-aware optimization strategy for our exploration method. 
Specifically, we oversample early timesteps during loss computation and apply a decreasing reward scale schedule to balance reward magnitudes across timesteps.

The contributions of this paper are summarized as follows:
\begin{enumerate}
    \item We propose a novel recipe to improve direct preference optimization for T2I preference alignment, by introducing a stable reference model update method combined with a timestep-aware optimization strategy.
    \item Our analysis provides new insights into reference model relaxation and timestep-dependent behavior of preference optimization in diffusion models, distinguishing from existing methods.
    \item By combining the reference model update strategy with a timestep-aware optimization strategy, our method significantly enhances Diffusion-DPO's alignment performance and achieves state-of-the-art performance. This success highlights that effective exploration is key to maximizing the performance of DPO for diffusion models.
\end{enumerate}

\section{Preliminaries}
\label{sec:preliminaries}
\noindent\textbf{Diffusion Models.} Diffusion models are a class of generative models that learn to reverse a gradual noising process applied to data.
Following the DDPM~\cite{ddpm} formulation, the forward process is defined as a Markov chain with a noise schedule ${\alpha_t}$, resulting in a sequence of latent variables $\x_{1:T}$:
\begin{equation}
\label{eq:forward}
\begin{aligned}
q(\x_{1:T} \mid \x_0) &= \prod_{t=1}^{T} q(\x_t \mid \x_{t-1}), \\
\text{where} \quad q(\x_t \mid \x_{t-1}) &= \mathcal{N}(\x_t; \sqrt{\alpha_t} \x_{t-1}, (1 - \alpha_t)\I).
\end{aligned}
\end{equation}

The goal of the diffusion model is to learn a reverse process parameterized by a neural network $p_\theta(\x_{0:T}) = p(\x_T)\prod_{t=1}^T p_\theta(\x_{t-1}|\x_t)$ to obtain generated samples $p_\theta(\x_0)$.
Given $\x_t \sim q(\x_t | \x_{0}) = \calN(\bar{\alpha}_t \x_0, (1-\bar{\alpha}_t) \I)$, where $\bar{\alpha}_t = \prod_{s=1}^{t} \alpha_s $, the model estimates the noise $\boldsymbol{\epsilon} \sim \calN(\boldsymbol{0},\I) $ via $\boldsymbol{\epsilon}_\theta(\x_t, t)$. The training objective is derived from the evidence lower bound (ELBO) on the data likelihood:
\begin{equation}
\label{eq:loss-dm}
\mathcal{L}_{\text{DDPM}} = \bbE_{\x_0, \boldsymbol{\epsilon}, t} \left[\lambda(t) \left|| \boldsymbol{\epsilon} - \boldsymbol{\epsilon}_\theta(\x_t, t) \right||_2^2\right],
\end{equation}
where $t \sim \mathcal{U}(0, T)$ and $\lambda(t)$ denotes timestep-wise weighting function.
Recent works~\cite{choi2022p2,Hang2023minsnr,Hu2024Debias} suggest advanced weighting schedules for $\lambda(t)$ to improve sample quality and convergence.

\noindent\textbf{Preference Optimization in Diffusion.} 
To align the conditional distribution $p_\theta(\x_0|\vc)$ with human preferences, where $c \sim \calD_c$ denotes the prompt condition, RLHF methods~\cite{ouyang2022human,xu2024imagereward,blacktraining} utilize a reward model $r(\vc, \x_0)$.
These methods aim to maximize the reward of the generated sample $\x_0$ while keeping the distribution close to a reference distribution $\pref$ in terms of KL-divergence regularization:
\begin{multline}
\label{eq:rlhf}
\max_{p_\theta} \bbE_{\vc\sim \calD_c,\x_0\sim p_\theta(\x_0|\vc)} \left[r(\vc,\x_0)\right] - \\
\beta \kl\left[p_\theta(\x_0|\vc)\|p_{\text{ref}}(\x_0|\vc)\right].
\end{multline}

The reward model is typically learned from preference-annotated datasets under the Bradley-Terry model, where each data entry consists of a triplet $(c, \xw_0, \xl_0)$, representing a prompt, a preferred image, and a dispreferred image, respectively.
Rather than training a reward model, Direct Preference Optimization (DPO)~\cite{rafailov2023direct} parametrizes the \textit{implicit reward} using the current and the reference model:
\begin{equation}\label{eq:orig-dpo-r}
    r(\vc,\x_0) = \beta \log \frac{p_{\theta}(\x_0|\vc)}{\pref(\x_0|\vc)},
\end{equation}
where we omit the partition function $Z(\vc)=\sum_{\x_0}\pref(\x_0|\vc)\exp\left(r(\vc,\x_0)/\beta\right)$ as it does not contribute to the loss formulation.
Diffusion-DPO~\cite{wallace2024diffusion} expands the RLHF objective (Eq. \ref{eq:rlhf}) into the diffusion trajectory $p_\theta(\x_{0:T})$, and then plugs the implicit reward into the Bradley-Terry model to obtain the following loss:
\begin{multline}
\label{eq:initial_dpo_loss}
\mathcal{L}(\theta) 
= -\mathbb{E}_{(\xw_0, \xl_0) \sim \mathcal{D}} 
\log \sigma \Biggl( \\
\beta \,
\mathbb{E}_{\substack{
\xw_{1:T} \sim p_\theta(\xw_{1:T} \mid \xw_0), 
\xl_{1:T} \sim p_\theta(\xl_{1:T} \mid \xl_0)
}} \\ \Biggl[
\log \frac{p_\theta(\xw_{0:T})}{\pref(\xw_{0:T})}
- \log \frac{p_\theta(\xl_{0:T})}{\pref(\xl_{0:T})}
\Biggr] \Biggr).
\end{multline}
This is intractable as the loss requires sampling from $p_\theta(\x_{0:T})$.
Note that we omit the prompt $c$ for simplicity.
Utilizing Jensen's inequality and approximating the reverse process $p_\theta$ with the forward process $q$, Diffusion-DPO derives the final tractable loss:
\begin{multline}
\label{eq:loss-dpo}
\mathcal{L}(\theta)
= - \mathbb{E}_{ 
(\xw_0, \xl_0) \sim \mathcal{D},\,
t \sim \mathcal{U}(0, T),\,
\xw_t \sim q(\xw_t \mid \xw_0),\,
\xl_t \sim q(\xl_t \mid \xl_0)
} \\
\left[ \log \sigma \big( \beta T \lambda(t) \left( r_t(\xw_0) - r_t(\xl_0) \right) \big) \right],
\end{multline}
where we denote $r_t(\x_t) =  -(||\noise -\noise_\theta(\x_{t},t)\|^2_2 - \|\noise - \noise_\text{ref}(\x_{t},t)\|^2_2) $.
From this approximation, we interpret $r_t(\x_t)$ as a timestep-wise implicit reward.
Thus, the above loss can be regarded as forcing the model to maximize the margin between $r_t(\x_t^w)$ and $r_t(\x_t^l)$.

\section{Method}
\label{sec:method}
Our goal is to improve the preference alignment of Diffusion-DPO by addressing two limitations: insufficient exploration in the model space and imbalance in timestep-level learning.
To encourage exploration of the model, we begin by replacing the fixed reference model with the training model. 
We find that na\'ively updating the reference model leads to error scaling behavior, which can result in model divergence. 

To mitigate this issue, we constrain the divergence of the reference model from the pre-trained model, which allows the model to explore new solutions while preserving its generative quality.
However, we observe that our exploration method learns the preference signal unevenly across timesteps.
To facilitate the preference learning in early steps, we introduce a timestep-aware training strategy to address the imbalance problem.
By integrating this strategy with our exploration method, we further improve the performance of Diffusion-DPO.

\subsection{Reference Model Update with Regularization}
\label{sec:relax}

In standard DPO, the reference model remains fixed to the initial pre-trained model $\pinit$. While this design maintains training stability, it limits the model’s capacity to explore diverse alignment solutions. 
Recent works~\cite{pang2024iterative,zhang2025itercomp} have challenged this constraint by proposing multiple training stages using reward models, where the reference model is updated at each stage to improve preference alignment.
In language model alignment, TR-DPO~\cite{gorbatovski2024learn} demonstrates that updating the reference model during training can mitigate overoptimization and improve performance. 
Motivated by these findings, we extend this reference update strategy to the diffusion setting by periodically replacing the reference model with the current training model. 

\begin{figure}[t]
    \centering
    \begin{subfigure}[b]{0.49\columnwidth}
        \centering
        \includegraphics[width=\linewidth]{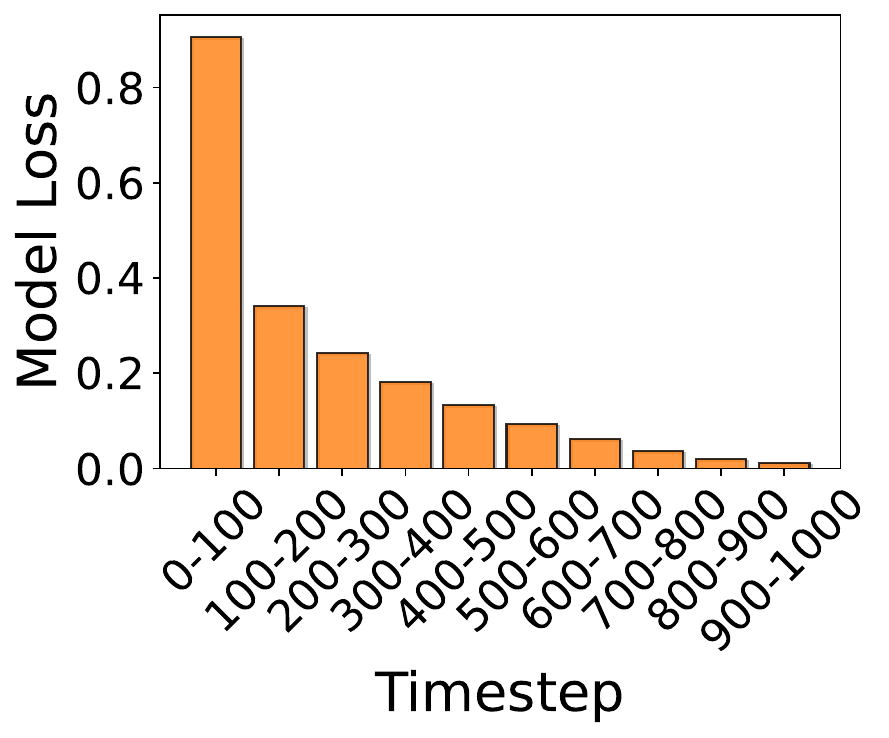}
        \caption{Model loss scale}
    \end{subfigure}
    \begin{subfigure}[b]{0.49\columnwidth}
        \centering
        \includegraphics[width=\linewidth]{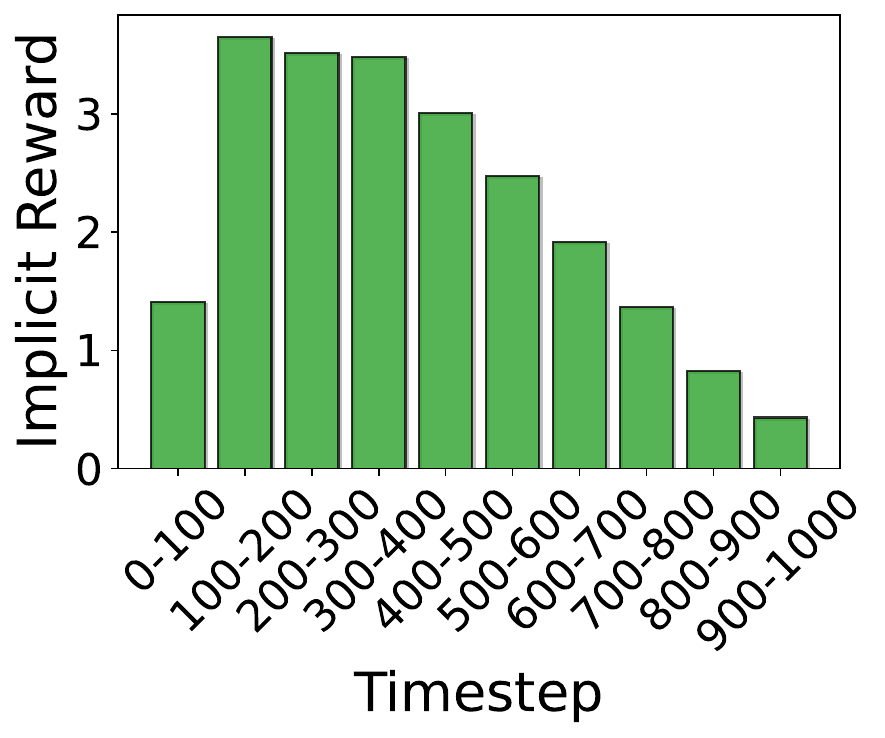}
        \caption{Implicit reward scale}
    \end{subfigure}
    \begin{subfigure}[b]{0.49\columnwidth}
        \centering
        \includegraphics[width=\linewidth]{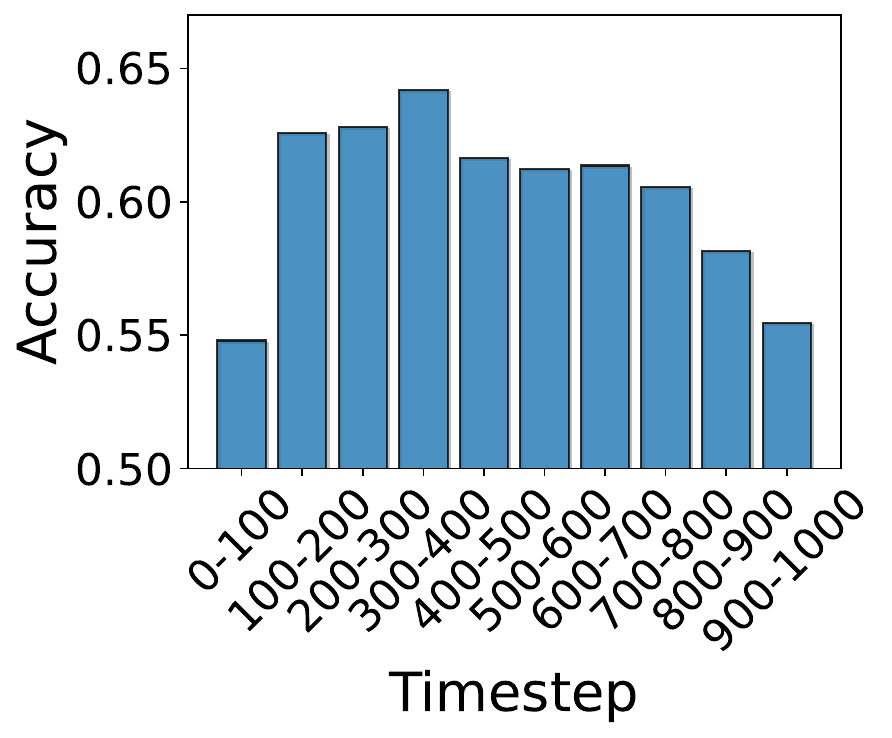} 
        \caption{Implicit reward accuracy}
    \end{subfigure}
    \begin{subfigure}[b]{0.49\columnwidth}
        \centering
        \includegraphics[width=\linewidth]{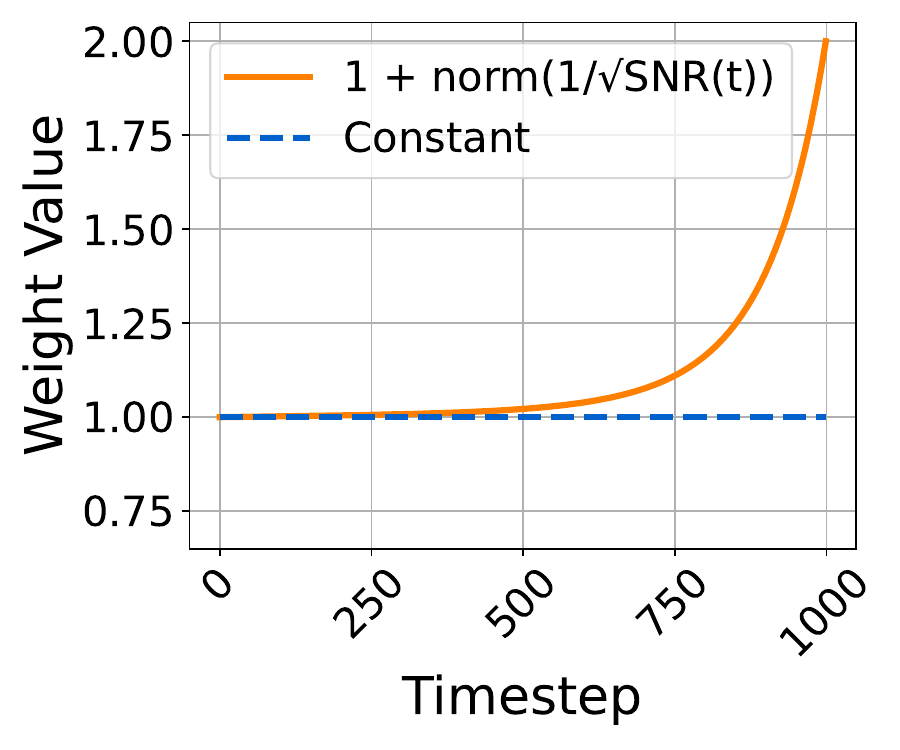}
        \caption{Reward scale schedule $\lambda(t)$}
    \end{subfigure}

    \caption{Imbalance problem in our reference update method. We present the scale of (a) model losses and (b) implicit rewards, (c) the preference accuracy of implicit rewards, (d) and our proposed reward scale schedule $\lambda(t)$.}
    \label{fig:timestep_balance}
\end{figure}

We observe that a na\'ive reference update strategy in diffusion models leads to a critical model divergence problem. 
To analyze this, we examine the training dynamics with a reference update period of $\tau = 32$ 
(see Appendix C for other values).
Figure~\ref{fig:overview}(b) shows the growing divergence of the reference model from the pre-trained model, along with the implicit reward margin $r_t(\x_t^w) - r_t(\x_t^l)$ (Update w/o Reg.).

As training progresses, both divergence and reward margin increase, indicating that the model is actively optimizing toward the DPO objective through exploration.
This also implies growing prediction error, quantified as $\left||\boldsymbol{\epsilon} - \boldsymbol{\epsilon}_\theta(\x_t, t)|\right|_2^2$.
Smaller $\tau$ values amplify this effect by reducing the gap between training and reference models, forcing the model to scale its prediction error more aggressively.
Although moderate exploration may help the model discover better solutions, the uncontrolled error explosion ultimately causes the model to diverge.
This behavior contrasts with observations in TR-DPO~\cite{gorbatovski2024learn} for language models, where updating the reference model tends to reset the reward margin toward zero during training. 
In the diffusion setting, however, excessive reference drift degrades image quality due to error scaling.

To balance exploration and training stability, we propose to \textit{regularize the reference model} by constraining its divergence from the pre-trained model (Figure~\ref{fig:overview}(c)).
Our key insight is that excessive divergence leads to increased prediction error.
By limiting this divergence, we can suppress error scaling while enabling controlled exploration.

We define a divergence metric $\mathbb{D}(\pref, \pinit)$, to quantify the deviation of the current reference model $\pref$ from the initial model $\pinit$.
Estimating this divergence requires computing the expectation under the joint distribution of $\pref$ or $\pinit$ across timesteps, which is intractable.
Instead, we approximate the divergence using the forward process $q$.
Specifically, the (reverse) KL divergence can be approximated as follows:
\begin{equation}
\kl(\pref,\pinit)
\approx
\mathbb{E}_{\x_{0}\sim\mathcal{D},\,
           \x_{1:T}\sim q(\x_{1:T}\mid\x_0)}
\bigl[\log\tfrac{\pref(\x_{0:T})}{\pinit(\x_{0:T})}\bigr].
\end{equation}

Using a similar derivation to equation in~\cite{wallace2024diffusion}, we obtain the following:

{\small
\begin{multline}
\label{eq:kl_approx}
\kl(\pref, \pinit) \approx 
T\, \mathbb{E}_{\x_0 \sim \mathcal{D},\, t,\, \x_t \sim q(\x_t \mid \x_0)} \\
\Bigl[
\kl\bigl(
q(\x_{t-1} \mid \x_{0,t}) \,\big\|\, \pref(\x_{t-1} \mid \x_t)
\bigr) \\
- \kl\bigl(
q(\x_{t-1} \mid \x_{0,t}) \,\big\|\, \pinit(\x_{t-1} \mid \x_t)
\bigr)
\Bigr].
\end{multline}
}

We empirically find that this approximation proves sufficient for divergence monitoring.
To reduce computational overhead, we evaluate the divergence on a small subset of preferred images $\x_0^w$ from the training batch.
After establishing the divergence metric, the next step is to choose a reference model that ensures training stability. 
When the divergence of the current reference model $\pref$ exceeds a threshold $\delta$, we freeze the reference model near the $\delta$-boundary to prevent further updates, as shown in Figure~\ref{fig:overview}(b) (Update w/ Reg.).
We also explore a re-initialization option in Section~\ref{sec:ablation}, which resets the reference model to the pre-trained model.

\subsection{Boosting with Timestep-Aware Optimization}
\label{sec:time}

Despite the benefits of our reference update method, we find that the learned preference signal is unevenly distributed across diffusion timesteps. 
This leads to suboptimal alignment, particularly in early steps where semantic structures are formed~\cite{balaji2022ediff, wang2024first}. 
In fact, several diffusion training studies \cite{Hu2024Debias, choi2022p2} discovered that optimization is more difficult in early timesteps and emphasizing these steps improves the output quality (Figure~\ref{fig:timestep_balance}(a)).

In the preference optimization setting, we observe a similar trend during our exploration. 
To investigate this, we analyze the implicit reward $r_t(\x_t)$ using a model trained with our reference update strategy. 
We randomly sample 5,000 image pairs from the Pick-a-Pic v2 validation set~\cite{kirstain2023pick} and compute both the average scale of implicit reward and preference accuracy (the number of cases where $r_t(\x_t^w)$ is greater than $r_t(\x_t^l)$) across 10 evenly partitioned intervals $[0, T]$.

As shown in Figure~\ref{fig:timestep_balance}(b) and (c), both the scale and accuracy of $r_t(\x_t)$ are marked lower at early timesteps. 
This finding indicates that the reward signal is weaker in early steps, leading to imbalanced preference learning difficulty.
Motivated by this observation, we aim to develop a timestep-aware preference optimization strategy that accounts for this imbalance.

To encourage preference learning in early steps, we apply an oversampling approach inspired by~\cite{yang2024dense}, drawing a single timestep $t$ instead of multisample expectations.
In this method, timesteps are drawn from a skewed categorical distribution $\text{Cat}(\gamma^t)$ towards early steps, with probability vector ${ \gamma^t / \sum_{t'} \gamma^{t'} }$, where $\gamma \in [0, 1]$. 
Moreover, we introduce a timestep-dependent reward scaling schedule $\lambda(t)$ to directly mitigate imbalance.
Although Eq.~\ref{eq:loss-dpo} already presents the weighting schedule, it has been ignored in previous works and treated as a constant in practice~\cite{wallace2024diffusion, zhu2025dspo}.
Instead, we design $\lambda(t)$ to decrease over timesteps, assigning larger values than the constant schedule during early steps.
As an example, we define $\lambda(t) = 1+ \text{norm}(1/\sqrt{SNR(t)})$, where $SNR(t)$ denotes the signal-to-noise ratio, $\text{norm}(\cdot)$ indicates the normalization operator over time (Figure~\ref{fig:timestep_balance}(d)).
As $\lambda(t)$ controls the implicit regularization via $\beta$, we also interpret this schedule as a means to reduce the risk of overfitting at early timesteps.
We explore other choices in Appendix C, verifying the advantage of the proposed schedule. 

We note that the timestep-aware strategy alone may not yield performance gains in isolation (Section~\ref{sec:experiment_results}).
Our key contribution lies in its synergy with our exploration method, which unlocks the potential of DPO for diffusion models.

\section{Experiment}
\label{sec:experiments}

\begin{table*}[t]
\centering
\begin{tabular}{l l c c c c c c }
\toprule
\textbf{Dataset} & \textbf{Model} & \textbf{PickScore} & \textbf{HPSv2} & \textbf{CLIP} & \textbf{Aesthetic} & \textbf{ImageReward} & \textbf{Average} \\
\cmidrule(lr){1-1} \cmidrule(lr){2-2} \cmidrule(lr){3-7} \cmidrule(lr){8-8} 
\multirow{5}{*}{PickV2} 
& vs. SD1.5*     & \textbf{89.96} & \textbf{83.84} & \textbf{64.56} & \textbf{78.04} & \textbf{77.76} & \textbf{78.83} \\
& vs. Diff-KTO*  & \textbf{74.52} & \textbf{52.16} & \textbf{56.16} & \textbf{56.00} & \textbf{51.80} & \textbf{58.13} \\
& vs. SFT        & \textbf{71.72} & \textbf{50.40} & \textbf{55.00} & 49.04 & \textbf{53.08} & \textbf{55.85} \\
& vs. Diff-DPO   & \textbf{75.20} & \textbf{70.80} & \textbf{53.64} & \textbf{69.16} & \textbf{66.36} & \textbf{67.03} \\
& vs. DSPO       & \textbf{71.36} & \textbf{51.76} & \textbf{53.72} & \textbf{51.32} & \textbf{51.16} & \textbf{55.86} \\

\midrule
\multirow{5}{*}{PartiPrompts} 
& vs. SD1.5*     & \textbf{84.25} & \textbf{84.31} & \textbf{60.66} & \textbf{81.00} & \textbf{80.82} & \textbf{78.21} \\
& vs. Diff-KTO*  & \textbf{71.57} & \textbf{56.80} & \textbf{53.80} & \textbf{65.32} & \textbf{62.56} & \textbf{62.01} \\
& vs. SFT        & \textbf{71.38} & \textbf{56.43} & \textbf{55.76} & \textbf{59.07} & \textbf{64.46} & \textbf{61.42} \\
& vs. Diff-DPO   & \textbf{72.18} & \textbf{75.80} & \textbf{53.19} & \textbf{75.37} & \textbf{73.47} & \textbf{70.00} \\
& vs. DSPO       & \textbf{69.73} & \textbf{56.56} & \textbf{53.74} & \textbf{60.48} & \textbf{62.68} & \textbf{60.64} \\

\midrule
\multirow{5}{*}{HPDv2} 
& vs. SD1.5*     & \textbf{91.44} & \textbf{89.34} & \textbf{63.62} & \textbf{82.66} & \textbf{84.22} & \textbf{82.26} \\
& vs. Diff-KTO*  & \textbf{73.12} & \textbf{53.69} & \textbf{52.75} & \textbf{56.59} & \textbf{55.31} & \textbf{58.29} \\
& vs. SFT        & \textbf{73.88} & \textbf{57.88} & \textbf{54.94} & \textbf{53.75} & \textbf{58.38} & \textbf{59.77} \\
& vs. Diff-DPO   & \textbf{77.22} & \textbf{77.81} & \textbf{53.87} & \textbf{69.62} & \textbf{73.50} & \textbf{70.40} \\
& vs. DSPO       & \textbf{72.28} & \textbf{57.75} & \textbf{53.97} & \textbf{53.09} & \textbf{57.44} & \textbf{58.91} \\
\bottomrule
\end{tabular}%
\caption{Win rates of our method against baseline preference optimization methods using SD1.5 as the base model. * indicates model checkpoints released by the original authors. Higher win rates indicate better alignment performance and win rates exceeding 50\% are marked in bold.}
\label{tab:model_comparison}
\end{table*}

\begin{table*}[t]
\centering
    \begin{tabular}{l c c c c c c}
    \toprule
    \textbf{Model} & \textbf{PickScore} & \textbf{HPSv2} & \textbf{CLIP} & \textbf{Aesthetic} & \textbf{ImageReward} & \textbf{Average} \\
    \cmidrule(lr){1-1} \cmidrule(lr){2-6} \cmidrule(lr){7-7}
    vs. SDXL* & \textbf{81.24} & \textbf{81.76} & \textbf{57.64} & \textbf{59.28} & \textbf{70.96} & \textbf{70.18} \\
    vs. MaPO* & \textbf{81.16} & \textbf{74.88} & \textbf{58.16} & 45.12 & \textbf{65.92} & \textbf{65.05} \\
    vs. InPO* & \textbf{64.80} & \textbf{56.56} & \textbf{54.76} & \textbf{55.00} & \textbf{56.76} & \textbf{57.58} \\
    vs. Diff-DPO & \textbf{68.40} & \textbf{73.76} &	\textbf{50.28} & \textbf{57.52} &\textbf{54.40} &\textbf{60.87} \\
    vs. DSPO & \textbf{60.88} & \textbf{64.68} & \textbf{51.44} & \textbf{55.52} & 49.28 & \textbf{56.36} \\
    \bottomrule
    \end{tabular}%
\caption{Win rates of our method using SDXL as the base model, evaluated on the Pick-a-Pic v2 test
set.} 
\label{tab:sdxl_winrates}
\end{table*}

\subsection{Experimental Setup}
\label{sec:experiental_setup}
\textbf{Dataset.} 
Following prior works~\cite{wallace2024diffusion,lialigning}, we use Pick-a-Pic v2 train dataset~\cite{kirstain2023pick} for training. 
For evaluation, we employ test set prompts from the Pick-a-Pic v2 dataset (500 entries), HPDv2 benchmark~\cite{wu2023human} (3,200 entries), and the PartiPrompts dataset~\cite{yu2022scaling} (1,632 entries). 
As Pick-a-Pic v2 has a small number of prompts, we generate a total of 2,500 images using five different seeds.


\noindent\textbf{Evaluation Protocol.}
To quantitatively evaluate the proposed method, we adopt five reward models as evaluation metrics: PickScore~\cite{kirstain2023pick}, HPSv2~\cite{wu2023human},  CLIP~\cite{radford2021learning}, Aesthetics Score~\cite{laionaes}, and ImageReward~\cite{xu2024imagereward}.
For each reward model, we compare the win rates of our method against the baseline approaches.
The win rate is the proportion of images with higher reward scores than those generated by the baseline model, under the same seed.

\noindent\textbf{Baseline Methods.} 
We evaluate our method against baseline preference optimization algorithms, Diffusion-DPO~\cite{wallace2024diffusion}, Diffusion-KTO~\cite{lialigning}, and DSPO~\cite{zhu2025dspo}.
We reproduce Diffusion-DPO and DSPO, and use a public checkpoint for Diffusion-KTO.
When reproducing the baseline methods, we maintain consistency by employing the same hyperparameters reported in the original paper. 
We also include supervised fine-tuning (SFT) as a baseline, but we exclusively use the preferred images. 

\noindent\textbf{Implementation Details.}
In this paper, we conduct experiments on Stable Diffusion v1.5 (SD1.5)~\cite{rombach2022high} and SDXL~\cite{sdxl}. 
We tune the reference model update period, $\tau$, by searching over $\{16, 32, 64\}$ steps and select the optimal value for each model.
The monitoring divergence threshold $\delta$ is empirically determined as $0.005$ for SD1.5 and $0.002$ for SDXL.
For the timestep-aware training strategy, we set the discount factor $\gamma$ for the timestep sampling to $0.9$ as the default.
Other details and hyperparameters are provided in Appendix A.

\subsection{Experiment Results}
\label{sec:experiment_results}
\textbf{Quantitative Results.} 
To verify the effectiveness of the proposed method, we compare our method with the original Diffusion-DPO and baseline preference optimization algorithms.
Table~\ref{tab:model_comparison} presents the experimental results, measured in win rates from five reward metrics and their average.
Notably, when comparing our method to Diffusion-DPO, the average win rate ranges from 67\% to 70\% across datasets, indicating significant improvement of alignment.
These findings underscore that model exploration plus the timestep-aware training strategy can unlock the potential of Diffusion-DPO.
We further report our results on SDXL in Table~\ref{tab:sdxl_winrates}, including public checkpoints of MAPO~\cite{hong2024marginaware} and InPO~\cite{lu2025inpo} as baselines.
Due to space constraints, we report results for the remaining test prompt sets and raw reward scores in Appendix B.

\begin{figure*}[h!]
    \centering
    \parbox{1.02 \textwidth}{
        \centering
        
3D digital illustration, Burger with wheels  speeding on the race track, supercharged, detailed, hyperrealistic, 4K
    }
    \begin{subfigure}[t]{0.19\textwidth} \includegraphics[width=\textwidth]{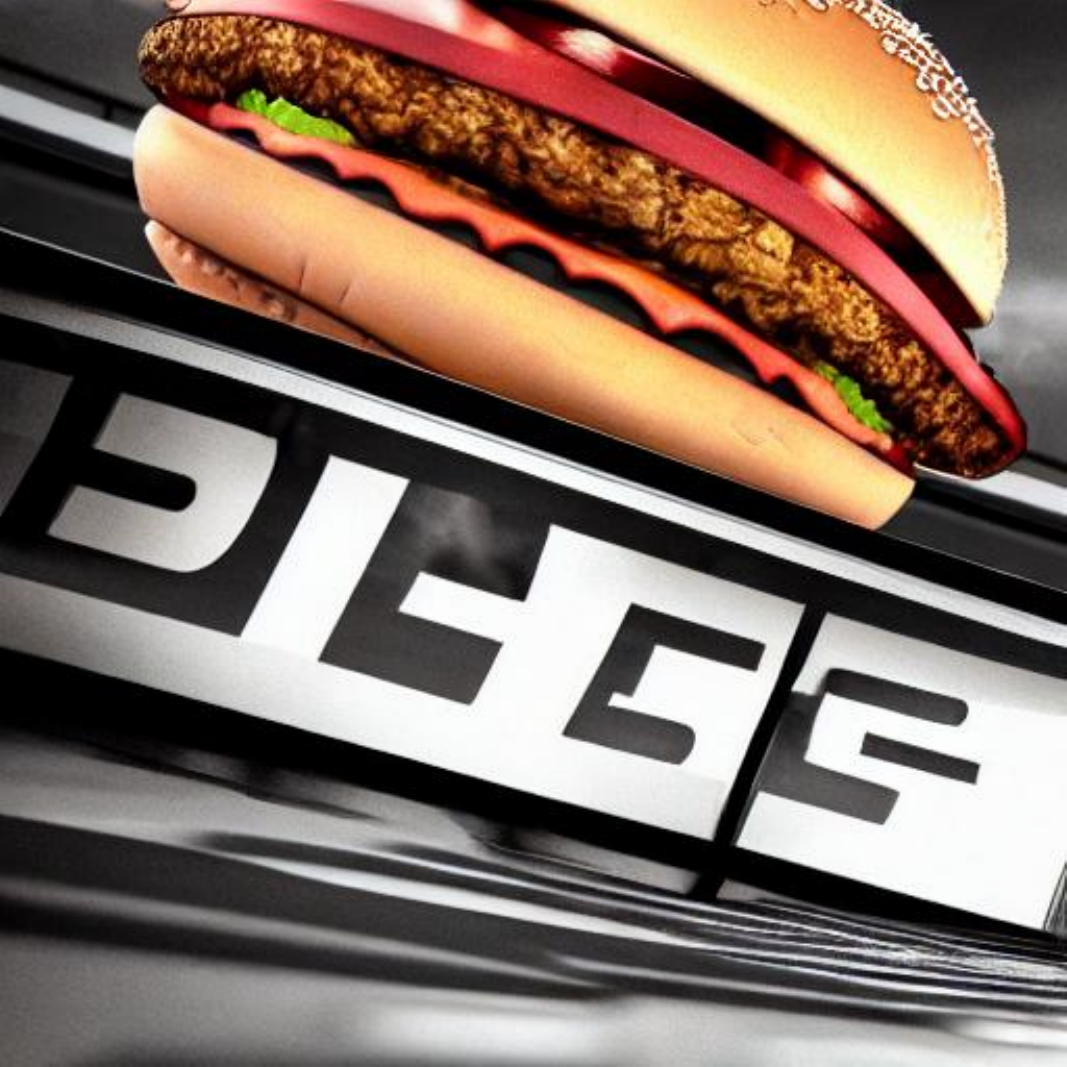} \end{subfigure}
    \begin{subfigure}[t]{0.19\textwidth} \includegraphics[width=\textwidth]{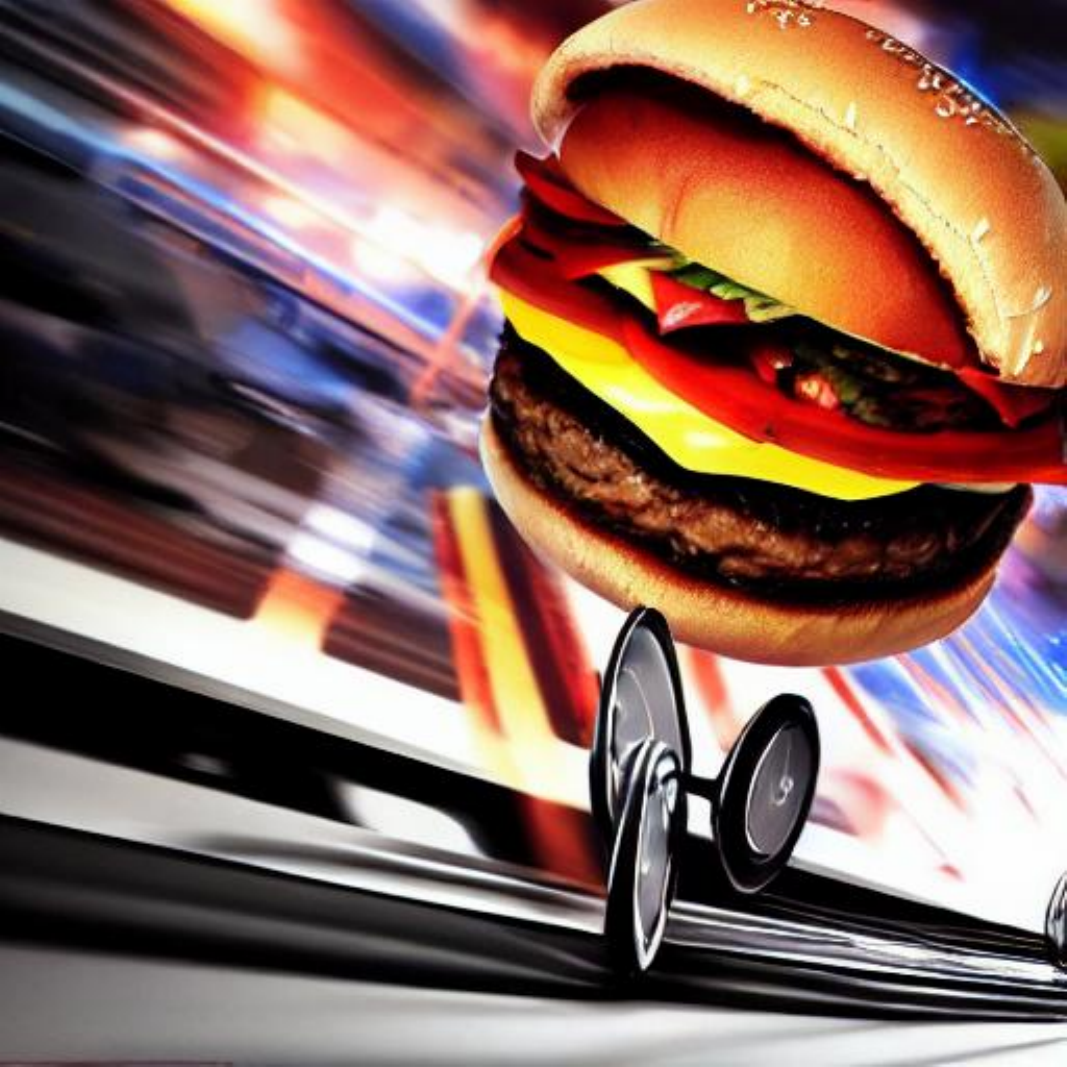} \end{subfigure}
    \begin{subfigure}[t]{0.19\textwidth} \includegraphics[width=\textwidth]{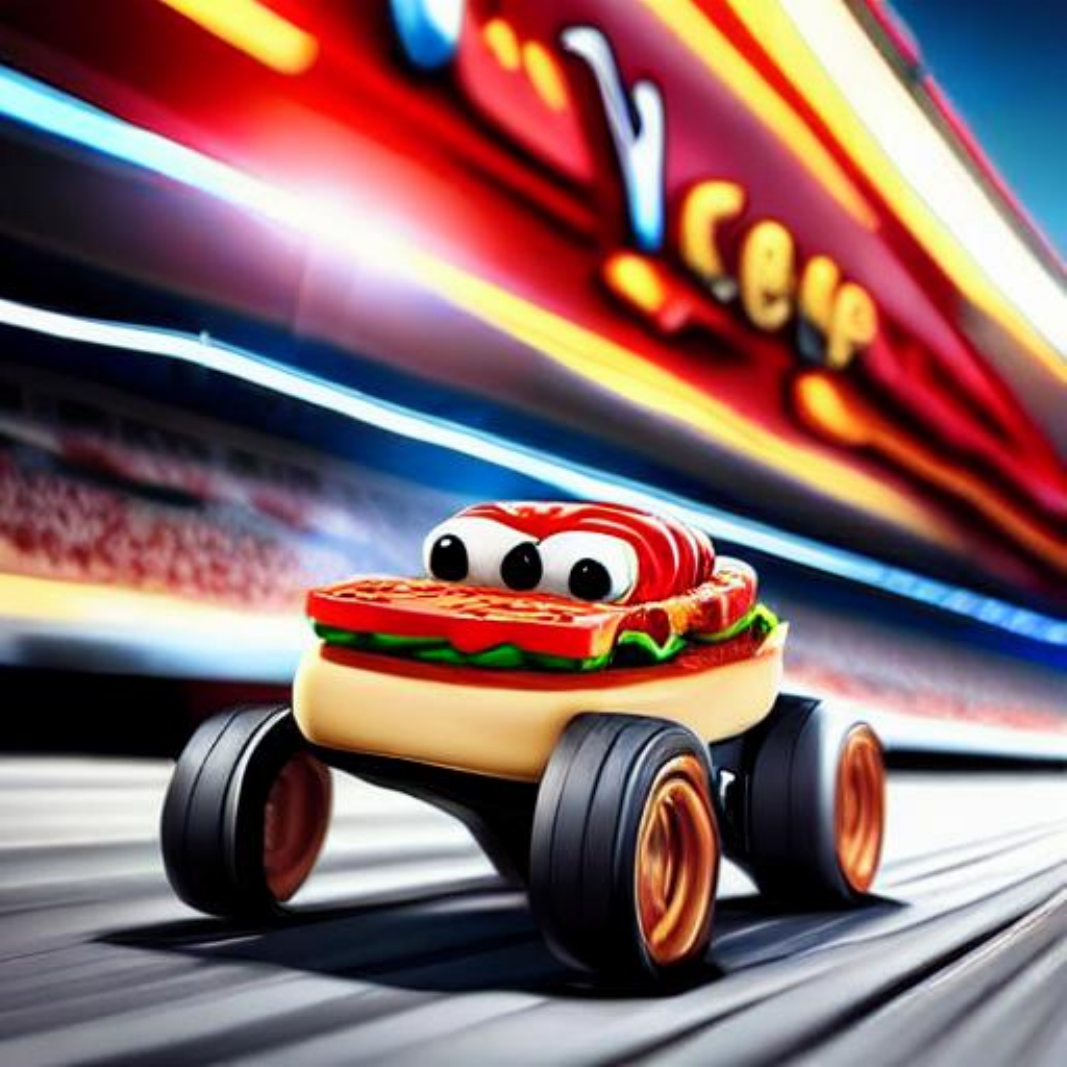} \end{subfigure}
    \begin{subfigure}[t]{0.19\textwidth} \includegraphics[width=\textwidth]{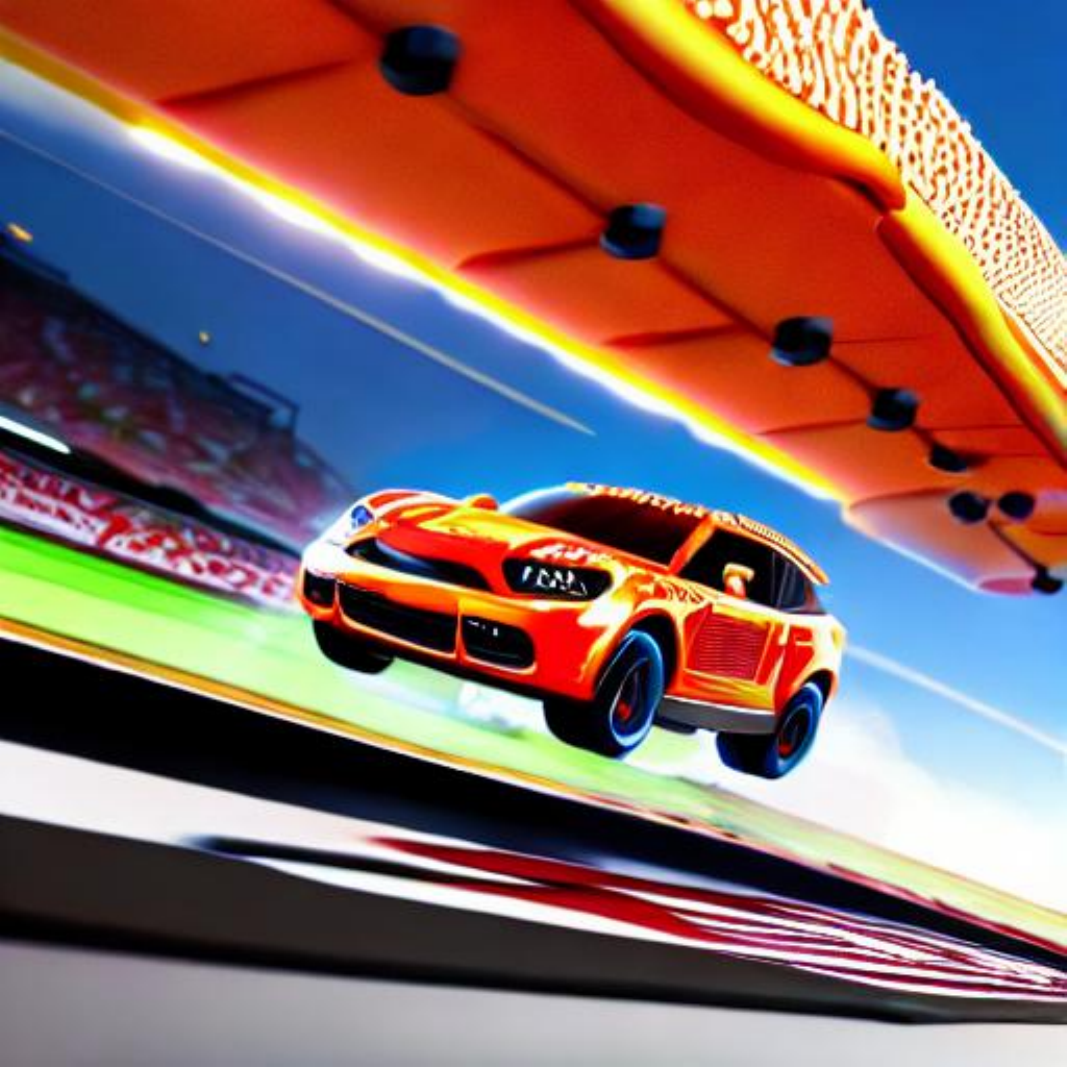} \end{subfigure}
    \begin{subfigure}[t]{0.19\textwidth} \includegraphics[width=\textwidth]{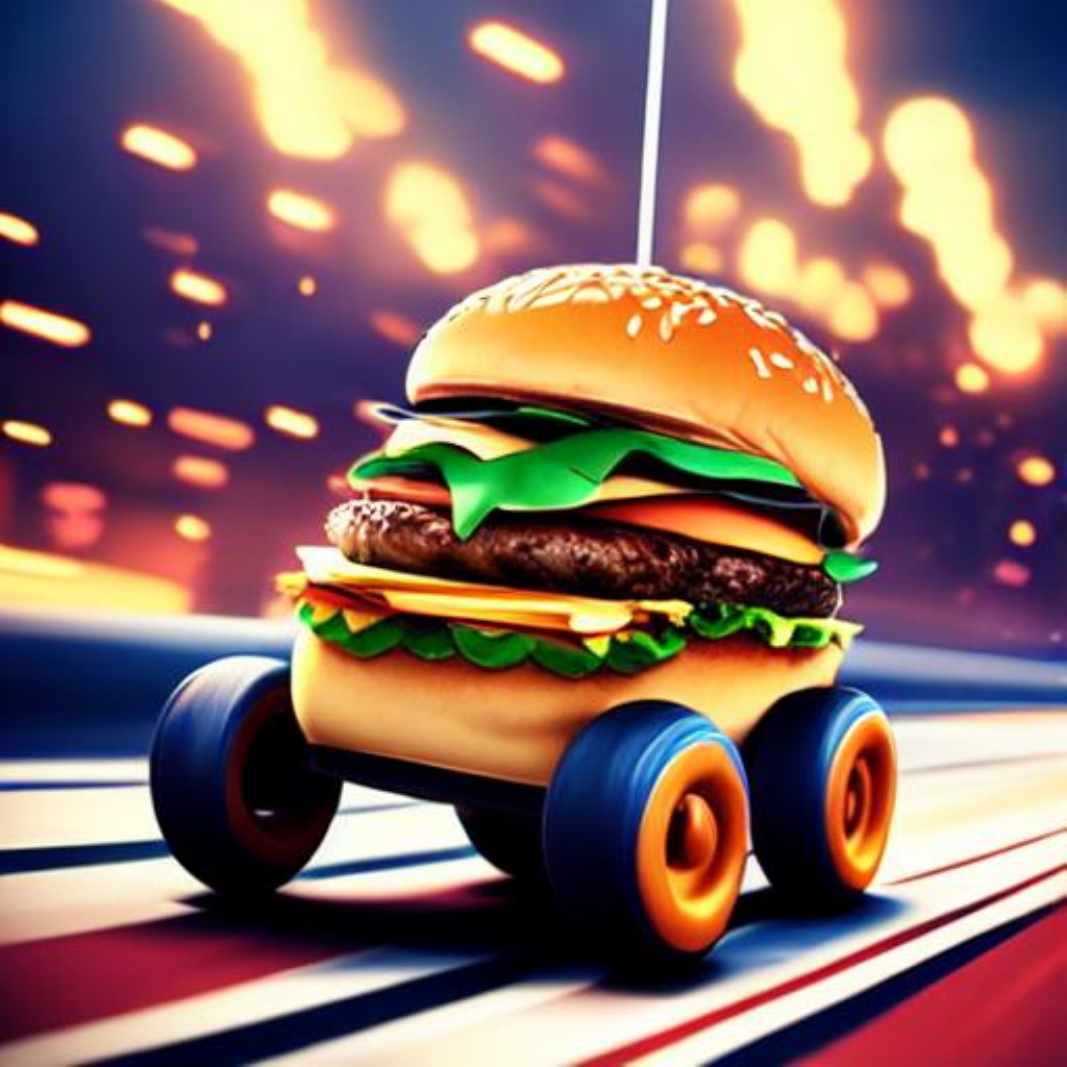} \end{subfigure}

    \parbox{1.02 \textwidth}{
        \centering
            Cyberpunk cat
    }
    \begin{subfigure}[t]{0.19\textwidth} \includegraphics[width=\textwidth]{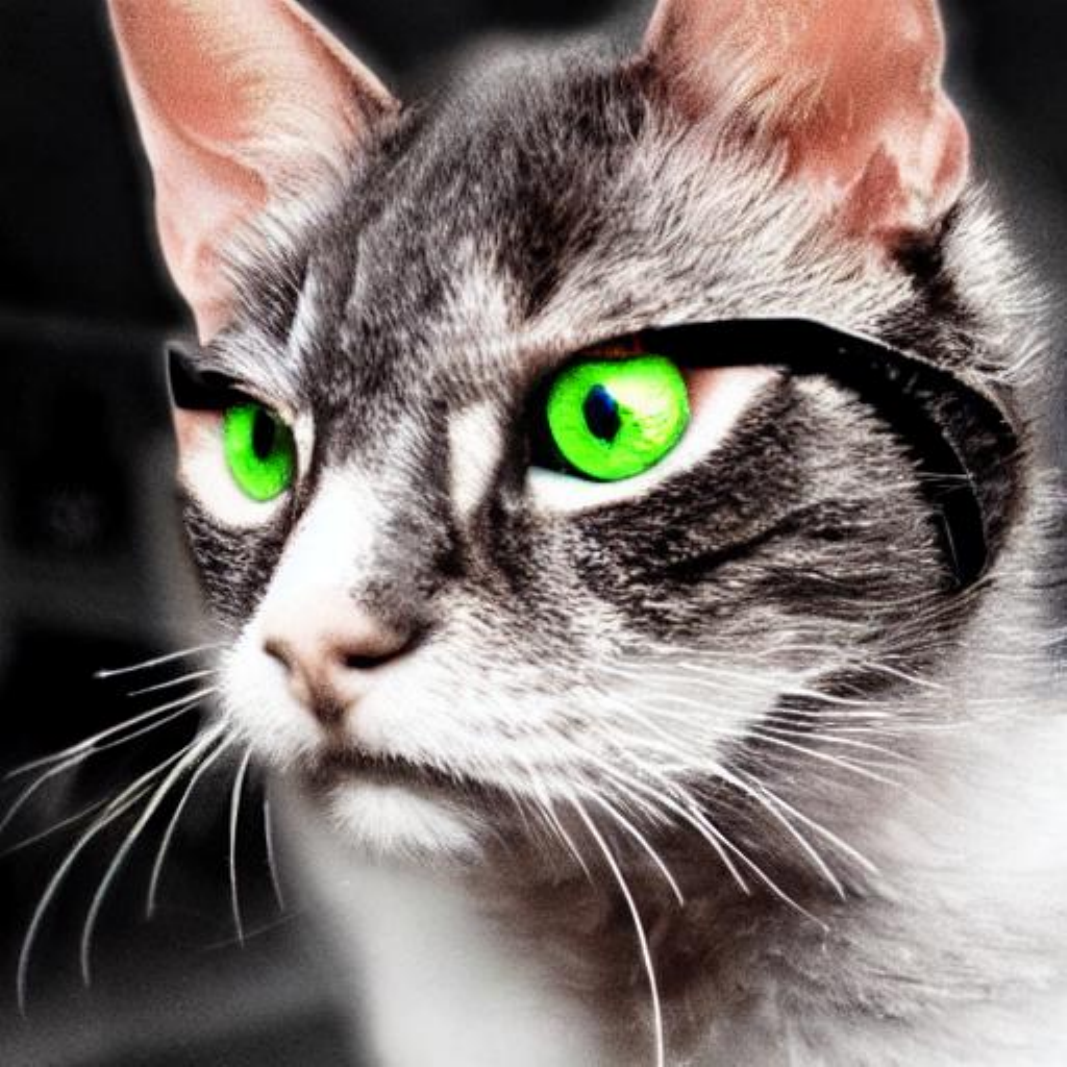} \end{subfigure}
    \begin{subfigure}[t]{0.19\textwidth} \includegraphics[width=\textwidth]{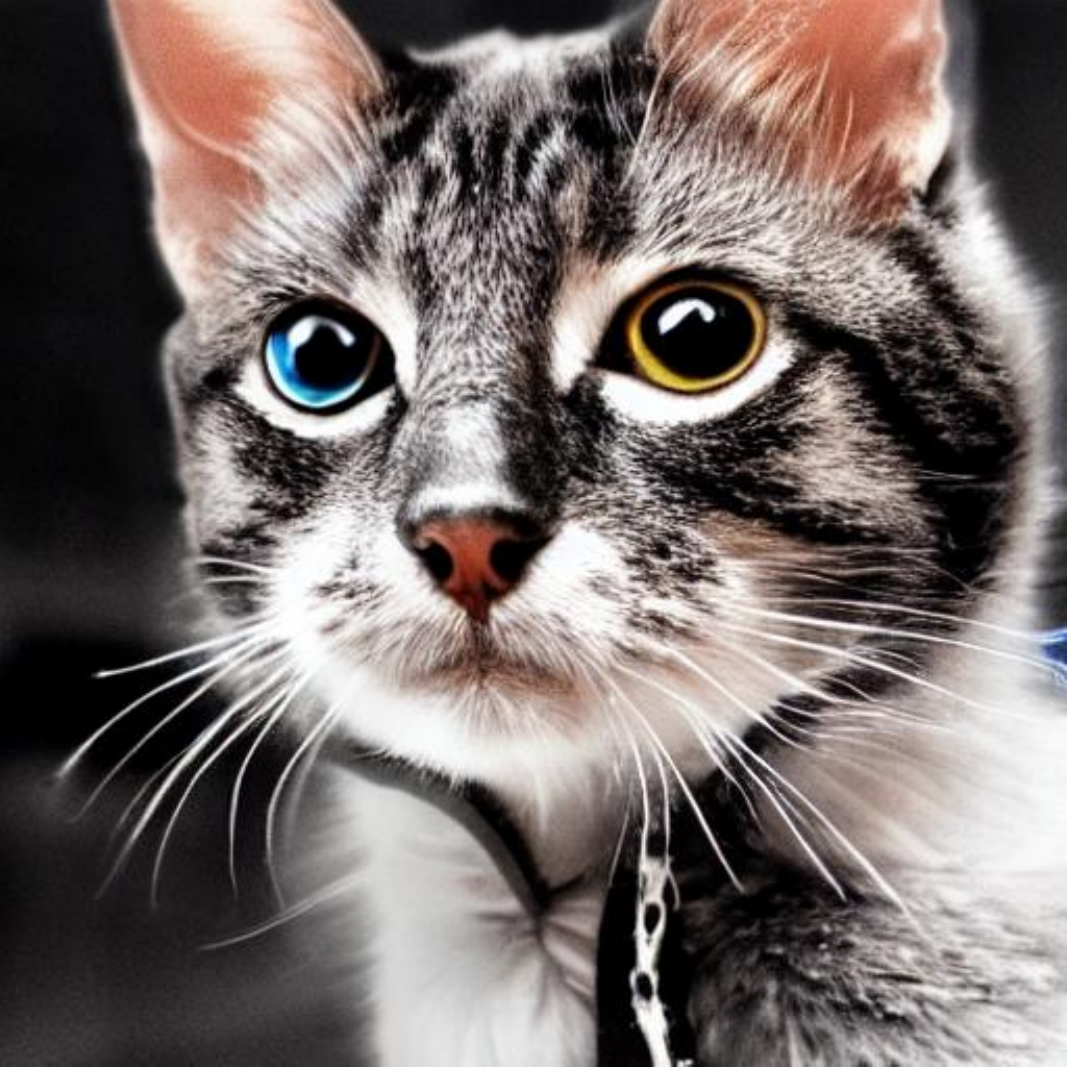} \end{subfigure}
    \begin{subfigure}[t]{0.19\textwidth} \includegraphics[width=\textwidth]{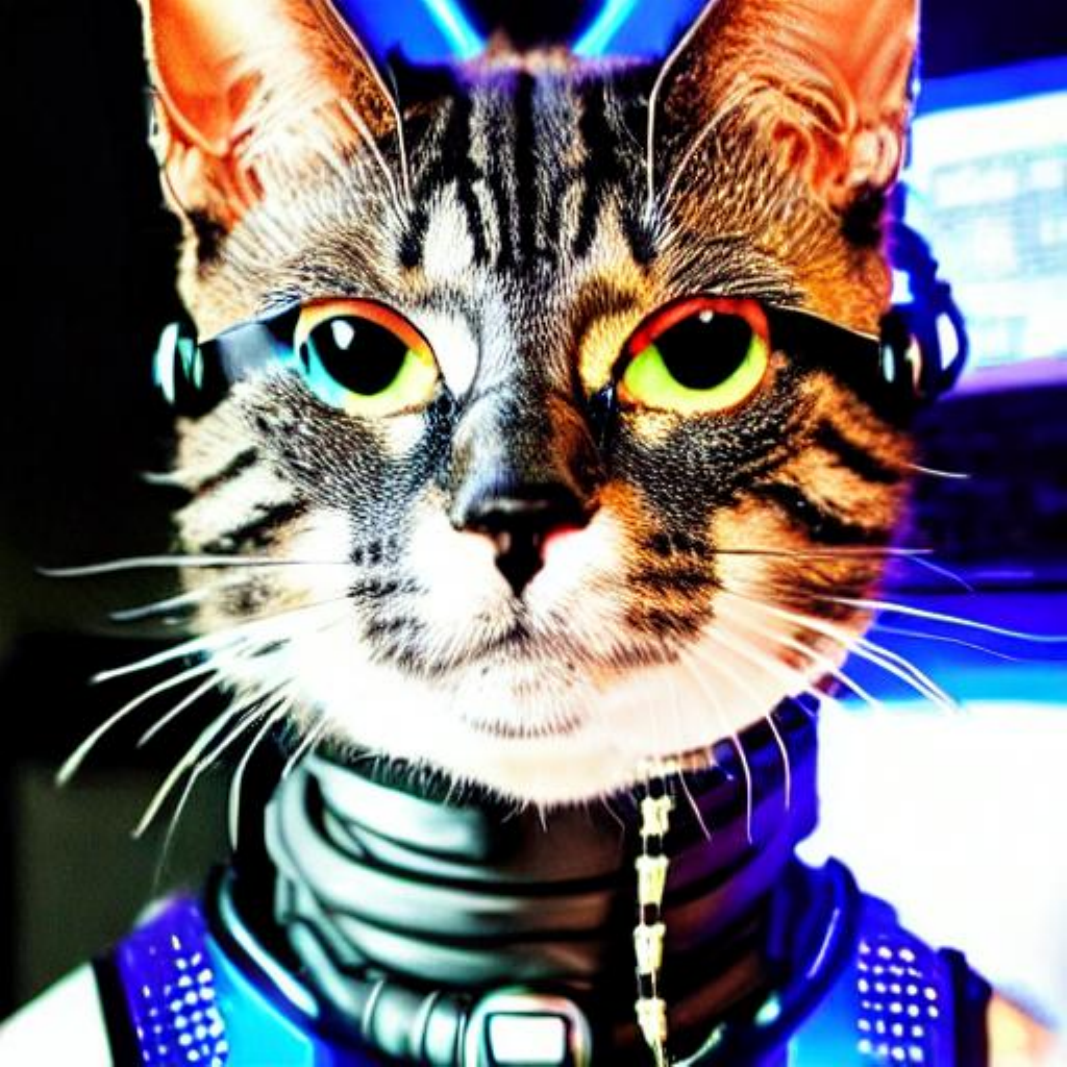} \end{subfigure}
    \begin{subfigure}[t]{0.19\textwidth} \includegraphics[width=\textwidth]{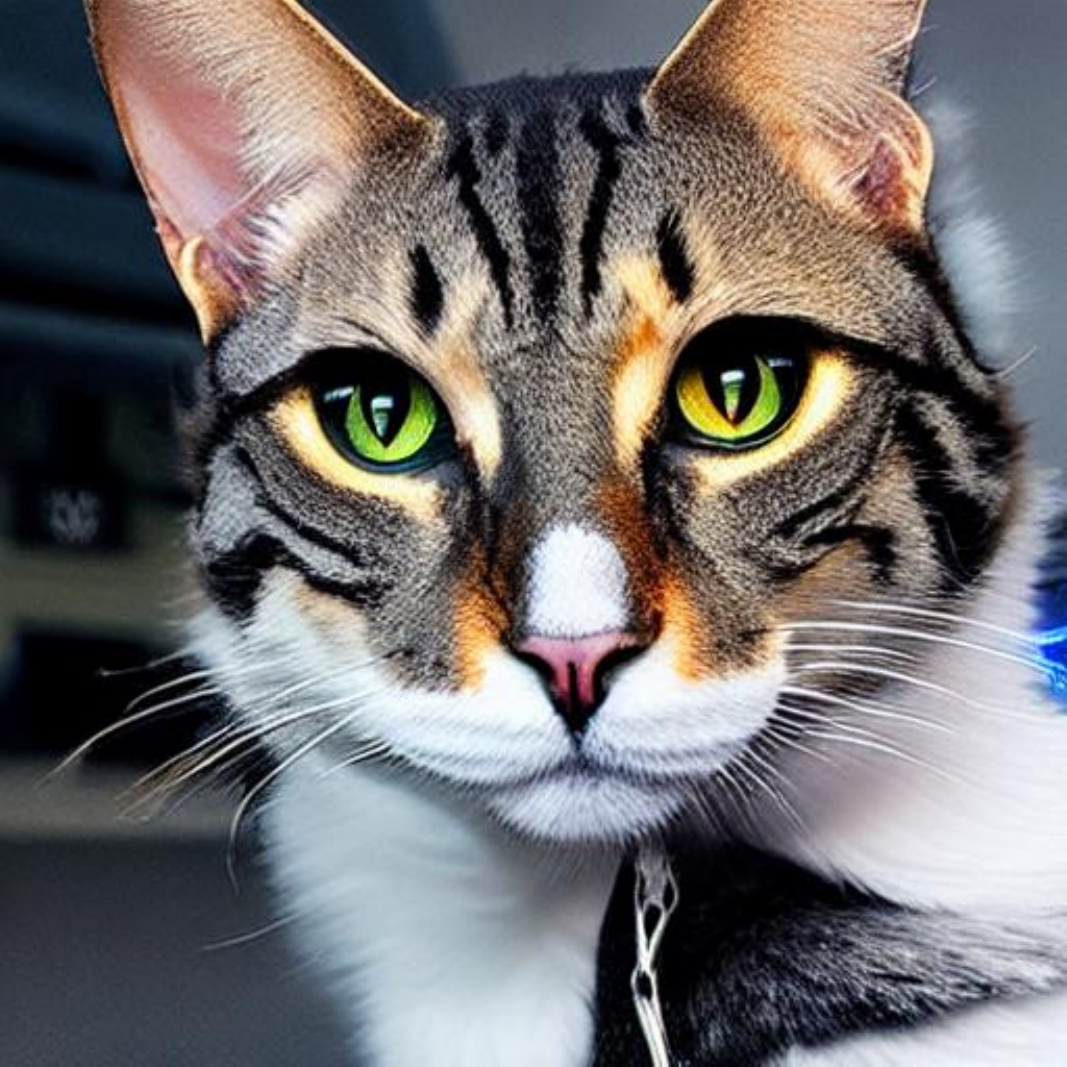} \end{subfigure}
    \begin{subfigure}[t]{0.19\textwidth} \includegraphics[width=\textwidth]{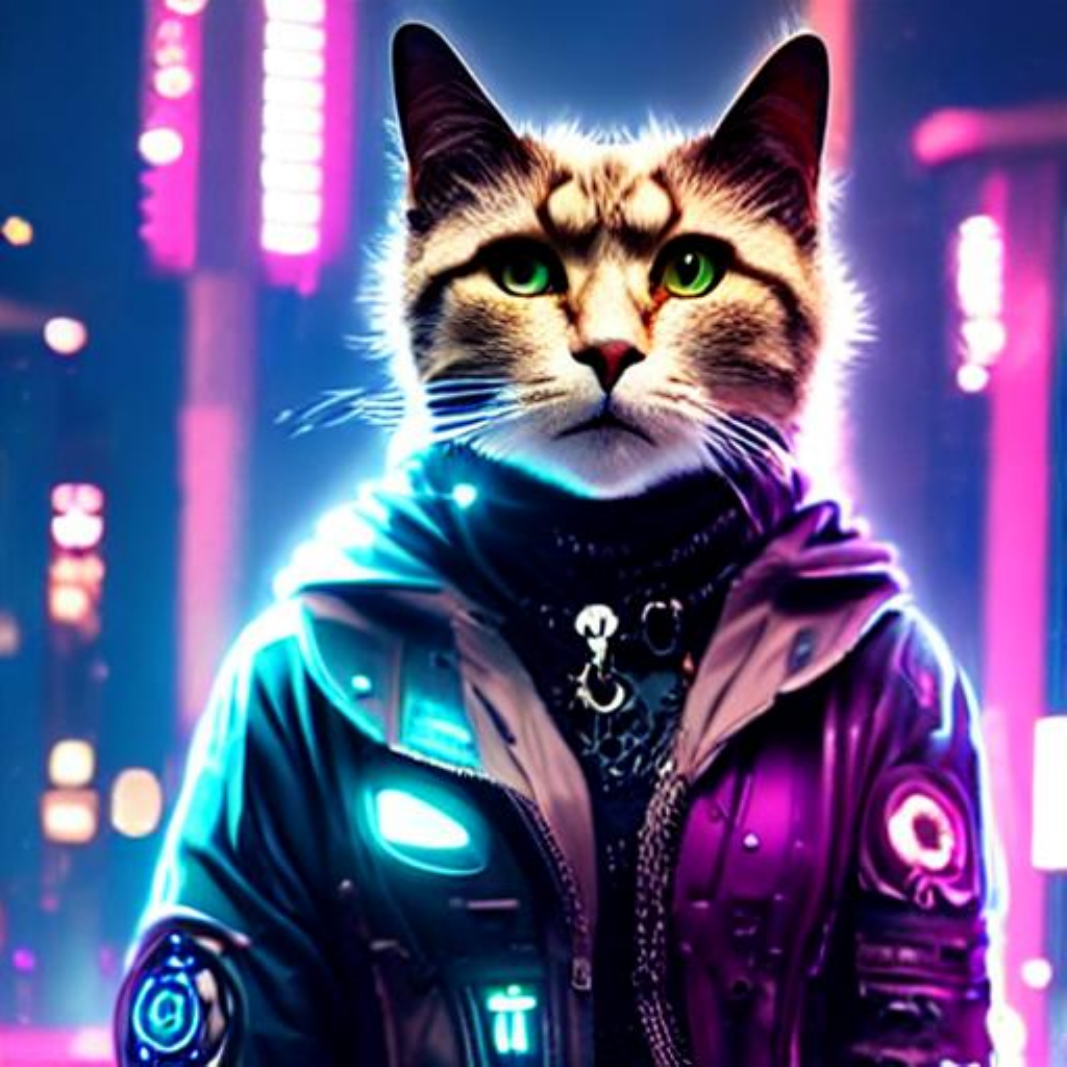} \end{subfigure}

    \parbox{1.02 \textwidth}{
        \centering
            <pixel art> gray French bulldog
    }
    \begin{subfigure}[t]{0.19\textwidth} \includegraphics[width=\textwidth]{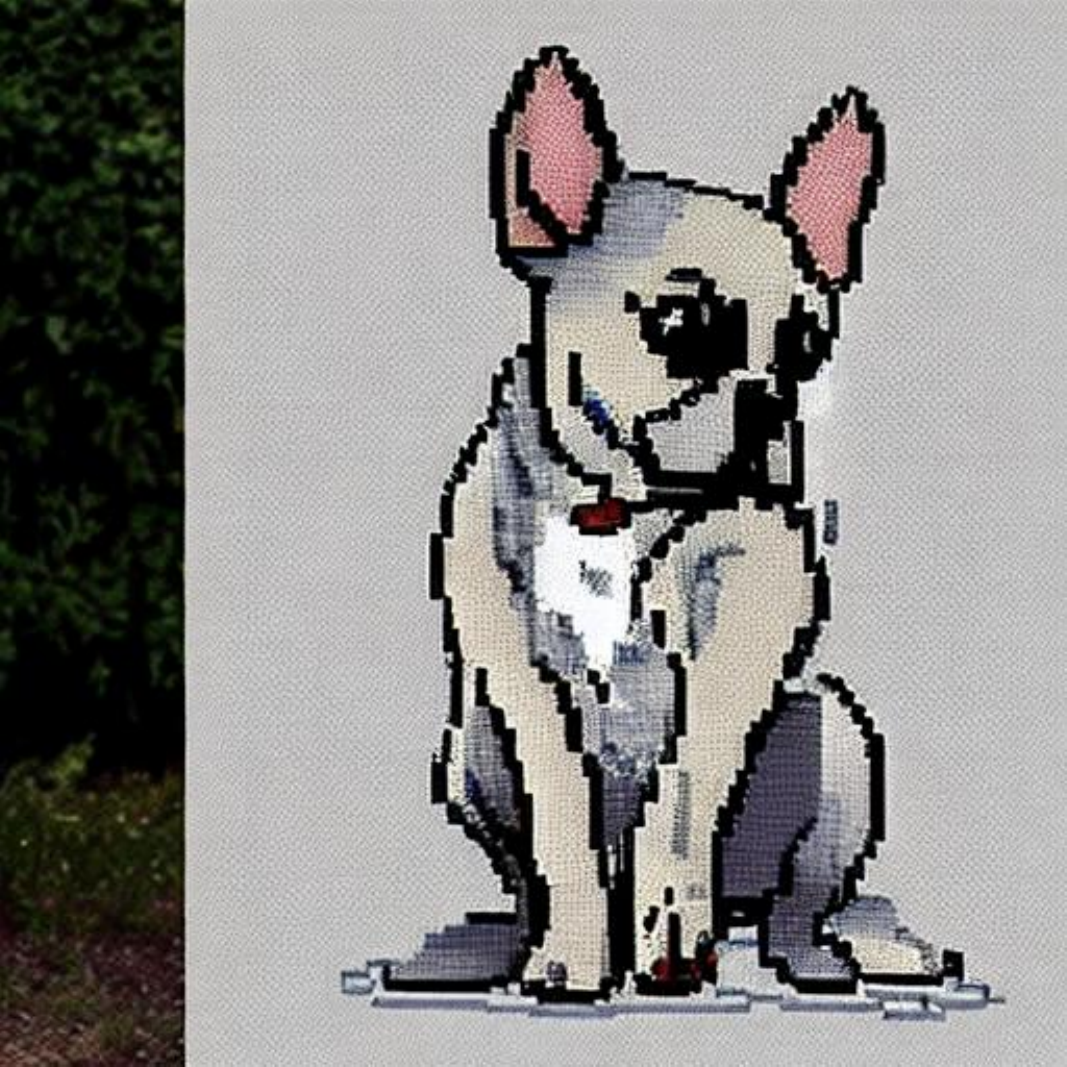} \end{subfigure}
    \begin{subfigure}[t]{0.19\textwidth} \includegraphics[width=\textwidth]{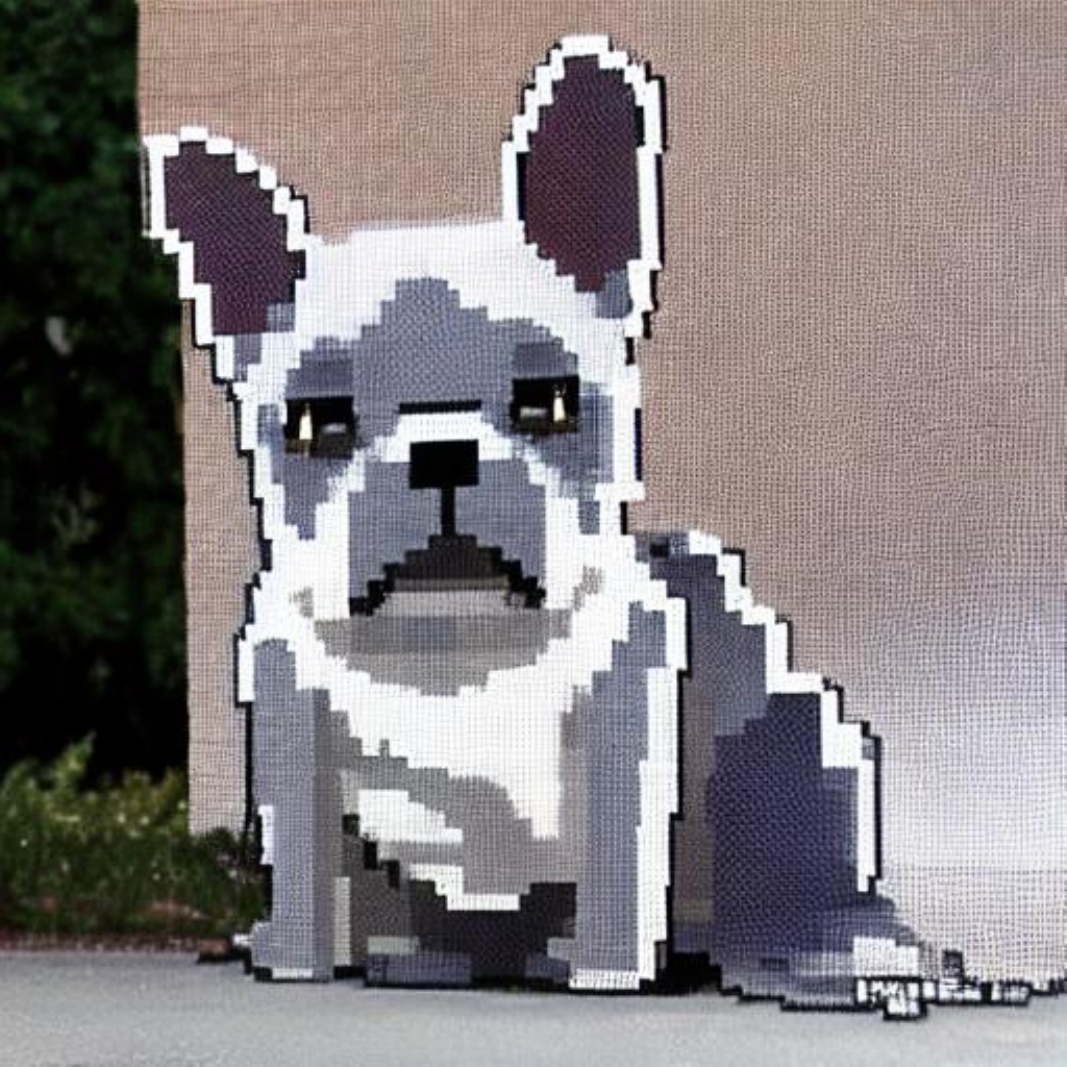} \end{subfigure}
    \begin{subfigure}[t]{0.19\textwidth} \includegraphics[width=\textwidth]{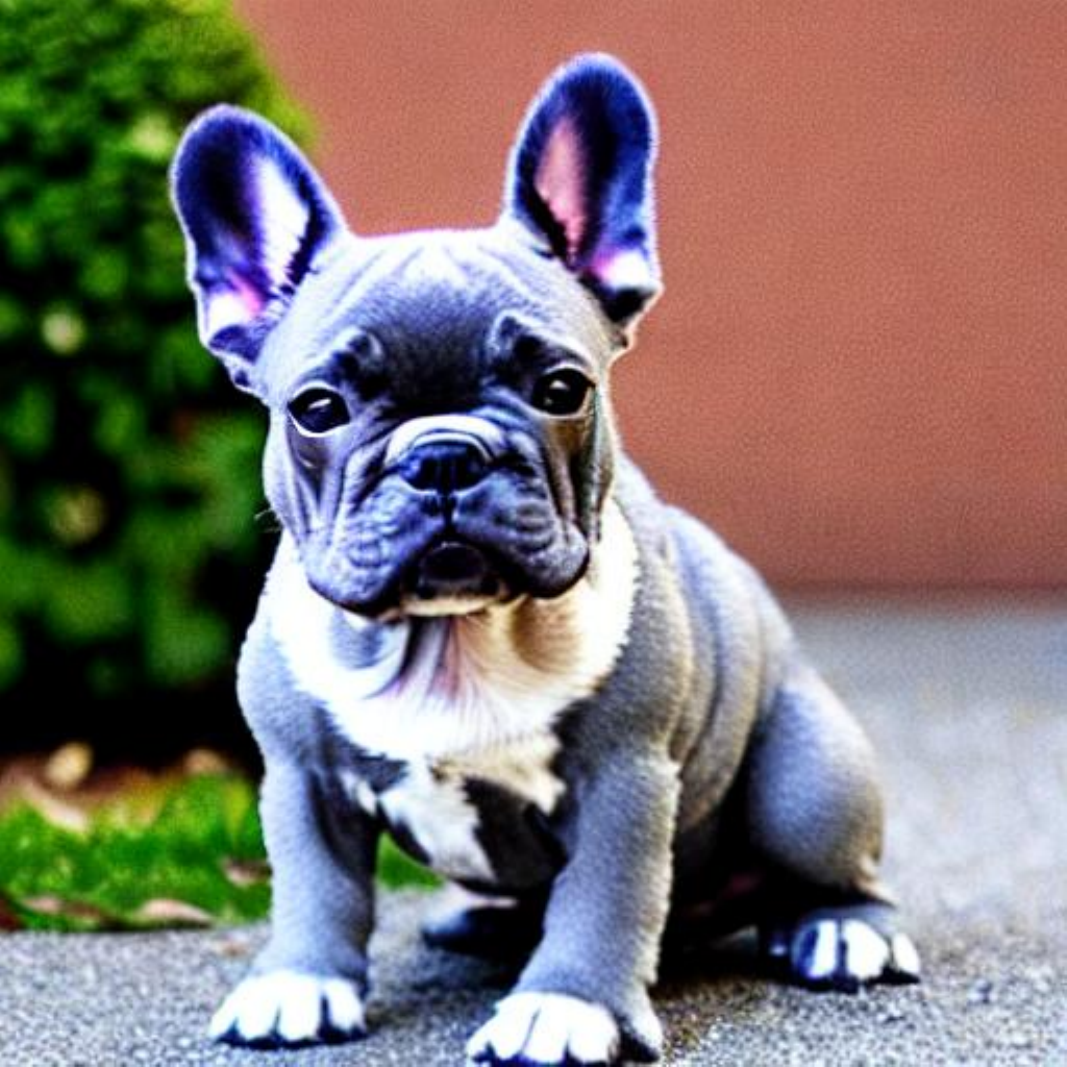} \end{subfigure}
    \begin{subfigure}[t]{0.19\textwidth} \includegraphics[width=\textwidth]{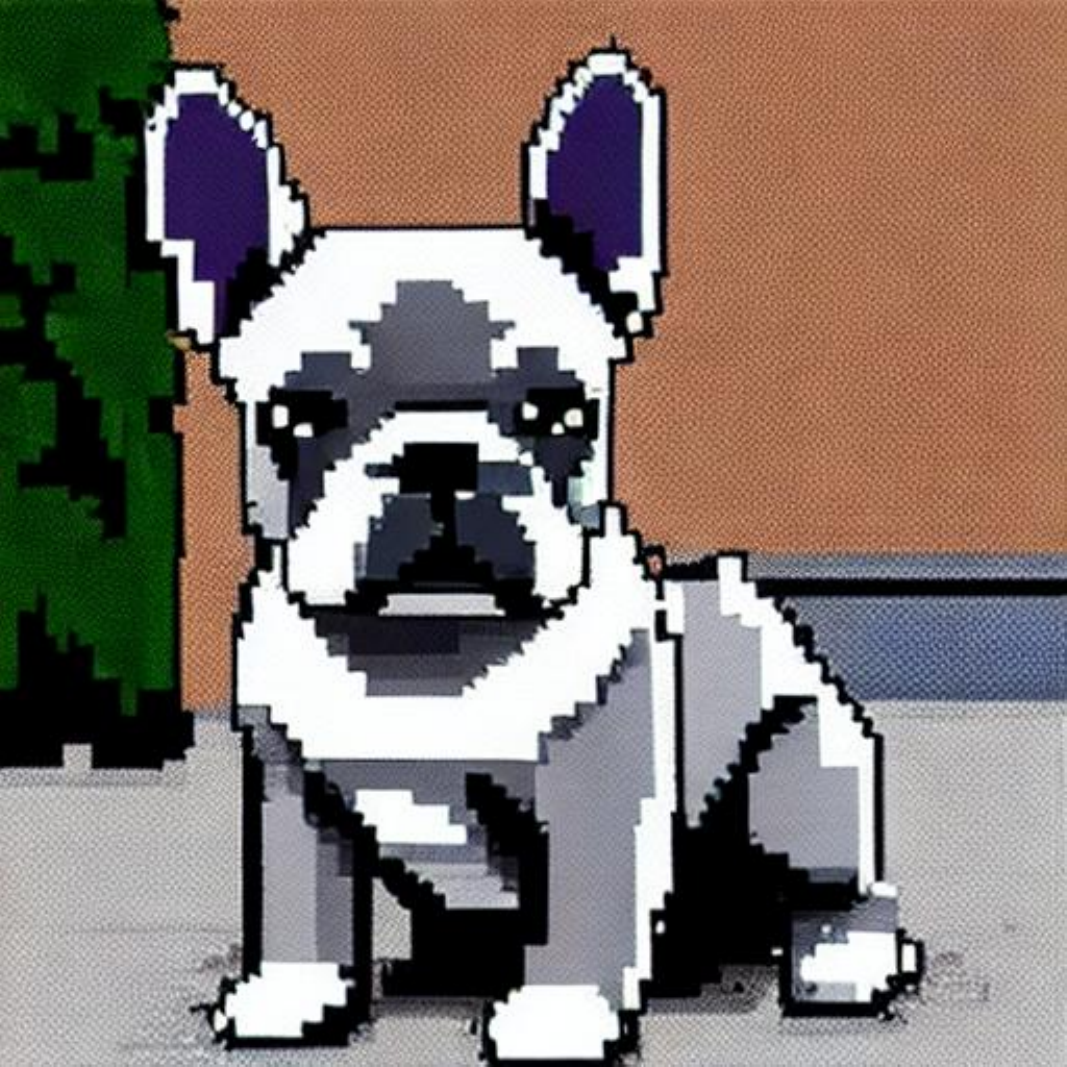} \end{subfigure}
    \begin{subfigure}[t]{0.19\textwidth} \includegraphics[width=\textwidth]{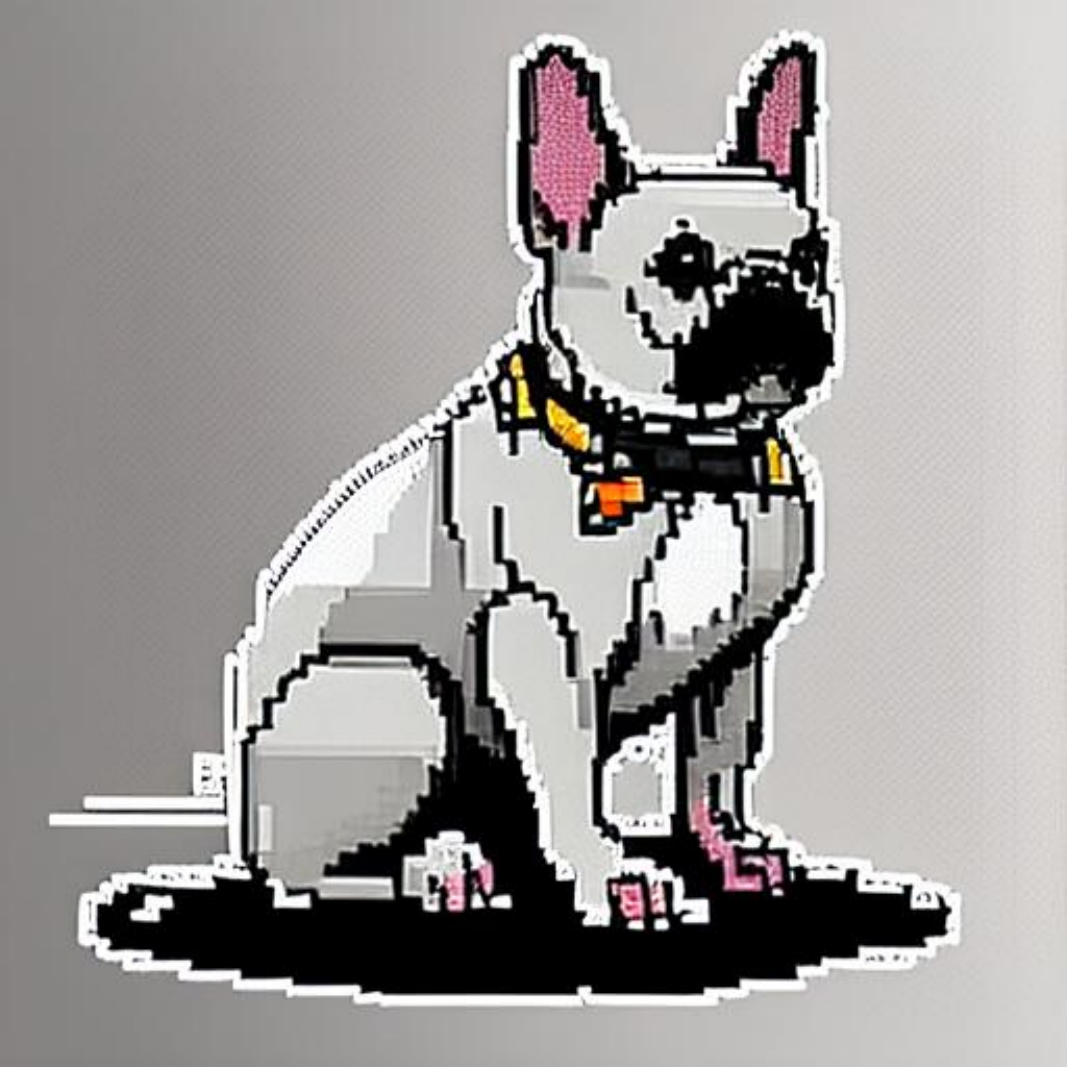} \end{subfigure}
    
    \parbox{1.02 \textwidth}{
        \centering
            a pair of headphones on a pumpkin
    }
    \begin{subfigure}[t]{0.19\textwidth} \includegraphics[width=\textwidth]{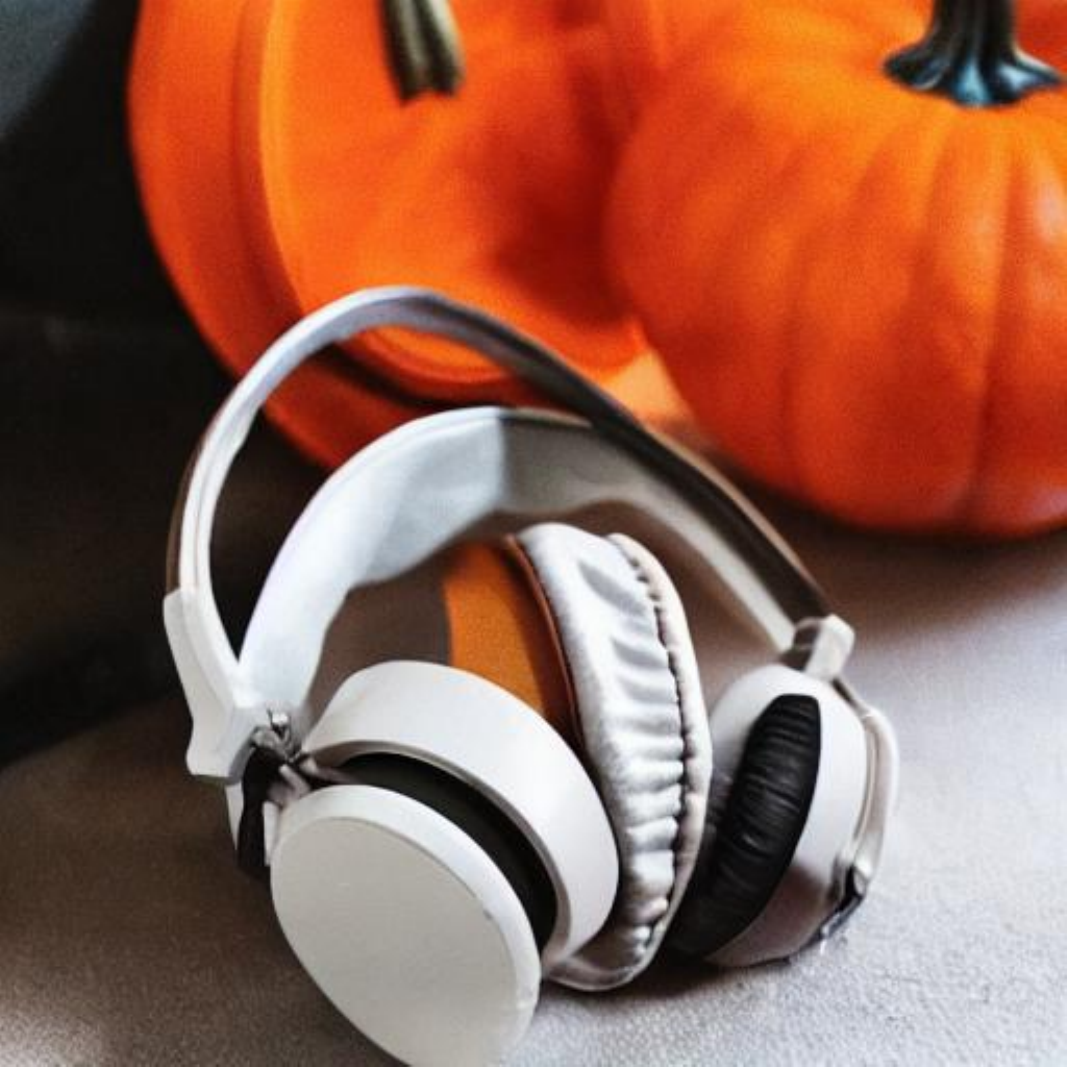} 
            \captionsetup{justification=centering}\caption{SD1.5~\cite{rombach2022high}}    \end{subfigure}
    \begin{subfigure}[t]{0.19\textwidth} \includegraphics[width=\textwidth]{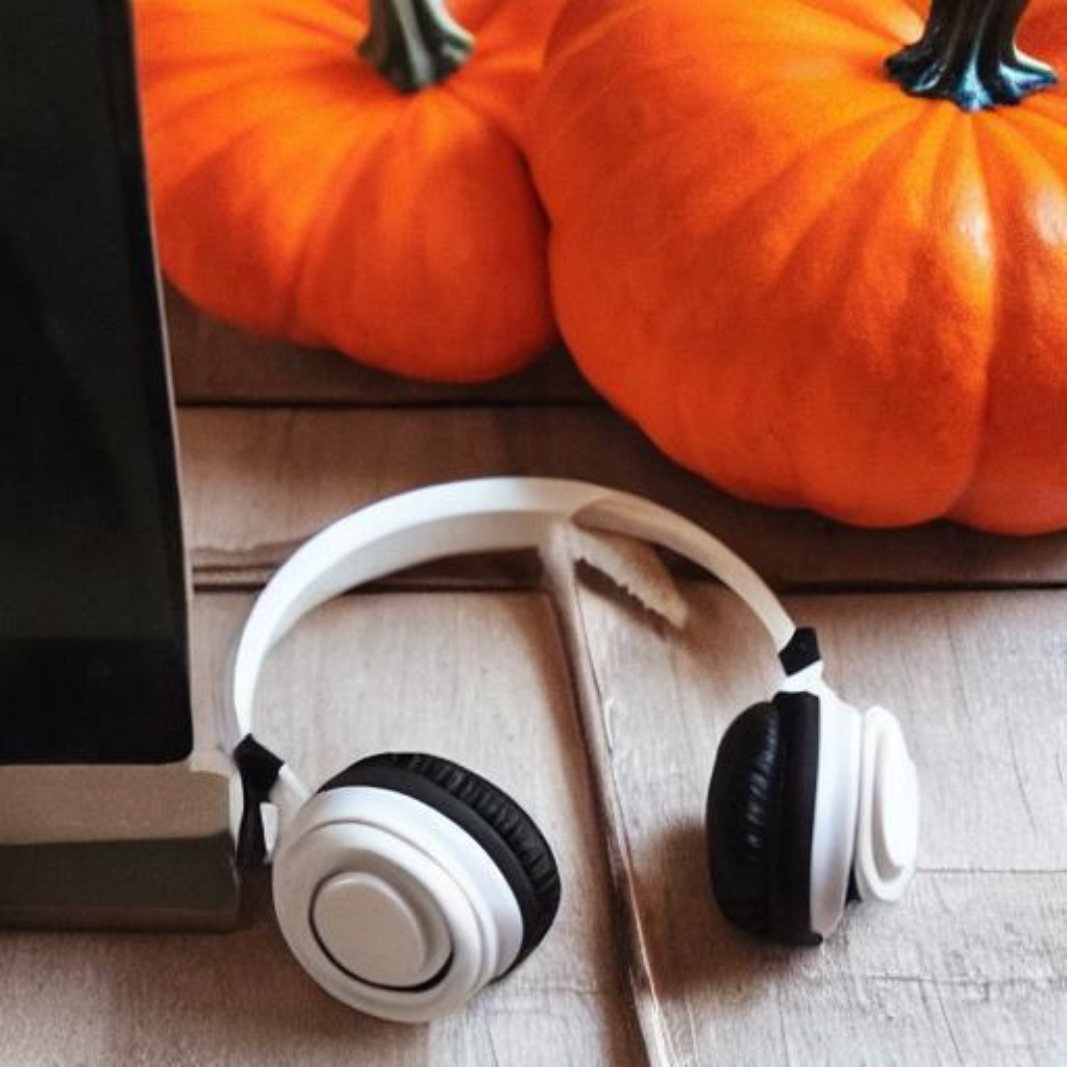} 
            \captionsetup{justification=centering}\caption{Diff-DPO~\cite{wallace2024diffusion}} \end{subfigure}
    \begin{subfigure}[t]{0.19\textwidth} \includegraphics[width=\textwidth]{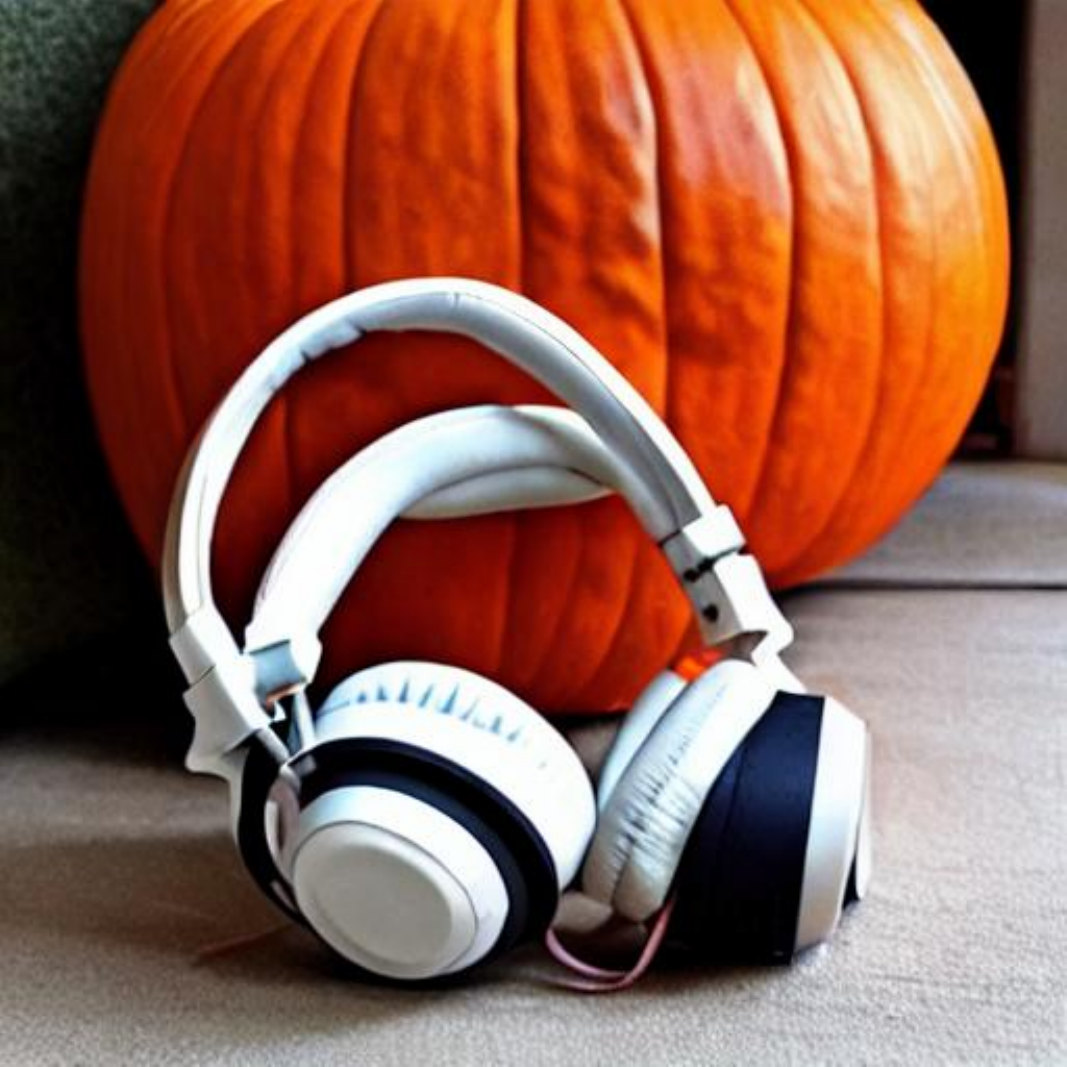}         \captionsetup{justification=centering}\caption{Diff-KTO~\cite{lialigning}} \end{subfigure}
    \begin{subfigure}[t]{0.19\textwidth} \includegraphics[width=\textwidth]{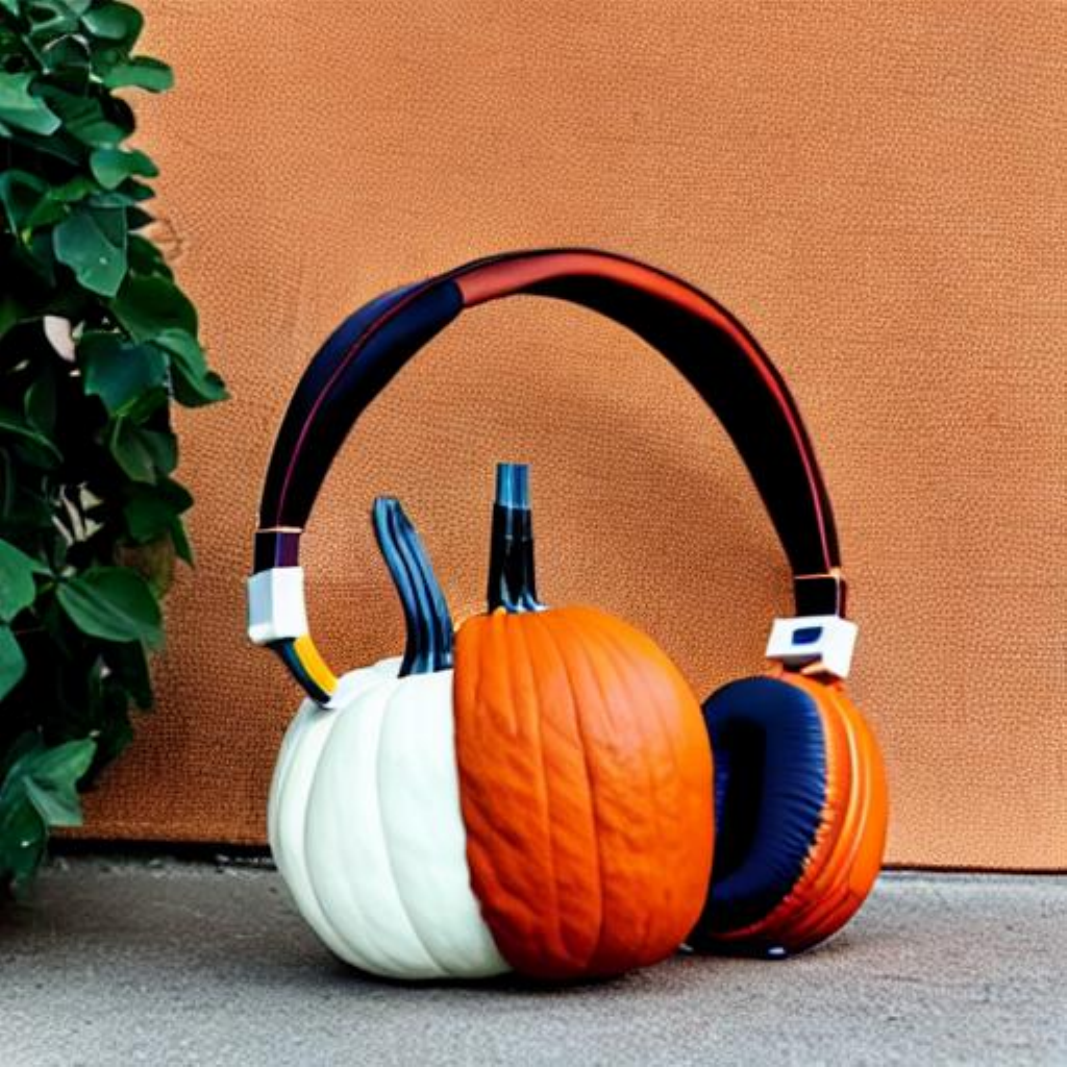}         \captionsetup{justification=centering}\caption{DSPO~\cite{zhu2025dspo}} \end{subfigure}
    \begin{subfigure}[t]{0.19\textwidth} \includegraphics[width=\textwidth]{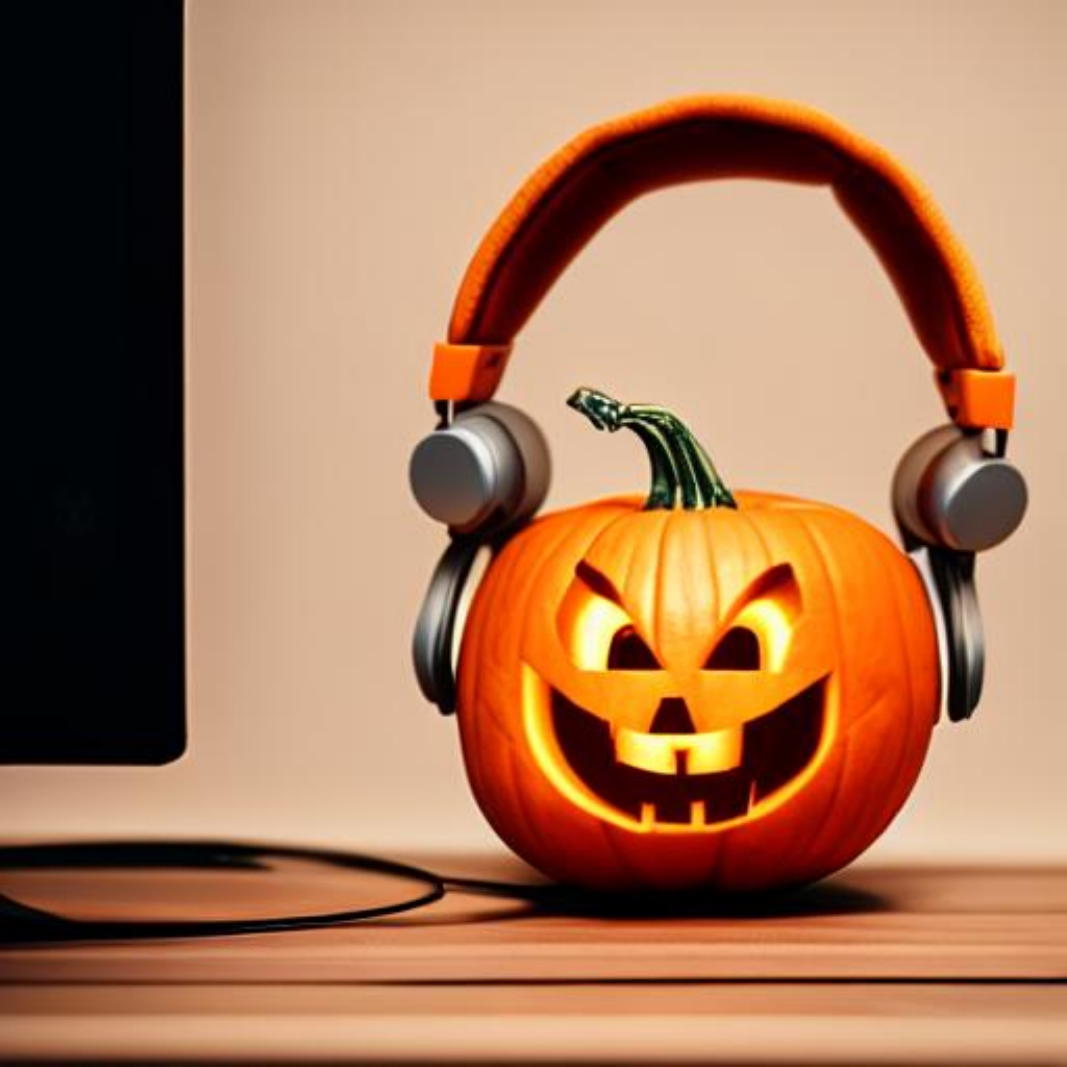}         \captionsetup{justification=centering}\caption{Ours} \end{subfigure}
    
    \caption{
        Qualitative comparison. We compare the generated outputs from various preference optimization algorithms based on SD1.5, including our method.
    }
    \label{fig:quali_comparison}
\end{figure*}

\begin{table}
    \setlength{\tabcolsep}{1mm}
    \begin{tabular}{cccccc}
    \toprule
    \textbf{Model} & \textbf{PickScore} & \textbf{HPSv2} & \textbf{CLIP} & \textbf{Aesthetic} & \textbf{IR}  \\
    \cmidrule(lr){1-1} \cmidrule(lr){2-6} 
    Diff-DPO & 21.36 & 27.19 & 33.84 & 5.53 & 0.32 \\
    LR=5e-8 & 19.75 & 24.95 & 29.81 & 4.79 & -0.39 \\
    $\beta$=1000 & 21.26 & 27.16 & 33.55 & 5.55 & 0.23 \\
    Re-Init & 21.50 & 27.16 & 34.17 & 5.57 & 0.38 \\
    Ours & \textbf{21.93} & \textbf{27.84} & \textbf{34.42} & \textbf{5.75} & \textbf{0.65} \\
    \bottomrule
    \end{tabular}%

\caption{Ablation study of alternative exploration strategies. Raw scores for each reward metric are reported. The highest value for each metric is displayed in bold.}
\label{tab:ablation_online}
\vspace{-1em}
\end{table}

\begin{table}
\setlength{\tabcolsep}{1mm}
\centering
\begin{small}
    \begin{tabular}{lccccc}
    \toprule
    \textbf{Model} & \textbf{Pick} & \textbf{HPSv2} & \textbf{CLIP} & \textbf{Aesthetic} & \textbf{IR} \\
    \cmidrule(lr){1-1} \cmidrule(lr){2-6} 
    Diff-DPO & 21.36 & 27.19 & 33.84 & 5.53 & 0.32 \\
    Time. only & 21.10 & 26.57 & 33.56 & 5.54 & 0.16 \\
    Exploration only & 21.88 & 27.63 & 34.40 & 5.70 & 0.58 \\
    \cmidrule(lr){1-1} \cmidrule(lr){2-6} 
$\gamma = 0.8$ & 21.66 & 27.49 & 34.25 & 5.66 & 0.52 \\
$\gamma = 0.8$ + Scale & 21.90 & 27.65 & 34.42 & 5.69 & 0.61 \\
    $\gamma = 0.9$ & 21.82 & 27.71 & 34.39 & 5.71 & 0.64\\
    \cmidrule(lr){1-1} \cmidrule(lr){2-6} 
    $\gamma = 0.9$ + Scale (Ours)  & \textbf{21.93} & \textbf{27.84} & \textbf{34.42} & \textbf{5.75} & \textbf{0.65} \\
    \bottomrule
    \end{tabular}%
\end{small}
\caption{Ablation study of the timestep-aware optimization strategy. (Top) Our timestep-aware strategy shows a synergistic effect when combined with exploration. (Bottom) The reward scale schedule further enhances performance. Raw scores for each reward metric are reported.}
\label{tab:ablation_time}
\end{table}

\noindent\textbf{Qualitative Results.} 
Figure~\ref{fig:quali_comparison} presents images generated by baselines and by our method. 
We find that Diffusion-DPO tends to show only subtle changes compared to the original model, due to limited exploration.
Diffusion-KTO and DSPO also struggle to produce images faithful to the text prompt.
For example, they fail to generate \textit{burgers} in the first row, and miss compositional objects such as \textit{cyberpunk} + \textit{cat} (DSPO) or \textit{pixel} + \textit{bulldog} (Diffusion-KTO).
Overall, our method correctly identifies objects and compositional relationships described in the text prompts and generates aesthetically appealing images compared to other models. 
We display more qualitative results in Appendix D.

\subsection{Ablation Study}
\label{sec:ablation}

\textbf{Comparison with Alternative Exploration Strategies.} 
Table~\ref{tab:ablation_online} compares our method (SD1.5) on the Pick-a-Pic v2 test set with alternative exploration strategies: (1) increasing learning rates, (2) reducing the implicit regularization coefficient $\beta$, and (3) re-initializing the reference model in the update strategy, when its divergence exceeds the threshold.
(1) Increasing the learning rate from 1e-8 to 5e-8 leads to model collapse and a substantial drop in all metrics. 
(2) Reducing $\beta$ from 5,000 to 1,000 does not improve performance.
(3) Re-initializing scheme yields a minor improvement, since the strong constraint of the initial model restricts exploration.

\begin{figure}
  \begin{center}
    \includegraphics[width=0.95\columnwidth]{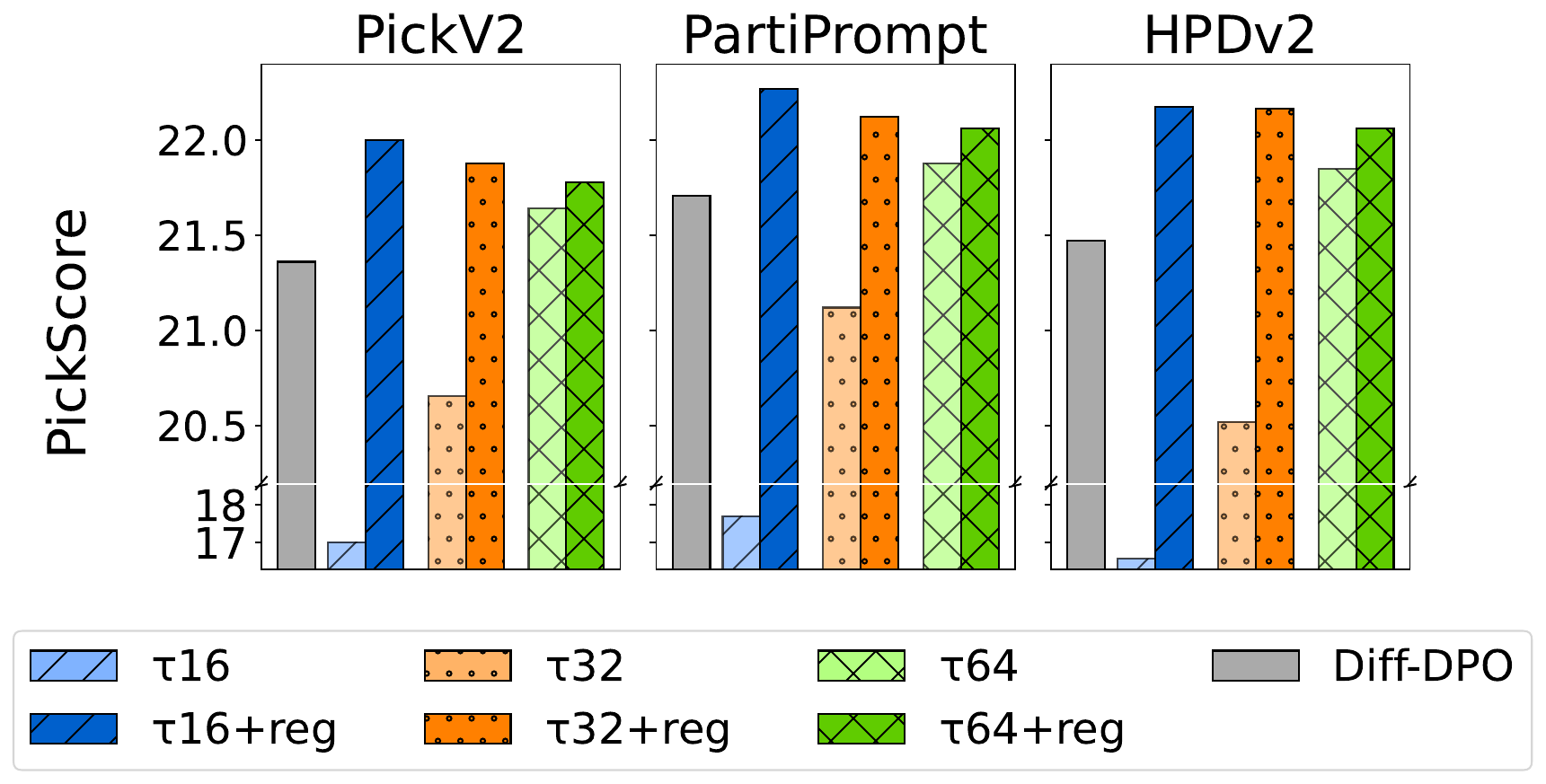}
  \end{center}
  \caption{Results of reference model regularization with $\tau \in \{16, 32, 64\}$, evaluated using the PickScore reward.}
  \label{fig:reference_ablation}
\end{figure}

\noindent\textbf{Effect of Reference Update Period.}
Figure \ref{fig:reference_ablation} illustrates that,
without our reference regularization, frequent model update ($\tau$ decreases) causes model divergence, leading to a sharp performance drop.
By constraining the update boundary with the divergence monitoring, our method consistently outperforms Diffusion-DPO, reducing the sensitivity to the update period $\tau$.

\begin{figure}
  \begin{center}
    \includegraphics[width=0.85\columnwidth]{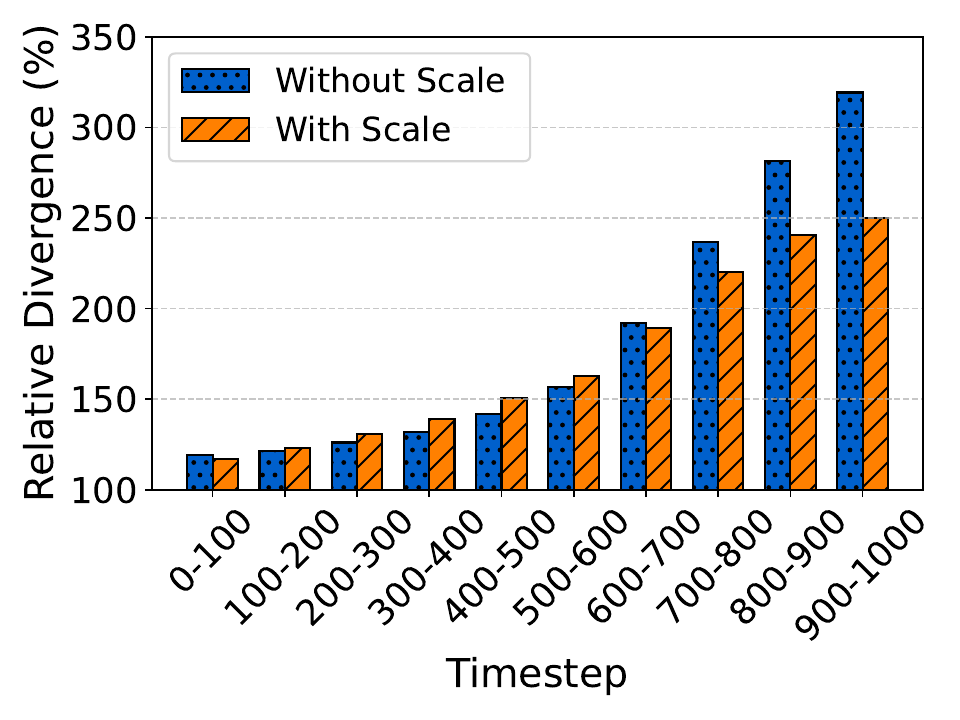}
  \end{center}
  \caption{Relative increase in divergence with and without reward scale scheduling. In each interval, 100\% represents the divergence of our reference update method.
  }
  \label{fig:kl_bar}
\end{figure}

\noindent\textbf{Effect of Timestep-Aware Strategy.}
Table \ref{tab:ablation_time} shows that combining timestep-aware optimization with exploration improves performance, while using it alone may degrade Diffusion-DPO.
This suggests that exploration is critical for enabling effective preference learning at early timesteps, highlighting the synergistic effect between the two components.
We also find that reward scale scheduling further enhances oversampling.
Figure~\ref{fig:kl_bar} presents the relative increase in model divergence induced by our timestep-aware strategy, compared to using only the reference update. 
The scheduled method exhibits a lower divergence budget in early timesteps, indicating a regularization effect that helps prevent overfitting and leads to better performance.

\section{Related Work}
\subsection{RLHF in Diffusion Models}
Reinforcement Learning from Human Feedback (RLHF) has proven highly effective in aligning human preference in the large language model domain~\cite{ouyang2022human,openai2024gpt4technicalreport}. Recently, similar approaches have been explored in the T2I diffusion domain, leveraging human feedback and various quality metrics as reward signals.
Previous works in RLHF to diffusion models have re-formulated the diffusion process as a Markov Decision Process (MDP). 
DDPO~\cite{blacktraining} and DPOK~\cite{fan2023dpok} compute rewards at the final timestep and apply the policy gradient method to fine-tune the model.
Alternatively, methods such as ReFL~\cite{xu2024imagereward}, DRaFT~\cite{clarkdirectly}, and AlignProp~\cite{prabhudesai2023aligning} propose differentiable reward frameworks, enabling direct policy updates through backpropagation.

\subsection{Direct Preference Optimization}
Direct Preference Optimization (DPO)~\cite{rafailov2023direct} 
has emerged as a promising alternative to RLHF, because it obviates the need to train a separate reward model. 
Building on the success of DPO, numerous variants have recently been explored in the language domain~\cite{azar2024general, gorbatovski2024learn, meng2024simpo, wu2024beta, hong2024orpo, zhao2025rainbowpo}.
DPO has also been extended to diffusion models to enhance alignment between generated images and human preferences. Notably, Diffusion-DPO~\cite{wallace2024diffusion} and D3PO~\cite{yang2024using} adapt the DPO loss to diffusion models.
Diffusion-KTO~\cite{lialigning} substitutes the standard DPO loss with Kahneman-Tversky Optimization (KTO), training with single-instance data without requiring pairwise comparisons. 
Meanwhile, some recent works consider the innate structure of diffusion models instead of naively applying the language model losses. 
Yang et al.,~\cite{yang2024dense} modify the uniform timestep sampling in Diffusion-DPO, deriving the loss from the densely defined rewards across timesteps. 
InPO~\cite{lu2025inpo} introduces DDIM inversion in Diffusion-DPO instead of random noise injection for training efficiency, and DSPO~\cite{zhu2025dspo} fine-tunes diffusion models by aligning with human preferences using score matching principles.

\section{Conclusion} 
\label{sec:conclusion}
We present a novel training framework for enhancing DPO in diffusion models. Our method enables the stable model exploration by updating the reference model under a divergence constraint and addressing reward scale imbalance across denoising steps to further improve exploration. Experiments show that our strategy significantly improves the alignment performance of Diffusion-DPO across multiple benchmarks, achieving new state-of-the-art results.
We believe our work opens for future research on the training dynamics of preference optimization and motivates further development of DPO-based methods in diffusion models.

\section*{Acknowledgements}
\label{sec:ack}
This research was supported by the Basic Science Research Program through the 
National Research Foundation of Korea (NRF), 
funded by the MSIP (RS-2025-00520207, RS-2023-00219019),
IITP grant funded by the Korean government (MSIT) (RS-2024-00457882, RS-2025-02217259),
KEIT grant funded by the Korean government (MOTIE) (No. 2022-0-00680, No. 2022-0-01045),
and Artificial Intelligence Graduate School Program (KAIST) (RS-2019-II190075).

\bibliography{aaai2026}

\clearpage

\renewcommand{\thesection}{\Alph{section}}
\setcounter{section}{0}
\renewcommand{\thetable}{S\arabic{table}}
\setcounter{figure}{0}  
\renewcommand{\thefigure}{S\arabic{figure}}
\setcounter{table}{0}   
\renewcommand{\theequation}{S\arabic{equation}}
\setcounter{equation}{0}   
\begin{figure*}[ht!]
    \centering
    \begin{subfigure}[b]{0.33\textwidth}
        \centering
        \includegraphics[width=\linewidth]{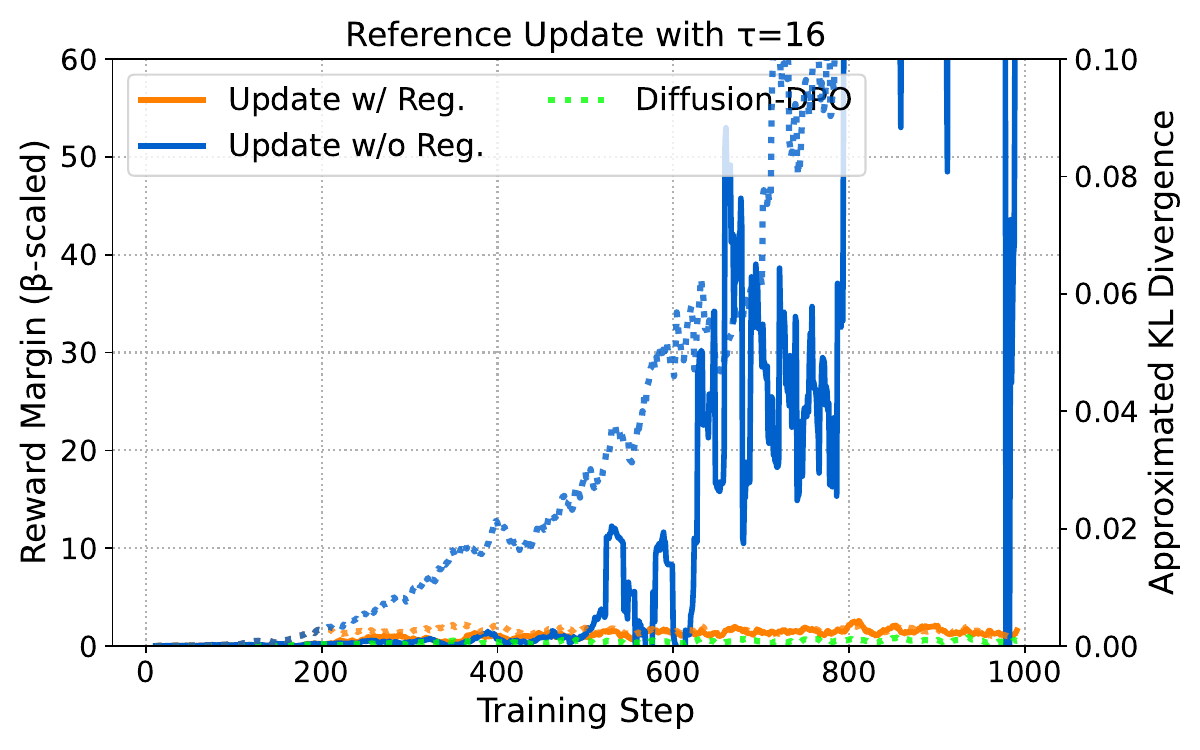}
    \end{subfigure}
    \begin{subfigure}[b]{0.33\textwidth}
        \centering
        \includegraphics[width=\linewidth]{fig/dynamicsanalysis_32.pdf}
    \end{subfigure}
    \begin{subfigure}[b]{0.33\textwidth}
        \centering
        \includegraphics[width=\linewidth]{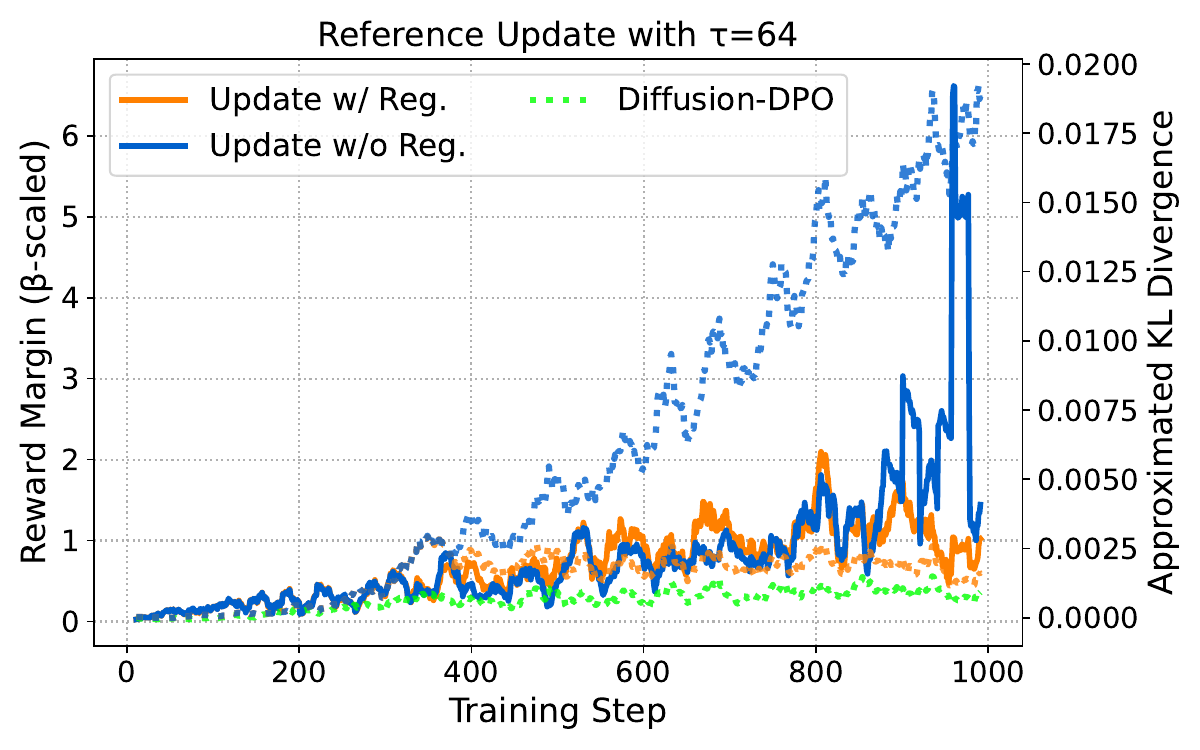}
    \end{subfigure}

    \caption{Training dynamics of reference update method, with $\tau = \{16, 32, 64\}$ (SD1.5). (solid lines) Implicit reward margin under the reference update strategy, with and without our regularization. (dotted lines) Approximated KL divergence between the training model and the pre-trained model (Diffusion-DPO), and between the reference model and the pre-trained model (ours).
    }
    \label{fig:training_dynamics_full}
\end{figure*}

\section*{Supplementary Materials}
\section{Further Implementation Details}
\label{appendix:implementation_details}

\begin{table*}[t]
\centering
\begin{tabular}{l l c c c c c c }
\toprule
\textbf{Dataset} & \textbf{Model} & \textbf{PickScore} & \textbf{HPSv2} & \textbf{CLIP} & \textbf{Aesthetic} & \textbf{ImageReward} & \textbf{Average} \\
\cmidrule(lr){1-1} \cmidrule(lr){2-2} \cmidrule(lr){3-7} \cmidrule(lr){8-8} 
\multirow{5}{*}{PickV2} 
& vs. SDXL*   & \textbf{81.24} & \textbf{81.76} & \textbf{57.64} & \textbf{59.28} & \textbf{70.96} & \textbf{70.18} \\
& vs. MaPO*        & \textbf{81.16} & \textbf{74.88} & \textbf{58.16} & 45.12 & \textbf{65.92} & \textbf{65.05} \\
& vs. InPO*  & \textbf{64.80} & \textbf{56.56} & \textbf{54.76} & \textbf{55.00} & \textbf{56.76} & \textbf{57.58} \\
& vs. Diff-DPO  & \textbf{68.40} & \textbf{73.76} & \textbf{50.28} & \textbf{57.52} & \textbf{54.40} & \textbf{60.87} \\
& vs. DSPO       & \textbf{60.88} & \textbf{64.68} & \textbf{51.44} & \textbf{55.52} & 49.28 & \textbf{56.36} \\

\midrule
\multirow{5}{*}{PartiPrompts} 
& vs. SDXL*   & \textbf{71.45} & \textbf{79.84} & \textbf{53.00} & \textbf{64.71} & \textbf{74.20} & \textbf{68.64} \\
& vs. MaPO*        & 
\textbf{73.77} & \textbf{79.47} & \textbf{58.82} & 47.30 & \textbf{69.12} & \textbf{65.70} \\
& vs. InPO*  & \textbf{53.86} & \textbf{57.90} & \textbf{52.39} & \textbf{56.50} & \textbf{56.92} & \textbf{55.51} \\
& vs. Diff-DPO  & 
\textbf{59.68} & \textbf{73.04} & 45.34 & \textbf{58.21} & \textbf{54.66} & \textbf{58.19} \\
& vs. DSPO       & 
\textbf{59.13} & \textbf{67.10} & 48.35 & \textbf{54.04} & \textbf{52.14} & \textbf{56.15} \\

\midrule
\multirow{5}{*}{HPDv2} 
& vs. SDXL*   & \textbf{78.91} & \textbf{84.38} & \textbf{51.88} & \textbf{58.38} & \textbf{73.22} & \textbf{69.35} \\
& vs. MaPO*  & \textbf{77.69} & \textbf{73.19} & \textbf{52.56} & 49.03 & \textbf{67.41} & \textbf{63.97} \\
& vs. InPO*  & \textbf{58.06} & \textbf{53.25} & 49.97 & \textbf{54.59} & \textbf{56.34} & \textbf{54.44} \\
& vs. Diff-DPO  & 
\textbf{62.59} & \textbf{73.72} & 47.09 & \textbf{57.41} & \textbf{55.34} & \textbf{59.23} \\
& vs. DSPO       & 
\textbf{58.84} & \textbf{56.00} & 42.84 & \textbf{64.34} & \textbf{51.41} & \textbf{54.69} \\
\bottomrule
\end{tabular}%
\caption{Win rates of our method against baseline preference optimization methods using SDXL as the base model. * indicates model checkpoints released by the original authors. Higher win rates indicate better alignment performance and win rates exceeding 50\% are marked in bold.}
\label{tab:sdxl_winrates_full}
\end{table*}

\begin{table*}[ht!]
\centering
\begin{tabular}{l l c c c c c }
\toprule
\textbf{Dataset} & \textbf{Model} & \textbf{PickScore} & \textbf{HPSv2} & \textbf{CLIP} & \textbf{Aesthetic} & \textbf{ImageReward} \\
\cmidrule(lr){1-1} \cmidrule(lr){2-2} \cmidrule(lr){3-7}
\multirow{6}{*}{PickV2} 
& SD1.5* & 20.66$\pm$0.03 & 26.52$\pm$0.04 & 32.59$\pm$0.12 & 5.39$\pm$0.01 & -0.07$\pm$0.02 \\
& Diff-KTO* & 21.37$\pm$0.03 & 27.77$\pm$0.04 & 33.83$\pm$0.12 & 5.69$\pm$0.01 & 0.59$\pm$0.02 \\
& SFT & 21.42$\pm$0.03 & 27.77$\pm$0.04 & 33.79$\pm$0.12 & \textbf{5.76}$\pm$0.01 & 0.57$\pm$0.02 \\
& Diff-DPO & 21.36$\pm$0.03 & 27.19$\pm$0.04 & 33.84$\pm$0.12 & 5.53$\pm$0.01 & 0.32$\pm$0.02 \\
& DSPO & \underline{21.46}$\pm$0.03 & \underline{27.78}$\pm$0.04 & \underline{34.00}$\pm$0.12 & 5.74$\pm$0.01 & \underline{0.61}$\pm$0.02 \\
& Ours & \textbf{21.93}$\pm$0.03 & \textbf{27.84}$\pm$0.04 & \textbf{34.42}$\pm$0.11 & \underline{5.75}$\pm$0.01 & \textbf{0.65}$\pm$0.02 \\
\midrule
\multirow{6}{*}{PartiPrompts} 
& SD1.5* & 21.31$\pm$0.03 & 26.96$\pm$0.04 & 32.70$\pm$0.14 & 5.28$\pm$0.01 & -0.08$\pm$0.03 \\
& Diff-KTO* & 21.79$\pm$0.03 & 28.10$\pm$0.04 & \underline{33.79}$\pm$0.14 & 5.54$\pm$0.01 & 0.49$\pm$0.03 \\
& SFT & 21.78$\pm$0.03 & 28.09$\pm$0.04 & 33.50$\pm$0.14 & 5.59$\pm$0.01 & 0.43$\pm$0.03 \\
& Diff-DPO & 21.71$\pm$0.03 & 27.46$\pm$0.04 & 33.57$\pm$0.14 & 5.38$\pm$0.01 & 0.22$\pm$0.03 \\
& DSPO & \underline{21.81}$\pm$0.03 & \underline{28.11}$\pm$0.04 & 33.74$\pm$0.14 & \underline{5.59}$\pm$0.01 & \underline{0.49}$\pm$0.03 \\
& Ours &\textbf{22.21}$\pm$0.03 & \textbf{28.31}$\pm$0.04 & \textbf{34.29}$\pm$0.14 & \textbf{5.66}$\pm$0.01 & \textbf{0.73}$\pm$0.02 \\
\midrule
\multirow{6}{*}{HPDv2} 
& SD1.5* & 20.73$\pm$0.02 & 26.63$\pm$0.03 & 33.95$\pm$0.09 & 5.38$\pm$0.01 & -0.25$\pm$0.02 \\
& Diff-KTO* & \underline{21.64}$\pm$0.02 & \underline{28.26}$\pm$0.03 & \underline{35.55}$\pm$0.09 & 5.76$\pm$0.01 & \underline{0.57}$\pm$0.02 \\
& SFT & 21.58$\pm$0.02 & 28.16$\pm$0.03 & 35.23$\pm$0.09 & 5.79$\pm$0.01 & 0.51$\pm$0.02 \\
& Diff-DPO & 21.47$\pm$0.02 & 27.49$\pm$0.03 & 35.31$\pm$0.09 & 5.60$\pm$0.01 & 0.21$\pm$0.02 \\
& DSPO & 21.63$\pm$0.02 & 28.18$\pm$0.03 & 35.36$\pm$0.10 & \underline{5.80}$\pm$0.01 & 0.54$\pm$0.02 \\
& Ours & \textbf{22.16}$\pm$0.02 & \textbf{28.38}$\pm$0.03 & \textbf{35.91}$\pm$0.09 & \textbf{5.83}$\pm$0.01 & \textbf{0.68}$\pm$0.02 \\
\bottomrule
\end{tabular}%
\caption{Average reward scores for each method on SD1.5, with 1-sigma error bars. The highest score in each column is shown in bold, and the second highest is underlined.}
\label{tab:raw_sd15}
\end{table*}

\begin{table*}[ht!]
\centering
\begin{tabular}{l l c c c c c }
\toprule
\textbf{Dataset} & \textbf{Model} & \textbf{PickScore} & \textbf{HPSv2} & \textbf{CLIP} & \textbf{Aesthetic} & \textbf{ImageReward} \\
\cmidrule(lr){1-1} \cmidrule(lr){2-2} \cmidrule(lr){3-7}
\multirow{6}{*}{PickV2} 
& SDXL & 22.16±0.07 & 27.98±0.09 & 36.09±0.29 & 6.01±0.03 & 0.57±0.05 \\
& MaPO & 22.25±0.07 & 28.32±0.09 & 36.23±0.28 & \textbf{6.15±0.02} & 0.70±0.04 \\
& InPO & \underline{22.68±0.03} & \underline{28.88±0.04} & 36.89±0.12 & 6.09±0.01 & \underline{0.98±0.02} \\
& Diff-DPO & 22.65±0.07 & 28.46±0.08 & 37.23±0.26 & 6.02±0.03 & 0.89±0.04 \\
& DSPO & 22.66±0.07 & 28.81±0.08 & \textbf{37.58±0.26} & 5.96±0.02 & 0.95±0.04 \\
& Ours & \textbf{22.94±0.07} & \textbf{29.06±0.08} & \underline{37.28±0.26} & \underline{6.09±0.02} & \textbf{1.01±0.04} \\
\midrule
\multirow{6}{*}{PartiPrompts} 
& SDXL & 21.31±0.03 & 26.96±0.04 & 32.70±0.14 & 5.28±0.01 & -0.08±0.03 \\
& MaPO & 22.62±0.03 & 28.58±0.05 & 35.35±0.14 & \textbf{5.91±0.01} & 0.79±0.02 \\
& InPO & \underline{23.01±0.03} & \underline{29.14±0.05} & 35.89±0.15 & 5.86±0.01 & 1.01±0.02 \\
& Diff-DPO & 22.94±0.03 & 28.80±0.04 & \underline{36.36±0.14} & 5.85±0.01 & 1.08±0.02 \\
& DSPO & 22.95±0.03 & 29.06±0.04 & \textbf{36.60±0.14} & 5.84±0.01 & \underline{1.16±0.02} \\
& Ours & \textbf{23.09±0.03} & \textbf{29.39±0.05} & 36.22±0.14 & \underline{5.90±0.01} & \textbf{1.17±0.02} \\
\midrule
\multirow{6}{*}{HPDv2} 
&SDXL & 22.78±0.02 & 28.63±0.03 & 38.16±0.09 & 6.13±0.01 & 0.78±0.01 \\
&MaPO & 22.84±0.02 & 29.01±0.03 & 38.14±0.09 & \textbf{6.22±0.01} & 0.88±0.01 \\
&InPO & \underline{23.27±0.02} & \underline{29.55±0.03} & 38.46±0.09 & 6.18±0.01 & 1.04±0.01 \\
&Diff-DPO & 23.20±0.02 & 29.08±0.03 & 38.59±0.09 & 6.17±0.01 & 1.06±0.01 \\
&DSPO & 23.24±0.02 & 29.48±0.03 & \textbf{38.94±0.08} & 6.11±0.01 & \underline{1.12±0.01} \\
&Ours & \textbf{23.41±0.02} & \textbf{29.63±0.03} & \underline{38.67±0.08} & \underline{6.21±0.01} & \textbf{1.13±0.01} \\
\bottomrule
\end{tabular}%
\caption{Average reward scores for each method in SDXL with 1-sigma error bar. The highest score in each column is shown
in bold, and the second highest is underlined.}
\label{tab:raw_sdxl}
\end{table*}

\begin{table*}[ht!]
    \setlength{\tabcolsep}{1mm}
    \centering
    \begin{tabular}{l c c c c c c} 
    \toprule
    \textbf{Model} & \textbf{PickScore} & \textbf{HPSv2} & \textbf{CLIP} & \textbf{Aesthetic} & \textbf{IR} & \textbf{Average} \\
    \midrule
    $\delta$=0.001 & 86.40 & 75.96 & 64.48 & 75.48 & 71.64 & 74.79 \\
    $\delta$=0.005  & \textbf{89.96} &\textbf{83.84} & \textbf{64.56} & \textbf{78.04} & \textbf{77.76} & \textbf{78.83} \\
    $\delta$=0.025  & 80.40 & 72.48 & 60.24 & 71.04 & 73.56 & 71.54 \\
    \bottomrule
    \end{tabular}%
\caption{Ablation study on different monitoring thresholds. Win rates are reported against SD1.5.}
\label{tab:appendix_delta}
\end{table*}

We train our models on the Pick-a-Pic v2 dataset~\cite{kirstain2023pick}.
The Pick-a-Pic v2 training set comprises about 900K image pairs with approximately 58k distinct prompts. 
Images were ranked by human evaluators, consisting of a preferred and a non-preferred image for each given input prompt.

During training, we follow the hyperparameter configurations of prior works~\cite{wallace2024diffusion, zhu2025dspo, lialigning}. 
We use the AdamW optimizer with a learning rate of $2.048\times10^{-8}$. 
Training is performed with a batch size of 4 per GPU, 128 gradient accumulation steps, and 4 NVIDIA A6000 GPUs, resulting in an effective batch size of 2048.
We train the models for 1000 iterations on SD1.5 and 600 iterations on SDXL. 
We set the base regularization coefficient (or signal scale)  $\beta=5000$ for Diffusion-DPO and DSPO (SDXL), and $\beta=0.001$ for DSPO (SD1.5), consistent with their original settings. 


For evaluation, we use 50 inference timesteps and set the classifier-free guidance scale to 7.5 (5.0 for SDXL).
To evaluate our method with a sufficient amount of images, we evaluate with 5 different random seeds on the Pick-a-Pic v2, generating a total of 2,500 images.
As the number of prompts in PartiPrompts and HPDv2 test dataset is large (1,632 and 3,200 prompts, respectively), the evaluation is conducted using a single seed.

\section{Quantitative Results}
\label{appendix:quantitative_results}
In this section, we provide more detailed quantitative results.
We include the win rate results for the SDXL model in the PartiPrompts and HPDv2 test prompts in Table~\ref{tab:sdxl_winrates_full}.
Our method consistently achieves average win rates above 50\% against Diffusion-DPO and baseline methods, with particularly strong performance on human preference metrics such as PickScore, HPSv2, and ImageReward.

Additionally, we present the raw reward scores from each method with 1-sigma error bars in Table~\ref{tab:raw_sd15} (SD1.5) and~\ref{tab:raw_sdxl} (SDXL).
In the SD1.5 results, our method significantly improves the reward scores of Diffusion-DPO, achieving the highest scores on most metrics.
For example, on the Pick-a-Pic v2 test set, PickScore and ImageReward increase by $0.57$ and $0.33$, respectively.
In SDXL results, while there are exceptions in the CLIP score (which was not trained on human preference prediction tasks) and the Aesthetic Score metric (which does not consider the text prompt), our method records the best performance in all other metrics.

\begin{table*}[]
\small
\centering
\begin{tabular}{l c c c c c c }
\toprule
\textbf{Dataset} & \textbf{PickScore} & \textbf{HPSv2} & \textbf{CLIP} & \textbf{Aesthetic} & \textbf{ImageReward} & \textbf{Average} \\
\cmidrule(lr){1-1} \cmidrule(lr){2-6} \cmidrule(lr){7-7} 
PickV2     & \textbf{70.80}&	\textbf{67.80}	&\textbf{55.00}	&\textbf{69.20}&	\textbf{64.60}	&\textbf{65.48} \\
PartiPrompts  & \textbf{58.21}&\textbf{62.01}&49.51&\textbf{52.02}&\textbf{55.82}	&\textbf{55.51} \\
HPDv2       & \textbf{70.69}&	\textbf{68.72}&	\textbf{51.47}&	\textbf{59.84}&\textbf{	62.81}&\textbf{62.71} \\

\bottomrule
\end{tabular}%

\caption{Win rates of our method against Diffusion-DPO using SD3 as the base model. Higher win rates indicate better alignment performance and win rates exceeding 50\% are marked in bold.}
\label{tab:sd3_comparison}
\end{table*}
To demonstrate the generalizability of our method, we also present results on Stable Diffusion 3, which modernizes diffusion models by introducing flow matching and a multimodal transformer.
Due to limited computational resources, we reduce the batch size from 2048 to 128 and train the model for 200 iterations with a learning rate of 3e-7.
The reference update period $\tau$ is set to 32, the monitoring threshold to 0.03, and all other hyperparameters remain the same as those used for SD1.5 and SDXL.

Table~\ref{tab:sd3_comparison} compares Diffusion-DPO with our method under the same training configuration, showing that our method again outperforms the standard Diffusion-DPO.
This result demonstrates that the effectiveness of our method is independent of the structural components of diffusion models.

\section{Further ablation study }
\label{appendix:hyperstudy_further}
\subsection{Analysis on Training Dynamics of $\tau$}

Figure~\ref{fig:training_dynamics_full} presents the training dynamics for the reference update period $\tau \in \{16, 32, 64\}$.
As discussed in Section~\ref{sec:method}, we observe a consistent increase in both the reference model's divergence and the implicit reward margin.
For smaller values of $\tau$, the reference model stays closer to the training model, resulting in more aggressive scaling of the prediction error.
Increasing $\tau$ can improve training stability by slowing reference updates, but it also reduces alignment performance due to limited exploration, as shown in Figure~\ref{fig:reference_ablation}.
In contrast, our reference regularization method effectively prevents divergence and error explosion, ensuring controlled exploration for the training model.

To gain deeper theoretical insight into the model divergence issue, we compare the DPO gradient behavior between diffusion models and autoregressive language models.
Suppose we sample a pairwise data $\xw_0,\xl_0$ and timestep $t$ for the Diffusion-DPO loss defined in Eq.~\ref{eq:loss-dpo}.
Then, the gradient of the loss is computed as:
\begin{multline}
    \nabla_\theta \mathcal{L}(\theta) = 
    2\beta \sigma( r_t(\xl_0) \\
    - r_t(\xw_0)) \cdot [ (\noise_\theta(\x_{t}^w,t) - \noise) \nabla_\theta\noise_\theta(\x_{t}^w,t) \\
    - (\noise_\theta(\x_{t}^l,t) - \noise) \nabla_\theta\noise_\theta(\x_{t}^l,t)  ],
\end{multline}
where $\beta$ absorbs constant terms for simplicity.
Empirically, we find that frequent updates to the reference model cause the gradient magnitude to become dominated by the model error term $\left|| \boldsymbol{\epsilon} - \boldsymbol{\epsilon}_\theta(\x_t, t) |\right|_2^2$.

In contrast, autoregressive language models are more robust to such error scaling, as reported in TR-DPO~\cite{gorbatovski2024learn}.
We hypothesize that this robustness stems from the fundamental modeling differences between language models and diffusion models.
Specifically, under the Lipschitz condition, the gradient of the DPO loss in language models is bounded.
\begin{theorem}
    Suppose $f = f_\theta(x) \in \mathbb{R}^V$ denote the output logits for a vocabulary of size $V$. Also, assume that $f_\theta$ is $K$-Lipschitz with respect to $\theta$.
    Let $y^w$ (preferred) and $y^l$ (dispreferred) be two responses for $x$, with the same length $T$.  
    Then, $\|\nabla_\theta \mathcal{L}_{\text{DPO}}\| \le 2\sqrt{2}\beta\cdot TK$.
\end{theorem}
\begin{proof}
We firstly show the upper bound of the logit.
Let $y \in \{1, \dots, V\}$ be a token.
The softmax distribution for $y$ is:
\begin{equation}
    \pi_\theta(y \mid x) = \frac{\exp(f_\theta(y))}{\sum_{j=1}^V \exp(f_\theta(j))},
\end{equation}
and the log-likelihood is computed as:
\begin{equation}
\log \pi_\theta(y \mid x) = f_\theta(y) - \log \left( \sum_{j=1}^V \exp(f_\theta(j)) \right).
\end{equation}
The gradient with respect to the logit is:
\begin{equation}
\nabla_f \log \pi_\theta(y \mid x) = \mathbf{e}_y - \pi_\theta(\cdot \mid x),
\end{equation}
where $\mathbf{e}_y$ is the one-hot vector.

Then,
\begin{multline}
    \|\nabla_f \log \pi_\theta(y \mid x)\|_2^2 
= \sum_{i=1}^V \left( \mathbf{e}_y(i) - \pi_\theta(i \mid x) \right)^2 \\
= (1 - 2\pi_\theta(y \mid x)) + \sum_{i=1}^V \pi_\theta(i \mid x)^2 \le
 1 + \sum_{i=1}^V \pi_\theta(i \mid x) = 2,
\end{multline}

where we use $\pi(i) \in [0,1]$. Hence, we have
\begin{equation}
    \|\nabla_f \log \pi_\theta(y \mid x)\|_2^2 \le 2.
    \label{eq:lm_logit_bound}
\end{equation}

Now consider the log probability for two responses $y^w$ and $y^l$:
\begin{equation}
\log \pi_\theta(y^i \mid x) = \sum_{t=1}^{T} \log \pi_\theta(y_t^i \mid x, y^i_{<t}), i \in \{w,l\}
\end{equation}
As the Equation \ref{eq:lm_logit_bound} holds for all $y_t$, and $f_\theta$ is $K$-Lipschitz with respect to $\theta$, it follows that:
\begin{equation}
\|\nabla_\theta \log \pi_\theta(y^i \mid x)\| \le \sqrt{2}TK.
\end{equation}
Now, consider the gradient of the DPO loss:
\begin{multline}
\nabla_\theta \mathcal{L}_{\text{DPO}} =\\
-\beta \cdot \sigma(-\beta z) \cdot \left( \nabla_\theta \log \pi_\theta(y^w \mid x) - \nabla_\theta \log \pi_\theta(y^l \mid x) \right),
\label{eq:lm_sequence_bound}
\end{multline}

where $z := \log \pi_\theta(y^w \mid x) - \log \pi_\theta(y^l \mid x) - (\log \pi_\text{ref}(y^w \mid x) - \log \pi_\text{ref}(y^l \mid x))$.

From the Equation \ref{eq:lm_sequence_bound}, the DPO gradient is bounded in norm by:
\begin{equation*}
    \|\nabla_\theta \mathcal{L}_{\text{DPO}}\| \le 2 \beta \cdot \sigma(-\beta z) \cdot \sqrt{2} TK.
\end{equation*}

Since $\sigma(\cdot) \le 1$, we arrive at the final upper bound.
\end{proof}

In Diffusion-DPO, even under the Lipschitz assumption for the model, the gradient can diverge due to the unbounded nature of the noise prediction error.
This again highlights the need for our reference model regularization, which controls model divergence while still allowing effective exploration.

\subsection{Analysis on Monitoring Threshold}
We conduct experiments across varying the monitoring threshold $\delta$ values, which defines the safe region for the reference model update strategy.
Table~\ref{tab:appendix_delta} presents results for three different $\delta$, based on Diffusion-DPO on SD1.5, evaluated on the Pick-a-Pic v2 test prompts.
The results show that $\delta = 0.005$ achieves the best performance, while either an excessively small or large $\delta$ leads to performance degradation.
A small $\delta$ does not allow the model to explore enough, while a large $\delta$ fails to provide adequate regularization for the reference model.
Although we experimentally choose $\delta$ based on this trade-off, the optimal value of $\delta$ may vary across models.
In future work, we hope to explore methods for the optimal selection of $\delta$.

\begin{figure}[t]
    \centering
    \includegraphics[width=0.85\columnwidth]{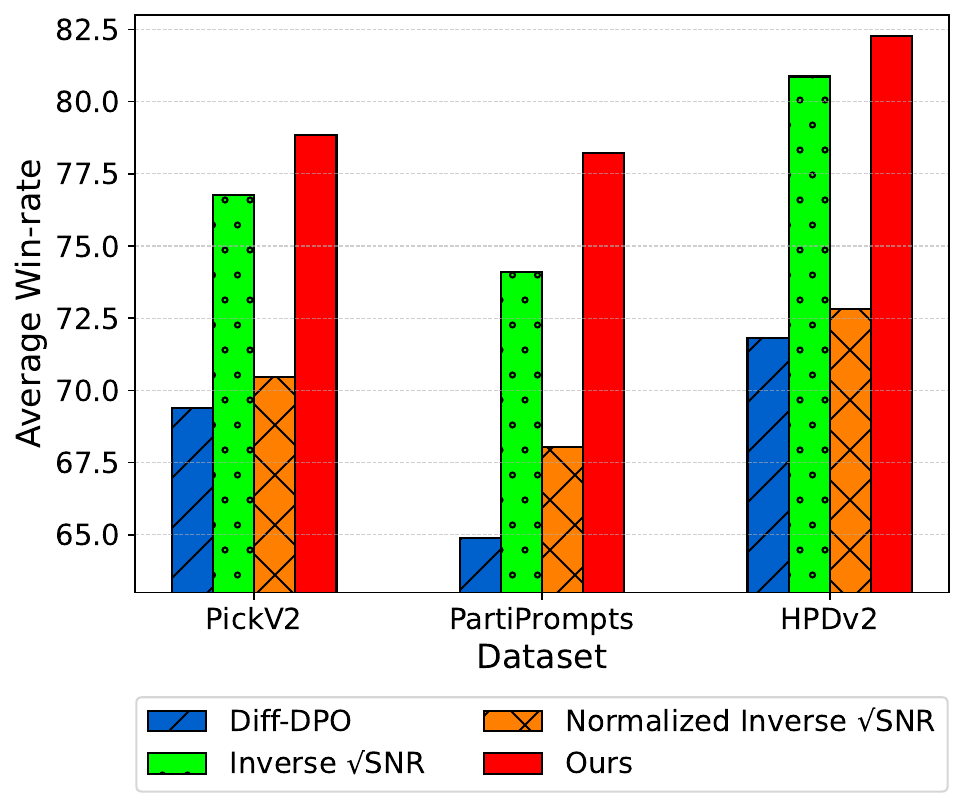}
    \caption{Ablation study on different timestep weights. Win rates are reported against SD1.5.}
    \label{fig:appendix_time_weight}
\end{figure}
\subsection{Analysis on Timestep-aware Training Strategy}
To further investigate the impact of time-step weighting strategies, we compare our method $\lambda(t)$ against other alternatives, including the square root of the inverse signal-to-noise value (SNR), $1/\sqrt{SNR(t)}$, and the normalized version, $\text{norm}(1/\sqrt{SNR(t)})$. 

The unnormalized weighting $1/\sqrt{SNR(t)}$ assigns large weights to highly noisy timesteps. 
Compared to our normalized version, its large value at early timesteps can impose overly excessive regularization, yielding suboptimal performance.
However, with the normalized weighting, we observe a sharp performance drop, as it assigns small regularization weights for later timesteps.
The additive offset of value $1$ in our method guarantees sufficient regularization for every timestep, while assigning higher weights to early steps.
Empirical results in Figure~\ref{fig:appendix_time_weight} validate the effectiveness of our design, achieving the highest average win rate across datasets.

\section{Additional Qualitative Results}
\label{appendix:qualitative_results}
\begin{figure*}[!hb!]
    \centering

    \parbox{1.02 \textwidth}{
        \centering
        Gothic cathedral in a stormy night
    }
    \begin{subfigure}[t]{0.19\textwidth} \includegraphics[width=\textwidth]{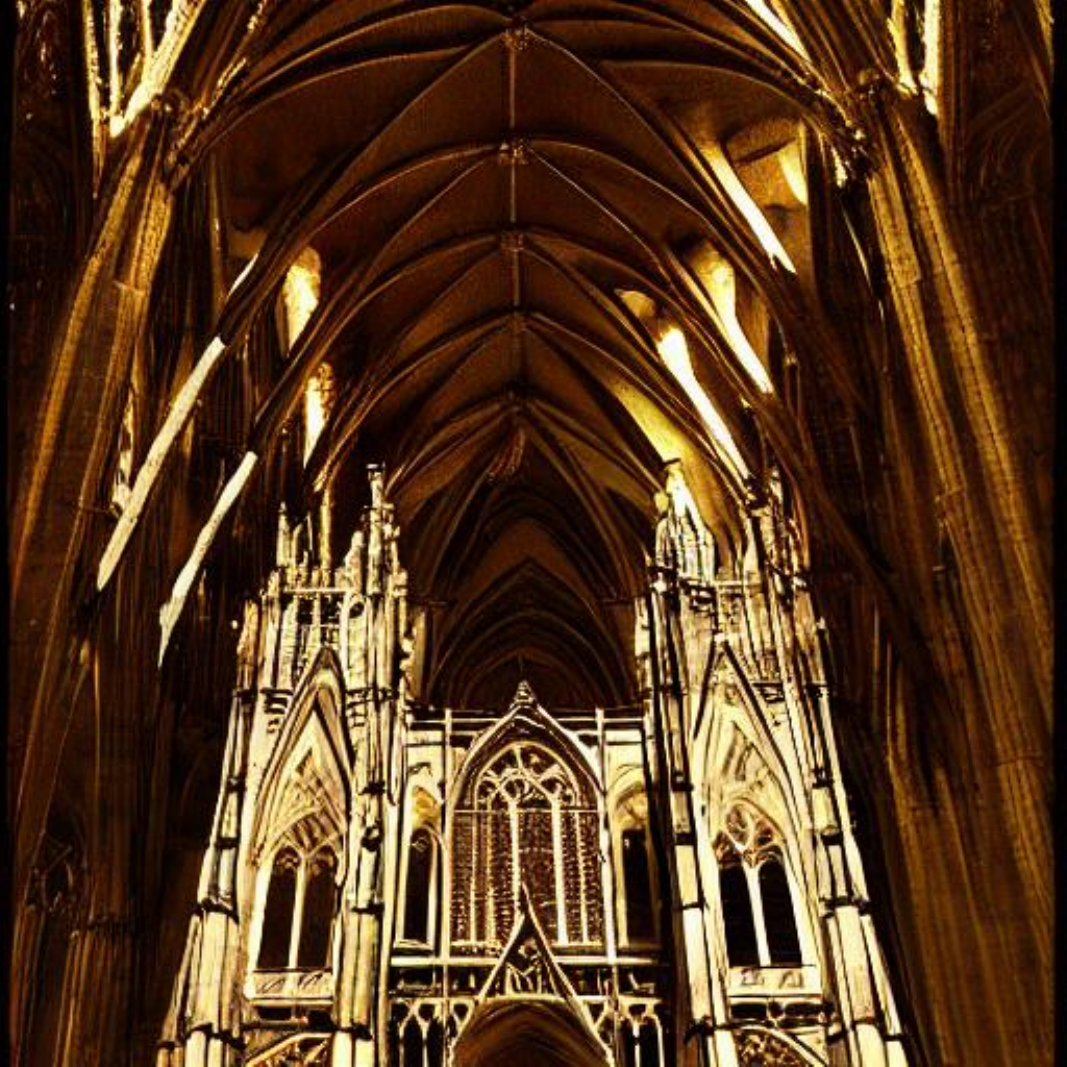} \end{subfigure}
    \begin{subfigure}[t]{0.19\textwidth} \includegraphics[width=\textwidth]{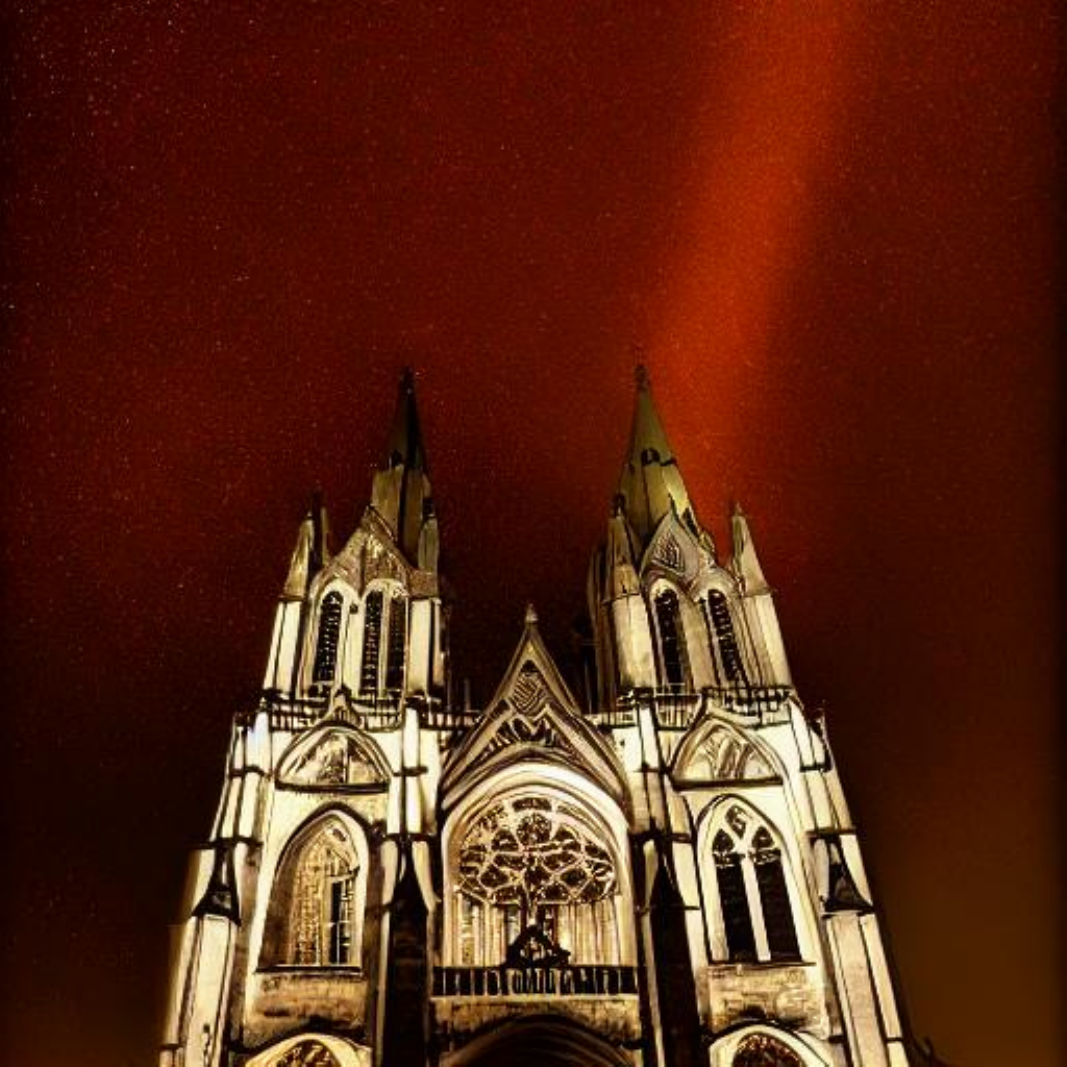} \end{subfigure}
    \begin{subfigure}[t]{0.19\textwidth} \includegraphics[width=\textwidth]{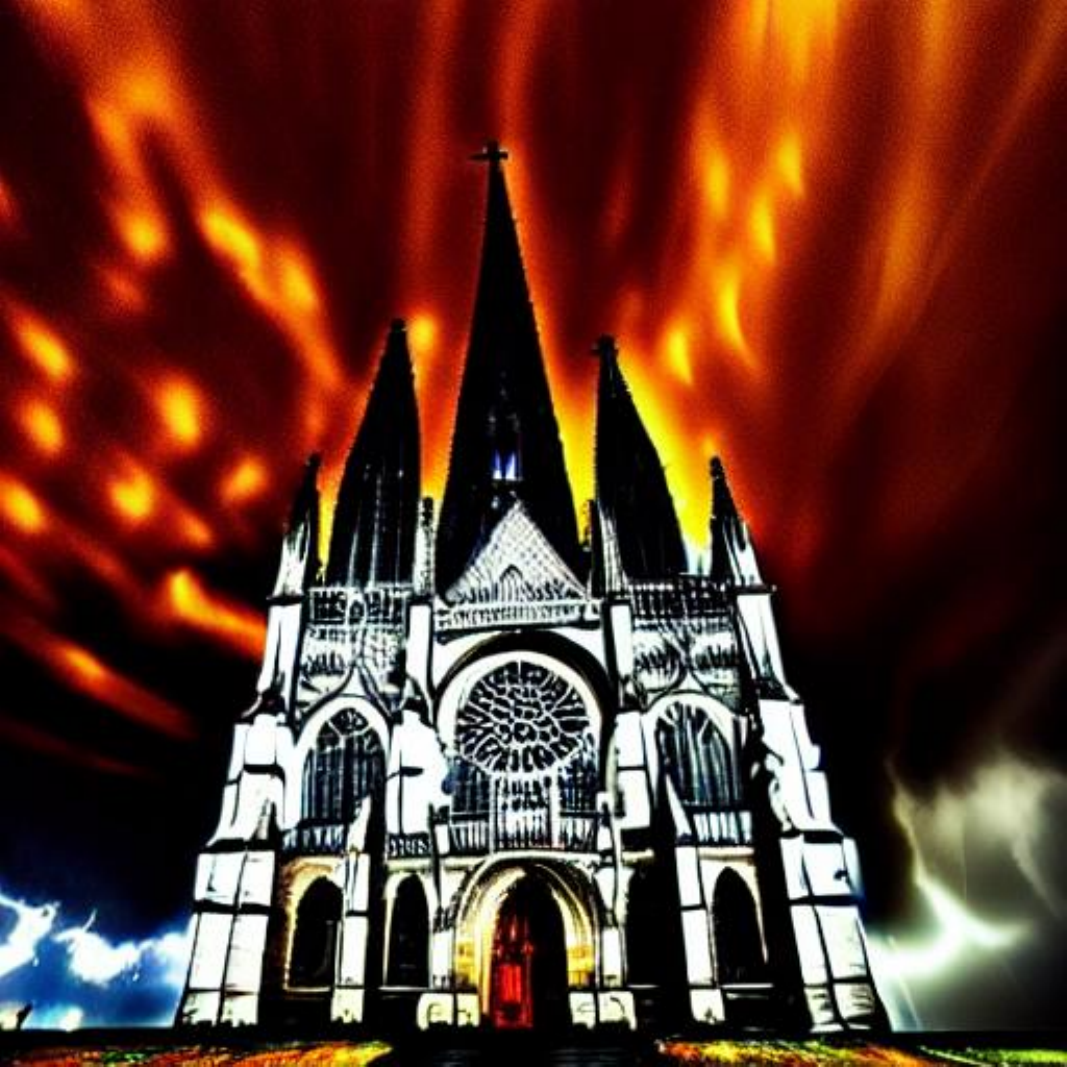} \end{subfigure}
    \begin{subfigure}[t]{0.19\textwidth} \includegraphics[width=\textwidth]{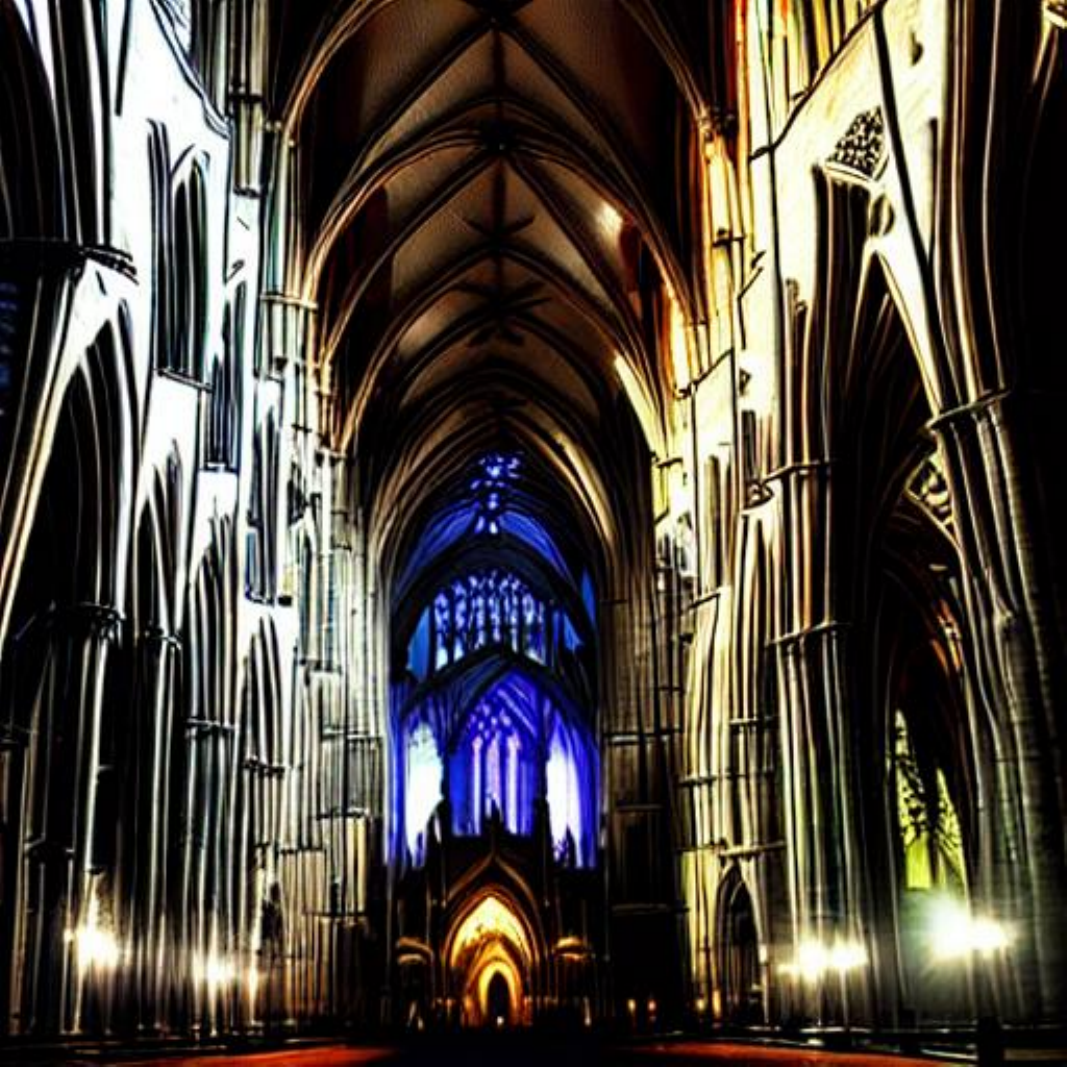} \end{subfigure}
    \begin{subfigure}[t]{0.19\textwidth} \includegraphics[width=\textwidth]{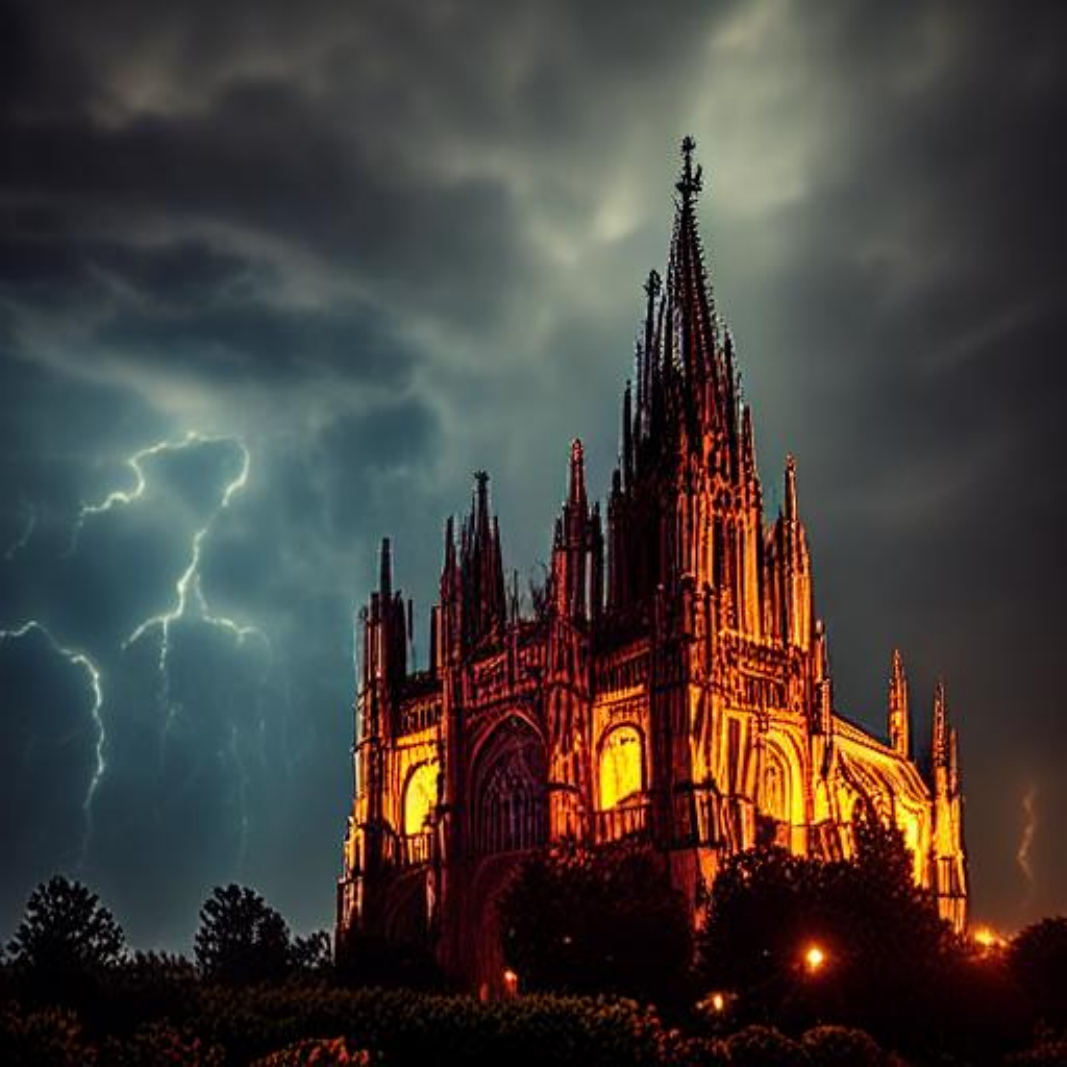} \end{subfigure}

    \parbox{1.02 \textwidth}{
        \centering
        
            75 years old, grandfather, bodybuilded too much, crush an apple in his hand
    }
    \begin{subfigure}[t]{0.19\textwidth} \includegraphics[width=\textwidth]{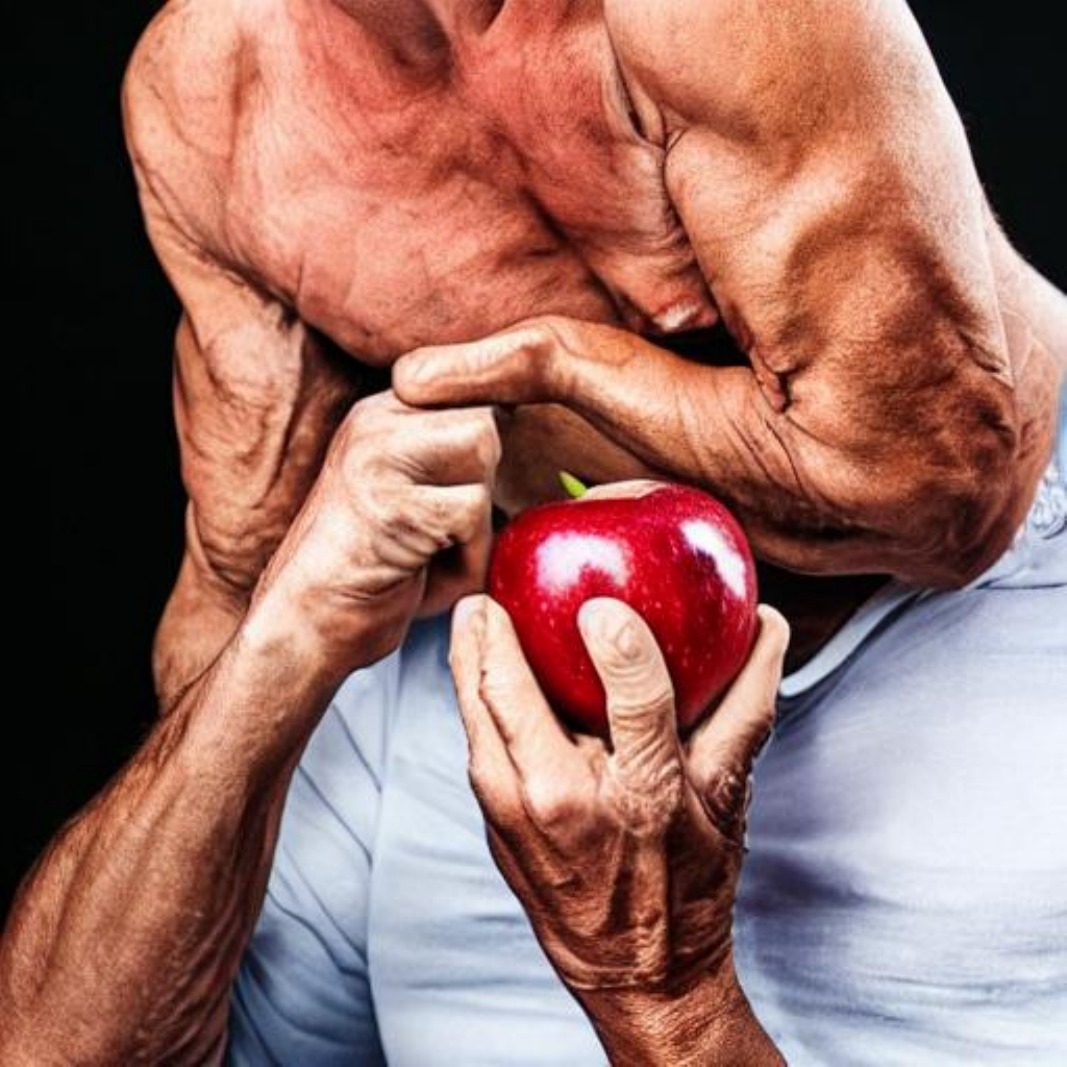} \end{subfigure}
    \begin{subfigure}[t]{0.19\textwidth} \includegraphics[width=\textwidth]{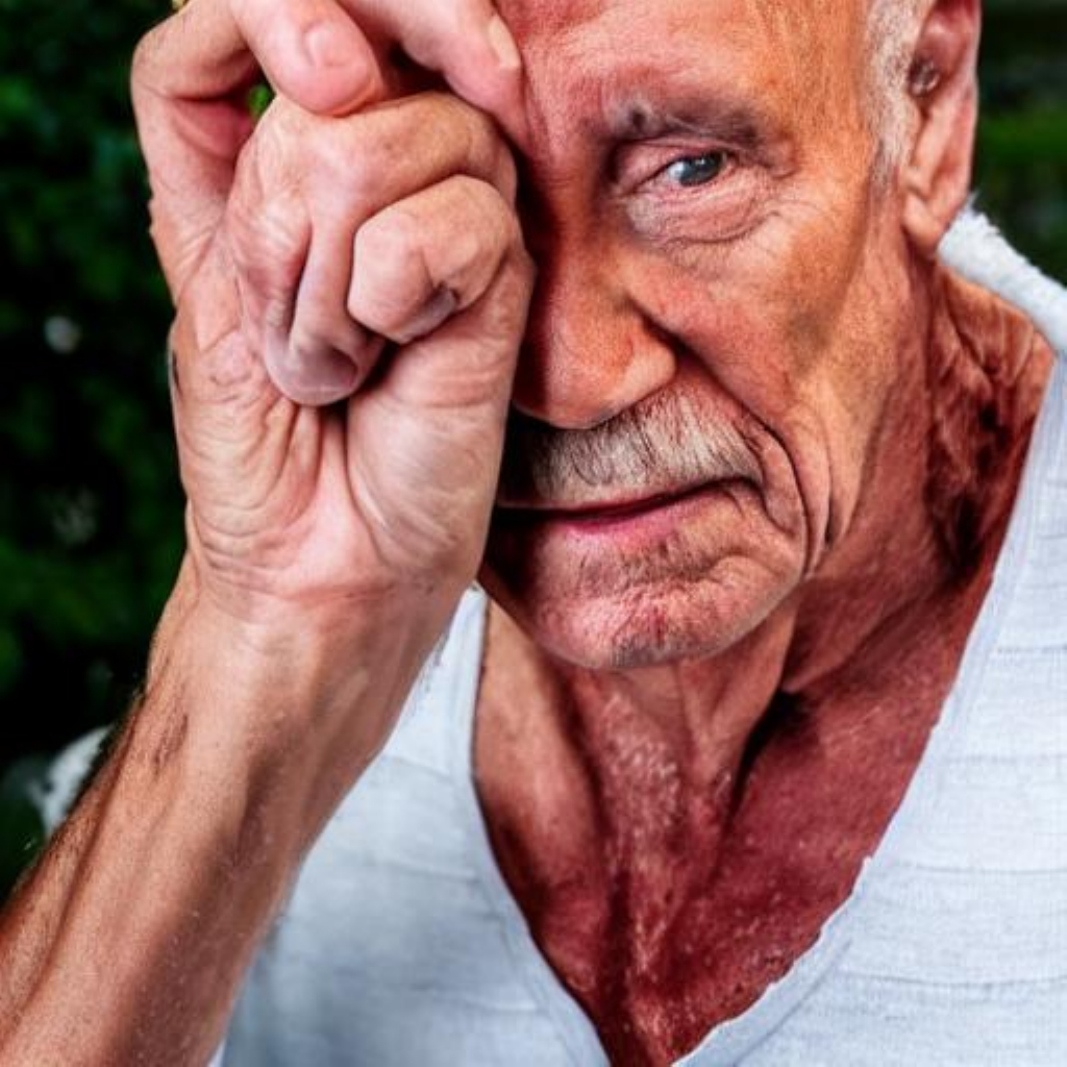} \end{subfigure}
    \begin{subfigure}[t]{0.19\textwidth} \includegraphics[width=\textwidth]{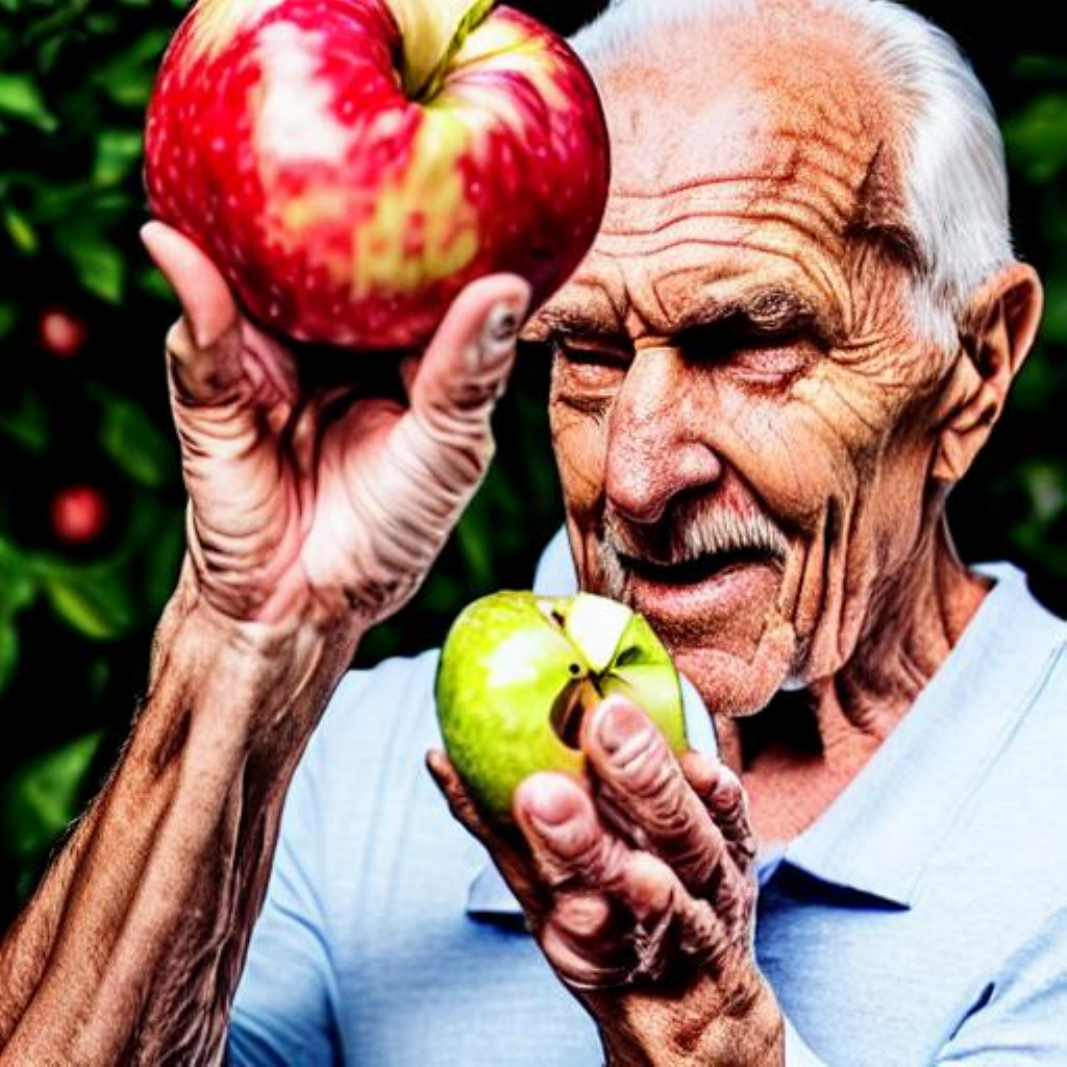} \end{subfigure}
    \begin{subfigure}[t]{0.19\textwidth} \includegraphics[width=\textwidth]{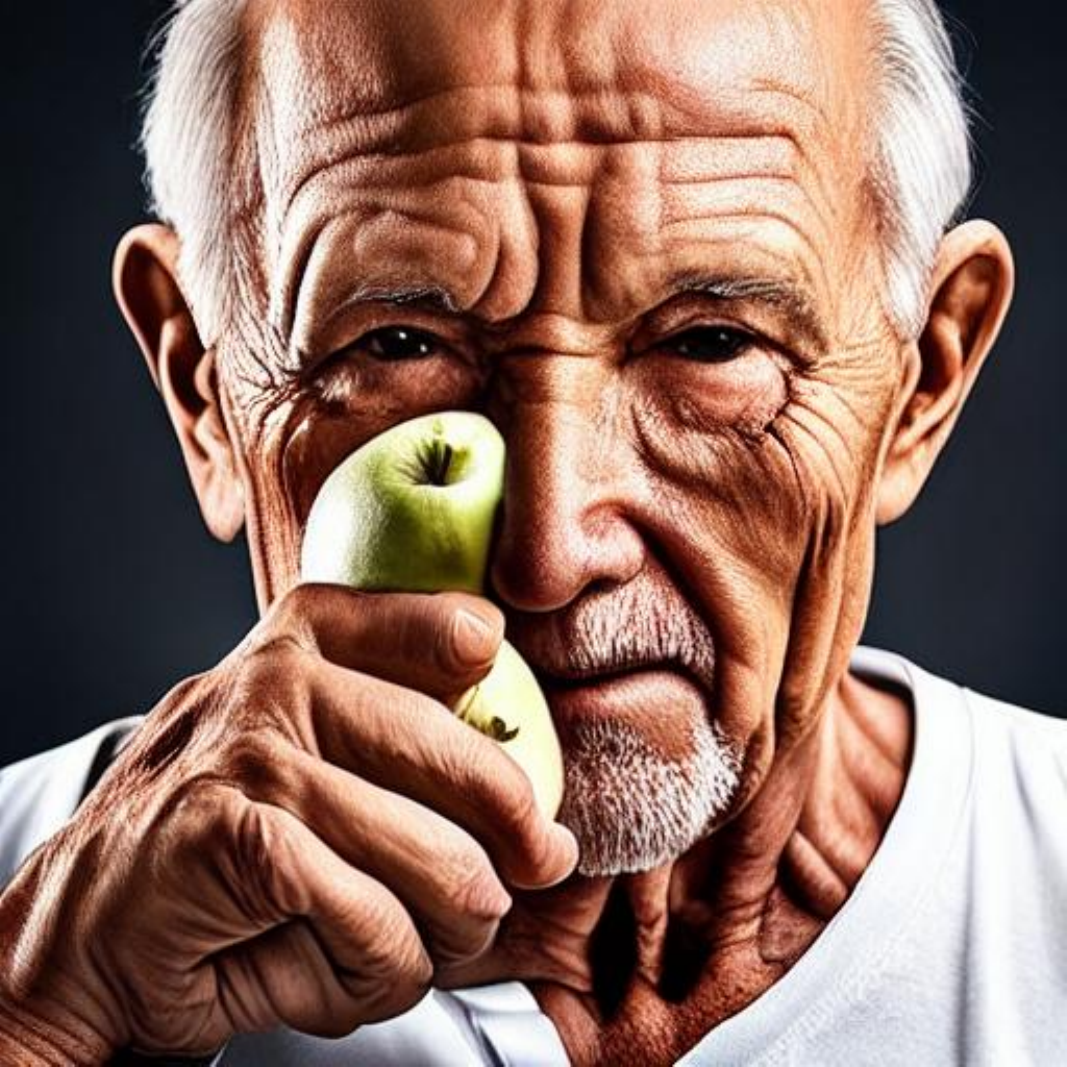} \end{subfigure}
    \begin{subfigure}[t]{0.19\textwidth} \includegraphics[width=\textwidth]{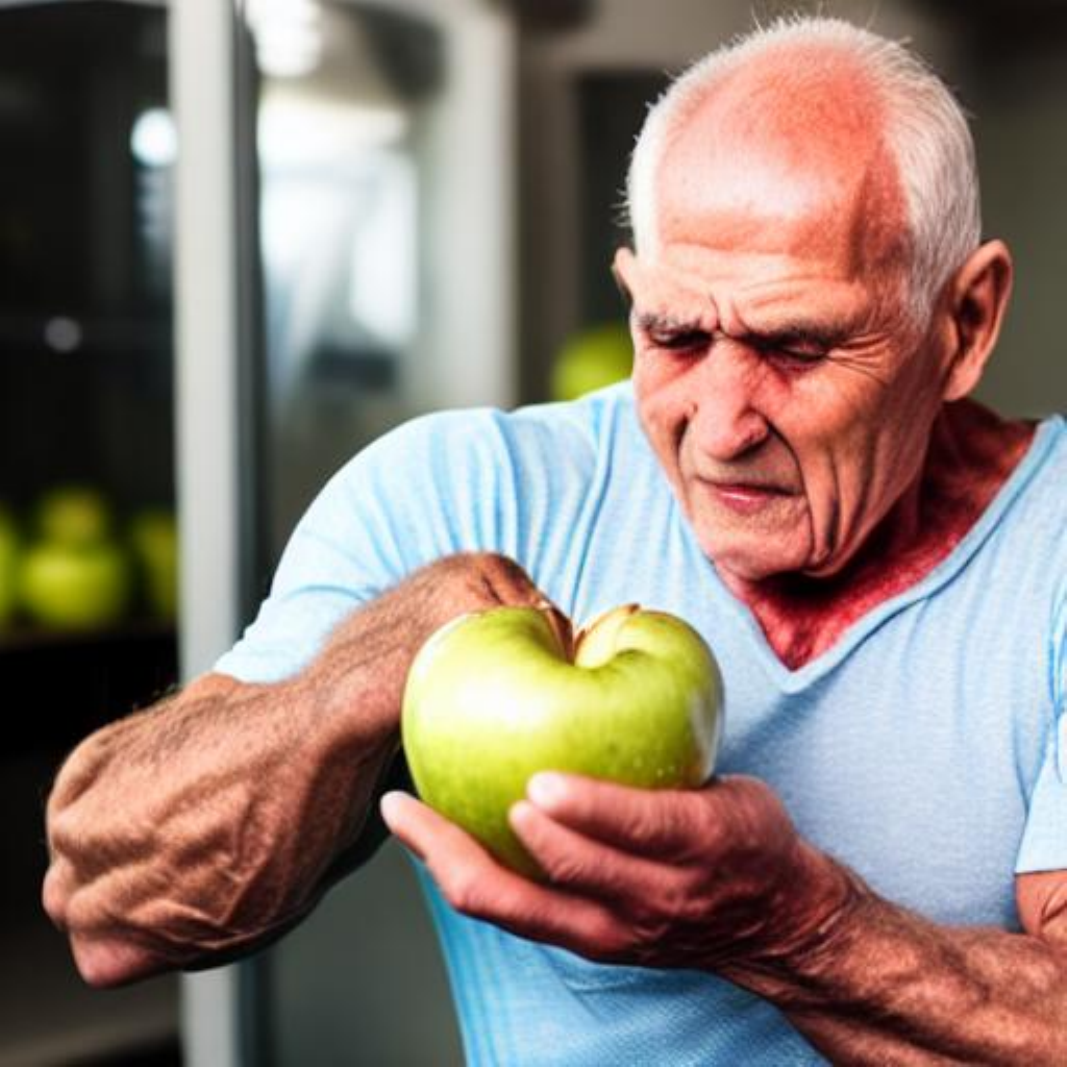} \end{subfigure}

    \parbox{1.02 \textwidth}{
        \centering
        A sail boat entering a majestic fjord landscape in winter
    }
    \begin{subfigure}[t]{0.19\textwidth} \includegraphics[width=\textwidth]{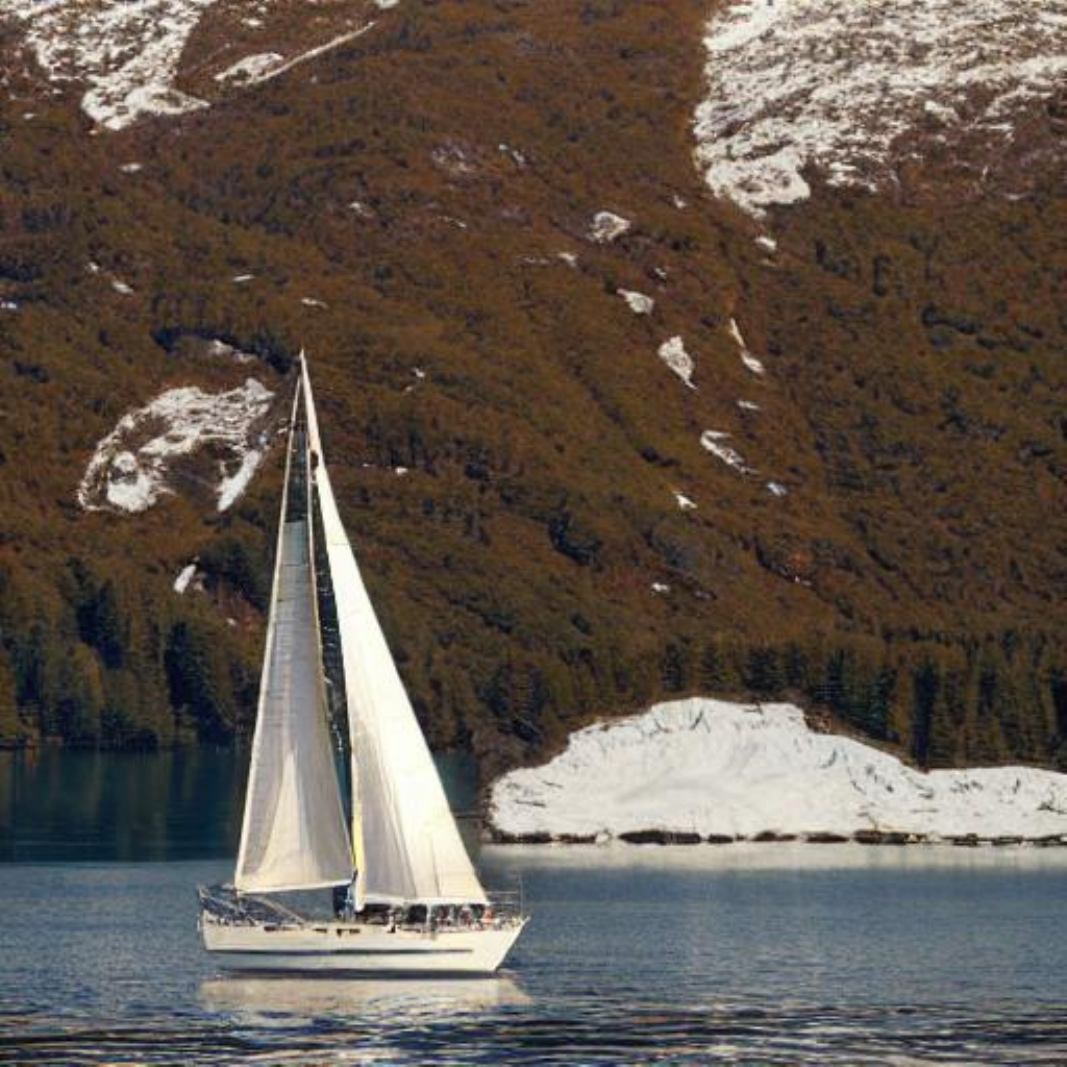} \end{subfigure}
    \begin{subfigure}[t]{0.19\textwidth} \includegraphics[width=\textwidth]{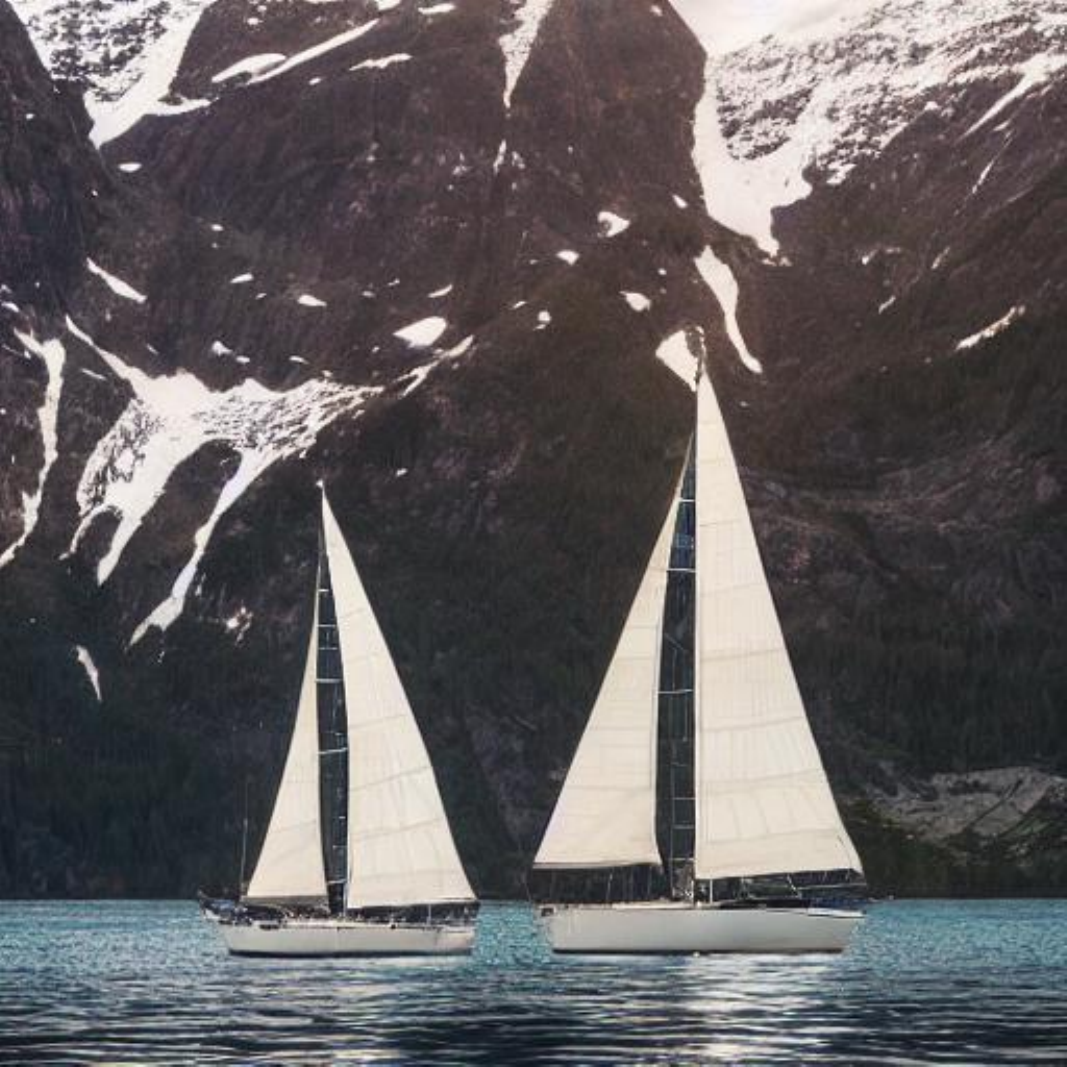} \end{subfigure}
    \begin{subfigure}[t]{0.19\textwidth} \includegraphics[width=\textwidth]{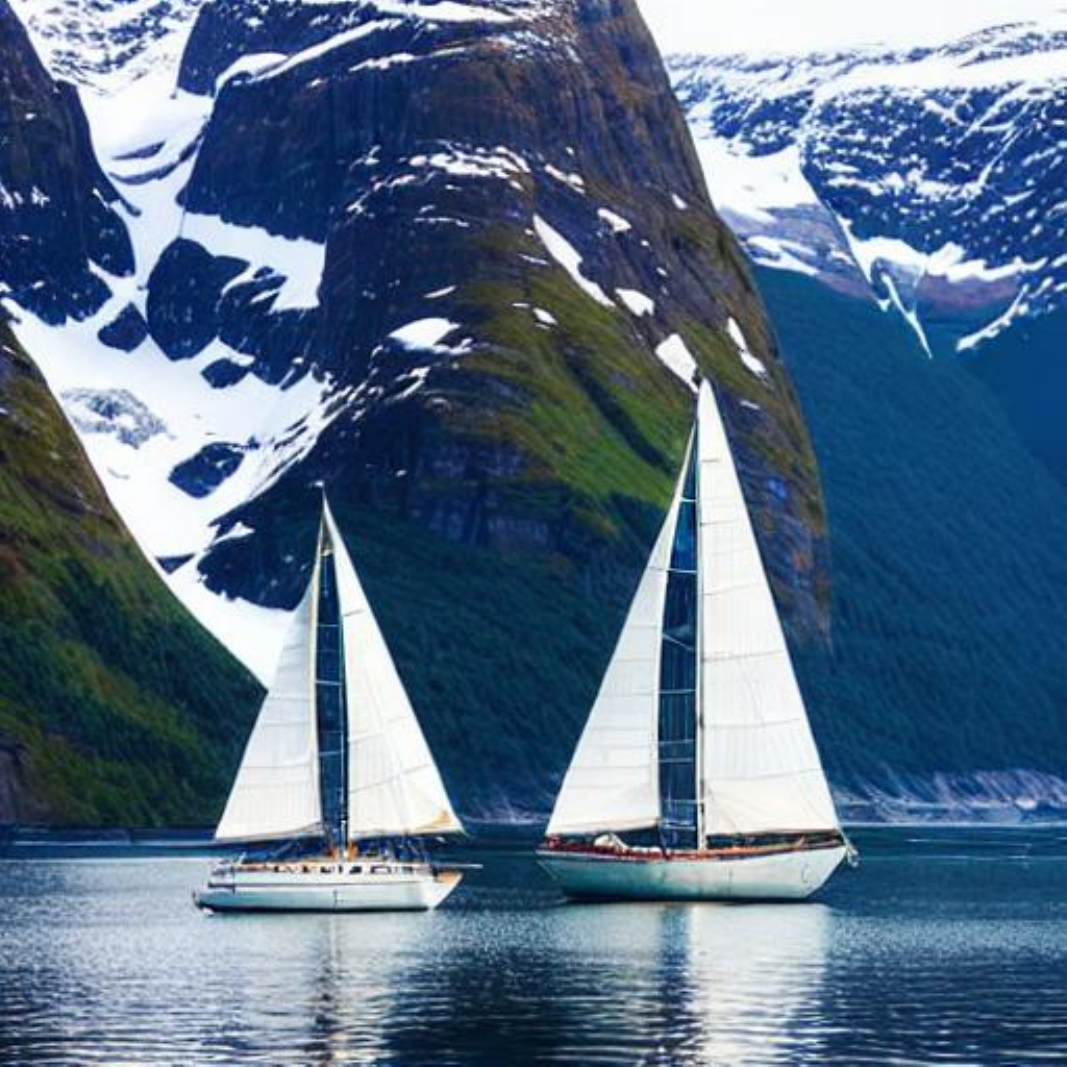} \end{subfigure}
    \begin{subfigure}[t]{0.19\textwidth} \includegraphics[width=\textwidth]{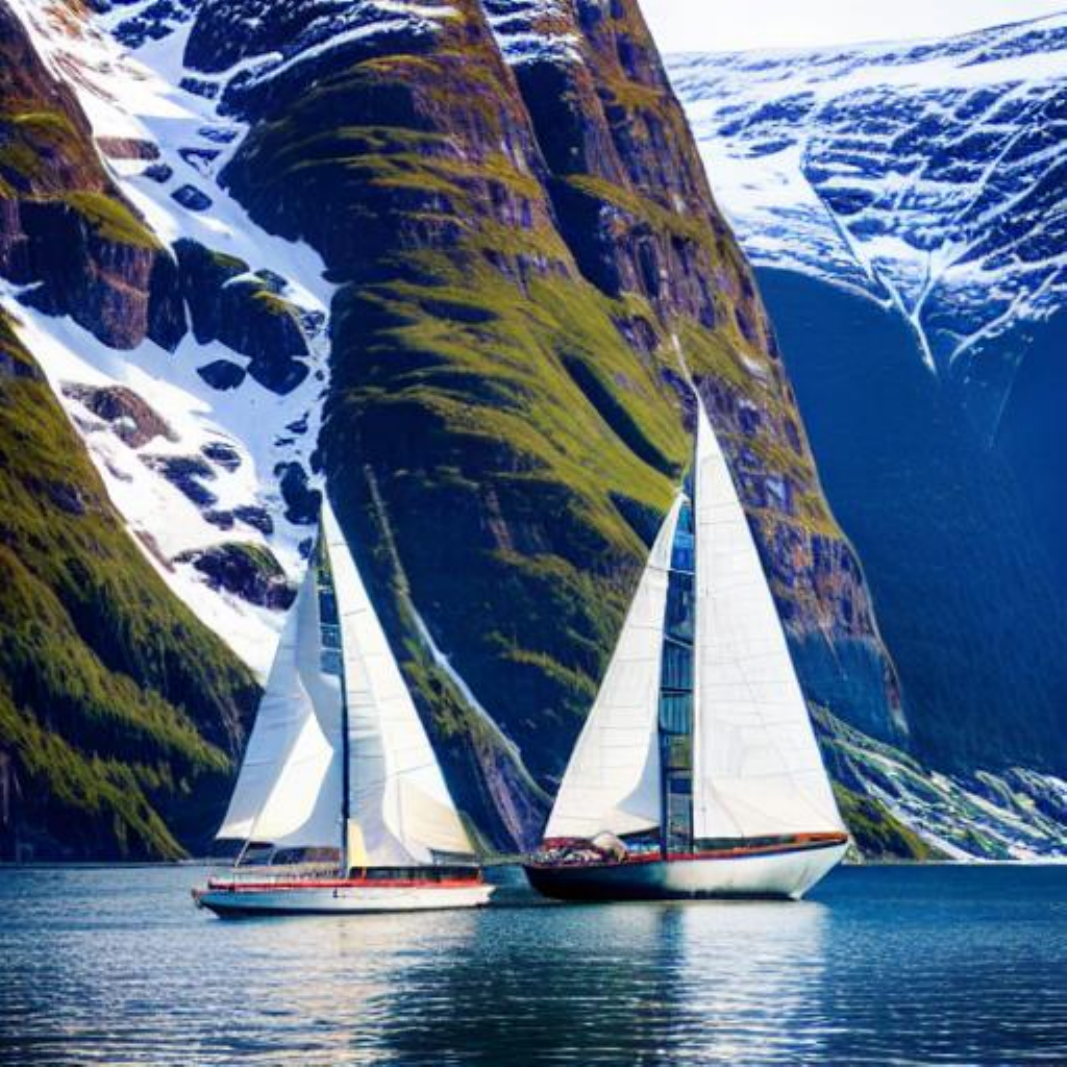} \end{subfigure}
    \begin{subfigure}[t]{0.19\textwidth} \includegraphics[width=\textwidth]{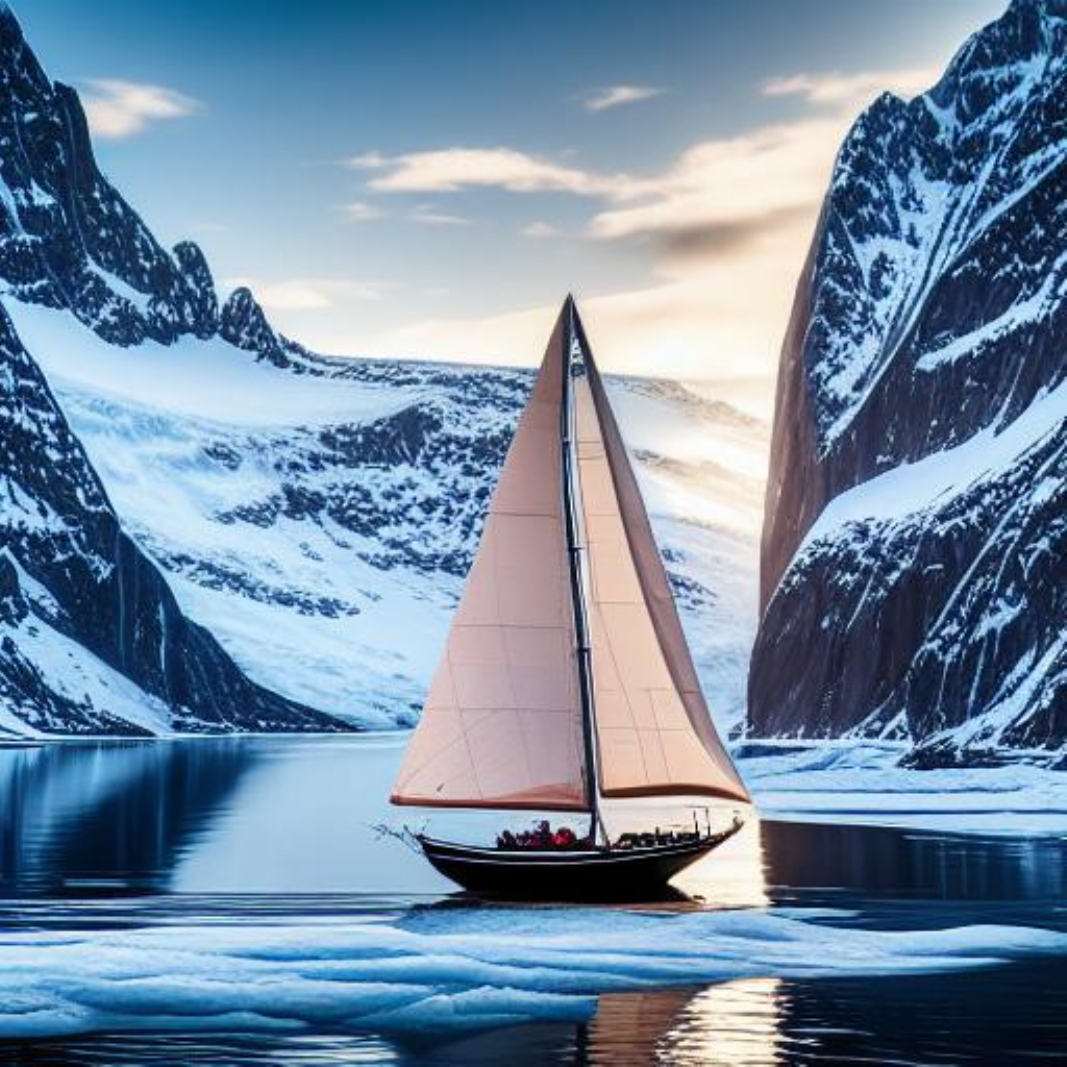} \end{subfigure}

    \parbox{1.02 \textwidth}{
        \centering
        Portrait of general with obscure hat
    }
    \begin{subfigure}[t]{0.19\textwidth} \includegraphics[width=\textwidth]{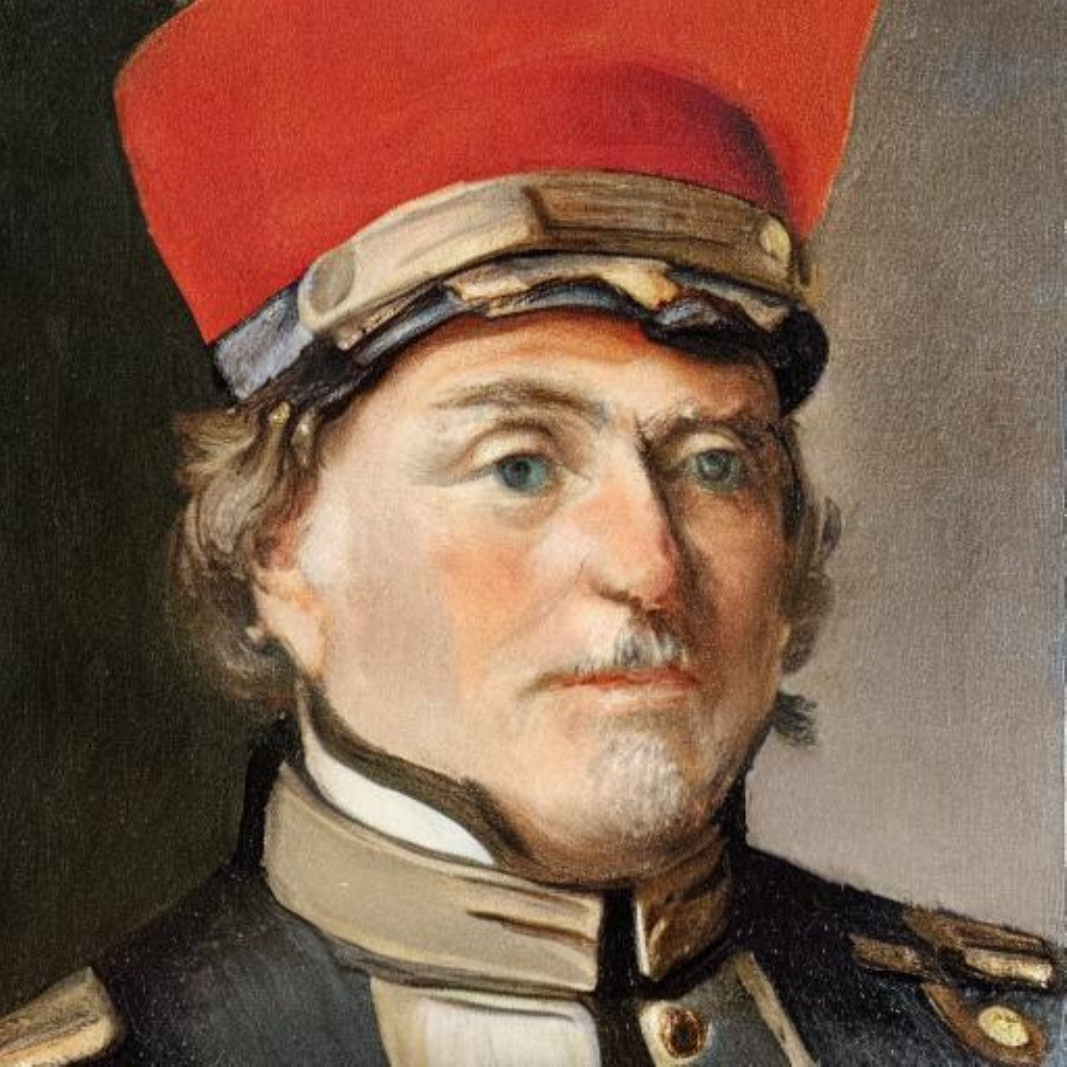} \end{subfigure}
    \begin{subfigure}[t]{0.19\textwidth} \includegraphics[width=\textwidth]{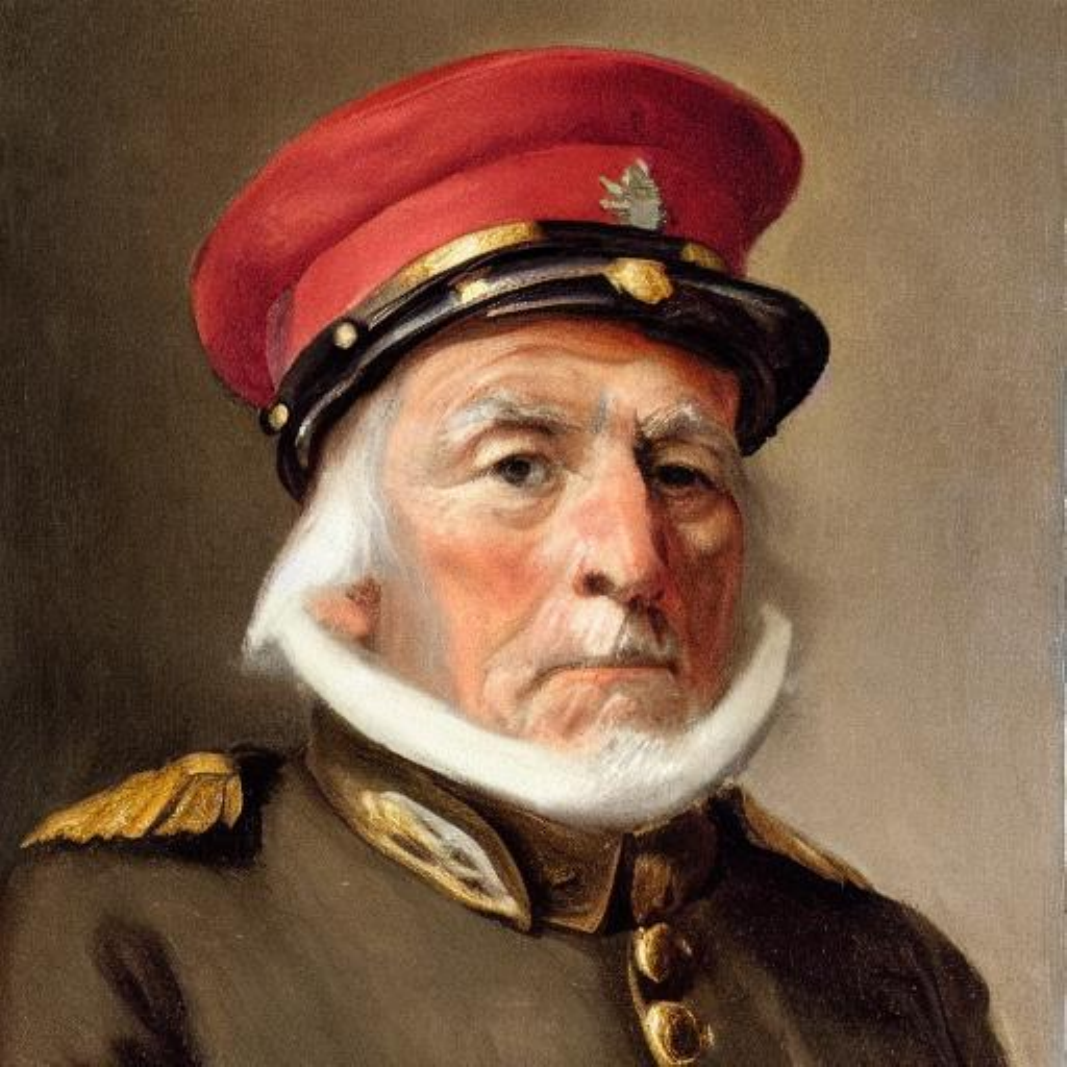} \end{subfigure}
    \begin{subfigure}[t]{0.19\textwidth} \includegraphics[width=\textwidth]{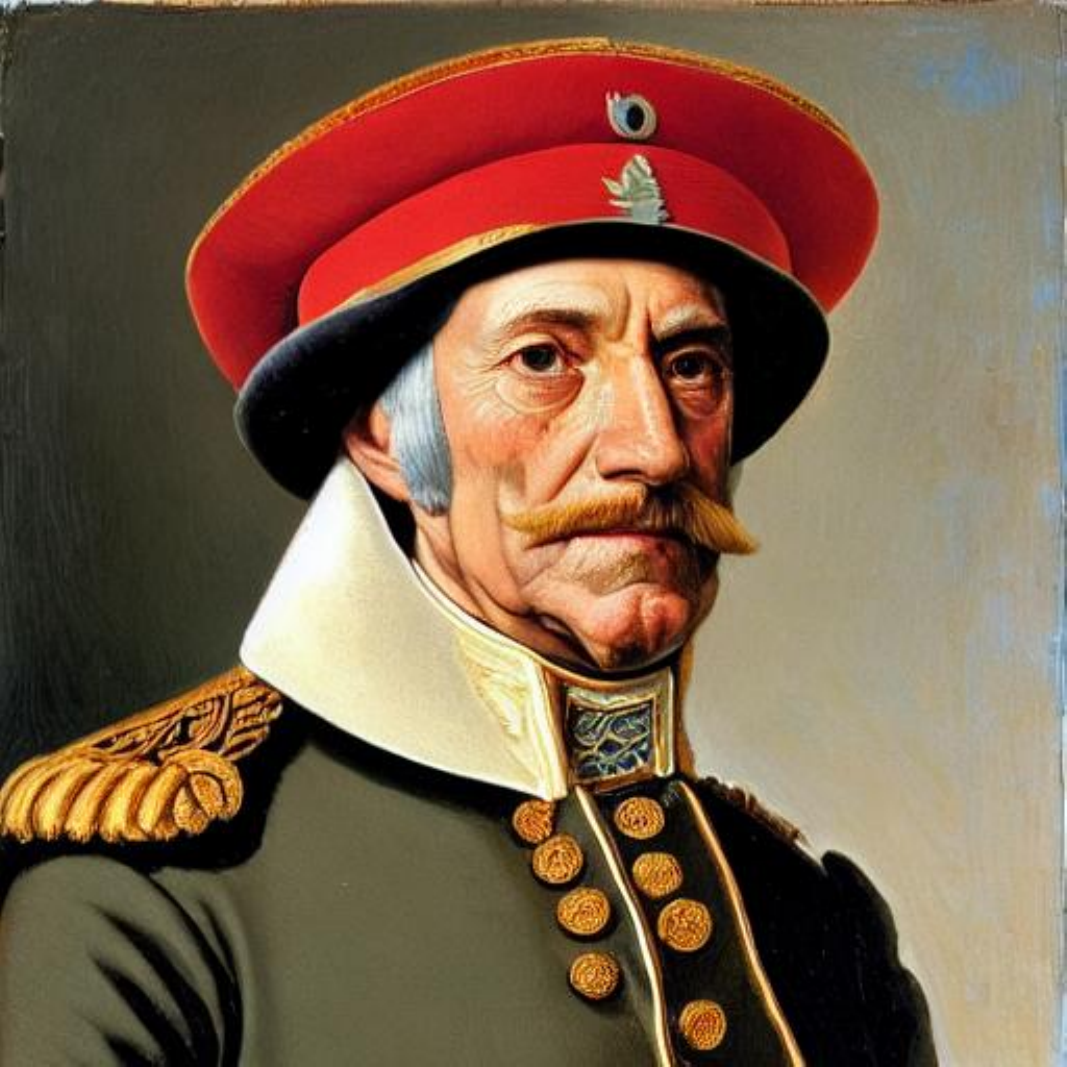} \end{subfigure}
    \begin{subfigure}[t]{0.19\textwidth} \includegraphics[width=\textwidth]{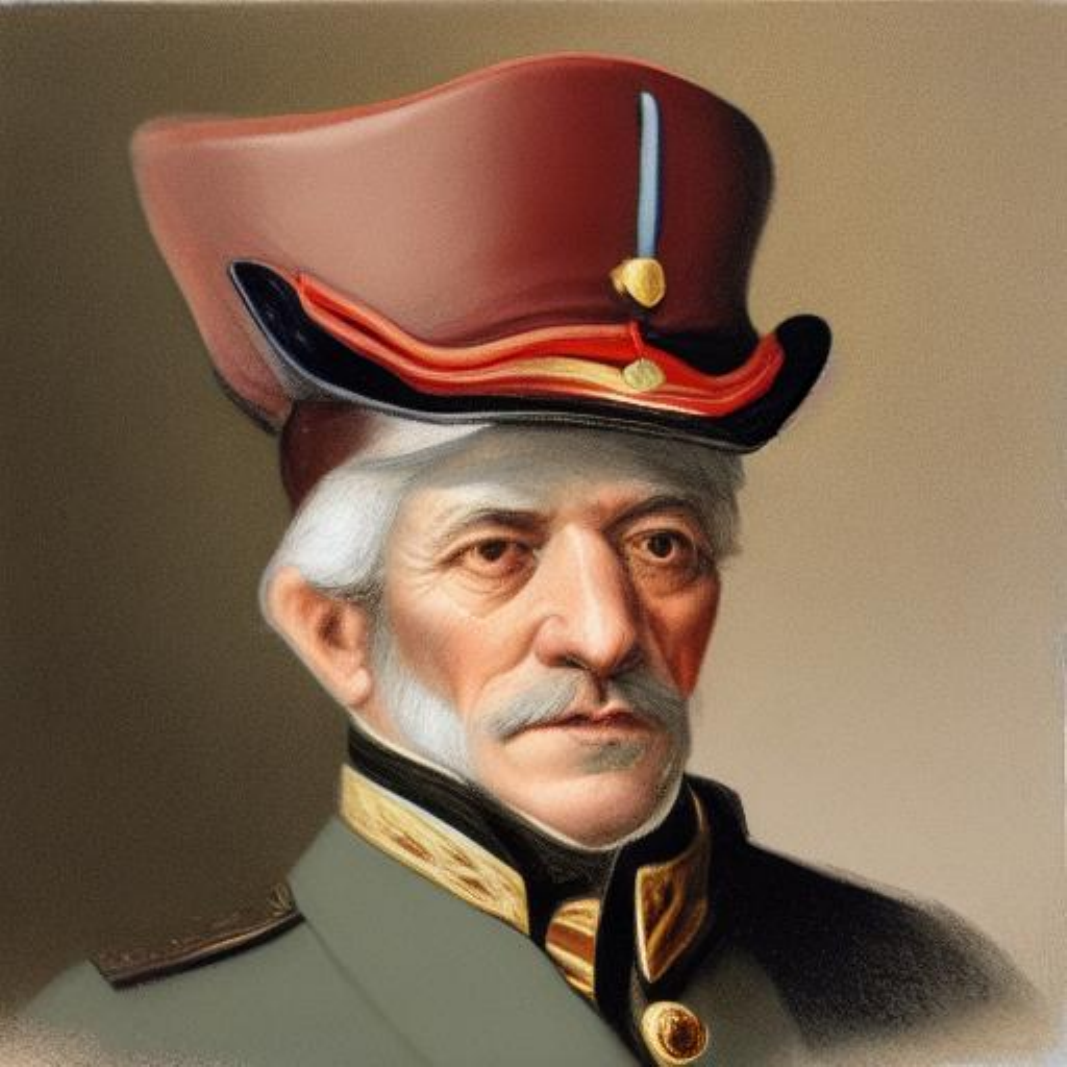} \end{subfigure}
    \begin{subfigure}[t]{0.19\textwidth} \includegraphics[width=\textwidth]{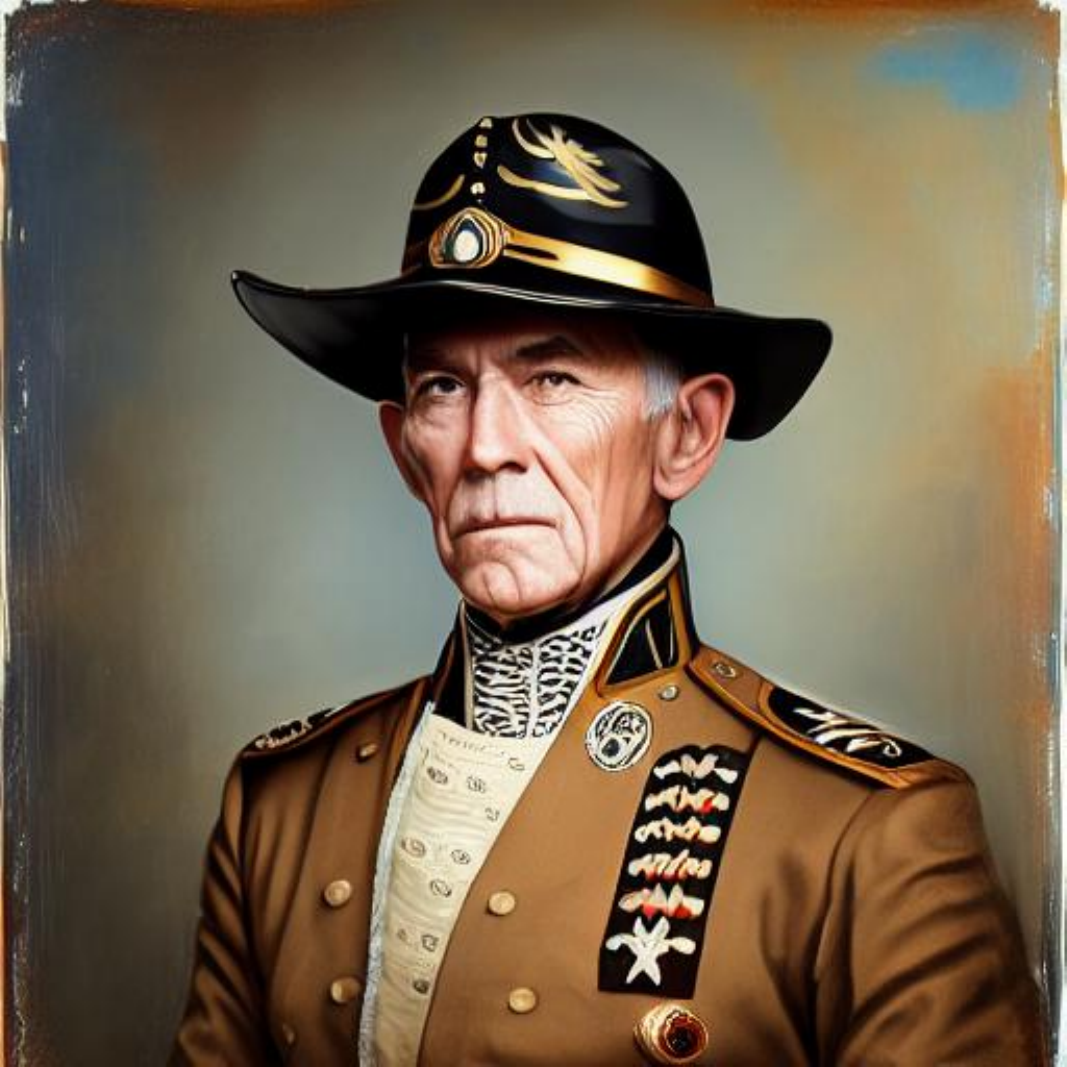} \end{subfigure}

    \parbox{1.02 \textwidth}{
        \centering
        pink eagle
    }
    \begin{subfigure}[t]{0.19\textwidth} \includegraphics[width=\textwidth]{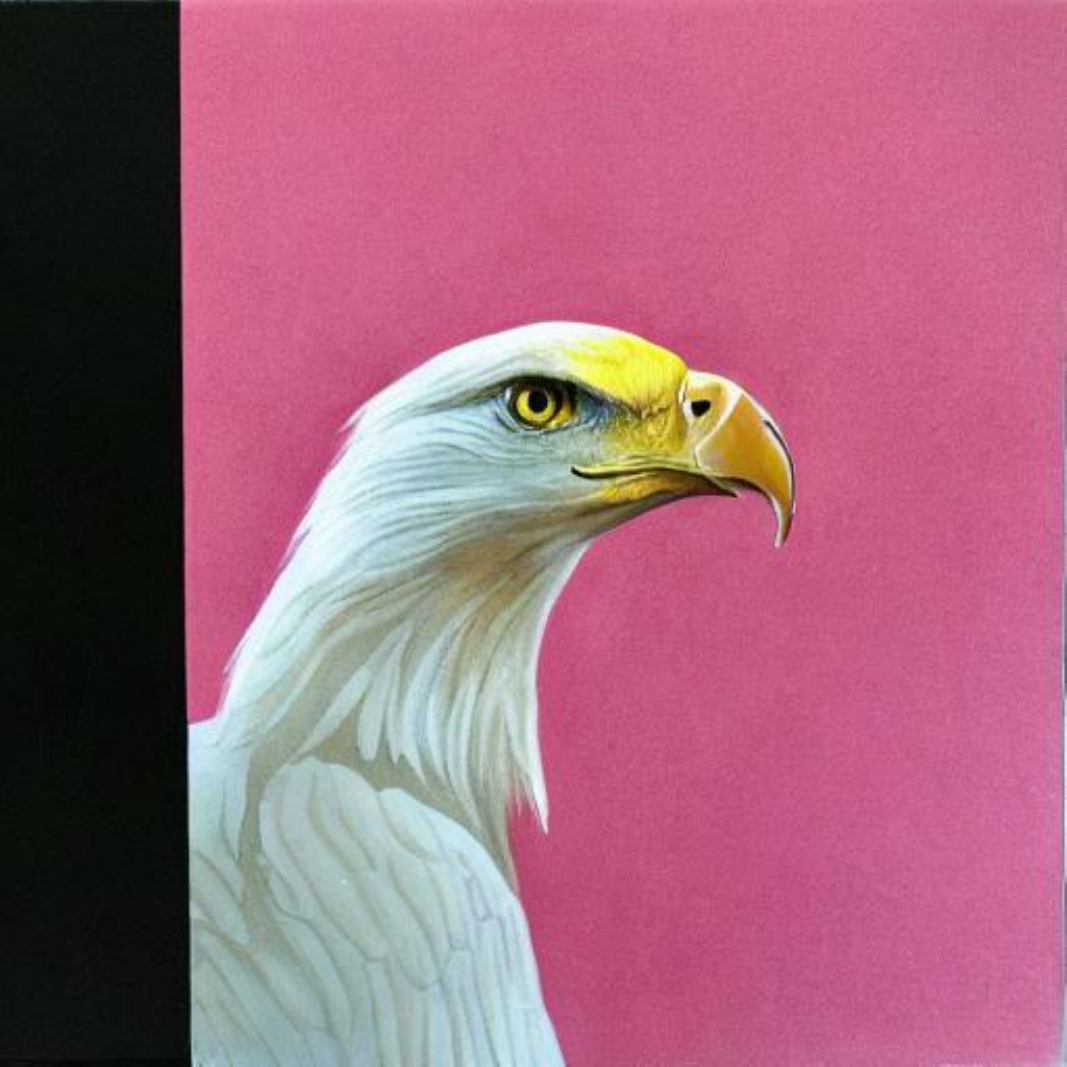} \end{subfigure}
    \begin{subfigure}[t]{0.19\textwidth} \includegraphics[width=\textwidth]{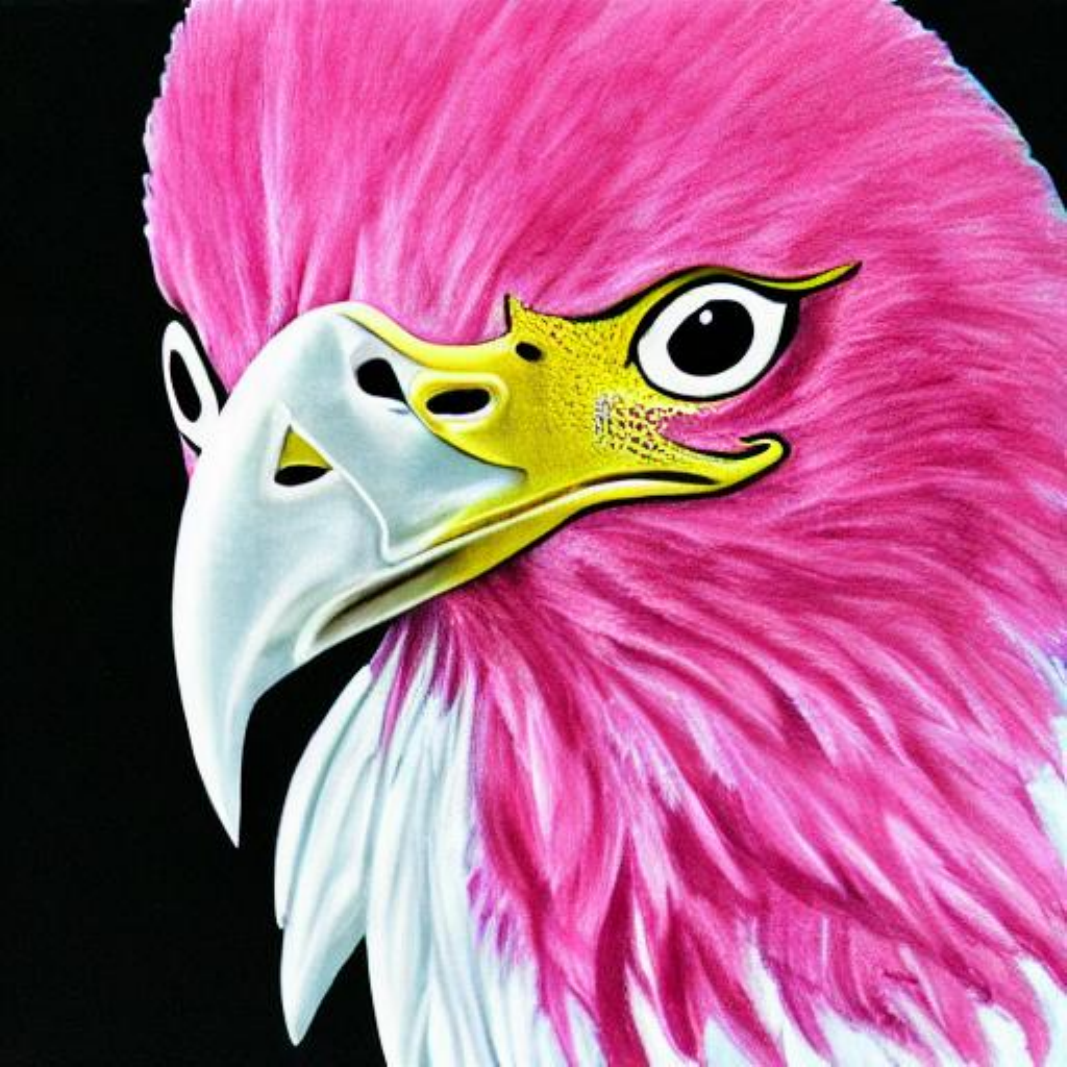} \end{subfigure}
    \begin{subfigure}[t]{0.19\textwidth} \includegraphics[width=\textwidth]{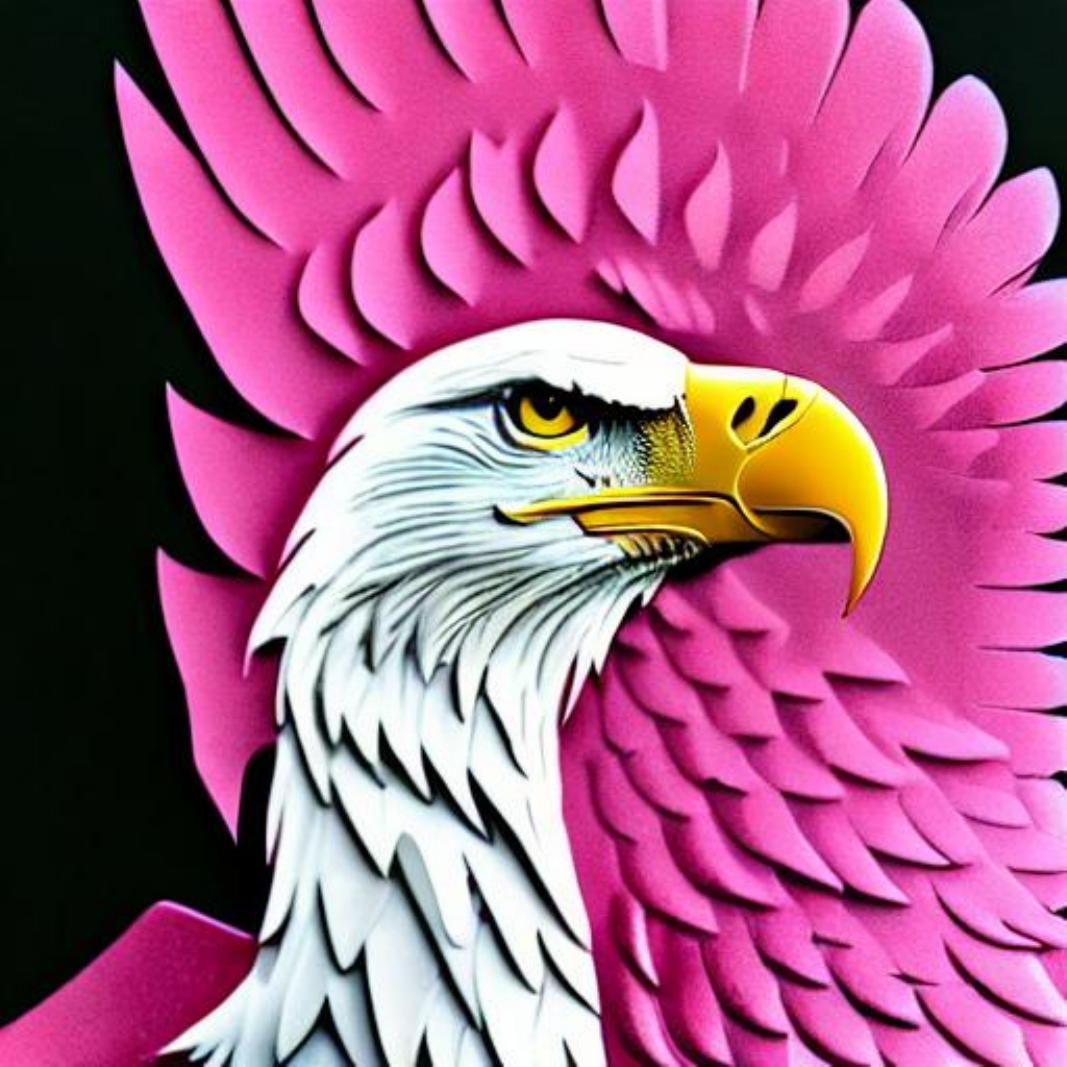} \end{subfigure}
    \begin{subfigure}[t]{0.19\textwidth} \includegraphics[width=\textwidth]{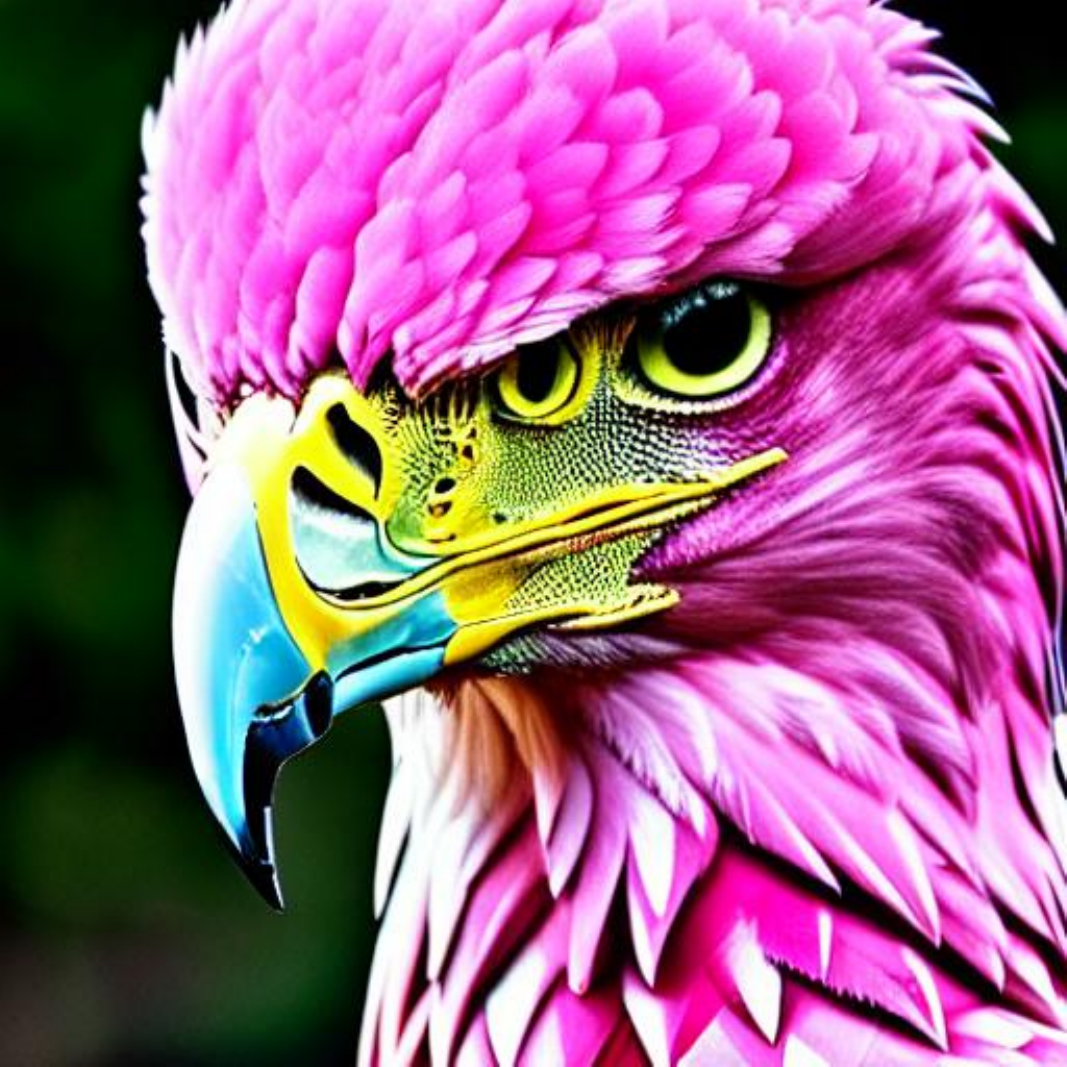} \end{subfigure}
    \begin{subfigure}[t]{0.19\textwidth} \includegraphics[width=\textwidth]{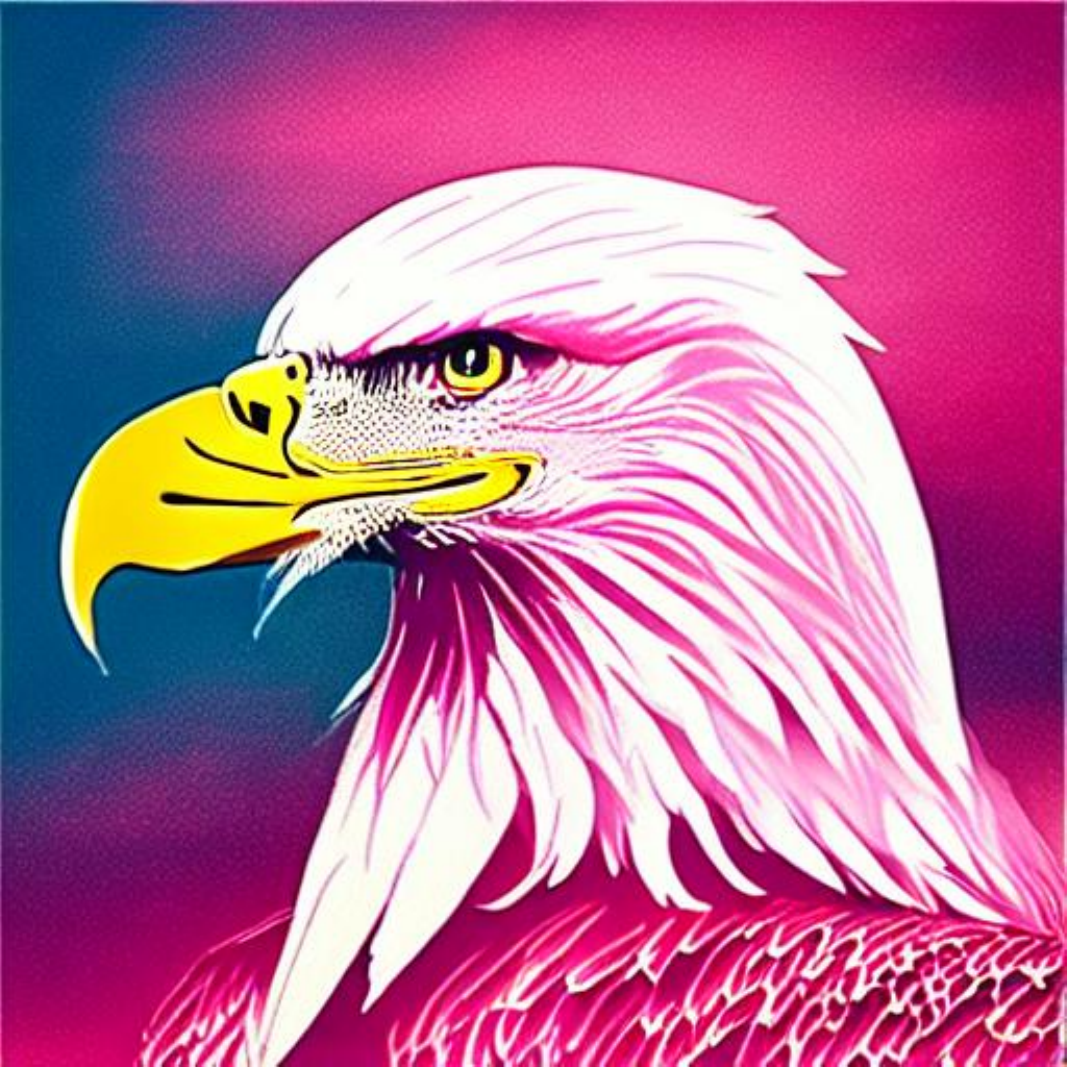} \end{subfigure}

    \parbox{1.02 \textwidth}{
        \centering
             A smooth purple octopus sitting on a rock in the middle of the sea, waves crashing,\\
             golden hour, sun reflections, high quality 3d render
        
    }
    \begin{subfigure}[t]{0.19\textwidth} \includegraphics[width=\textwidth]{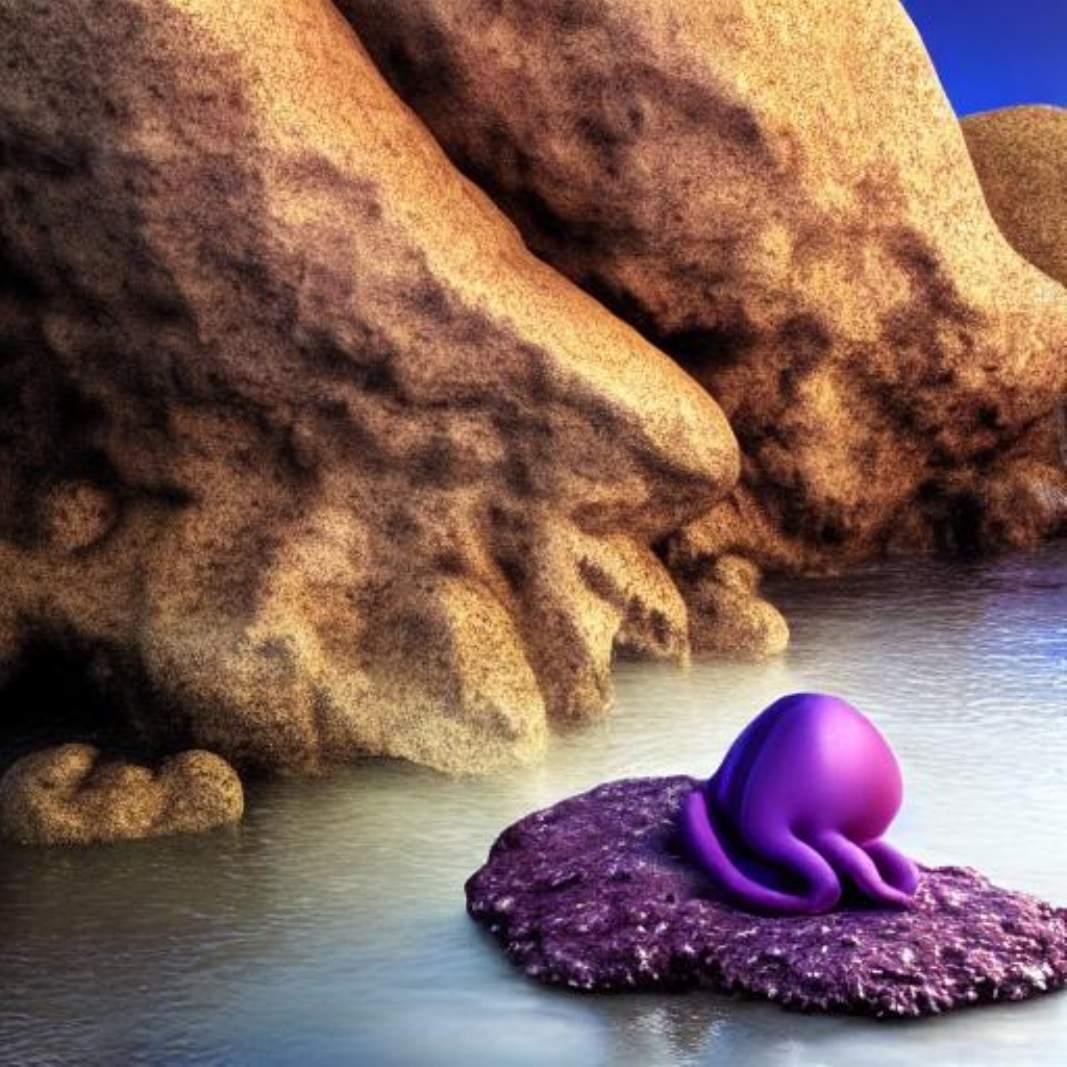} 
            \captionsetup{justification=centering}\caption{SD1.5~\cite{rombach2022high}}    \end{subfigure}
    \begin{subfigure}[t]{0.19\textwidth} \includegraphics[width=\textwidth]{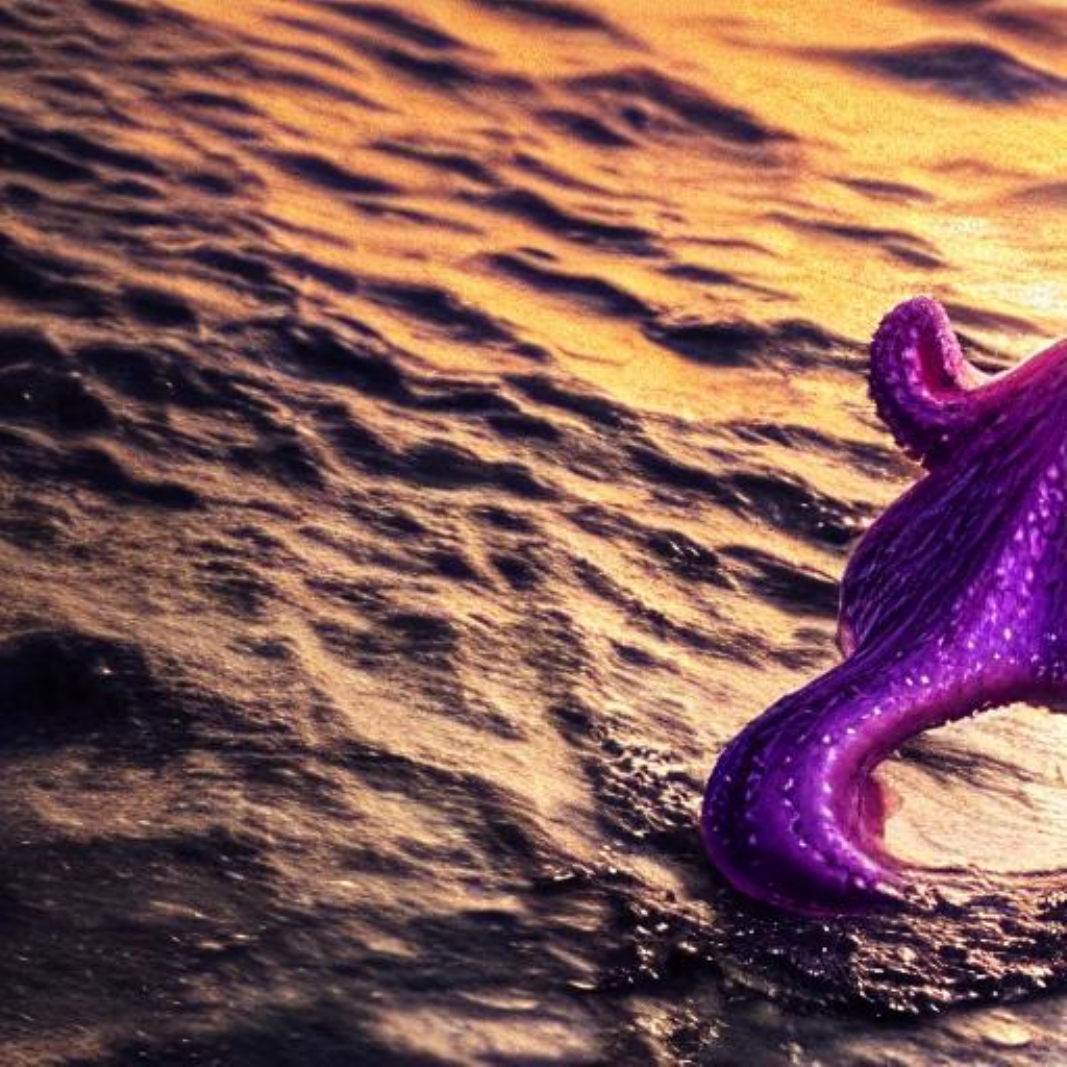} 
            \captionsetup{justification=centering}\caption{Diff-DPO~\cite{wallace2024diffusion}} \end{subfigure}
    \begin{subfigure}[t]{0.19\textwidth} \includegraphics[width=\textwidth]{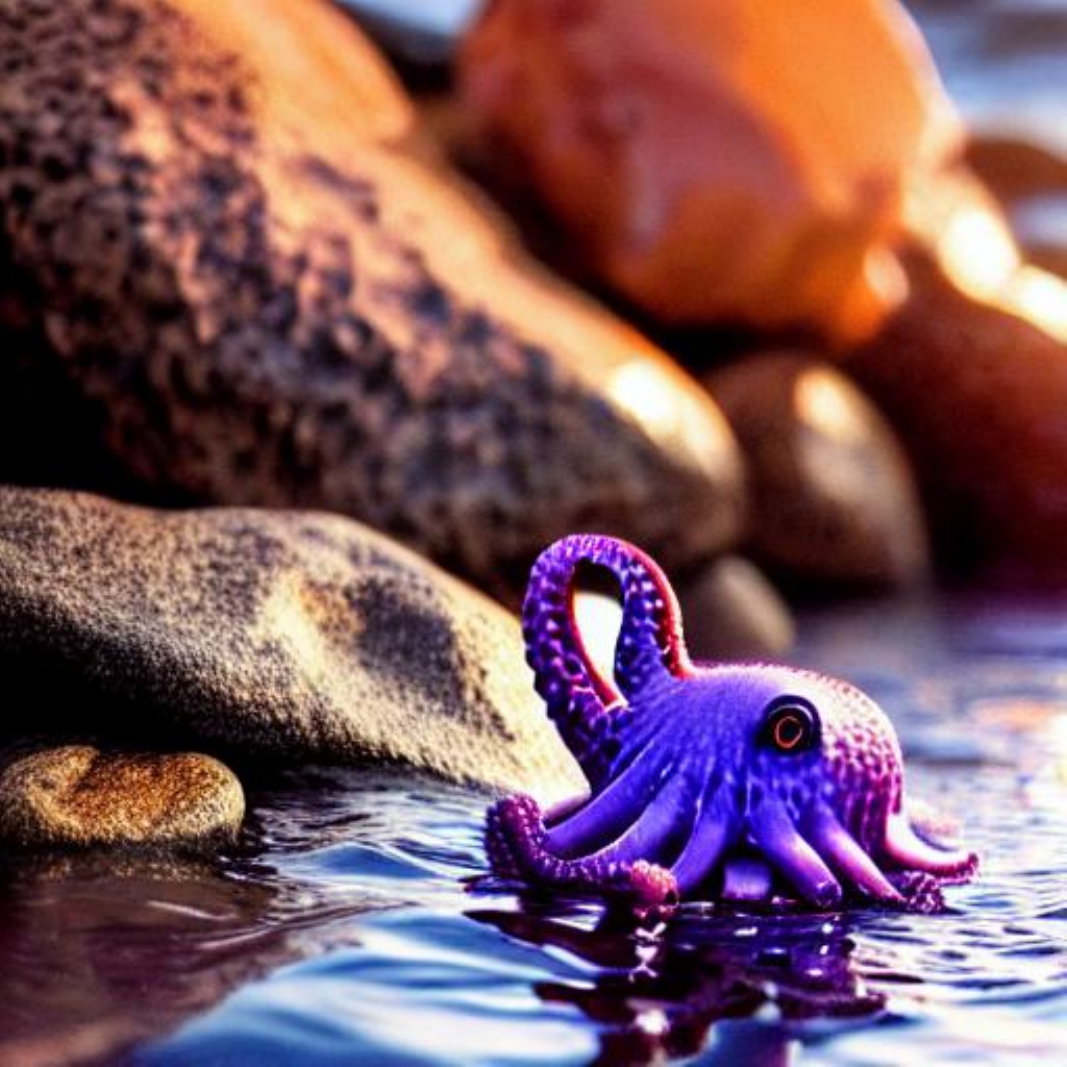}         \captionsetup{justification=centering}\caption{Diff-KTO~\cite{lialigning}} \end{subfigure}
    \begin{subfigure}[t]{0.19\textwidth} \includegraphics[width=\textwidth]{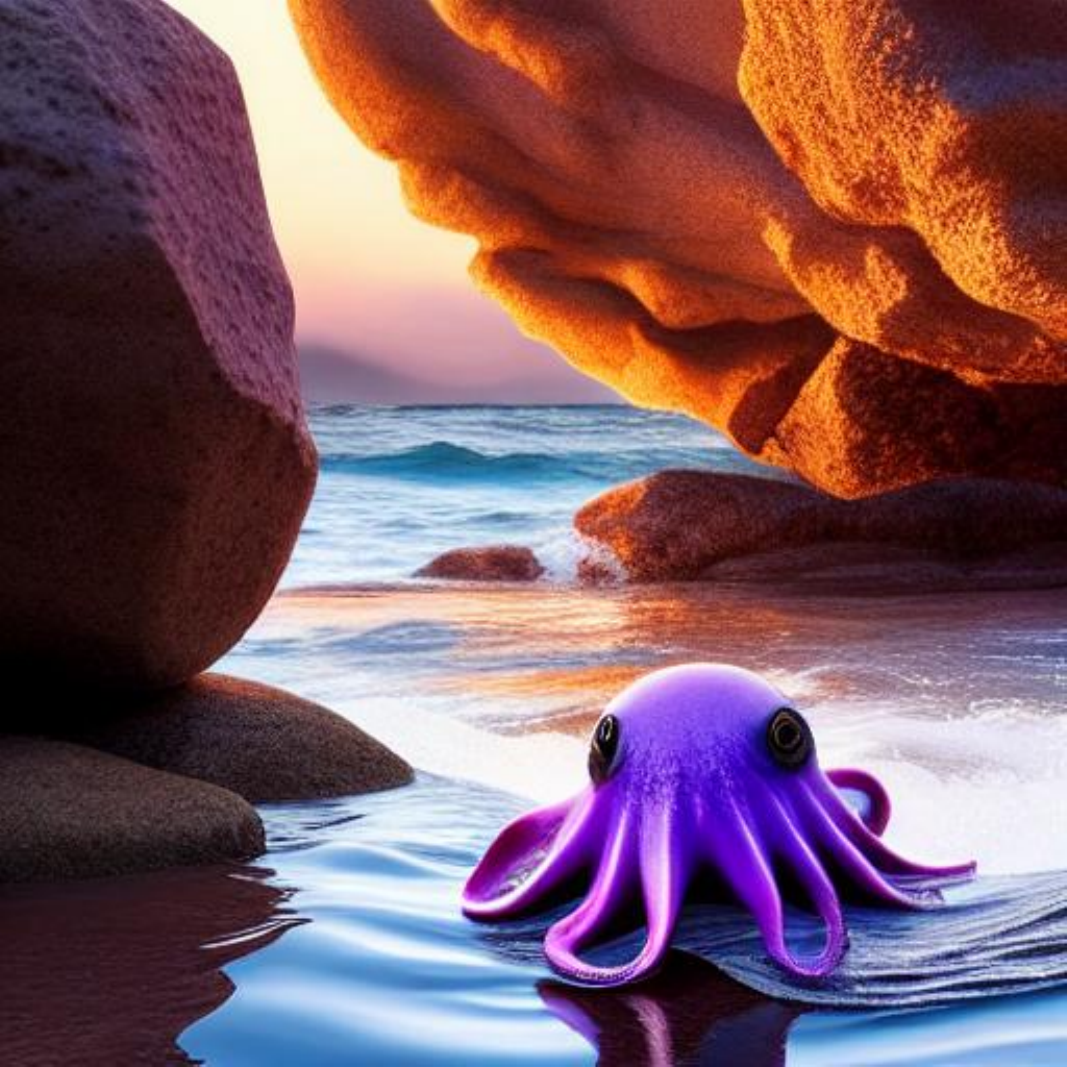}         \captionsetup{justification=centering}\caption{DSPO~\cite{zhu2025dspo}} \end{subfigure}
    \begin{subfigure}[t]{0.19\textwidth} \includegraphics[width=\textwidth]{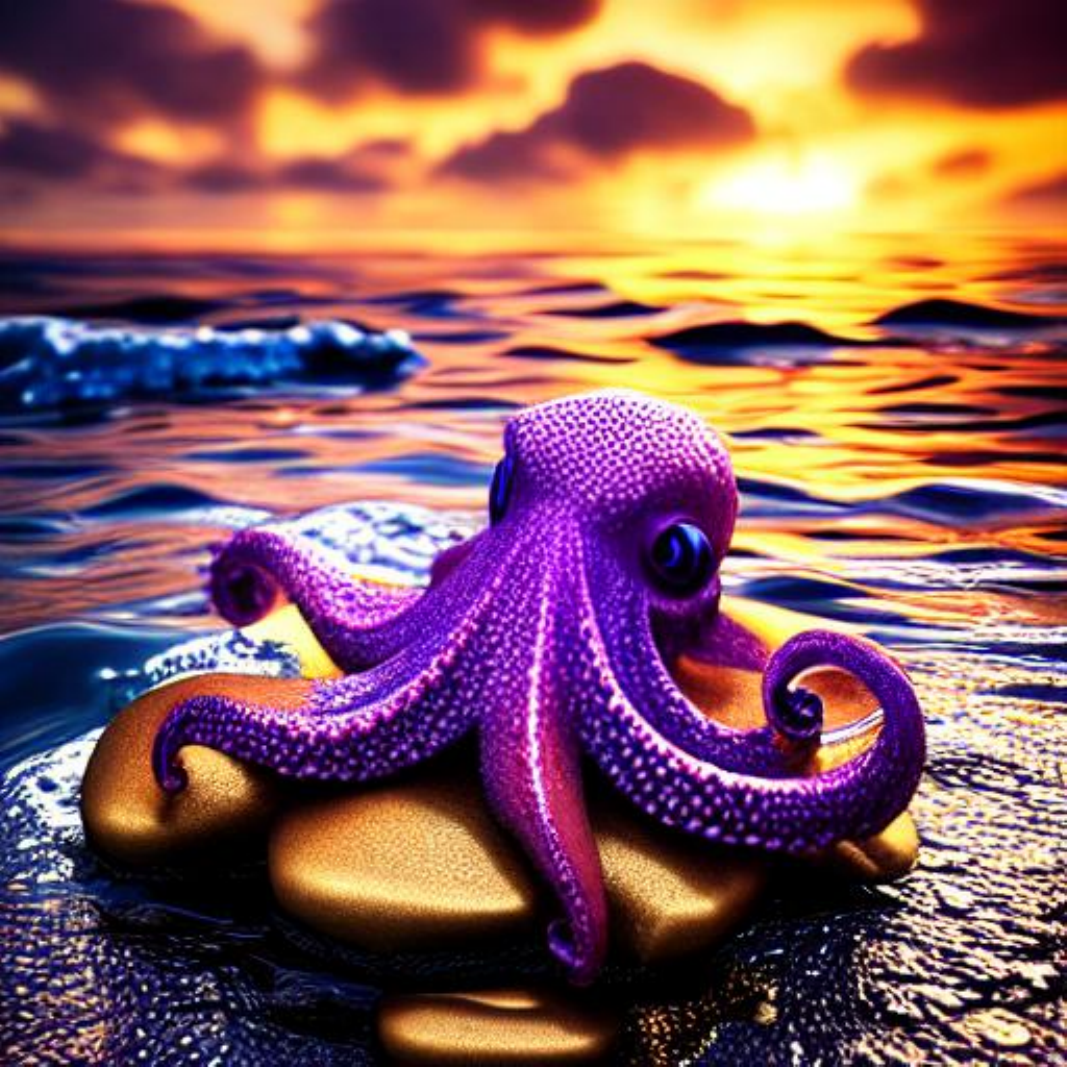}         \captionsetup{justification=centering}\caption{Ours} \end{subfigure}
    
    \caption{
        Qualitative comparisons on Pick-a-pic test set prompts.
    }
    \label{fig:appendix_qual_pick}
\end{figure*}

\begin{figure*}[hb!]
    \centering
    \parbox{1.02 \textwidth}{
        \centering
        A photograph of a portrait of a statue of a pharaoh wearing steampunk glasses, \\
            white t-shirt and leather jacket.
    }
    \begin{subfigure}[t]{0.19\textwidth} \includegraphics[width=\textwidth]{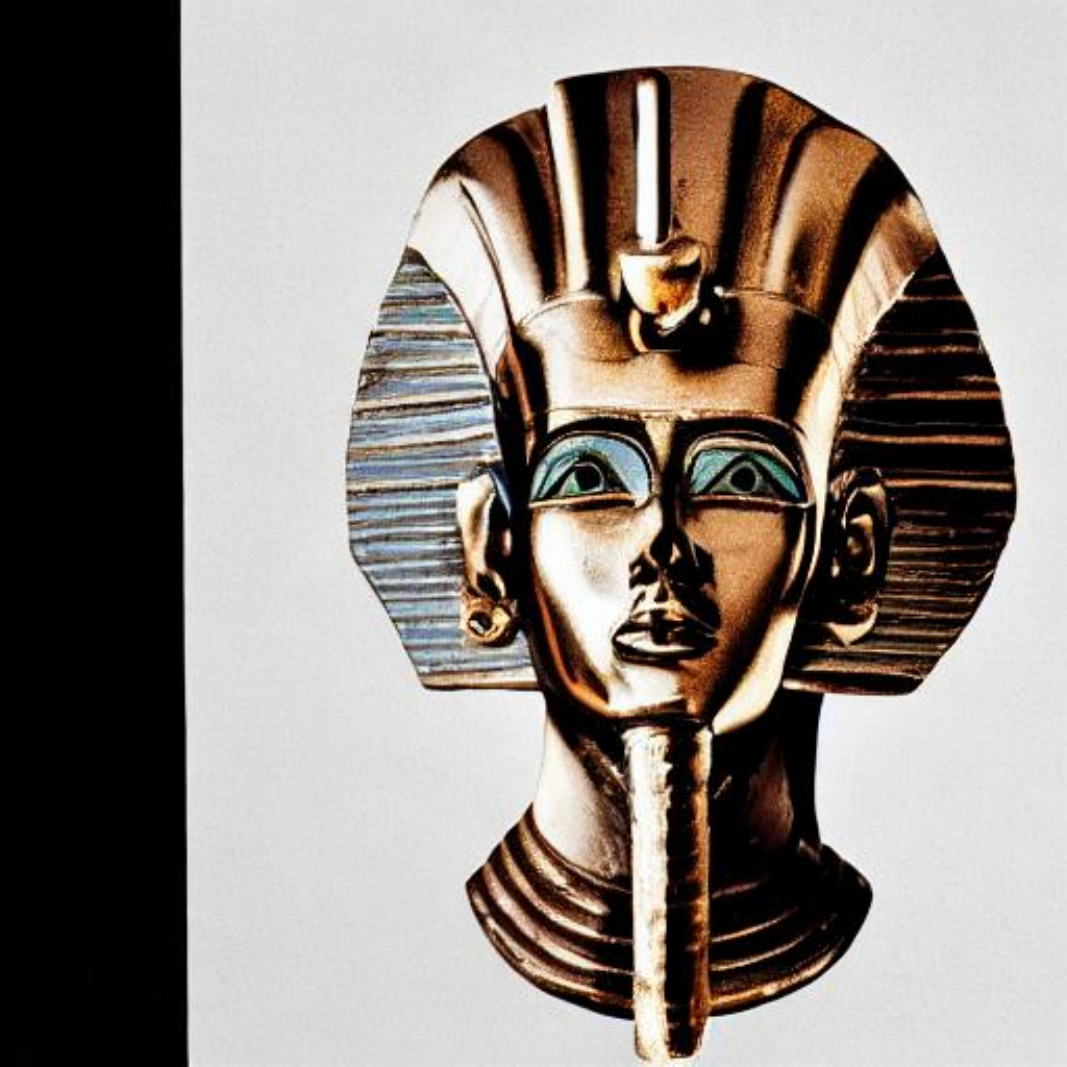} \end{subfigure}
    \begin{subfigure}[t]{0.19\textwidth} \includegraphics[width=\textwidth]{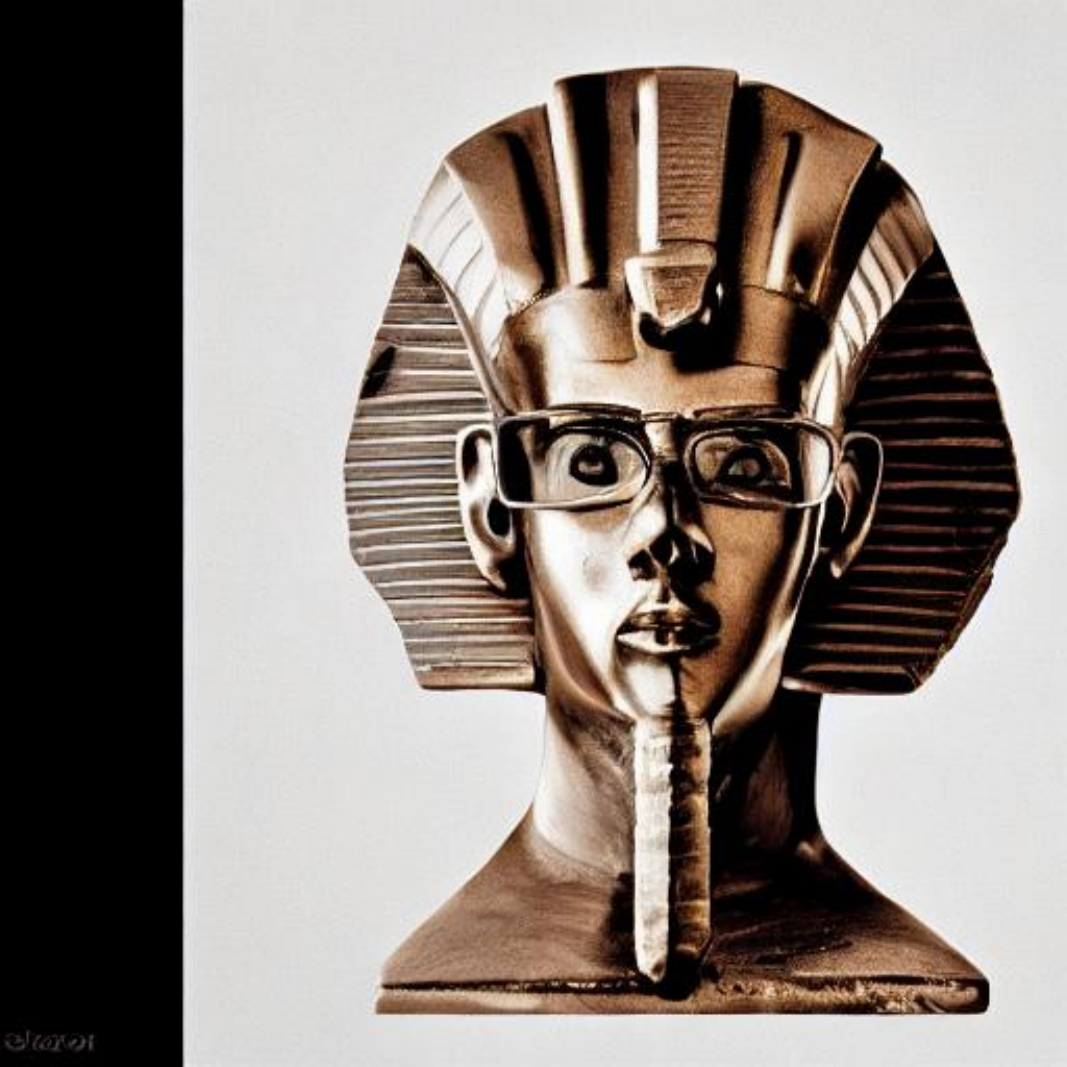} \end{subfigure}
    \begin{subfigure}[t]{0.19\textwidth} \includegraphics[width=\textwidth]{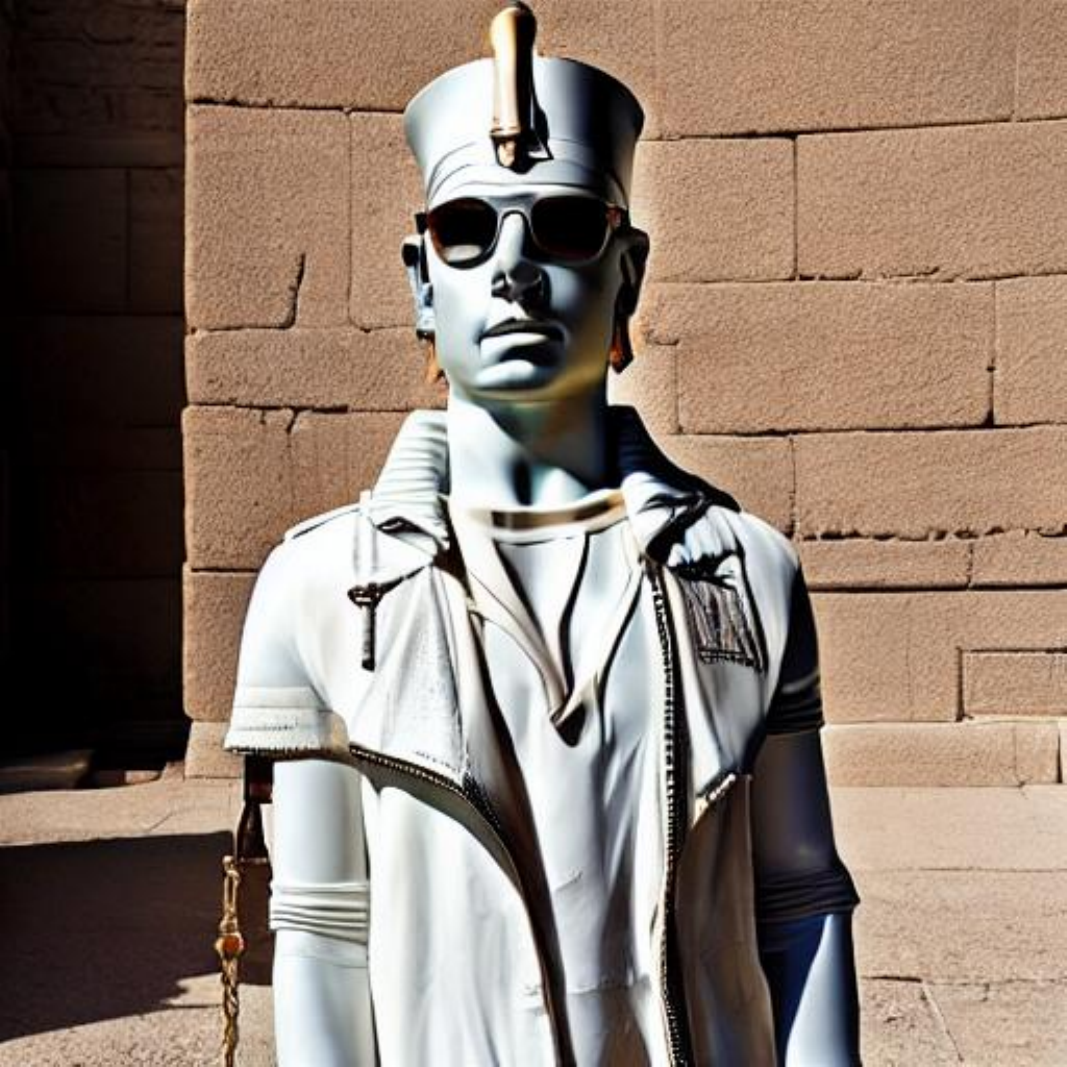} \end{subfigure}
    \begin{subfigure}[t]{0.19\textwidth} \includegraphics[width=\textwidth]{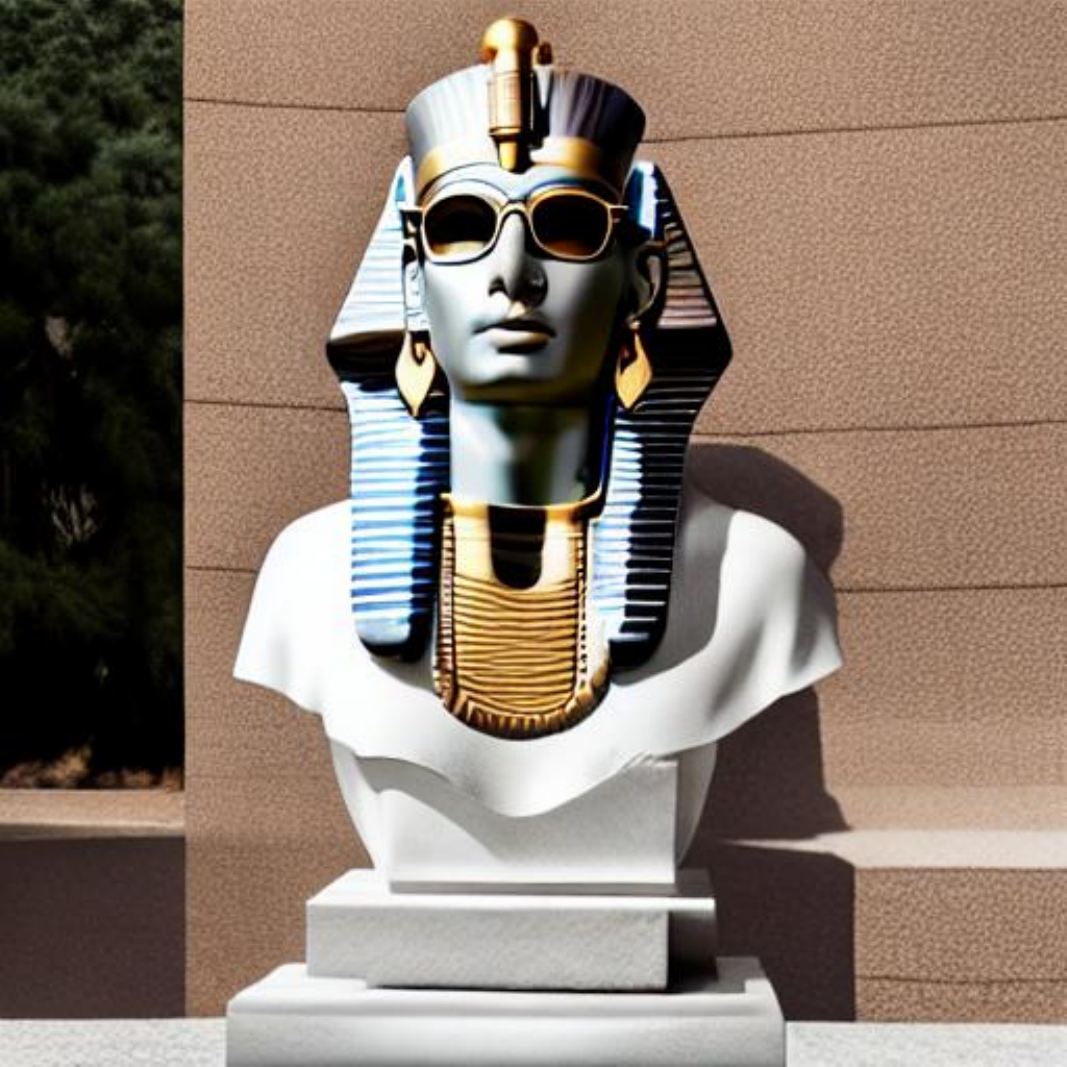} \end{subfigure}
    \begin{subfigure}[t]{0.19\textwidth} \includegraphics[width=\textwidth]{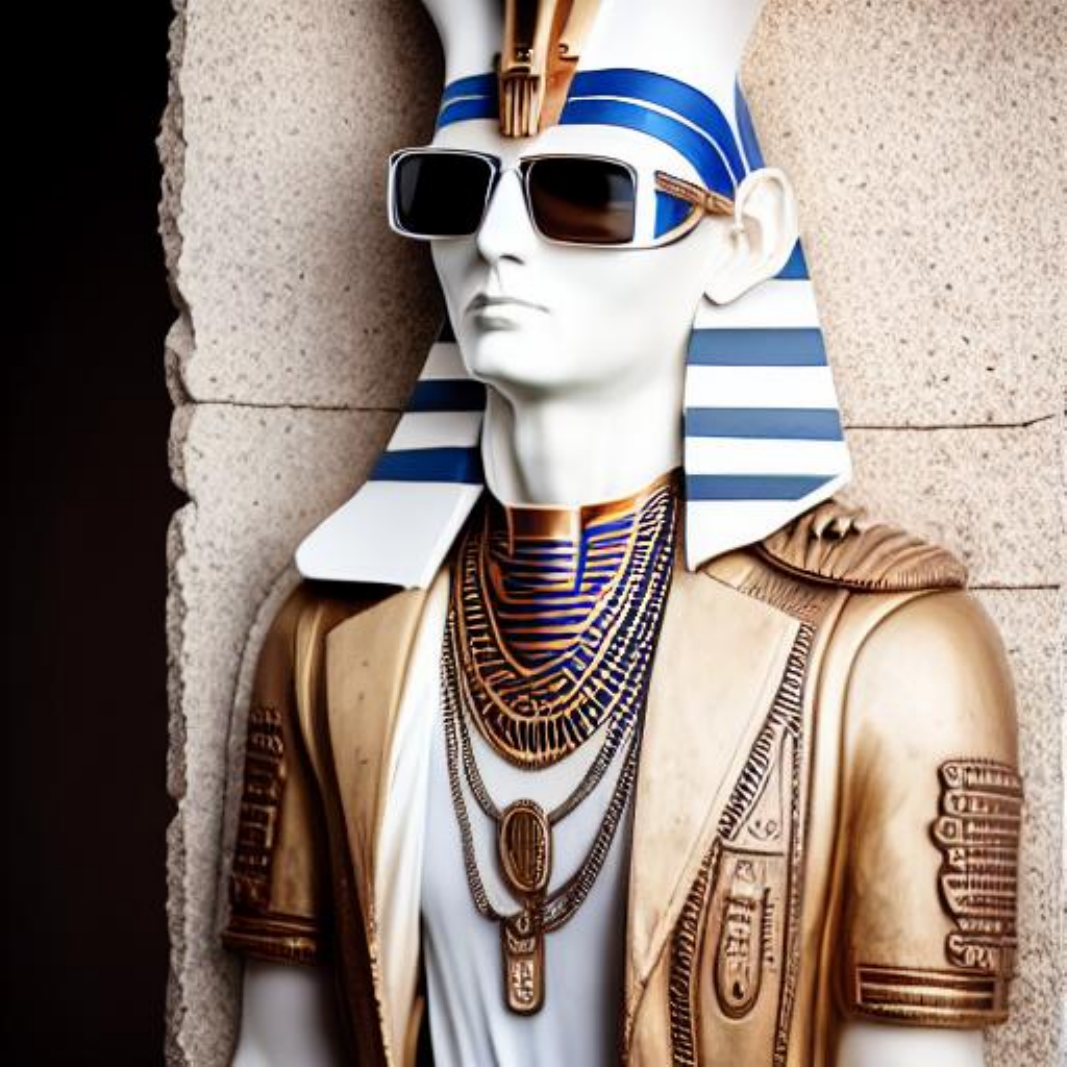} \end{subfigure}

    \parbox{1.02 \textwidth}{
        \centering
        a real flamingo reading a large open book. a big stack of books is piled up next to it. dslr photograph.
    }
    \begin{subfigure}[t]{0.19\textwidth} \includegraphics[width=\textwidth]{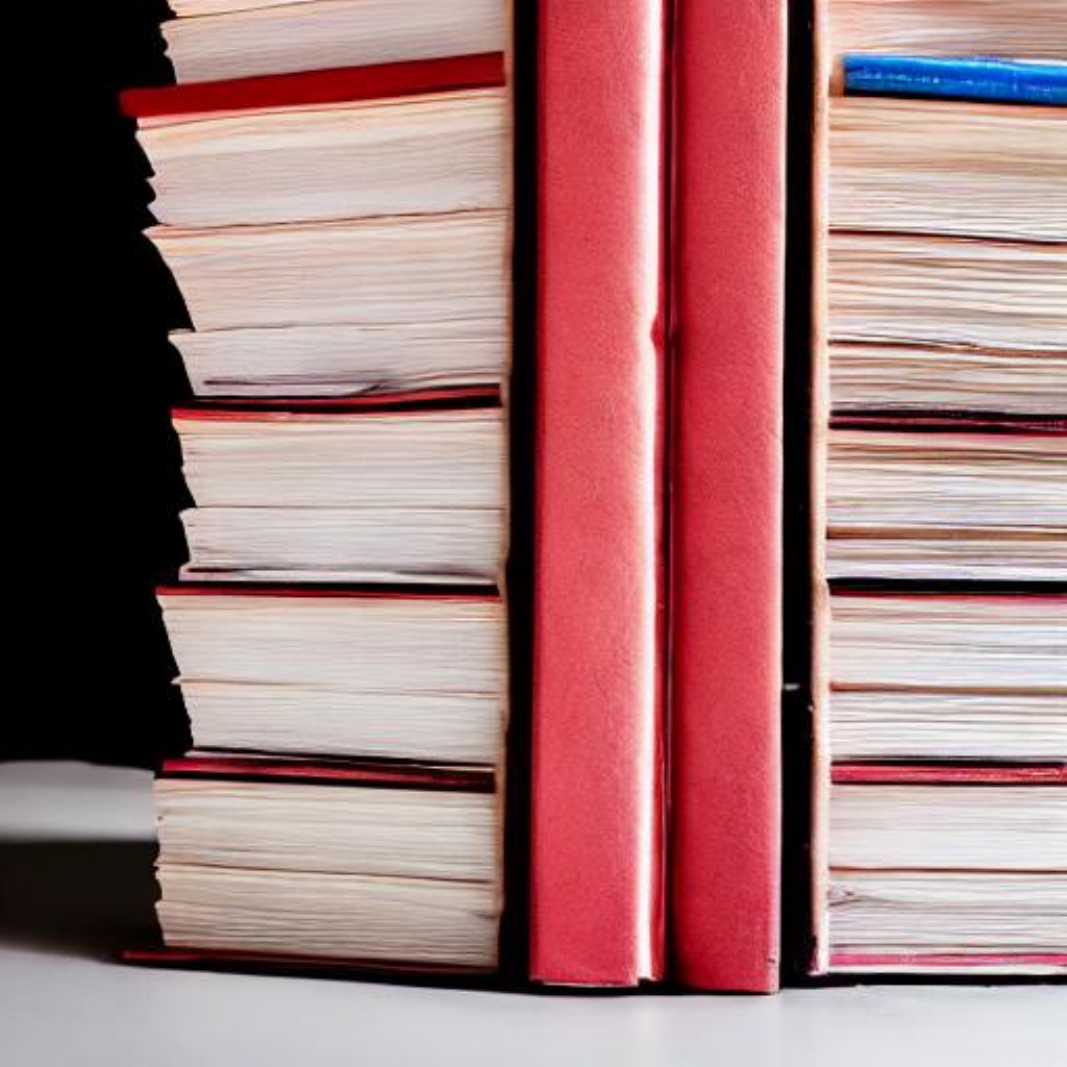} \end{subfigure}
    \begin{subfigure}[t]{0.19\textwidth} \includegraphics[width=\textwidth]{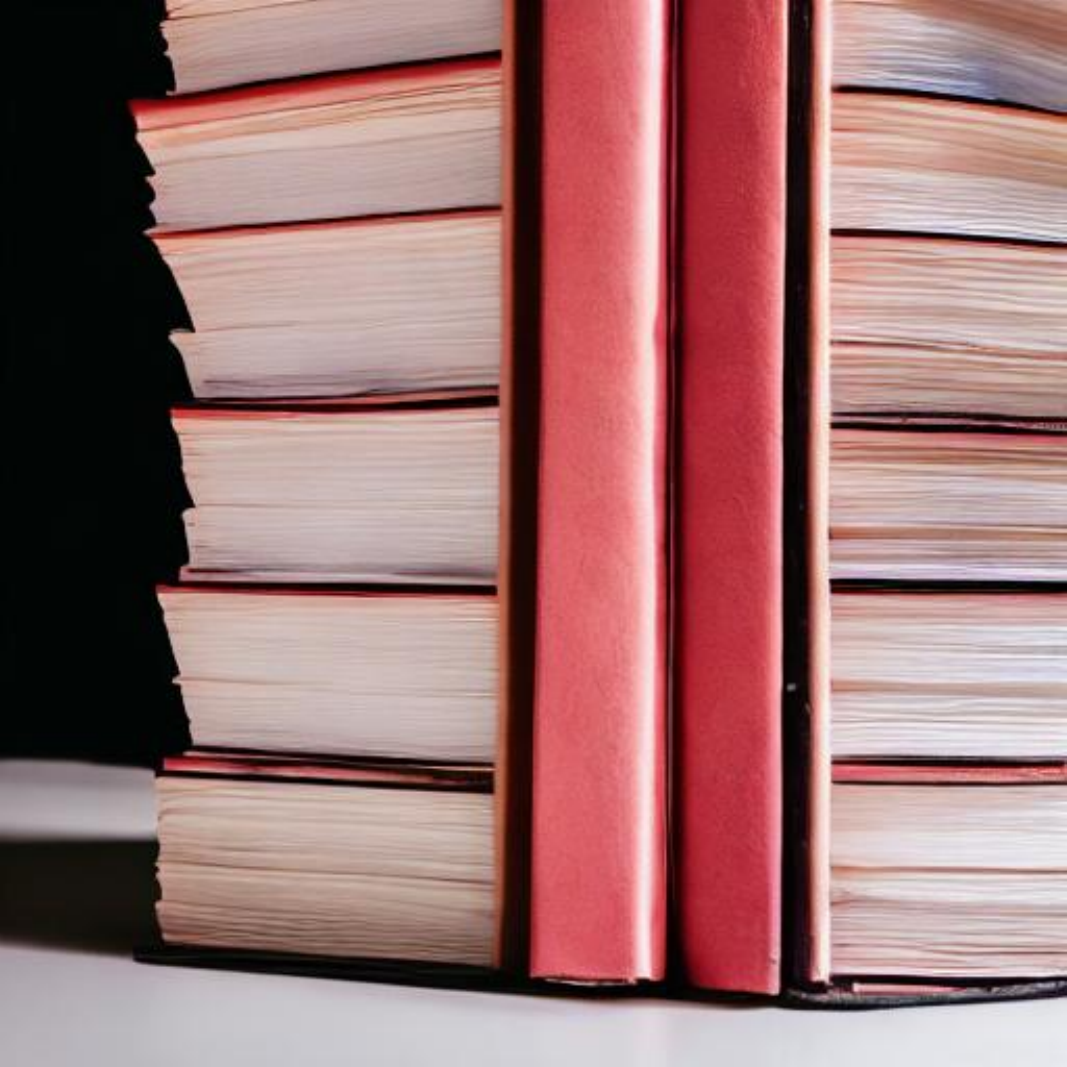} \end{subfigure}
    \begin{subfigure}[t]{0.19\textwidth} \includegraphics[width=\textwidth]{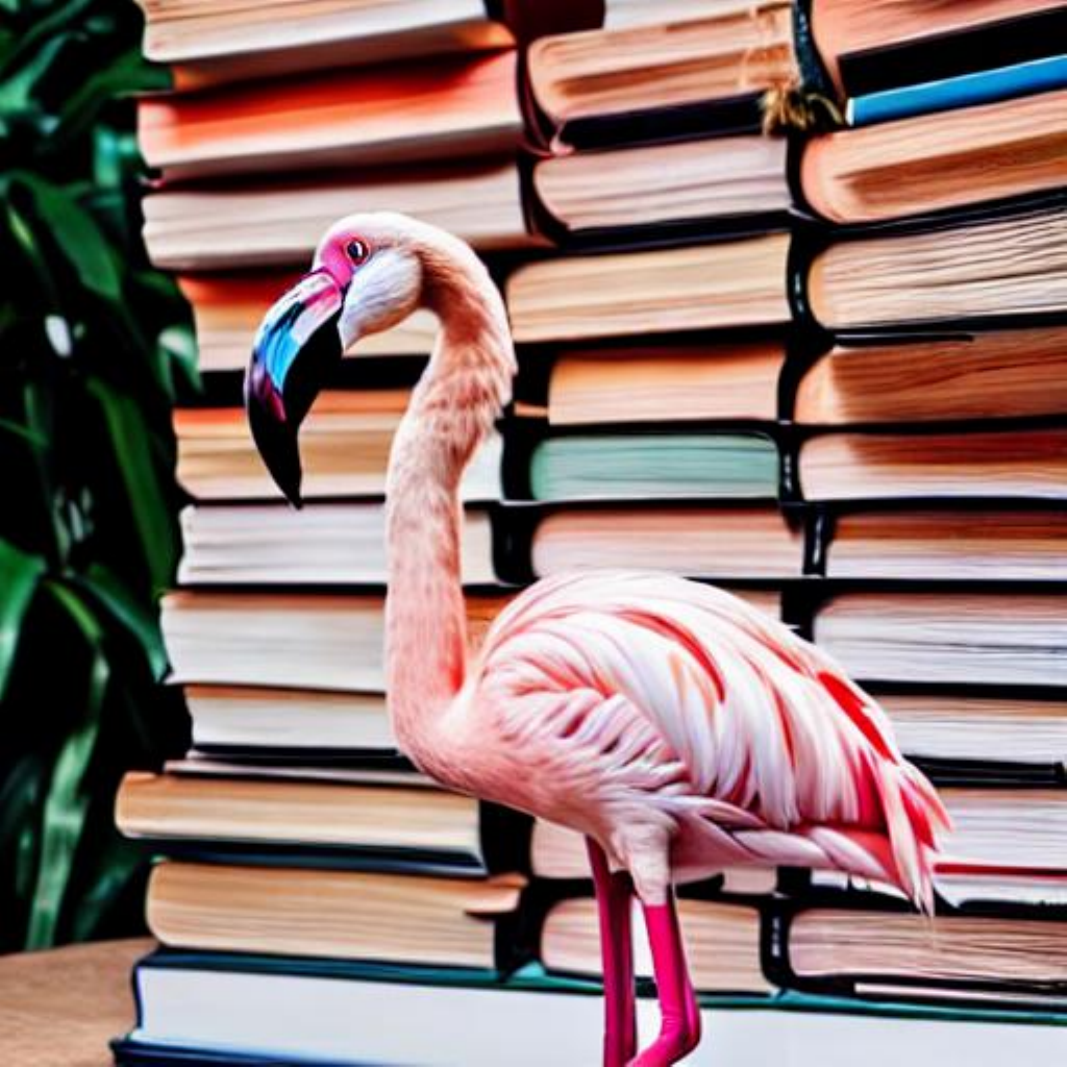} \end{subfigure}
    \begin{subfigure}[t]{0.19\textwidth} \includegraphics[width=\textwidth]{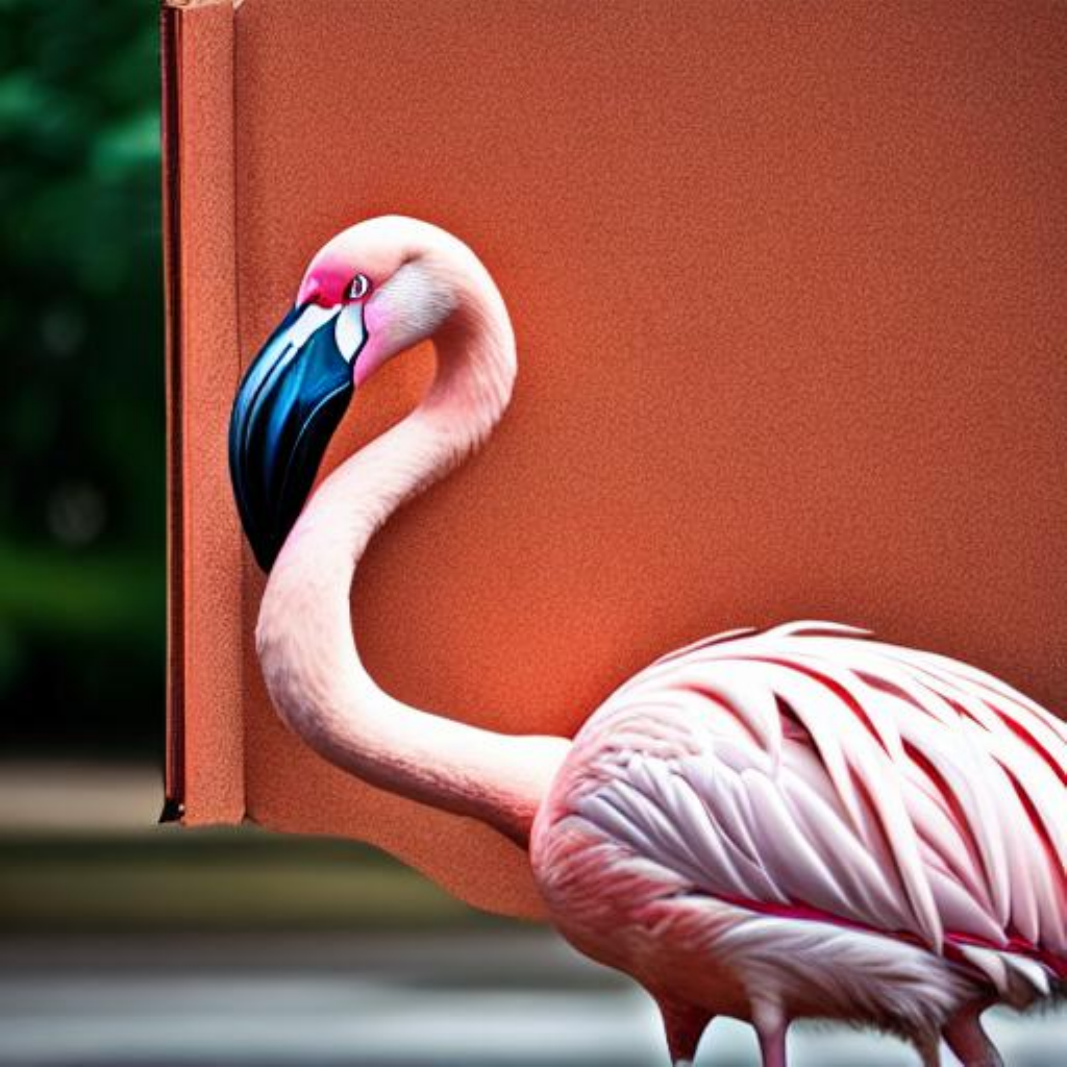} \end{subfigure}
    \begin{subfigure}[t]{0.19\textwidth} \includegraphics[width=\textwidth]{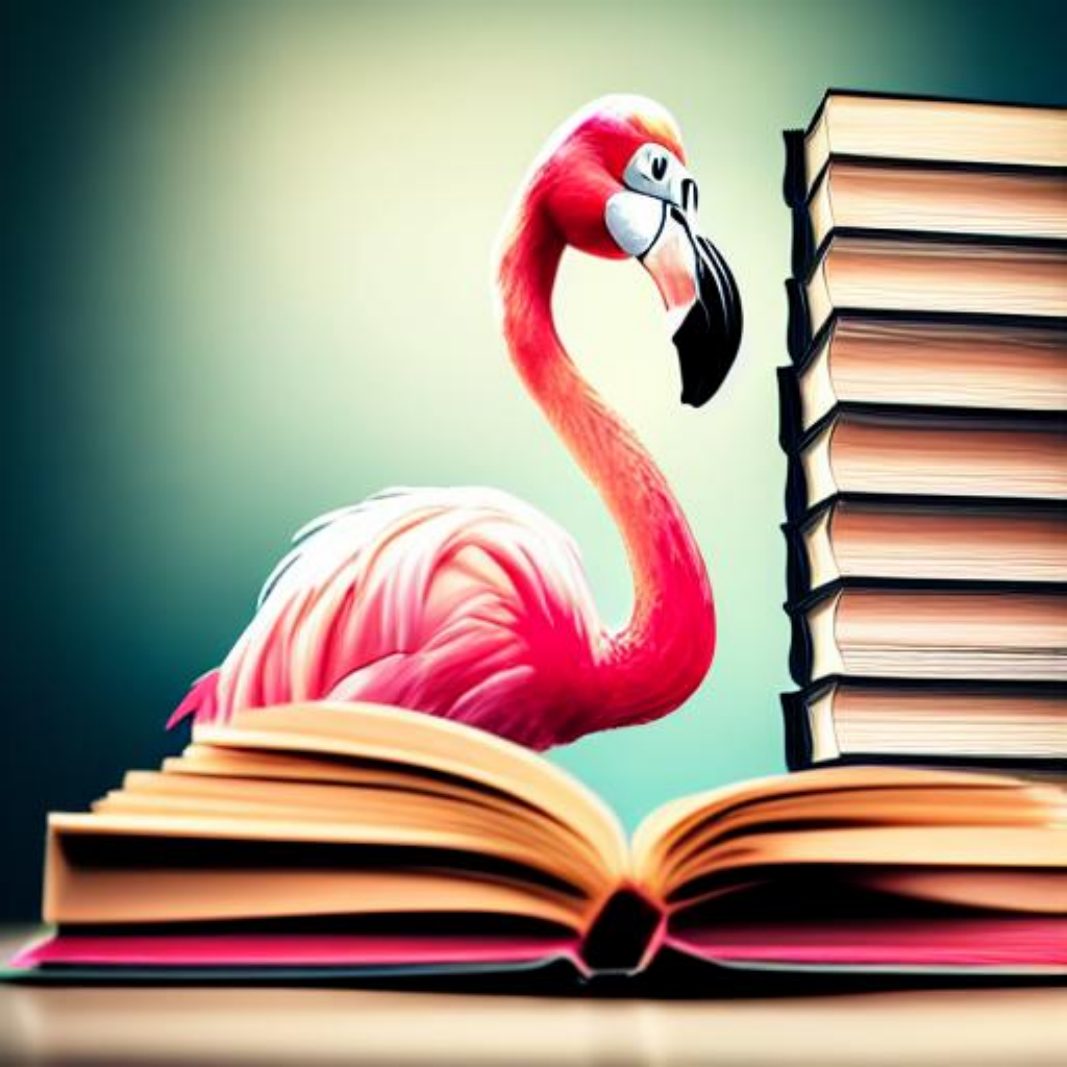} \end{subfigure}

    \parbox{1.02 \textwidth}{
        \centering
         A cozy living room with a painting of a corgi on the wall above a couch \\
            and a round coffee table in front of a couch and a vase of flowers on a coffee table
    }
    \begin{subfigure}[t]{0.19\textwidth} \includegraphics[width=\textwidth]{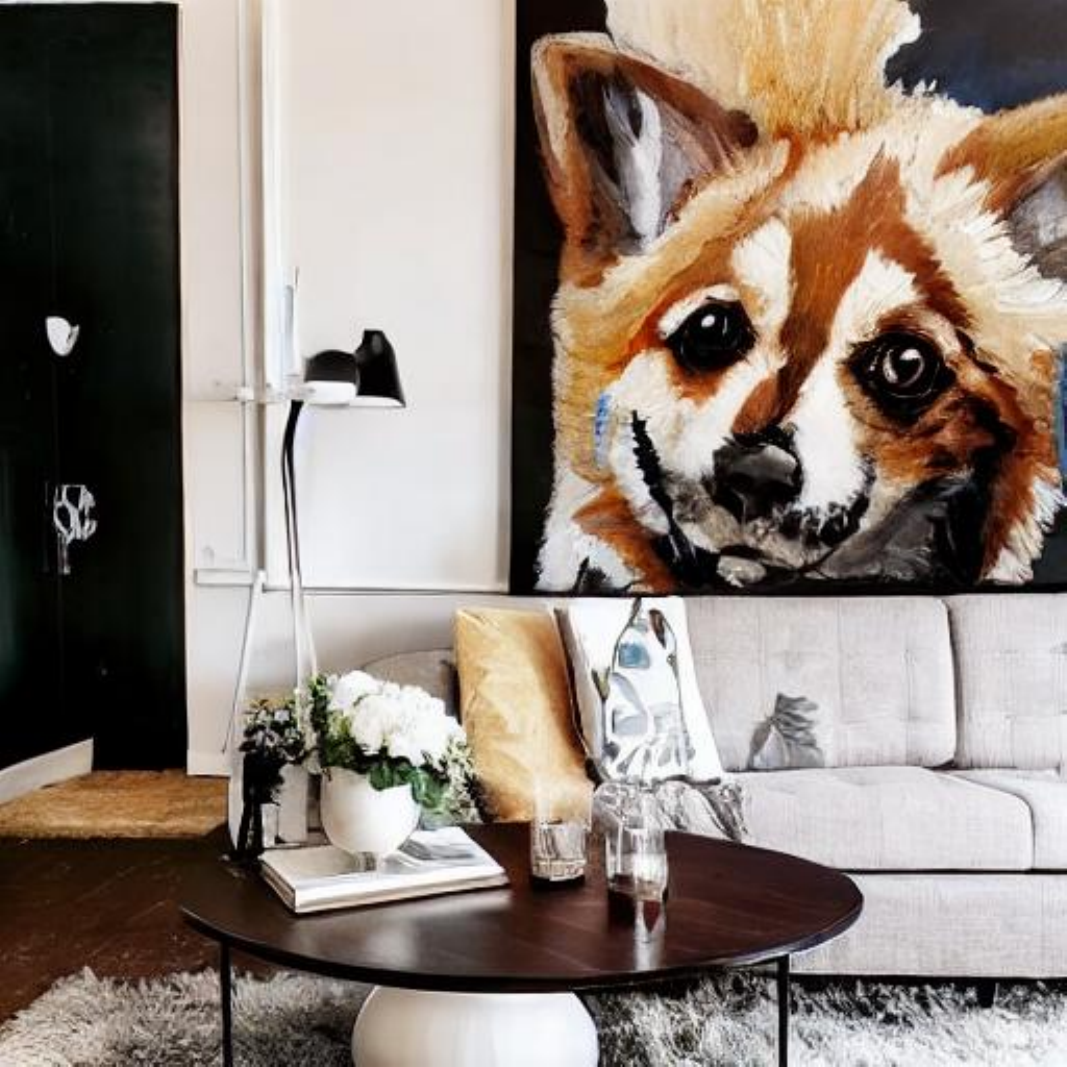} \end{subfigure}
    \begin{subfigure}[t]{0.19\textwidth} \includegraphics[width=\textwidth]{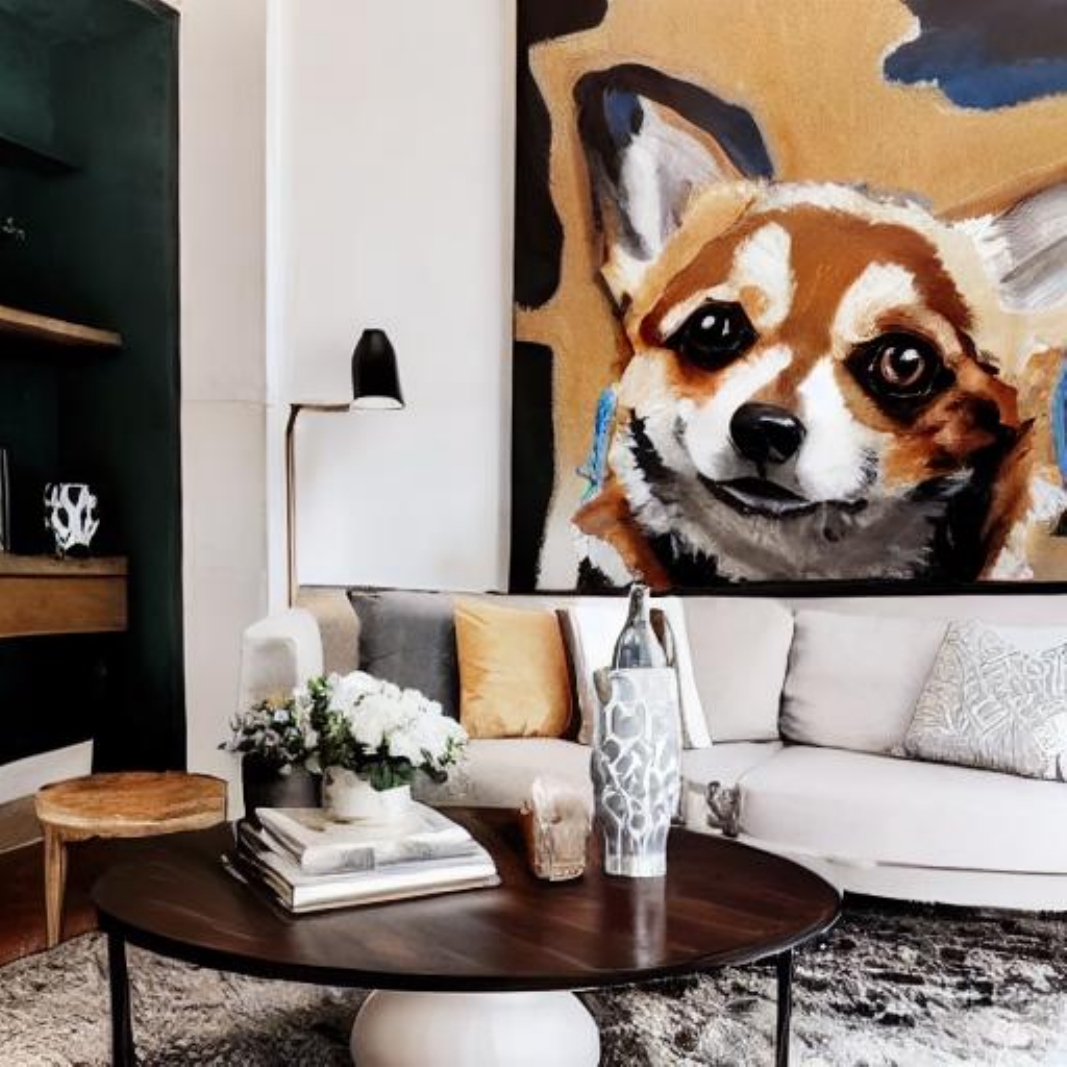} \end{subfigure}
    \begin{subfigure}[t]{0.19\textwidth} \includegraphics[width=\textwidth]{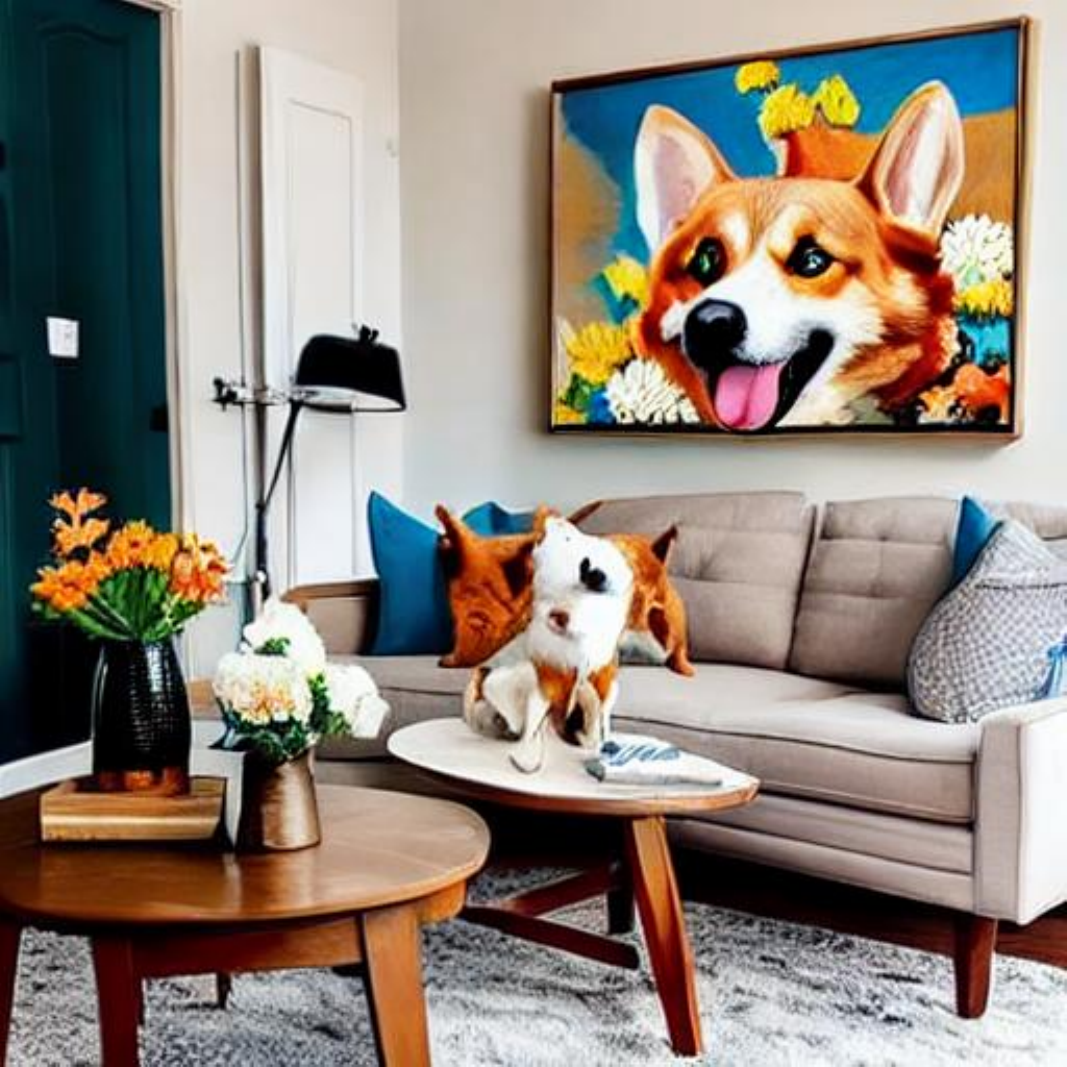} \end{subfigure}
    \begin{subfigure}[t]{0.19\textwidth} \includegraphics[width=\textwidth]{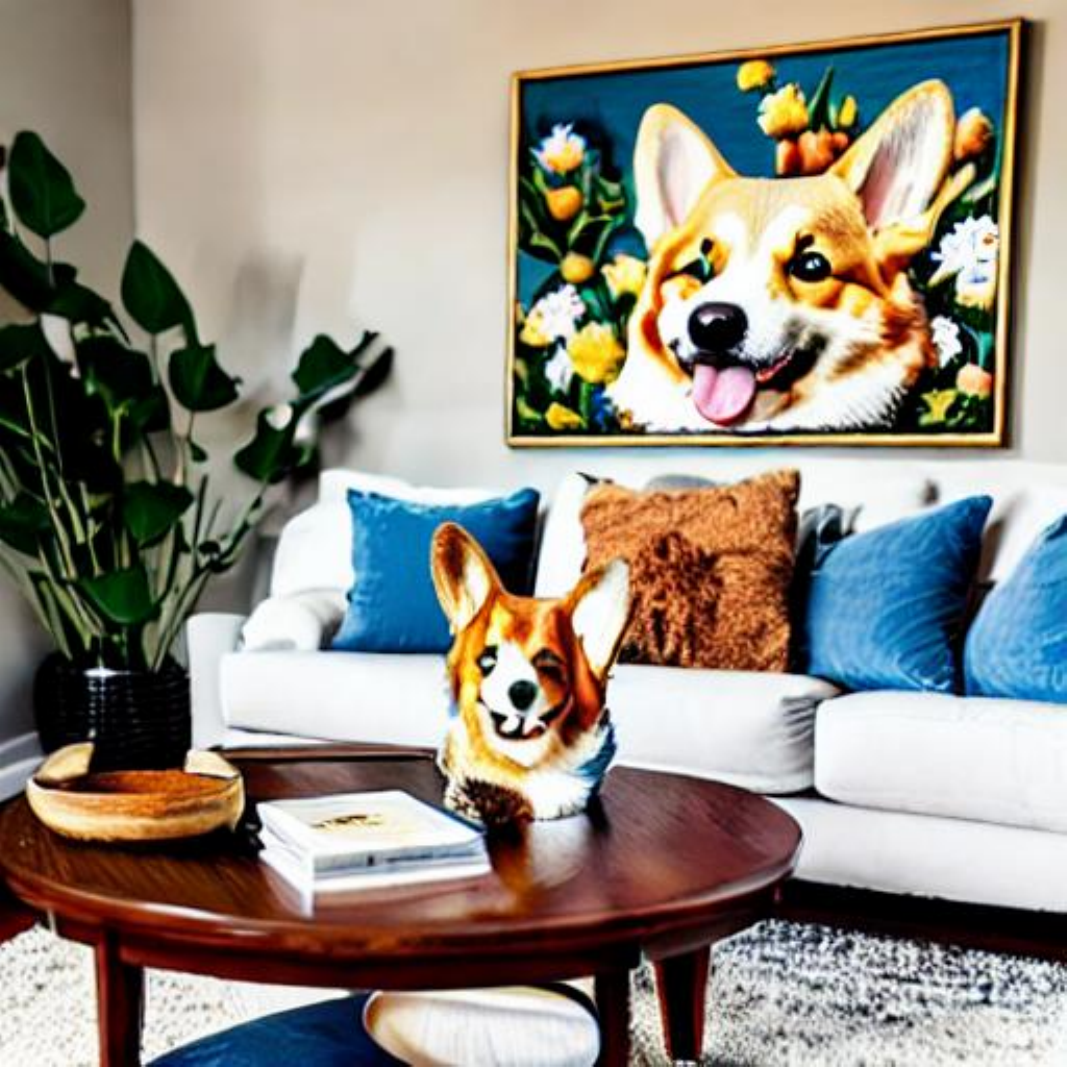} \end{subfigure}
    \begin{subfigure}[t]{0.19\textwidth} \includegraphics[width=\textwidth]{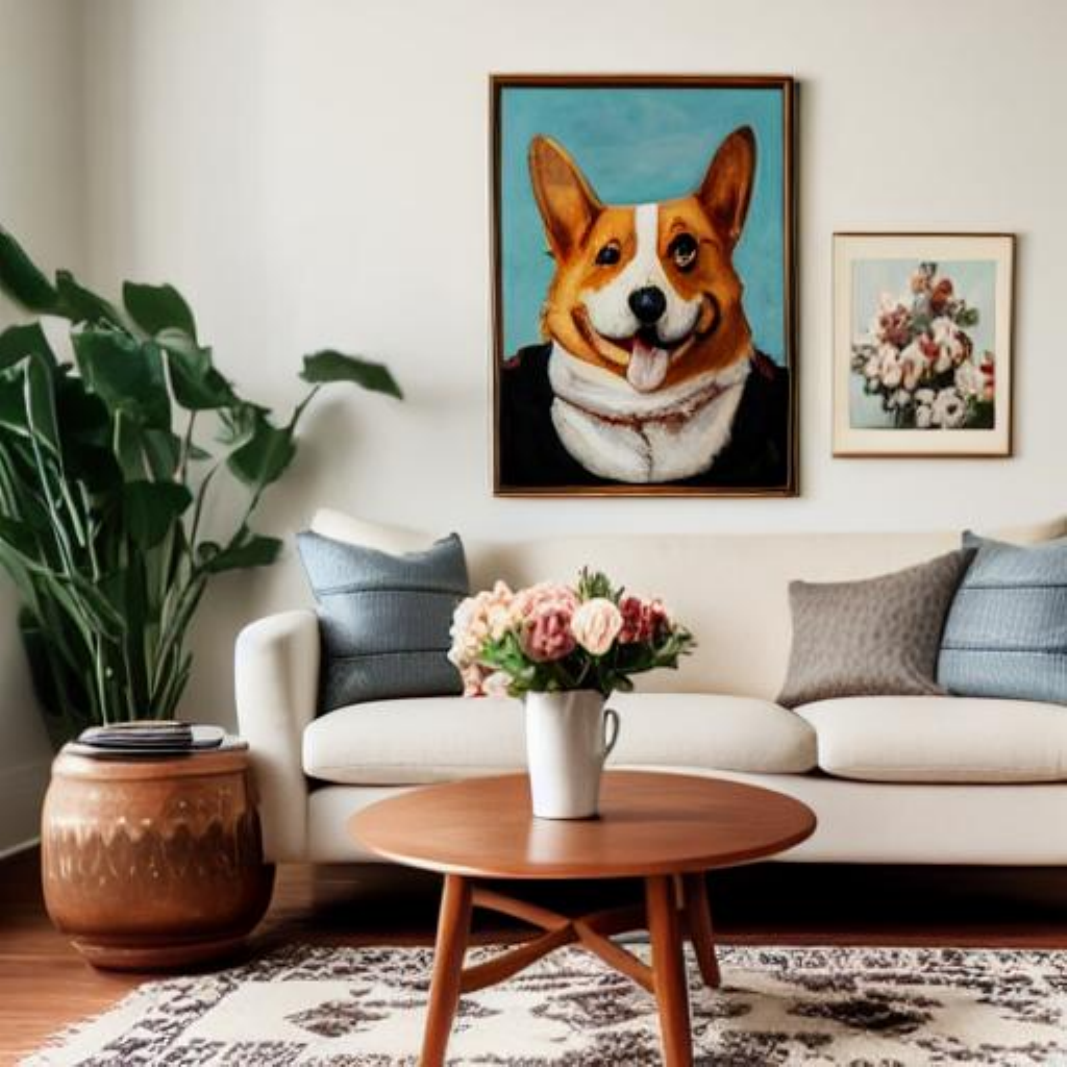} \end{subfigure}

    \parbox{1.02 \textwidth}{
        \centering
      a tree reflected in the hood of a blue car
    }
    \begin{subfigure}[t]{0.19\textwidth} \includegraphics[width=\textwidth]{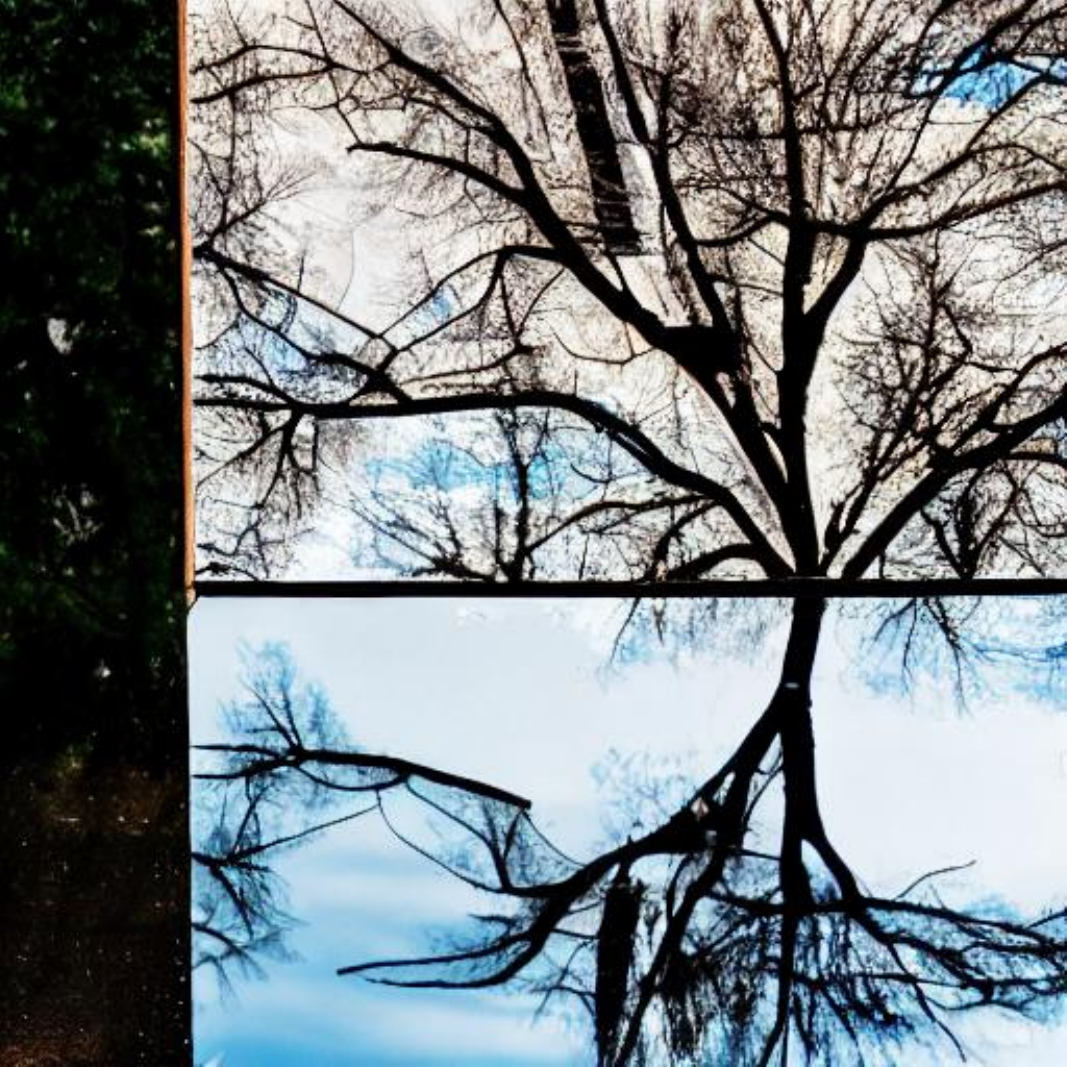} \end{subfigure}
    \begin{subfigure}[t]{0.19\textwidth} \includegraphics[width=\textwidth]{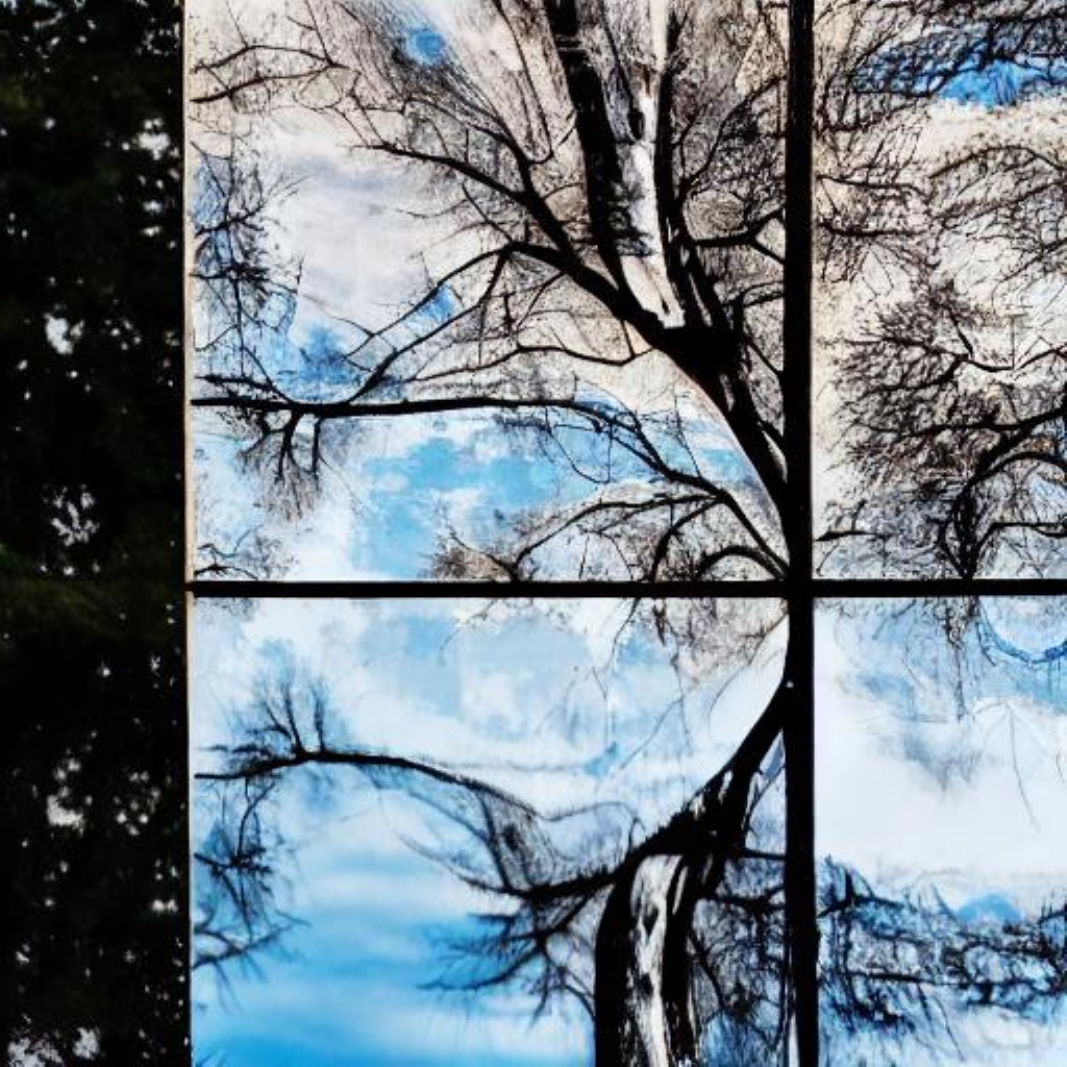} \end{subfigure}
    \begin{subfigure}[t]{0.19\textwidth} \includegraphics[width=\textwidth]{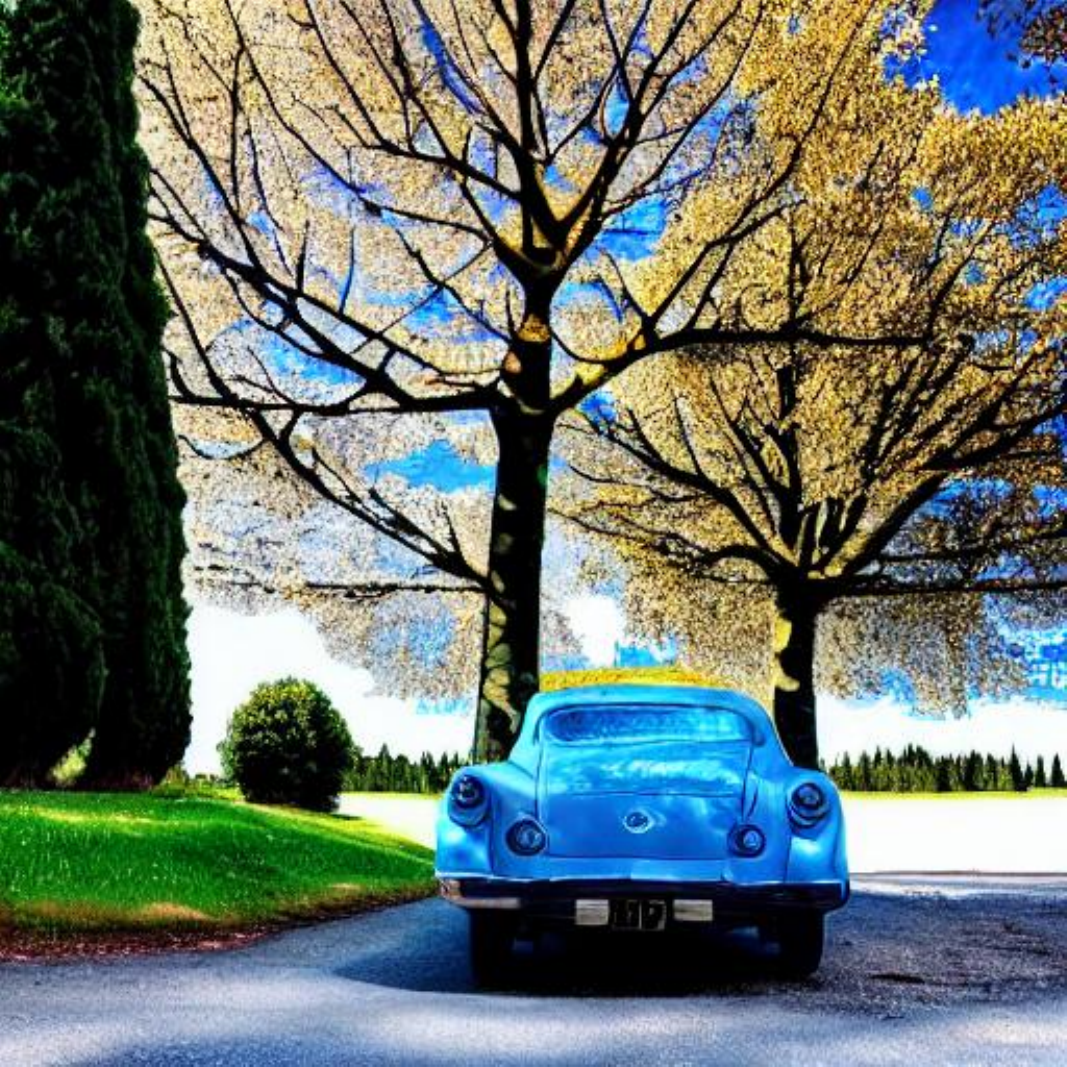} \end{subfigure}
    \begin{subfigure}[t]{0.19\textwidth} \includegraphics[width=\textwidth]{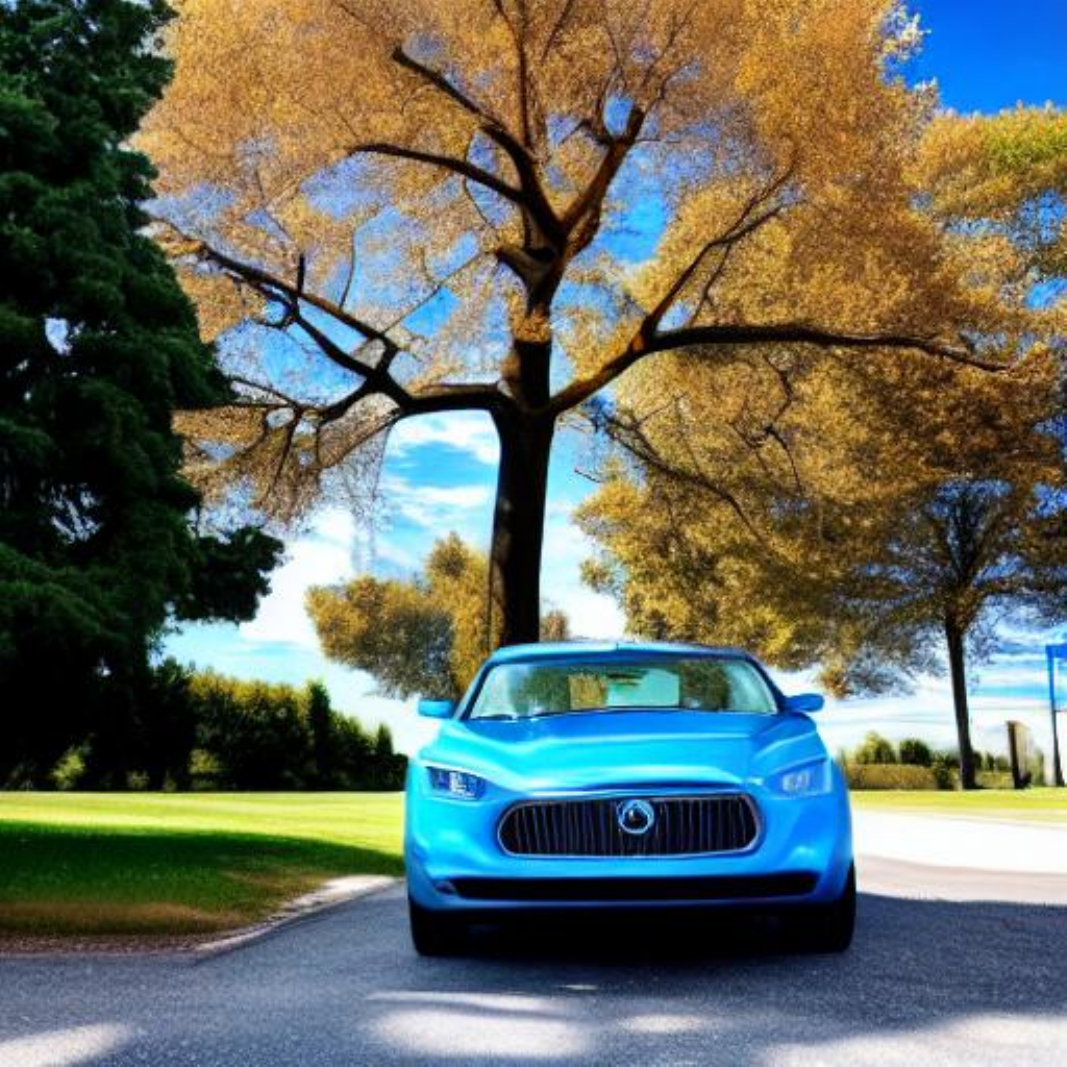} \end{subfigure}
    \begin{subfigure}[t]{0.19\textwidth} \includegraphics[width=\textwidth]{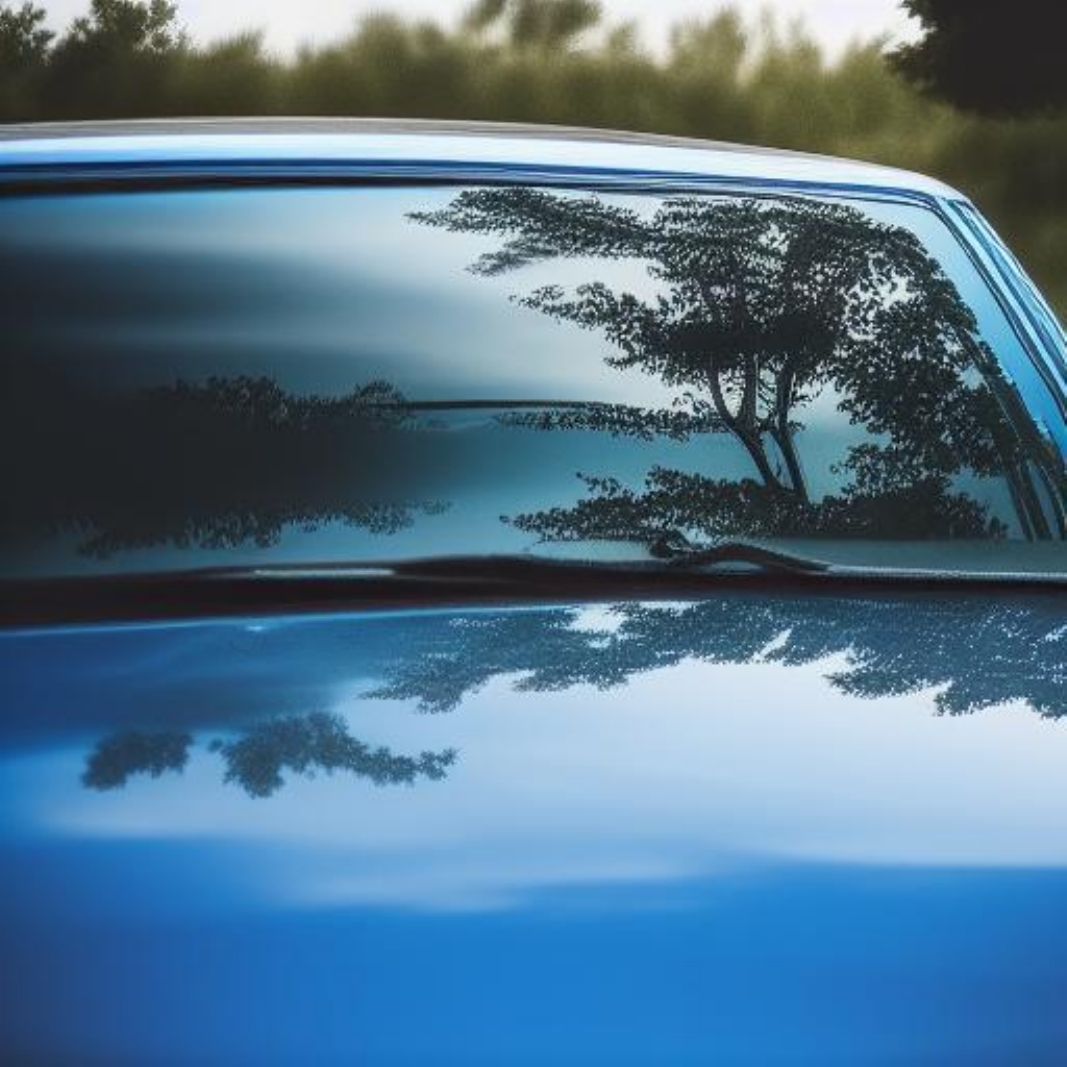} \end{subfigure}

    \parbox{1.02 \textwidth}{
        \centering
        a black dog jumping up to hug a woman wearing a red sweater
        }
    \begin{subfigure}[t]{0.19\textwidth} \includegraphics[width=\textwidth]{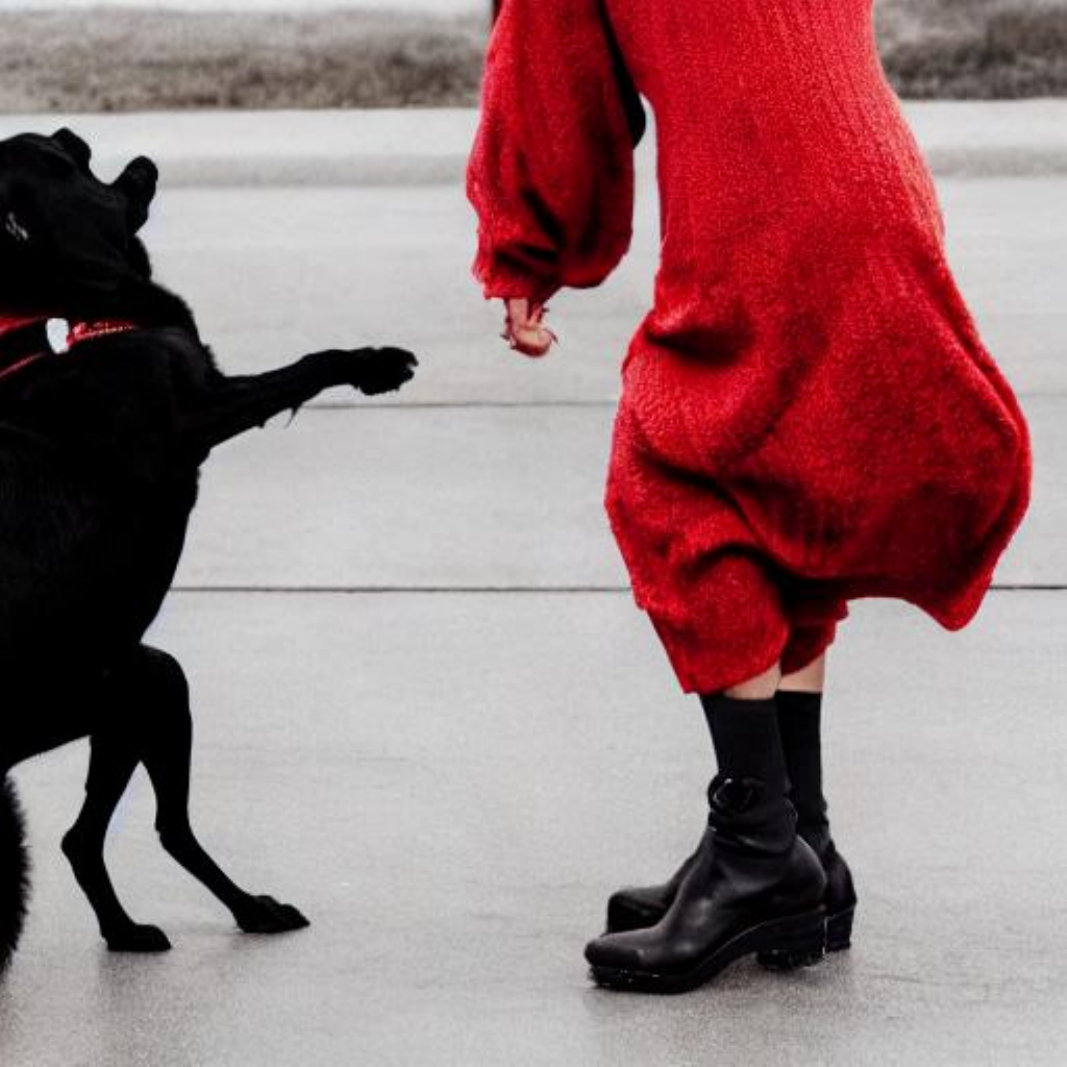} \end{subfigure}
    \begin{subfigure}[t]{0.19\textwidth} \includegraphics[width=\textwidth]{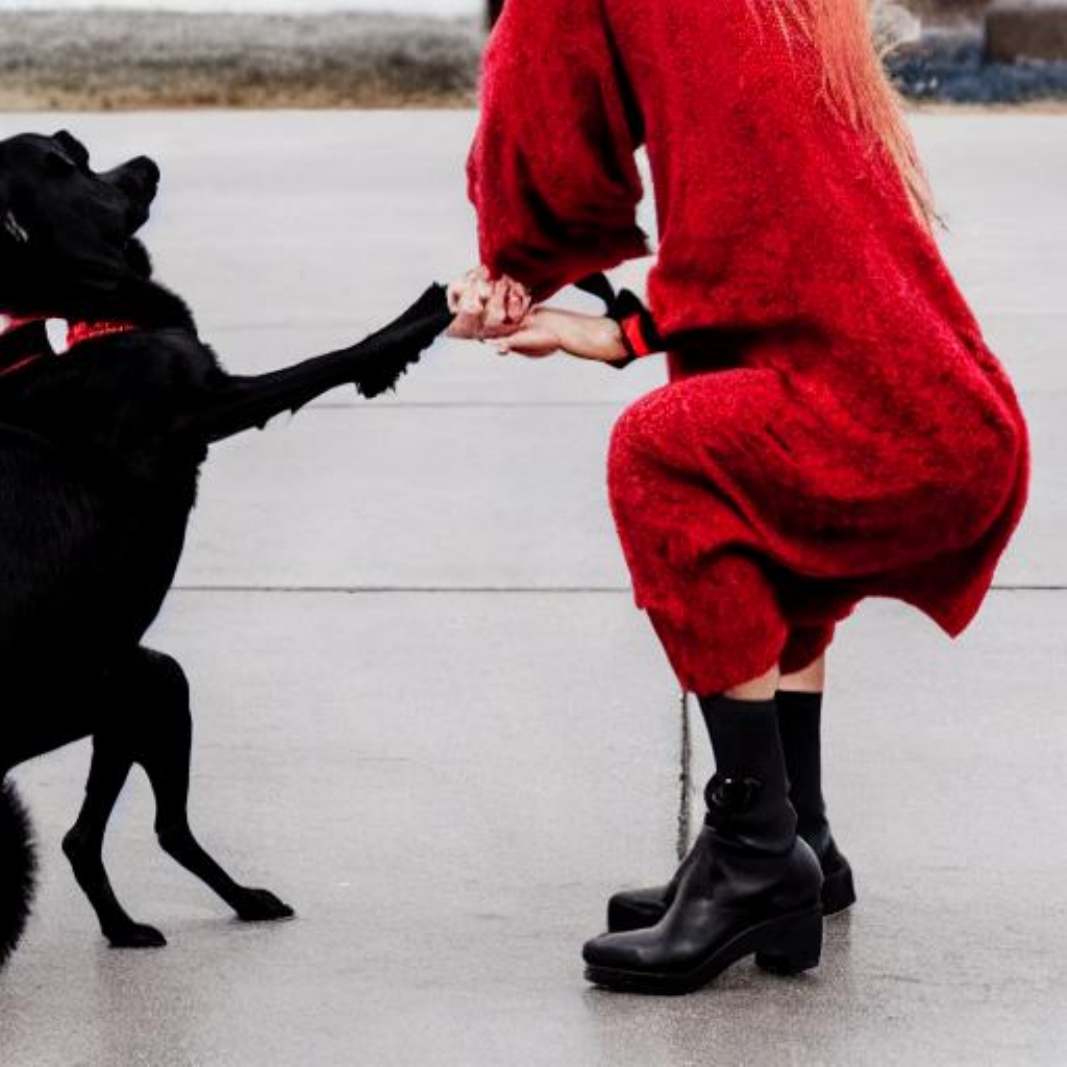} \end{subfigure}
    \begin{subfigure}[t]{0.19\textwidth} \includegraphics[width=\textwidth]{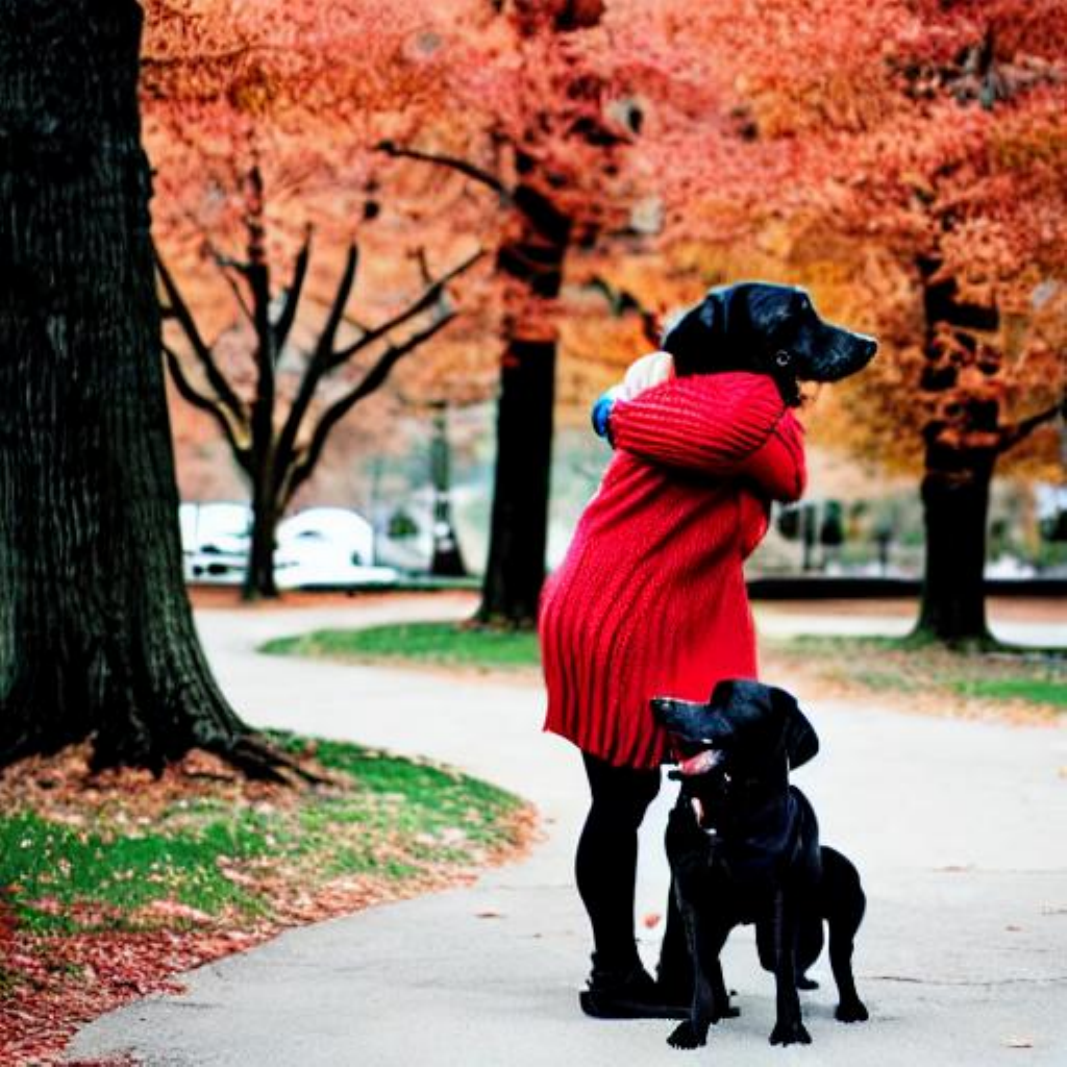} \end{subfigure}
    \begin{subfigure}[t]{0.19\textwidth} \includegraphics[width=\textwidth]{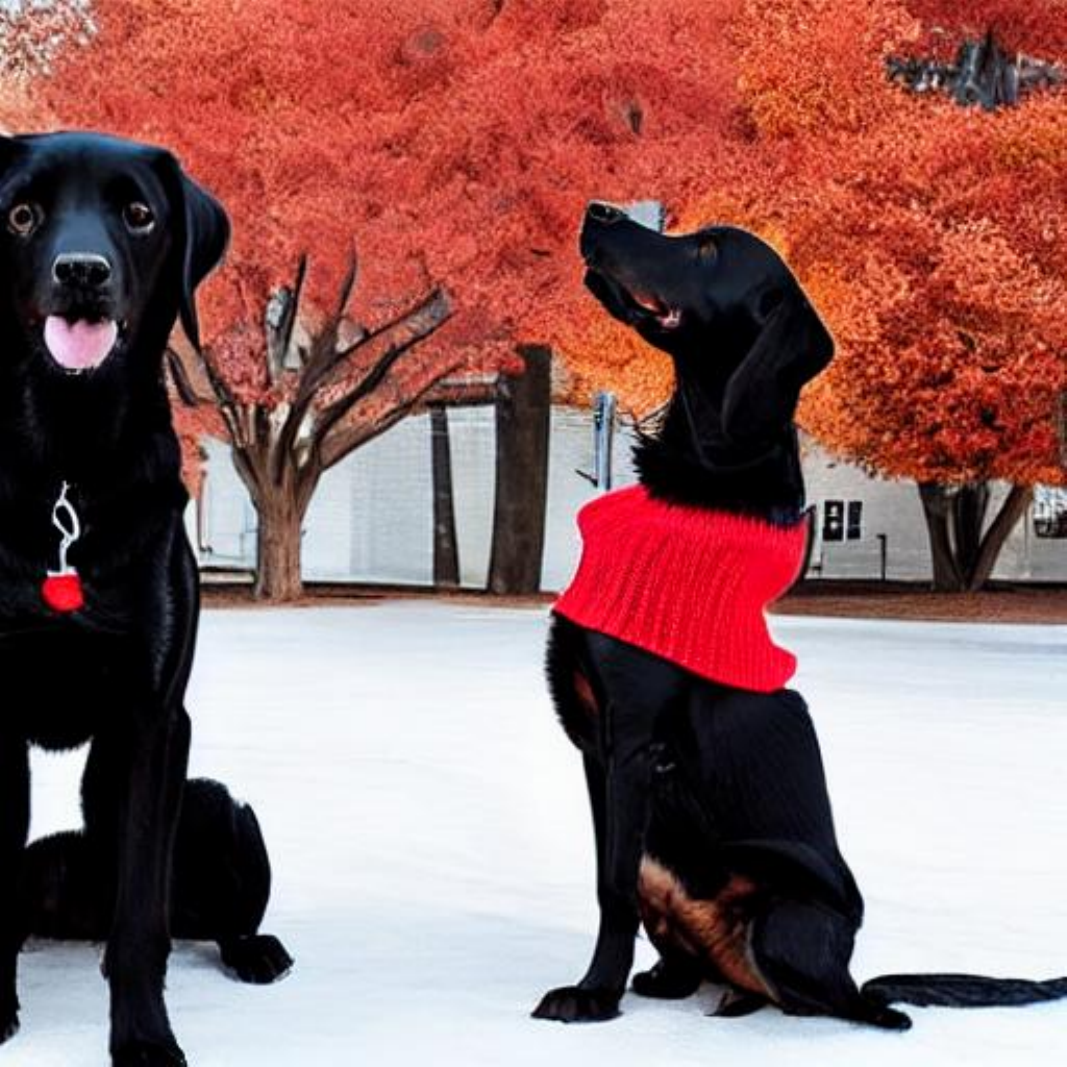} \end{subfigure}
    \begin{subfigure}[t]{0.19\textwidth} \includegraphics[width=\textwidth]{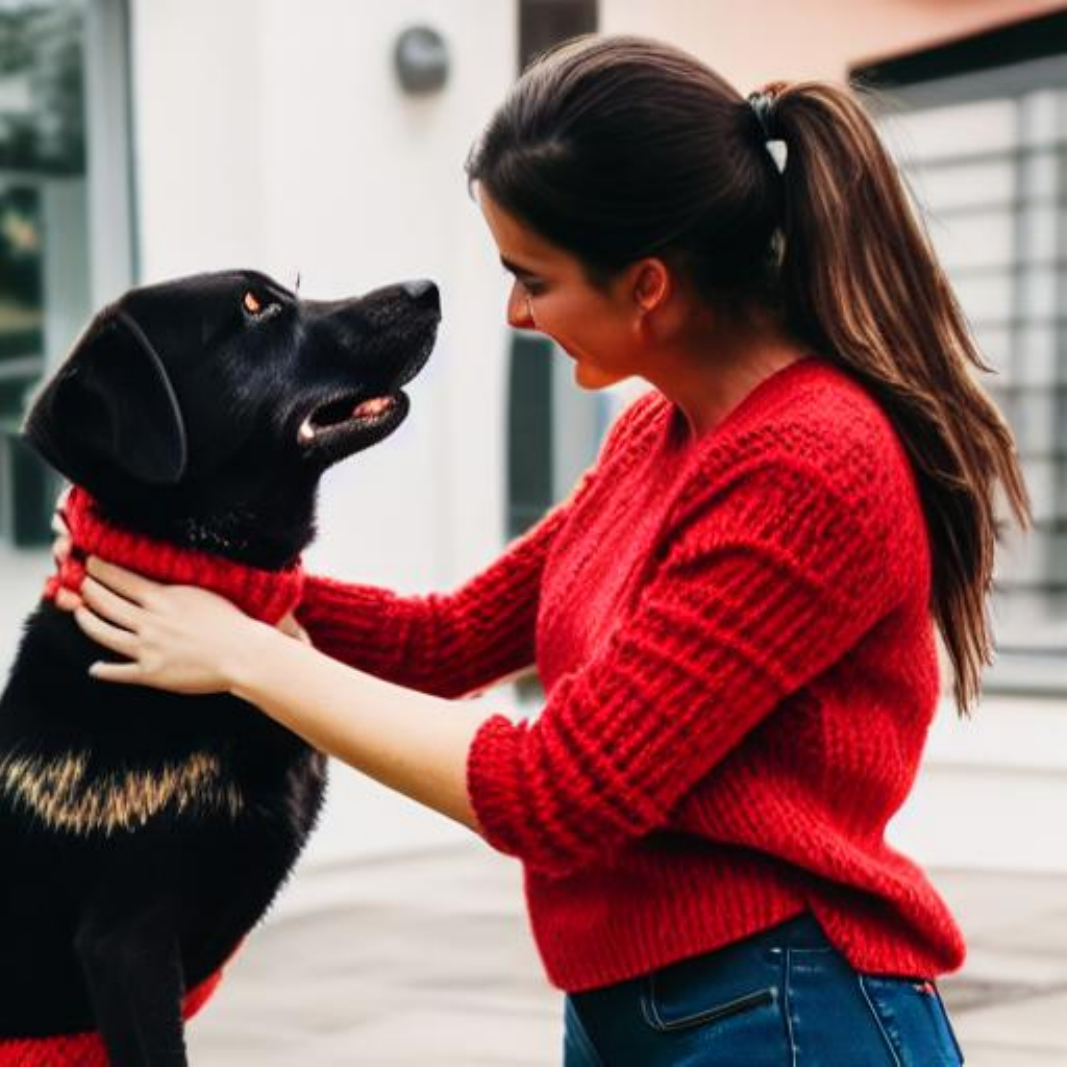} \end{subfigure}
    
    \parbox{1.02 \textwidth}{
        \centering
            a man looking at a distant mountain
        }
    \begin{subfigure}[t]{0.19\textwidth} \includegraphics[width=\textwidth]{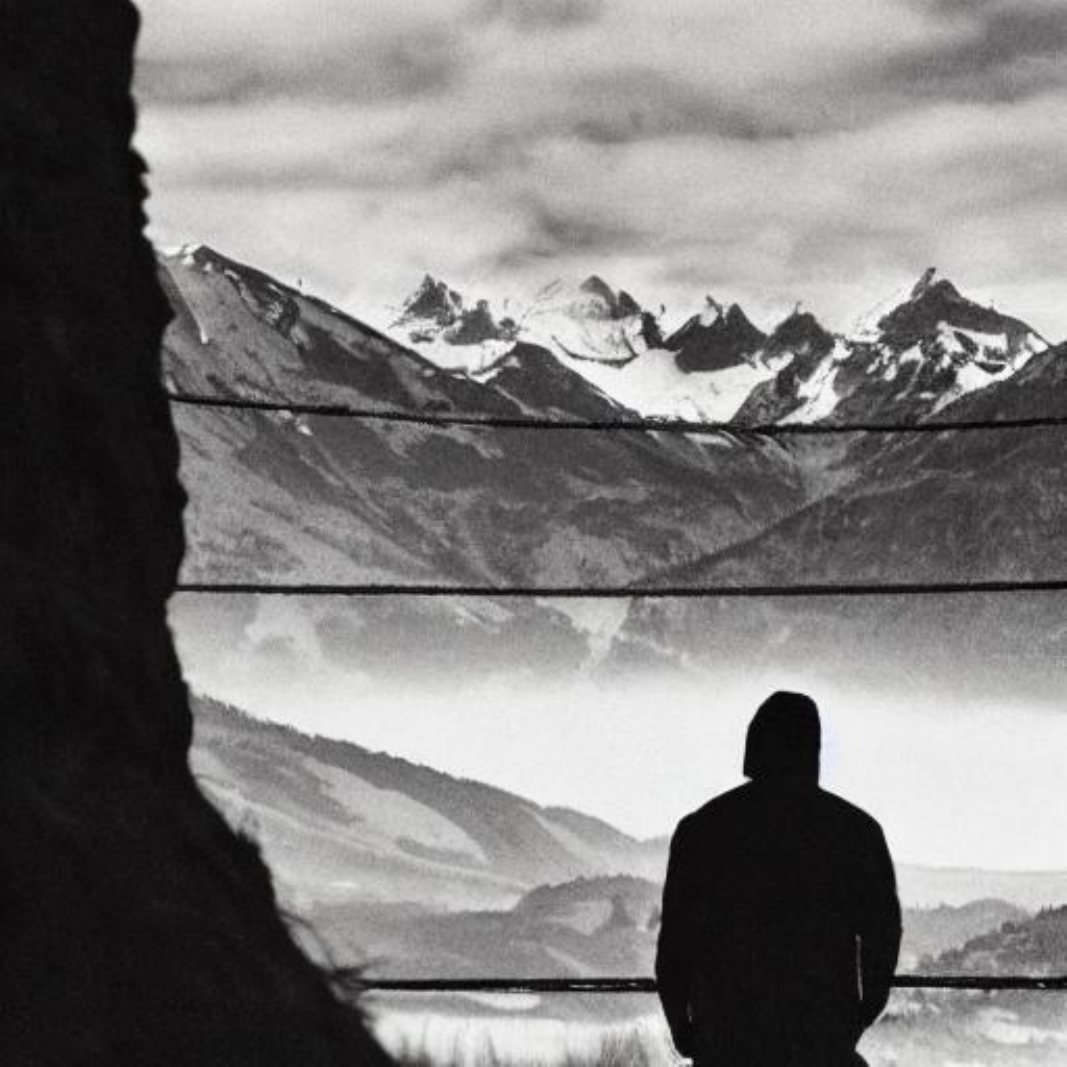} 
            \captionsetup{justification=centering}\caption{SD1.5~\cite{rombach2022high}}    \end{subfigure}
    \begin{subfigure}[t]{0.19\textwidth} \includegraphics[width=\textwidth]{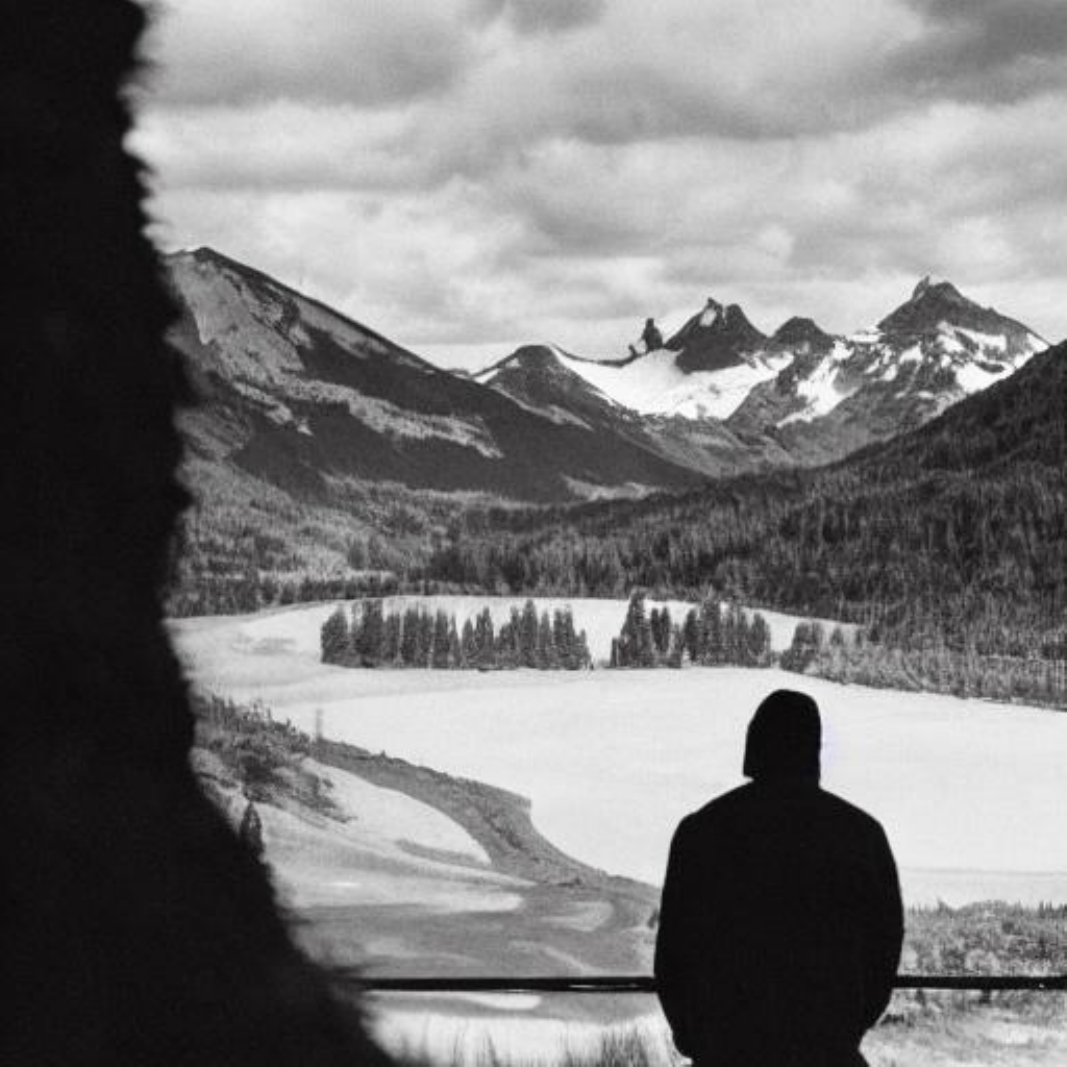} 
            \captionsetup{justification=centering}\caption{Diff-DPO~\cite{wallace2024diffusion}} \end{subfigure}
    \begin{subfigure}[t]{0.19\textwidth} \includegraphics[width=\textwidth]{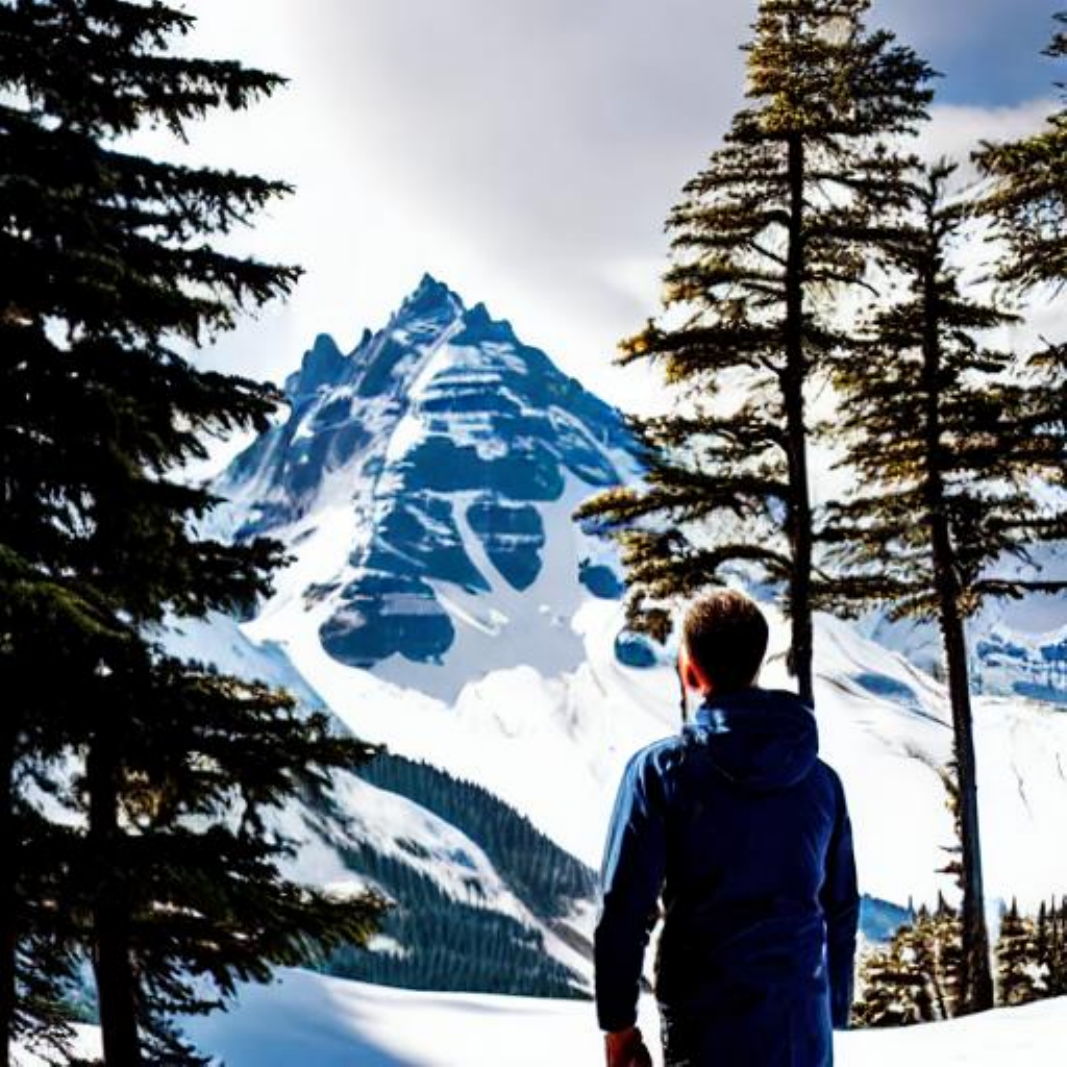}         \captionsetup{justification=centering}\caption{Diff-KTO~\cite{lialigning}} \end{subfigure}
    \begin{subfigure}[t]{0.19\textwidth} \includegraphics[width=\textwidth]{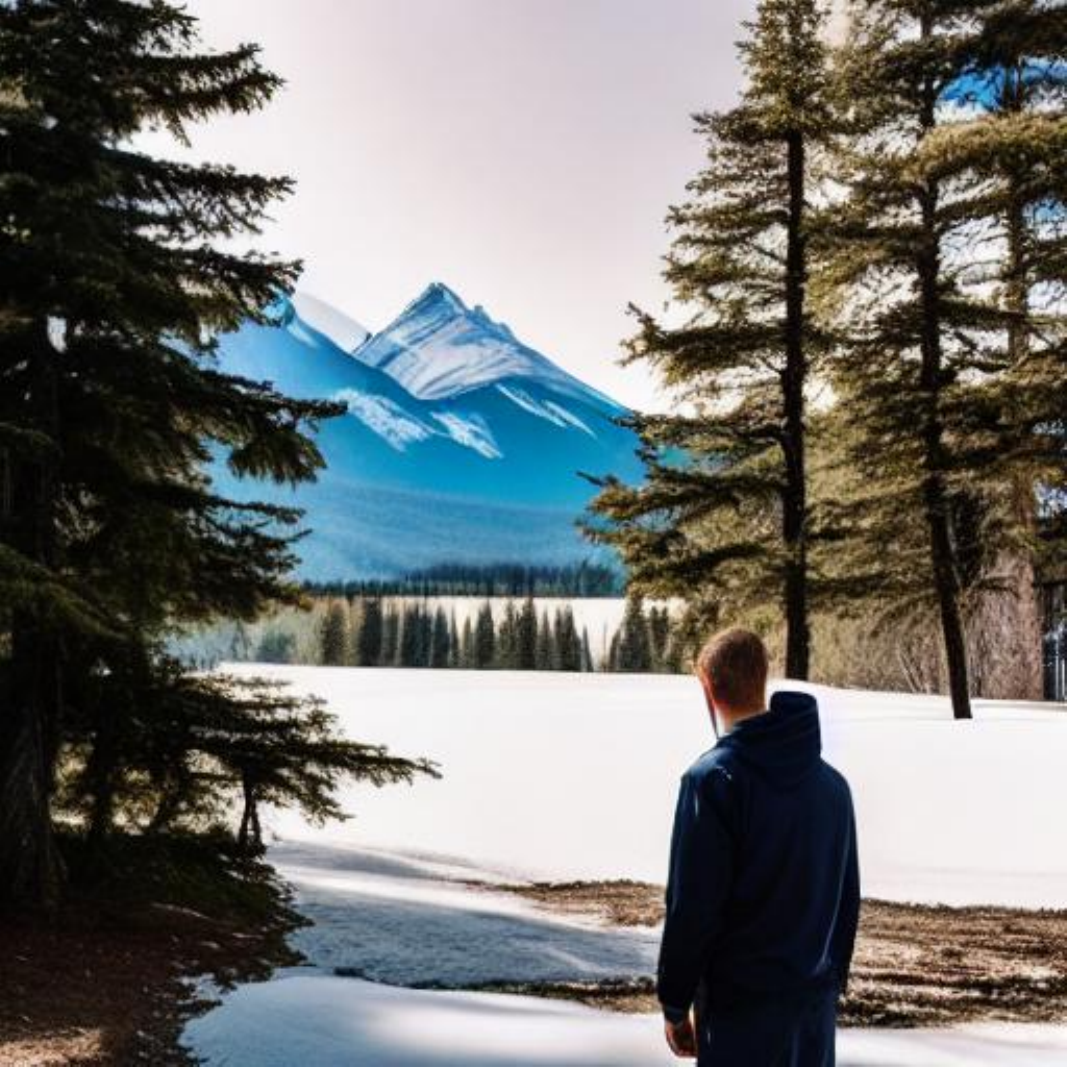}         \captionsetup{justification=centering}\caption{DSPO~\cite{zhu2025dspo}} \end{subfigure}
    \begin{subfigure}[t]{0.19\textwidth} \includegraphics[width=\textwidth]{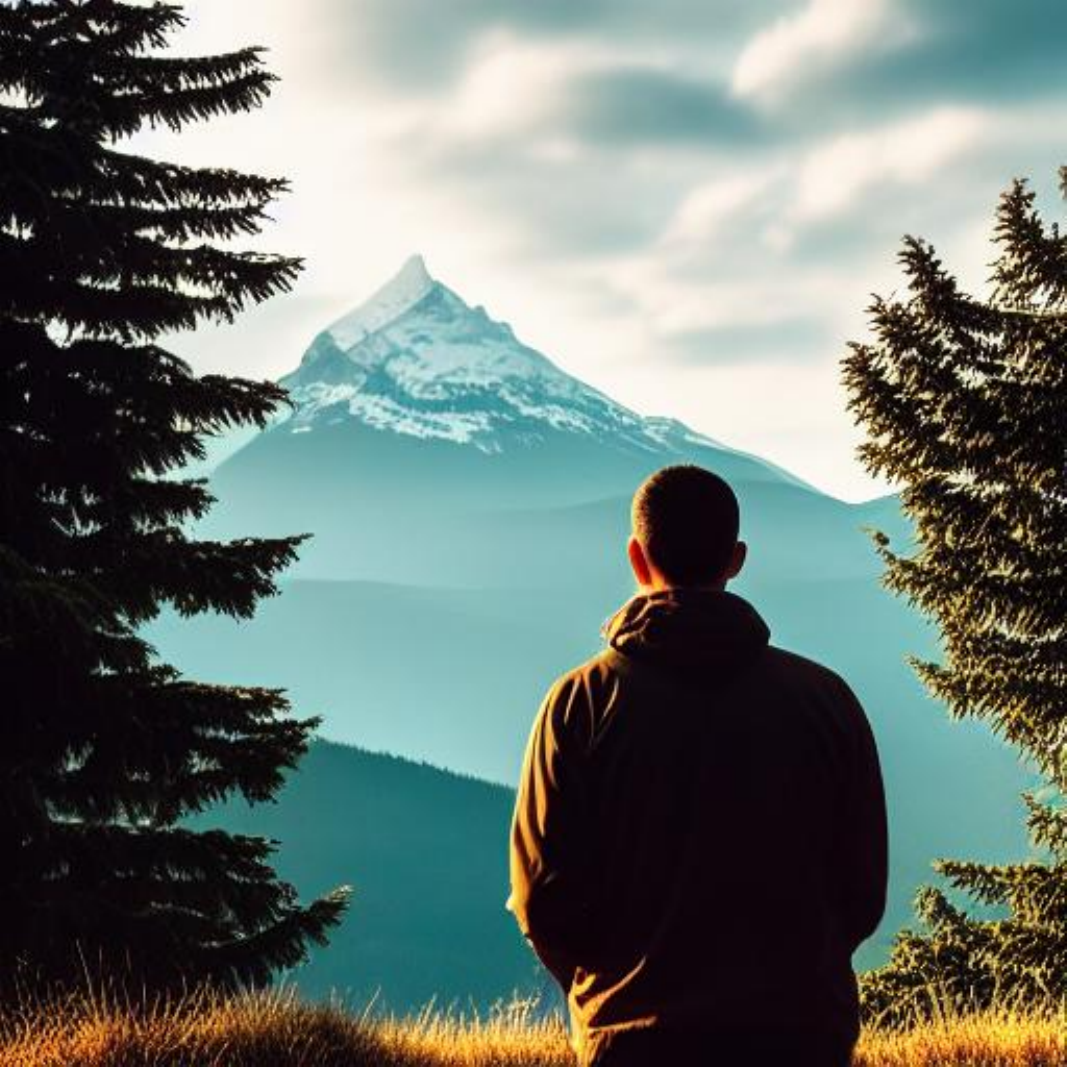}         \captionsetup{justification=centering}\caption{Ours} \end{subfigure}
    
    \caption{
       Qualitative comparisons on texts from PartiPrompts.
    }
    \label{fig:appendix_qual_parti}
\end{figure*}

\begin{figure*}[hb!]
    \centering
    \parbox{1.02 \textwidth}{
        \centering
        A bear in an astronaut suit sits on a rock on Mars surrounded by flowers under a starry sky.
    }
    \begin{subfigure}[t]{0.19\textwidth} \includegraphics[width=\textwidth]{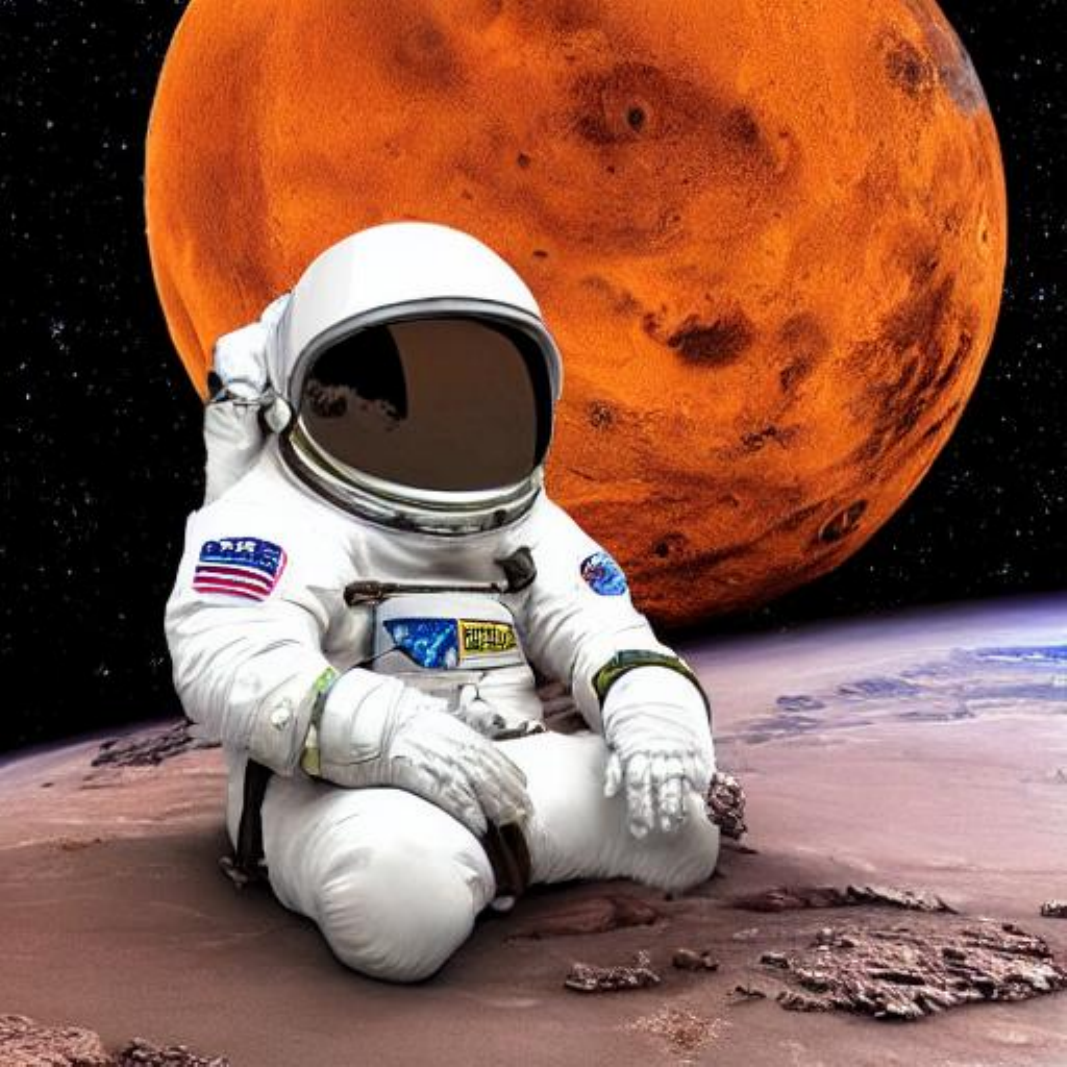} \end{subfigure}
    \begin{subfigure}[t]{0.19\textwidth} \includegraphics[width=\textwidth]{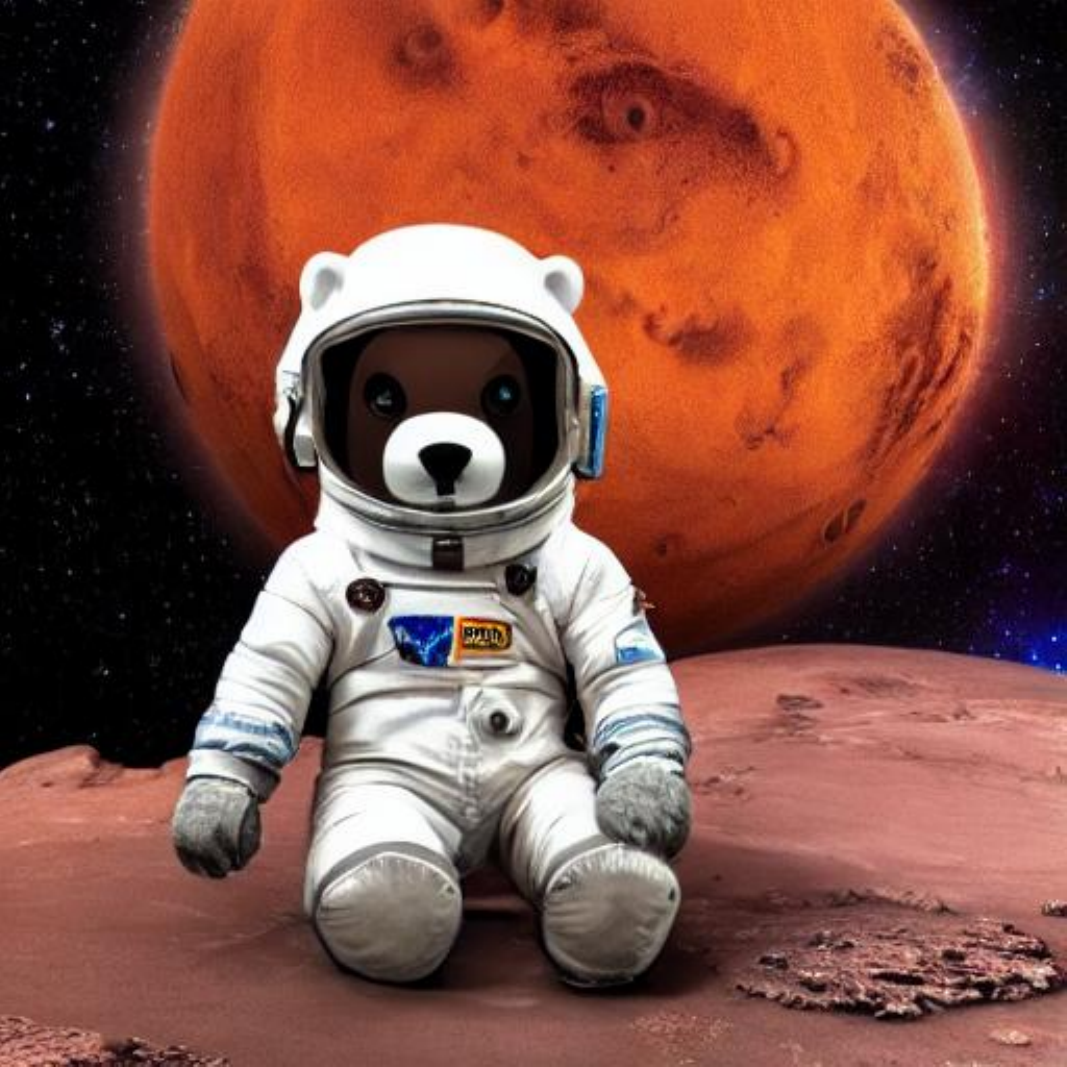} \end{subfigure}
    \begin{subfigure}[t]{0.19\textwidth} \includegraphics[width=\textwidth]{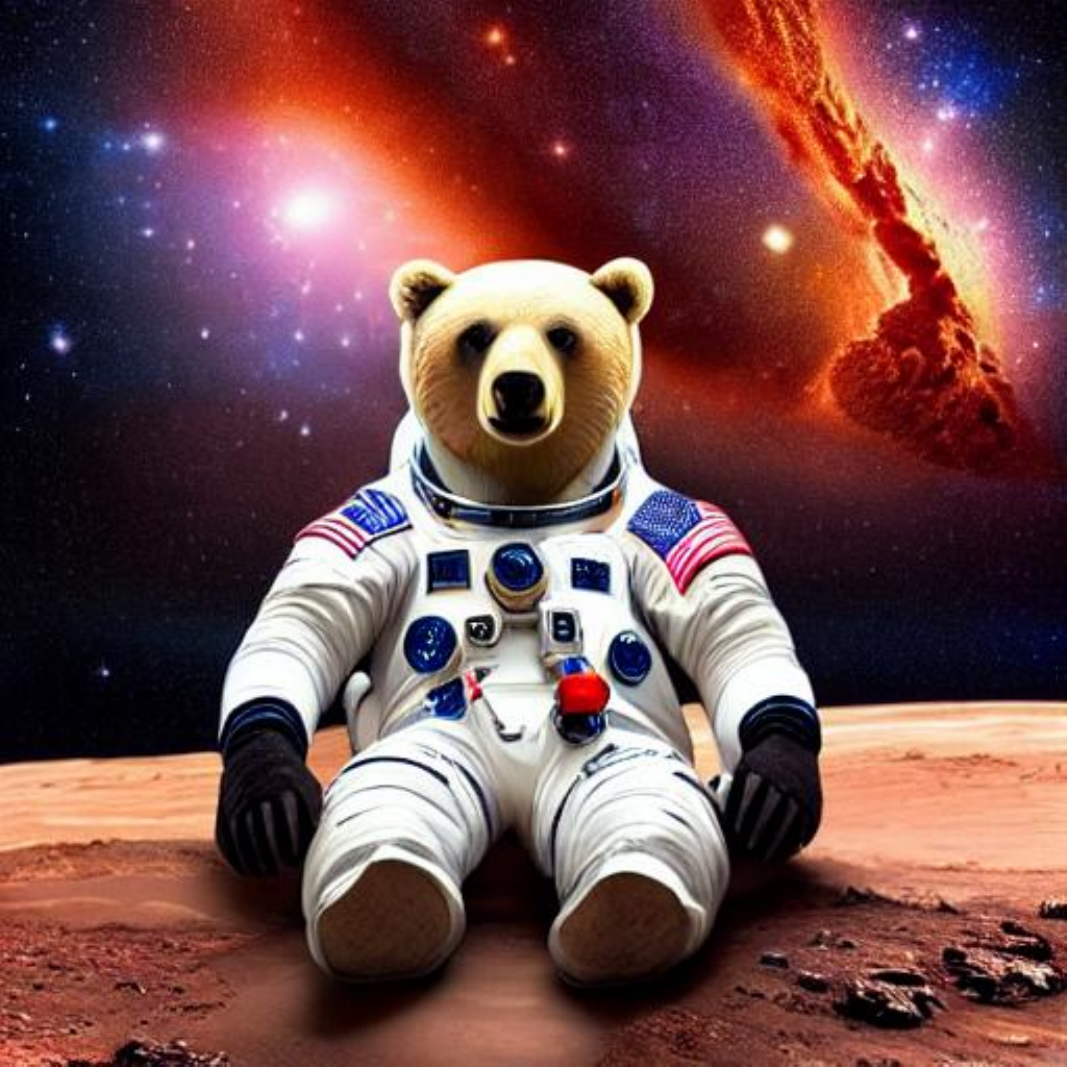} \end{subfigure}
    \begin{subfigure}[t]{0.19\textwidth} \includegraphics[width=\textwidth]{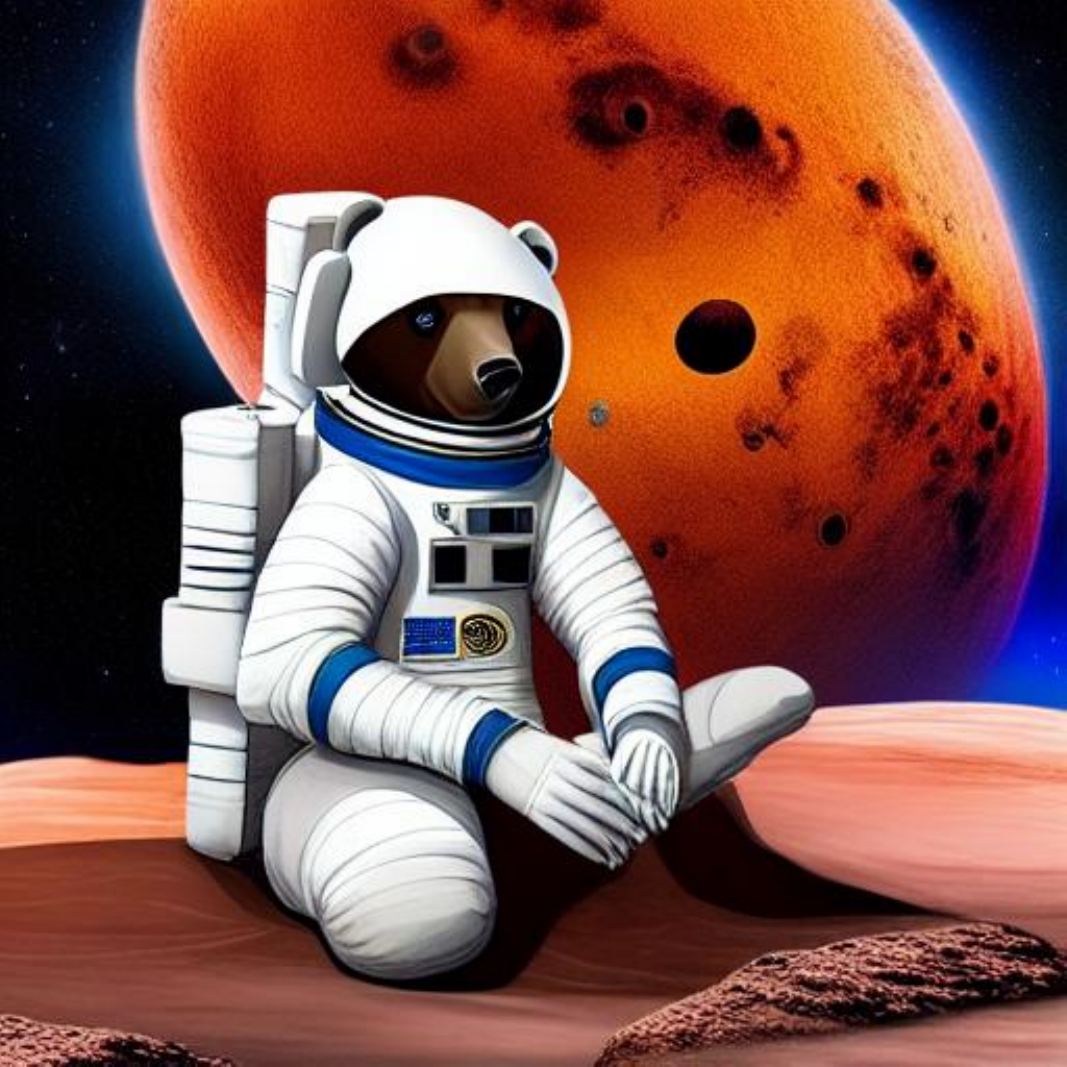} \end{subfigure}
    \begin{subfigure}[t]{0.19\textwidth} \includegraphics[width=\textwidth]{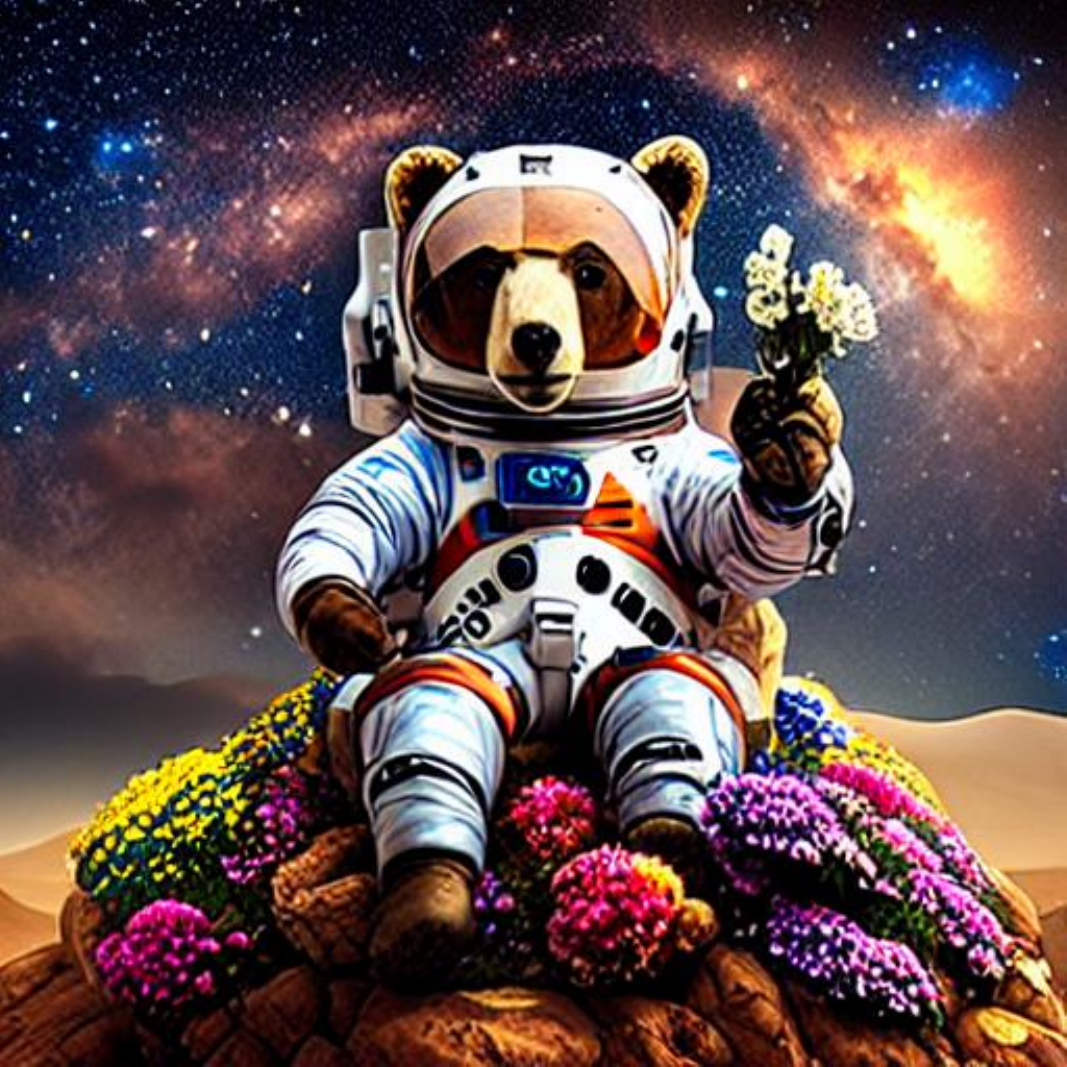} \end{subfigure}

    \parbox{1.02 \textwidth}{
        \centering
        A portrait of a stylized business cat in sharp focus with a medium shot perspective, resembling boxart.

    }
    \begin{subfigure}[t]{0.19\textwidth} \includegraphics[width=\textwidth]{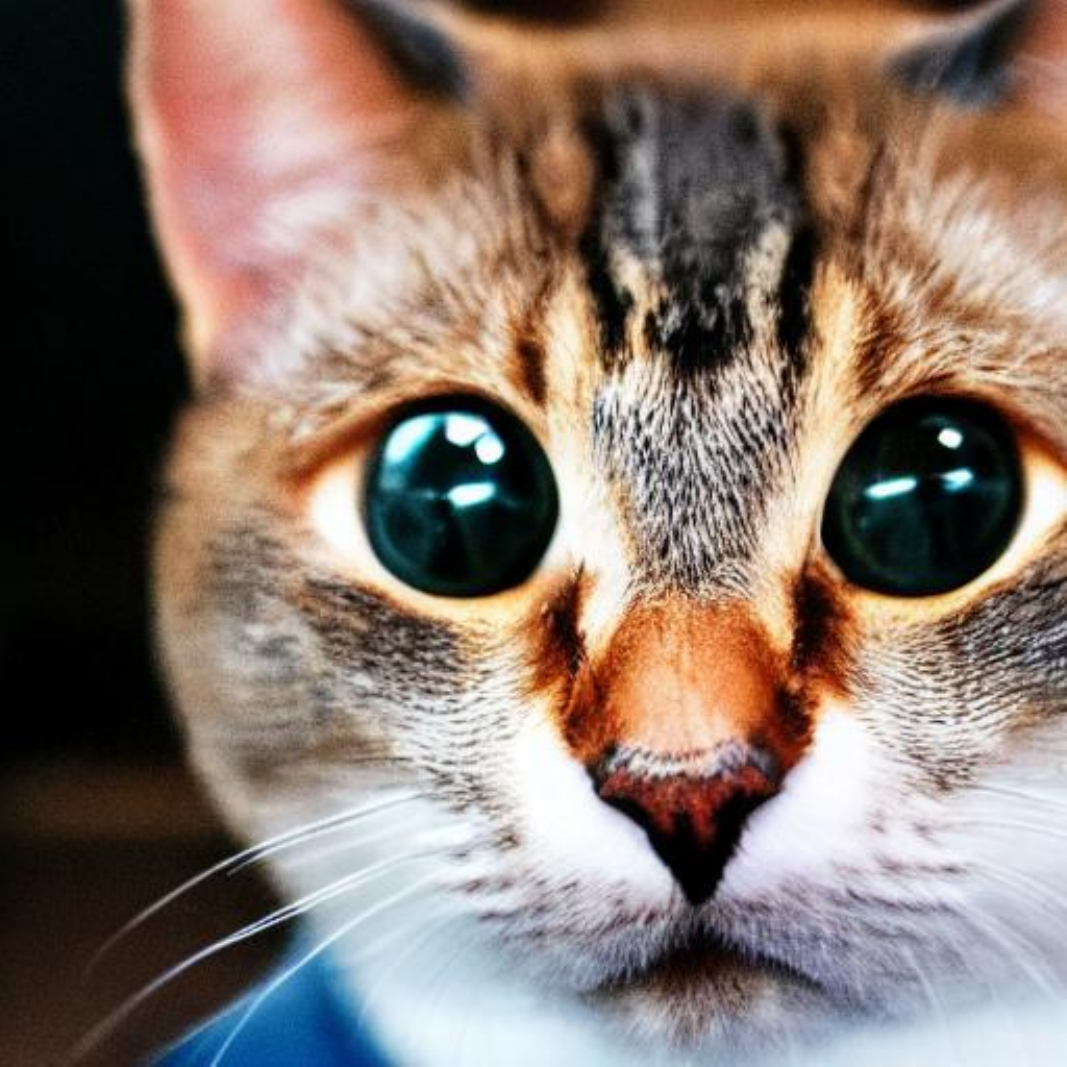} \end{subfigure}
    \begin{subfigure}[t]{0.19\textwidth} \includegraphics[width=\textwidth]{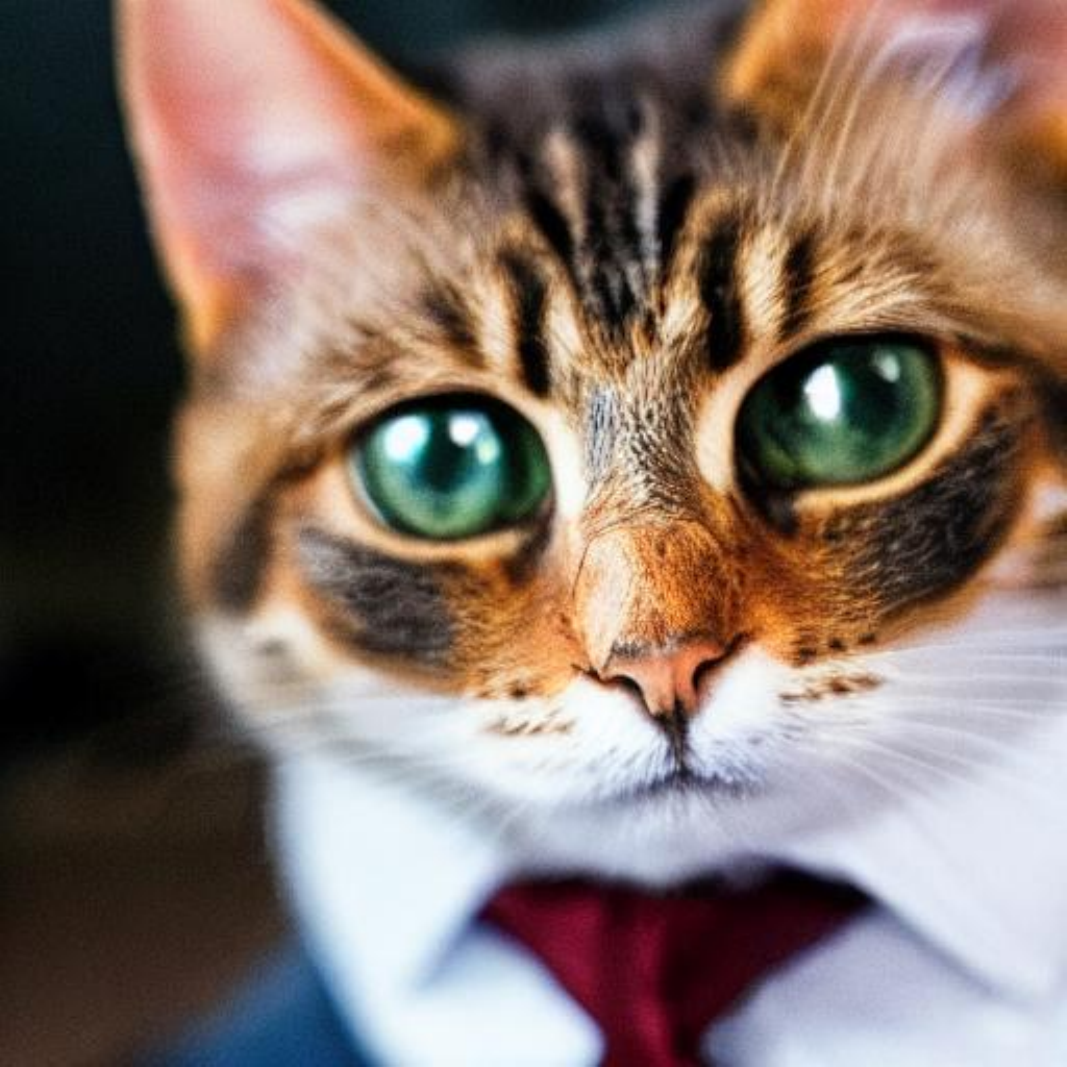} \end{subfigure}
    \begin{subfigure}[t]{0.19\textwidth} \includegraphics[width=\textwidth]{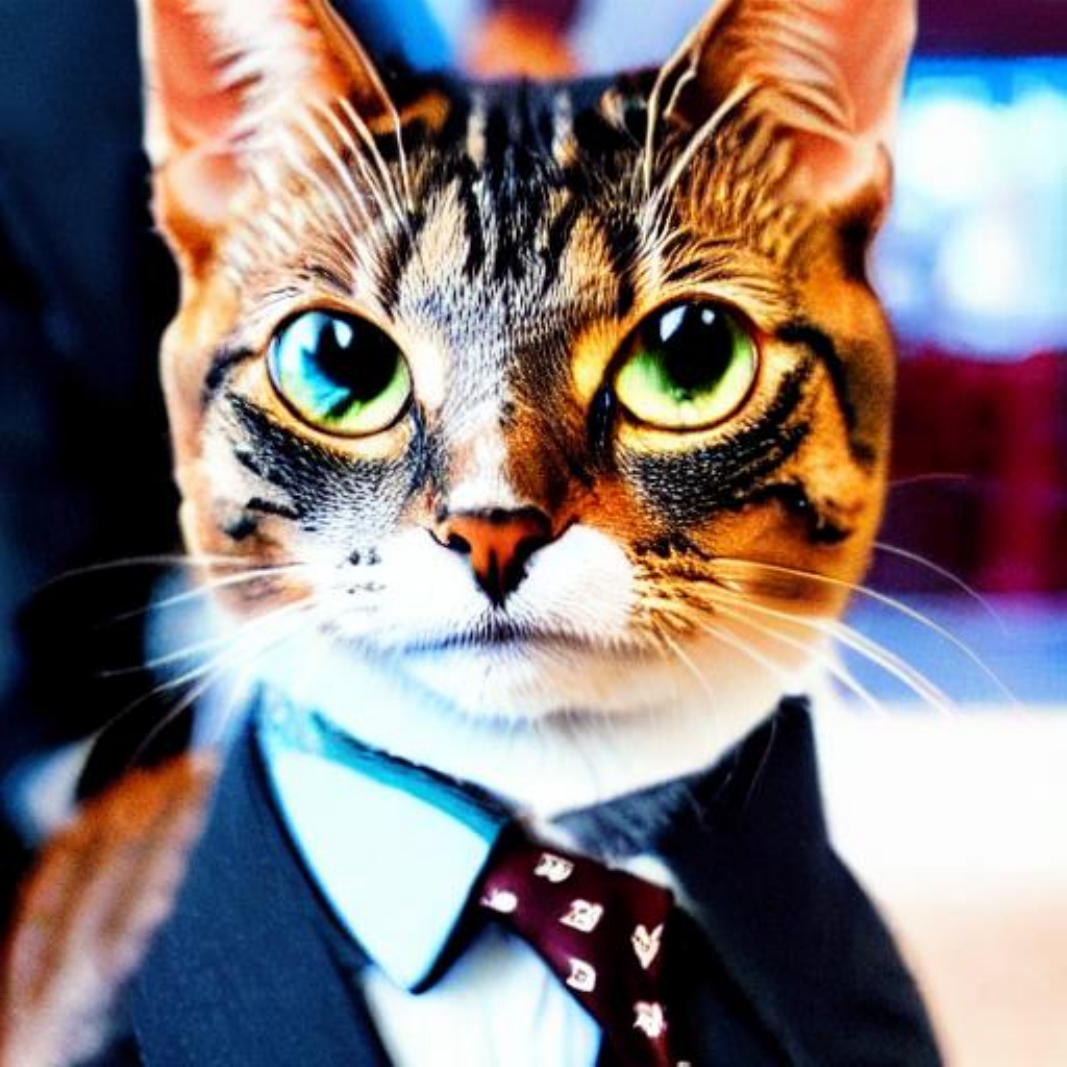} \end{subfigure}
    \begin{subfigure}[t]{0.19\textwidth} \includegraphics[width=\textwidth]{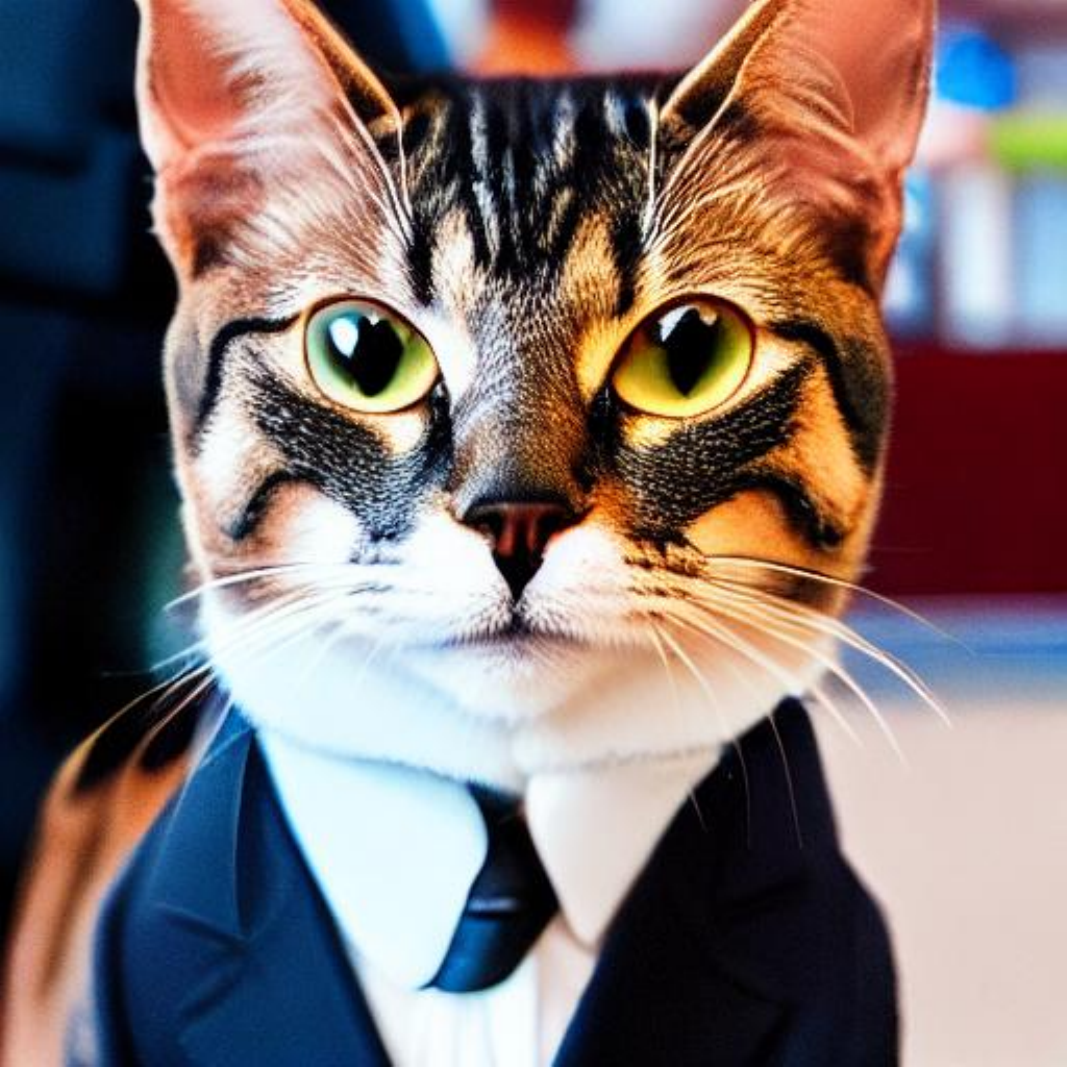} \end{subfigure}
    \begin{subfigure}[t]{0.19\textwidth} \includegraphics[width=\textwidth]{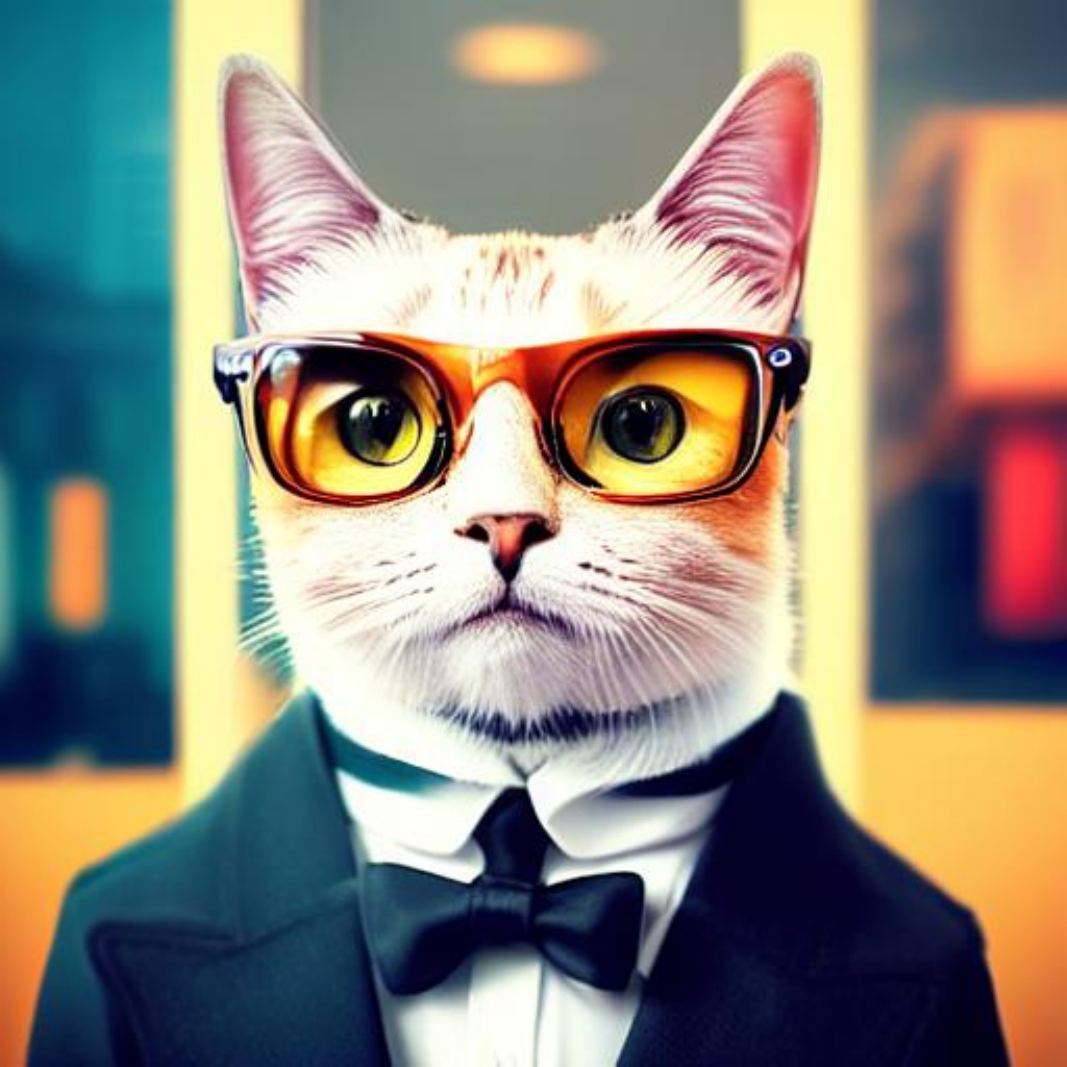} \end{subfigure}

    \parbox{1.02 \textwidth}{
        \centering
        A white bichon frise puppy dog riding a black motorcycle in Hollywood at sundown with palm trees in the background.

    }
    \begin{subfigure}[t]{0.19\textwidth} \includegraphics[width=\textwidth]{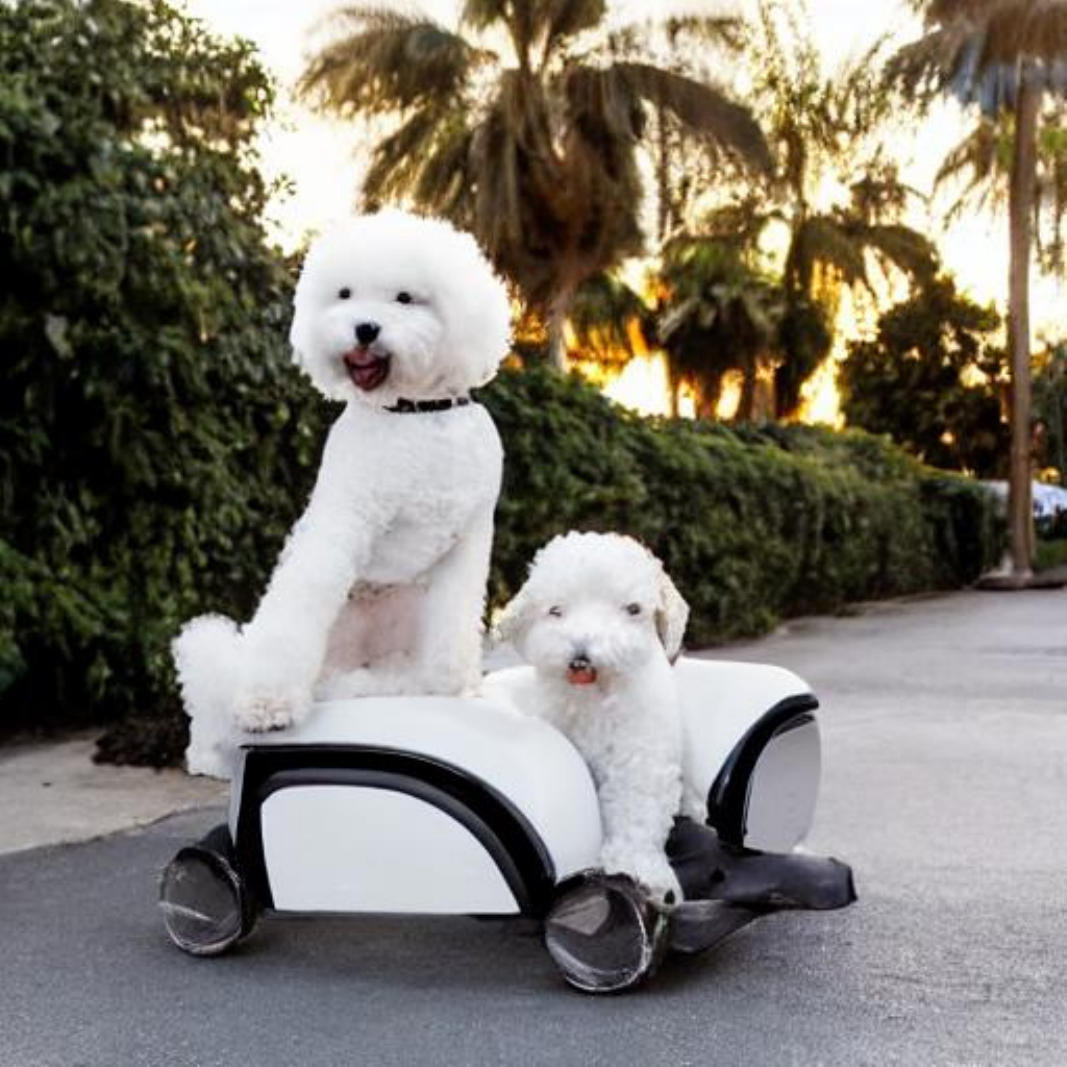} \end{subfigure}
    \begin{subfigure}[t]{0.19\textwidth} \includegraphics[width=\textwidth]{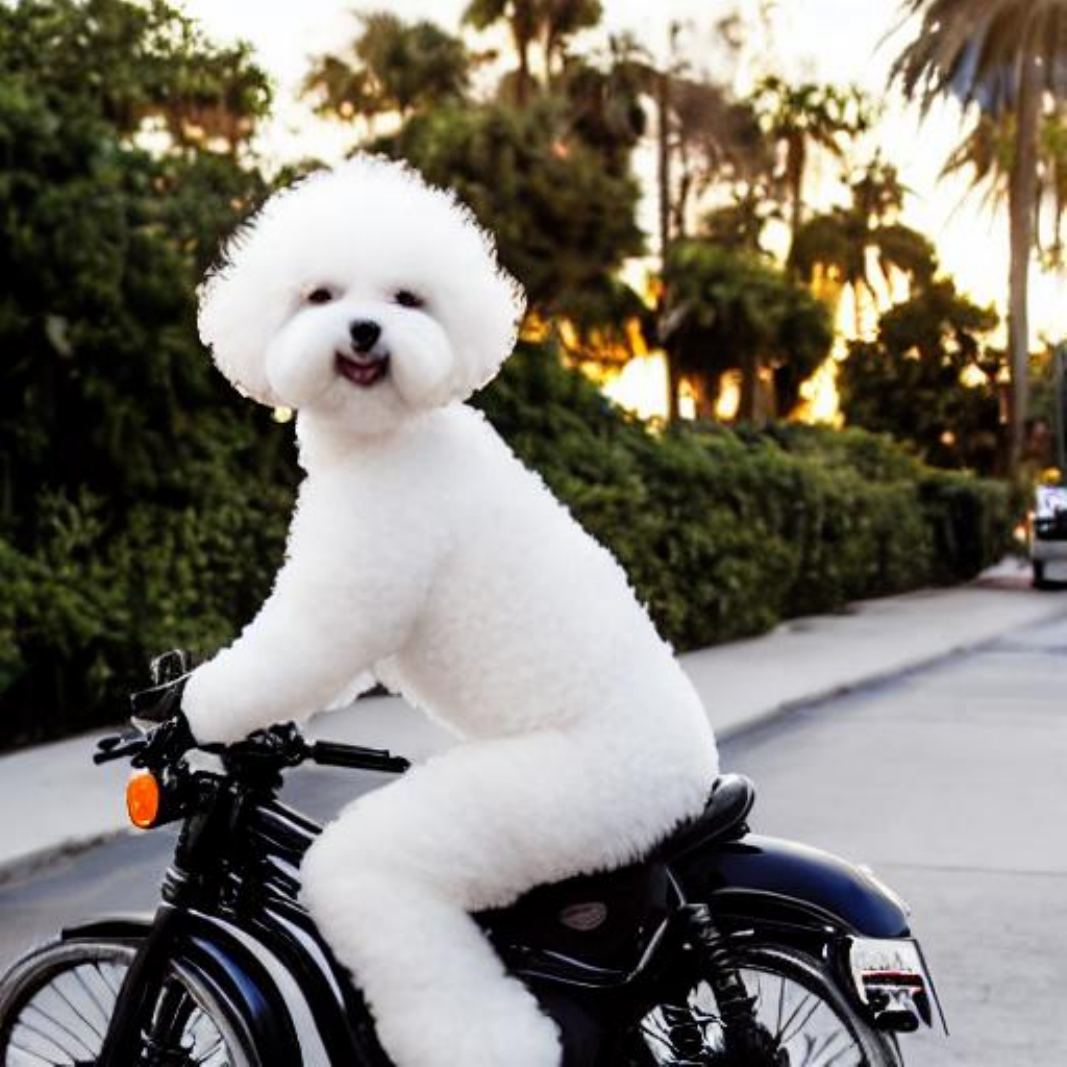} \end{subfigure}
    \begin{subfigure}[t]{0.19\textwidth} \includegraphics[width=\textwidth]{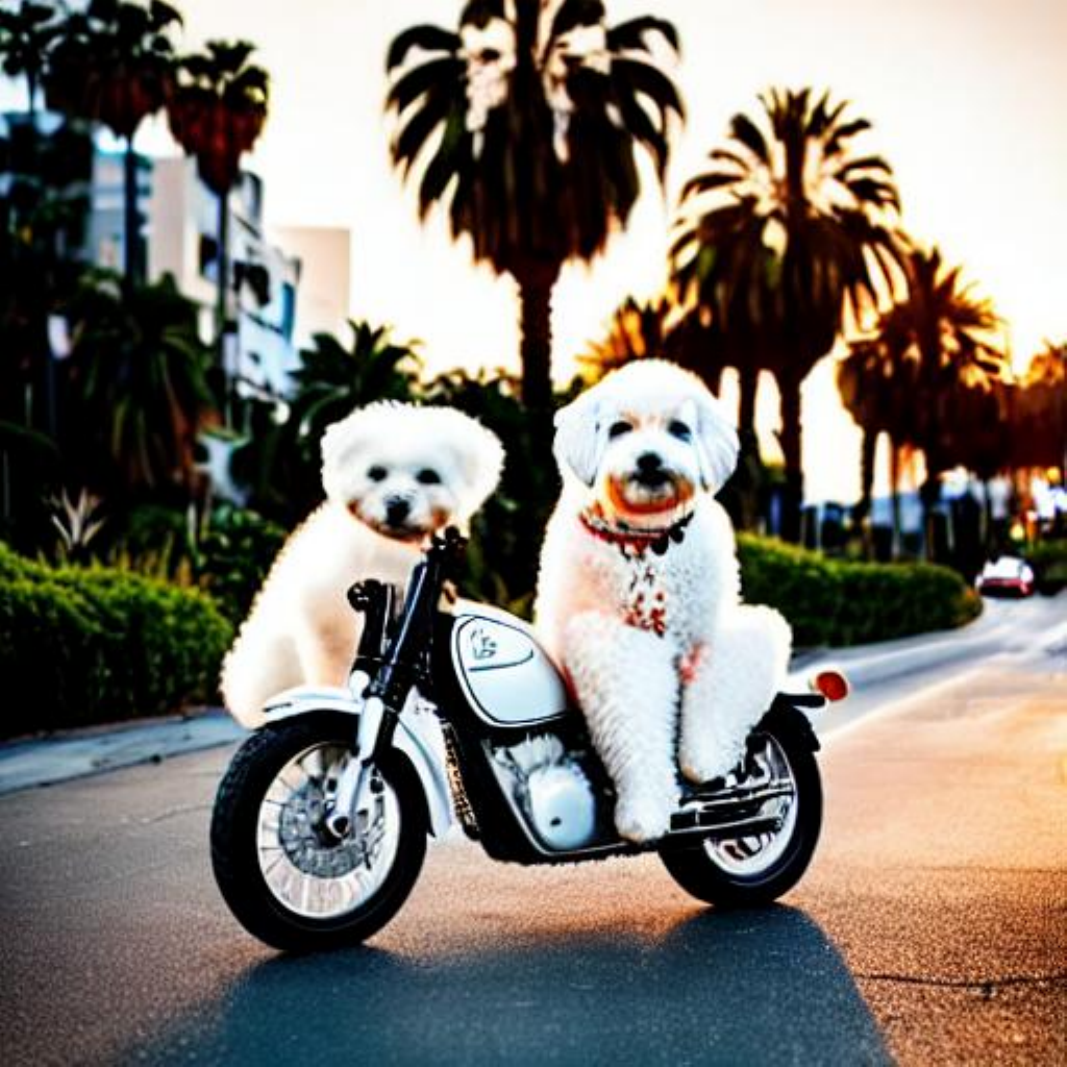} \end{subfigure}
    \begin{subfigure}[t]{0.19\textwidth} \includegraphics[width=\textwidth]{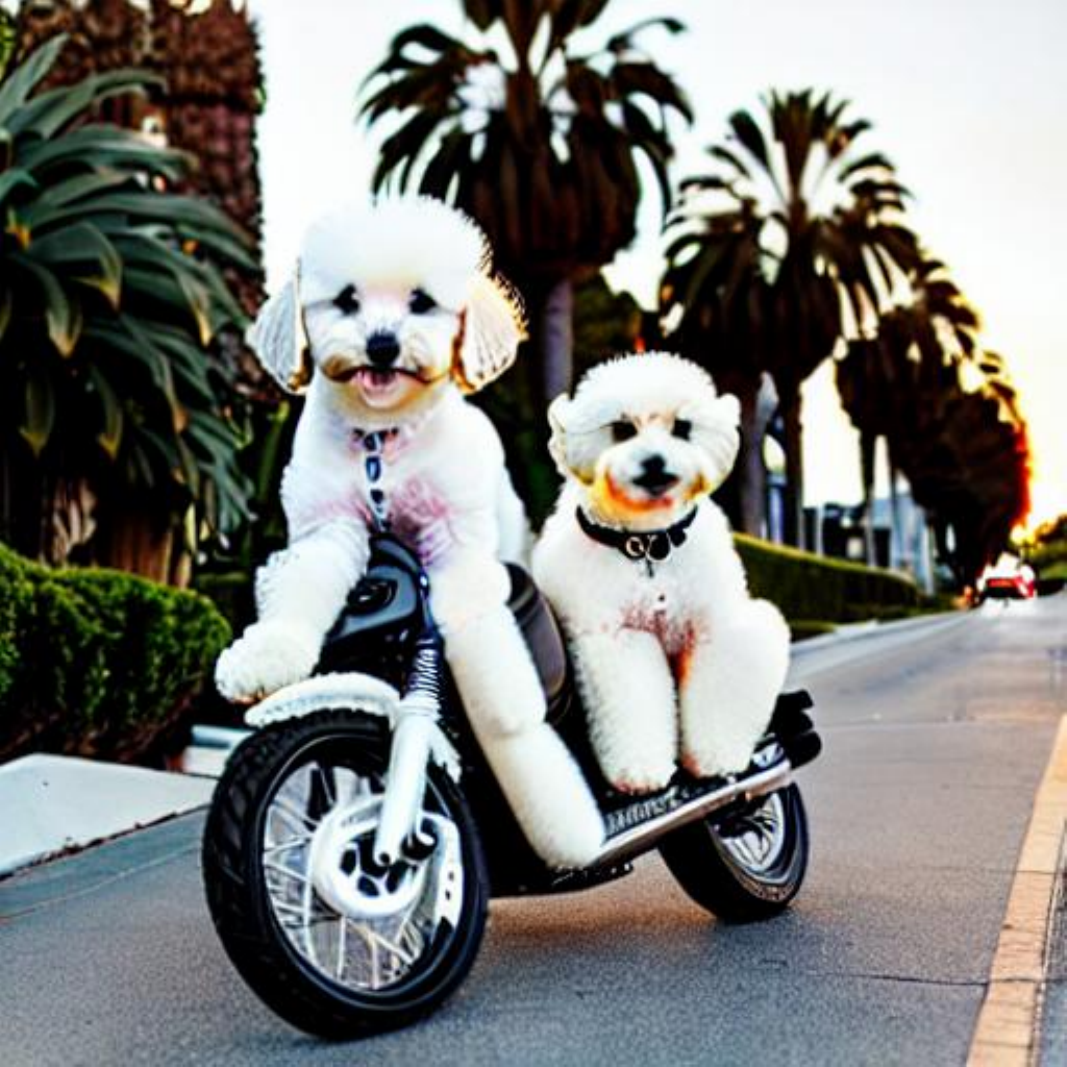} \end{subfigure}
    \begin{subfigure}[t]{0.19\textwidth} \includegraphics[width=\textwidth]{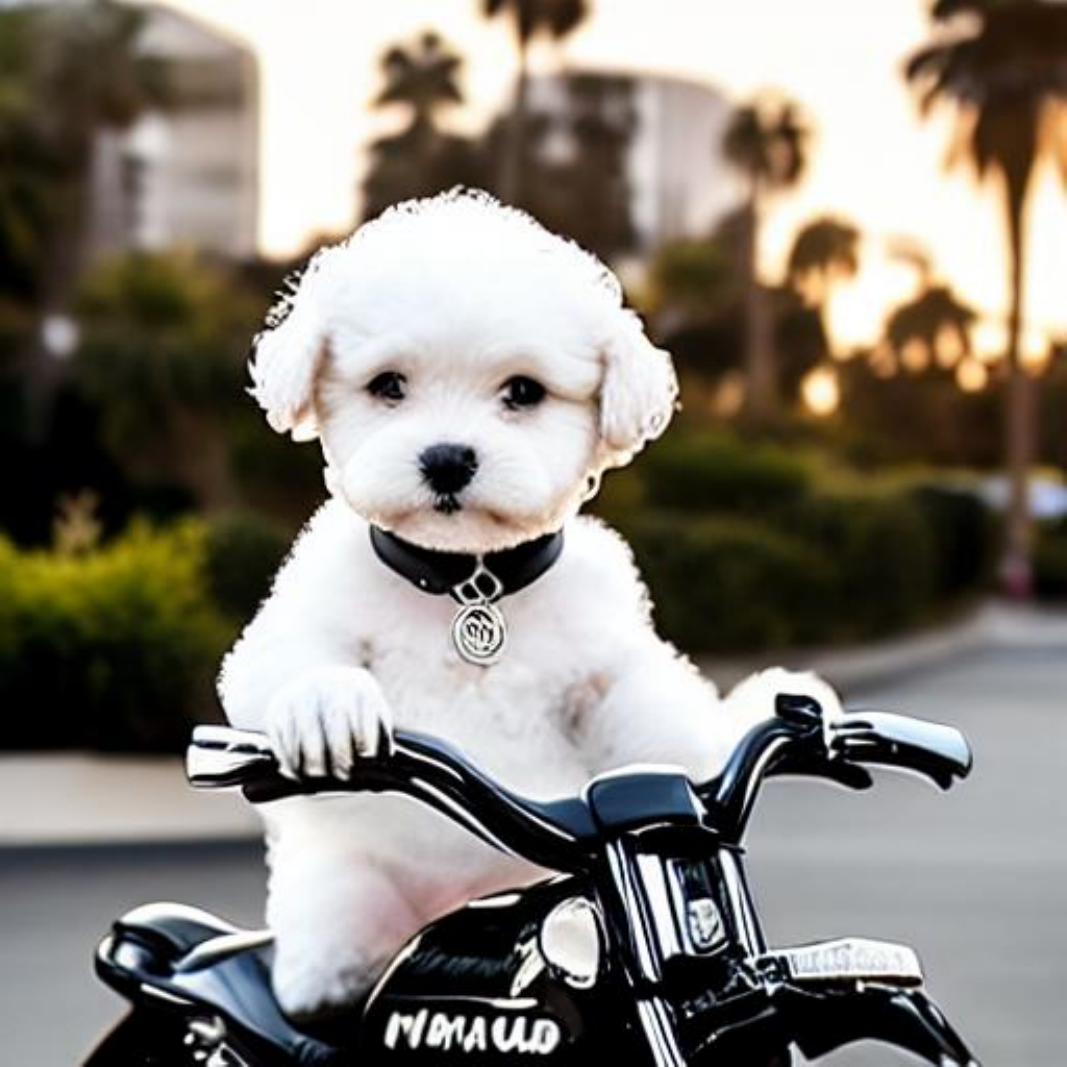} \end{subfigure}

    \parbox{1.02 \textwidth}{
        \centering
        A candy house on the ocean in a fantasy setting.

    }
    \begin{subfigure}[t]{0.19\textwidth} \includegraphics[width=\textwidth]{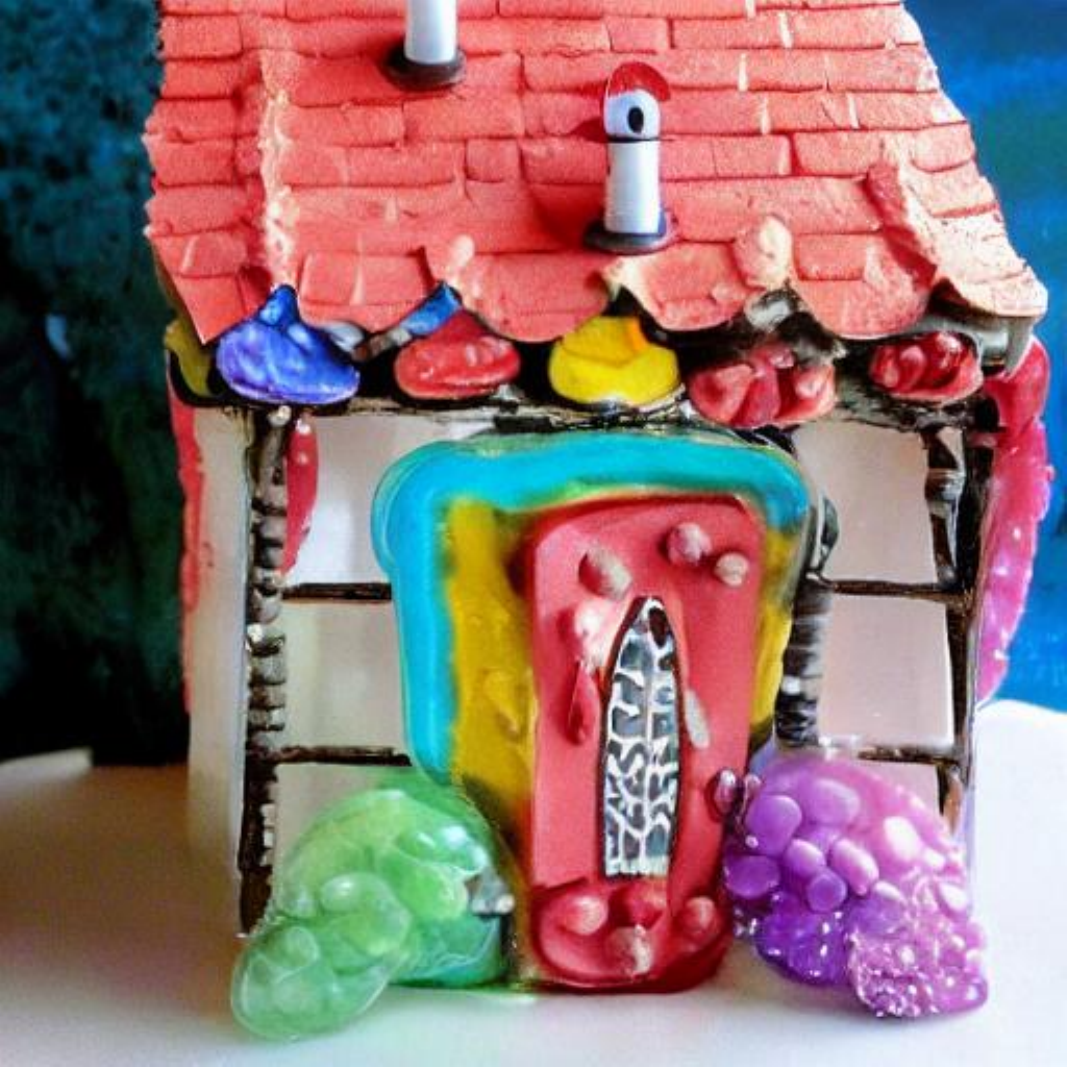} \end{subfigure}
    \begin{subfigure}[t]{0.19\textwidth} \includegraphics[width=\textwidth]{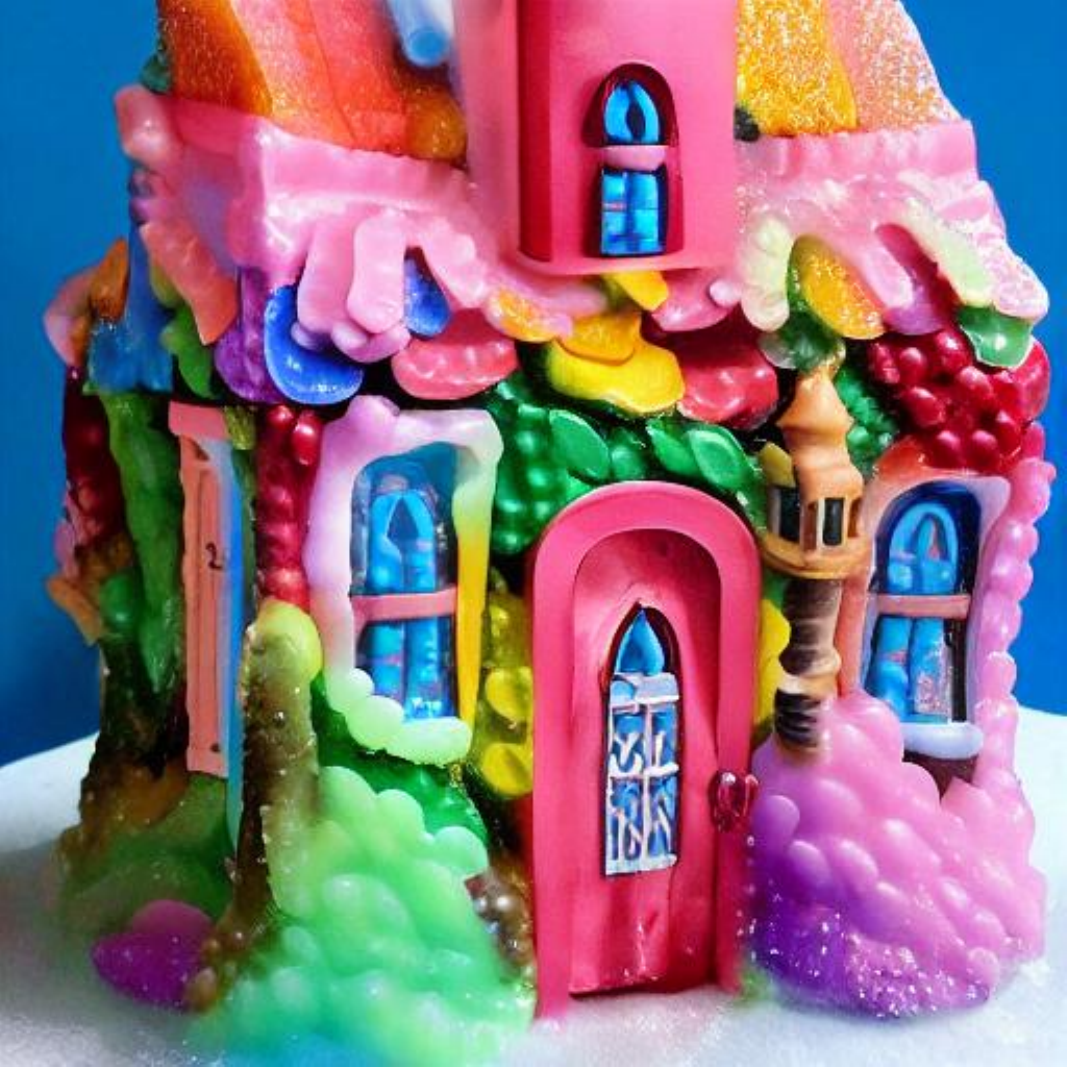} \end{subfigure}
    \begin{subfigure}[t]{0.19\textwidth} \includegraphics[width=\textwidth]{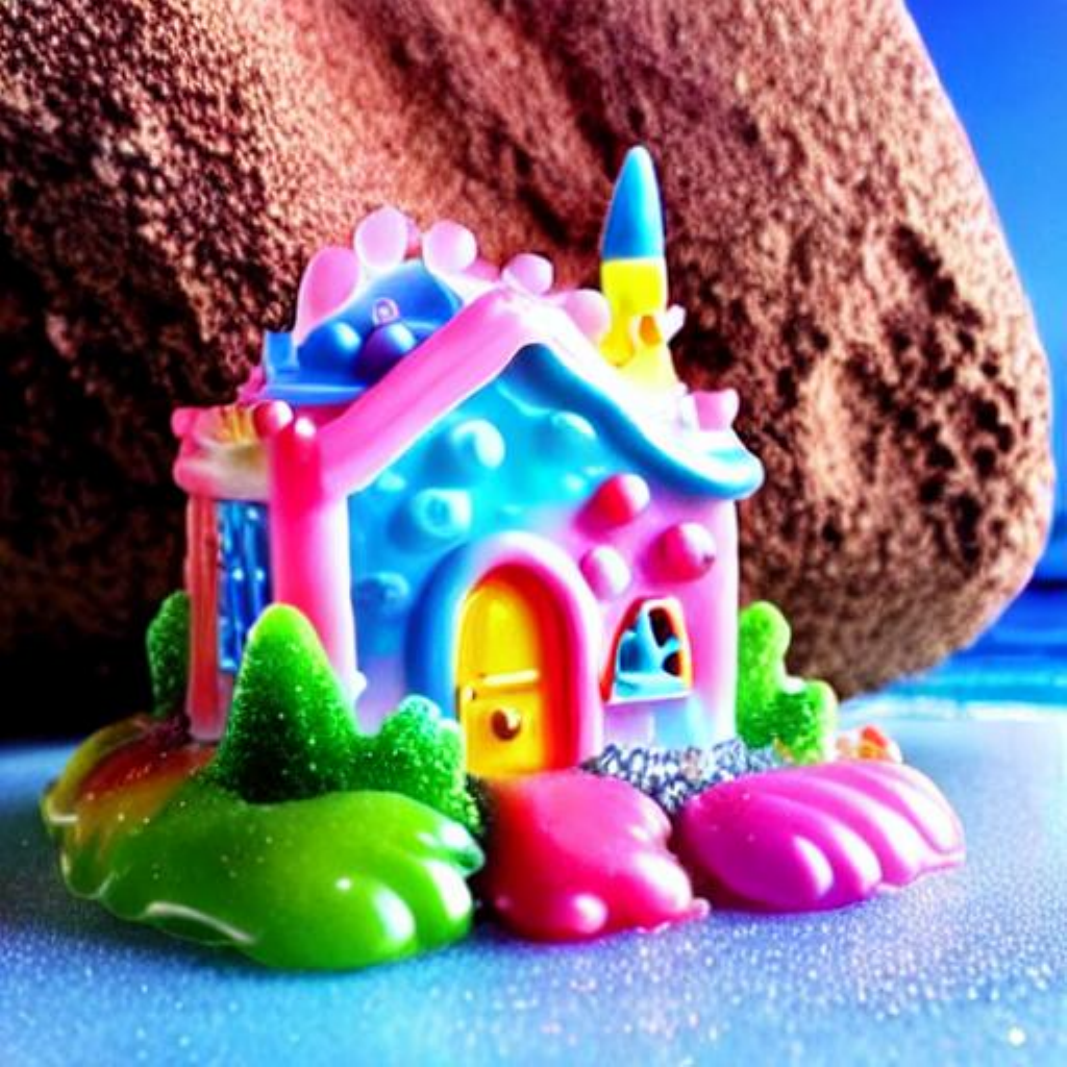} \end{subfigure}
    \begin{subfigure}[t]{0.19\textwidth} \includegraphics[width=\textwidth]{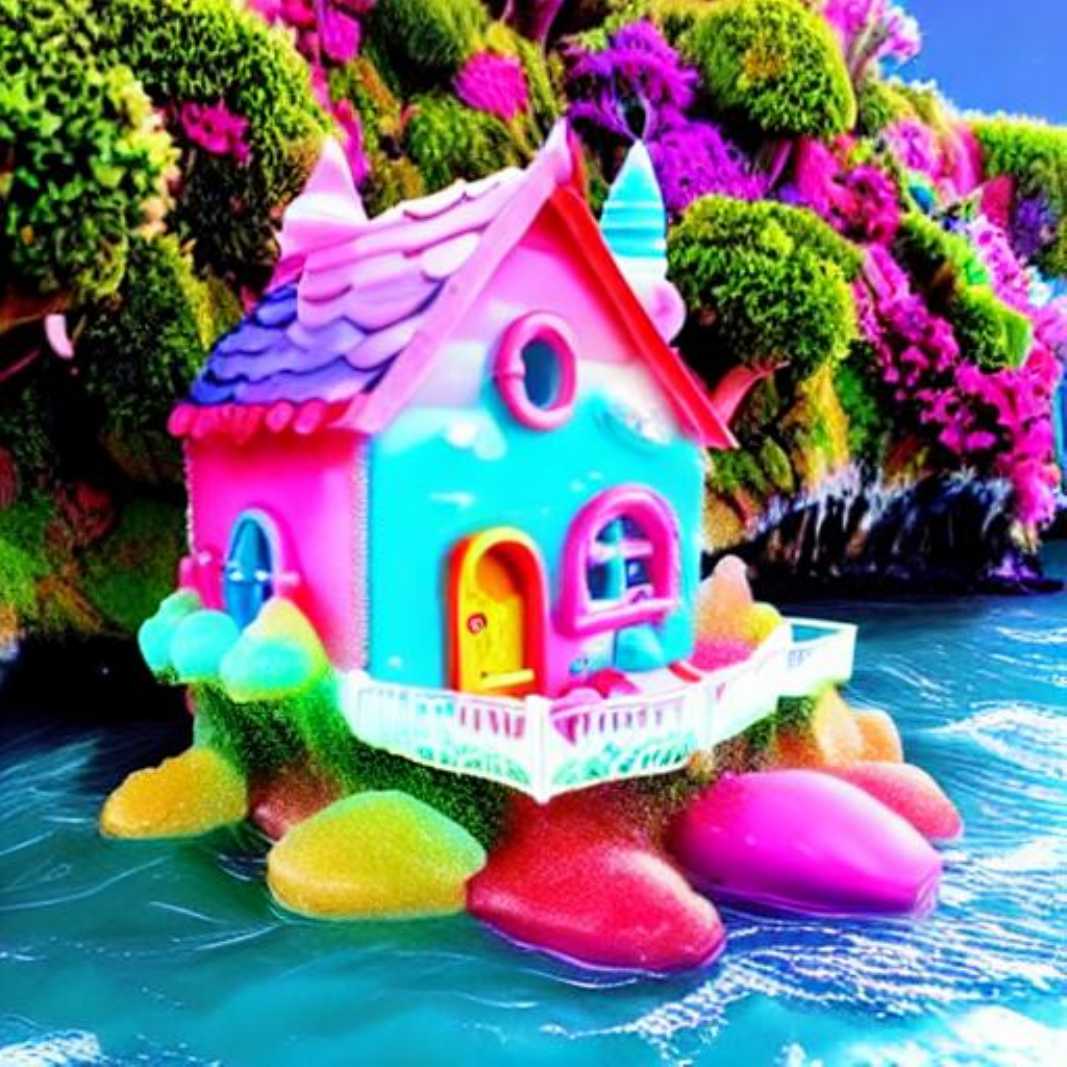} \end{subfigure}
    \begin{subfigure}[t]{0.19\textwidth} \includegraphics[width=\textwidth]{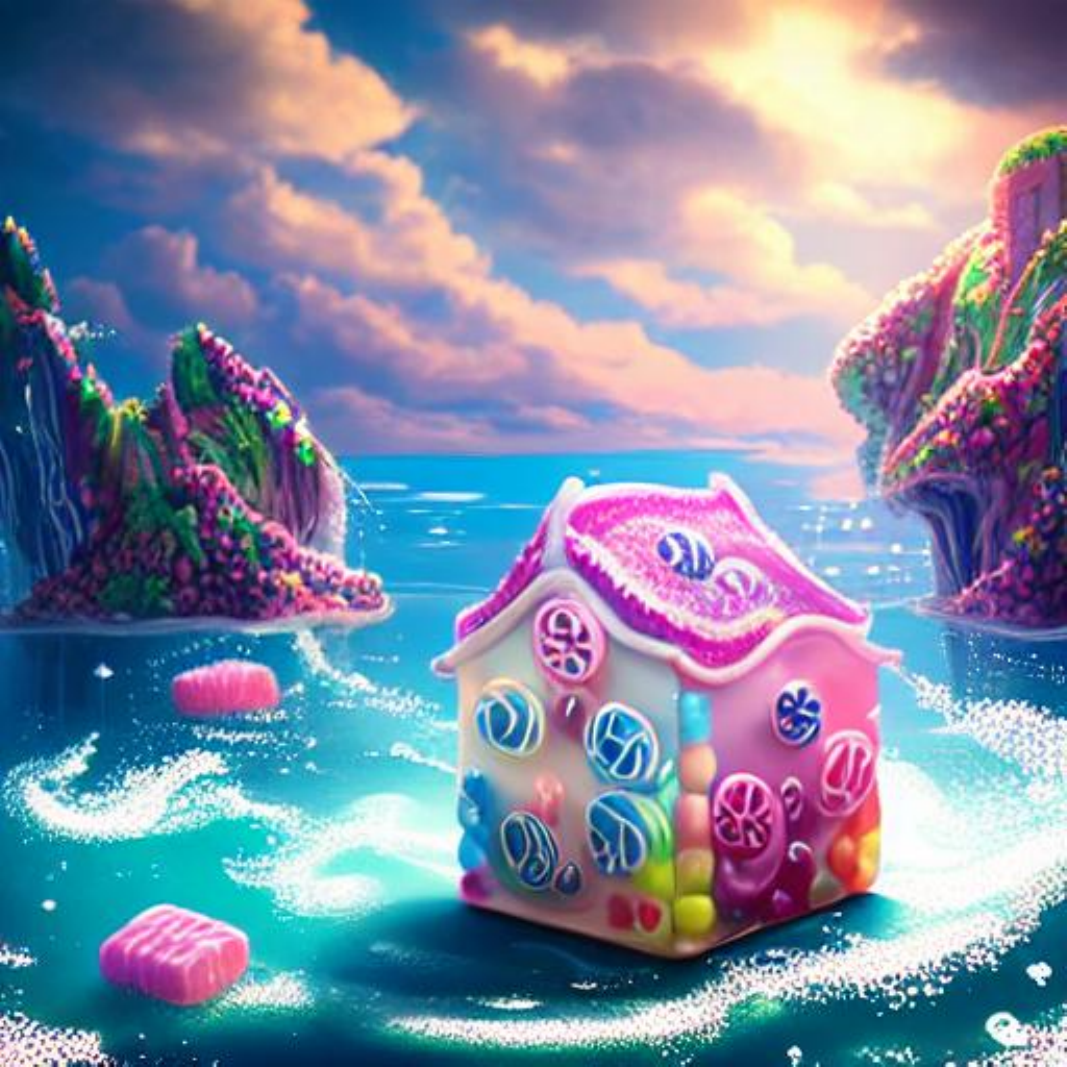} \end{subfigure}

    \parbox{1.02 \textwidth}{
        \centering
        A monkey wearing a jacket.

    }
    \begin{subfigure}[t]{0.19\textwidth} \includegraphics[width=\textwidth]{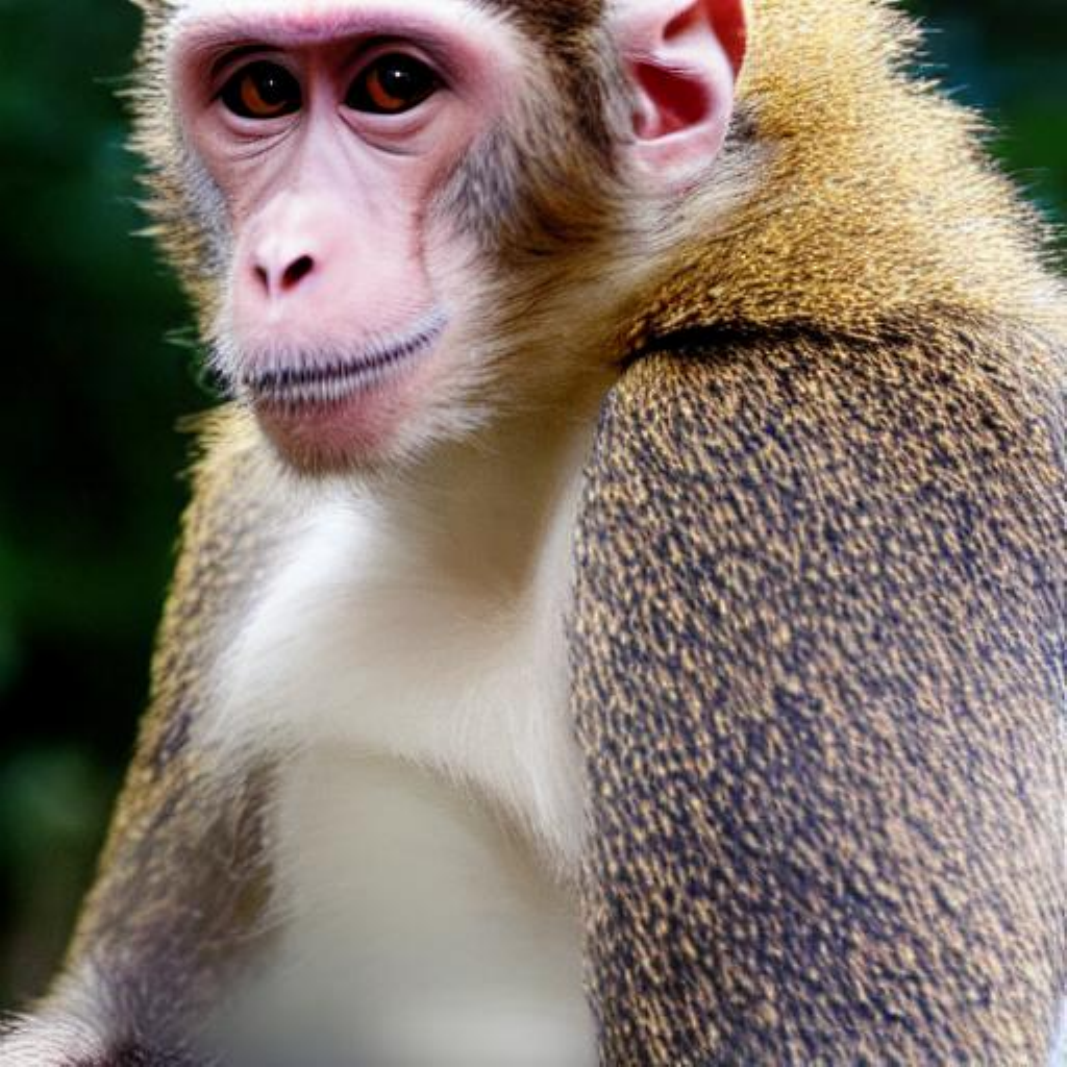} \end{subfigure}
    \begin{subfigure}[t]{0.19\textwidth} \includegraphics[width=\textwidth]{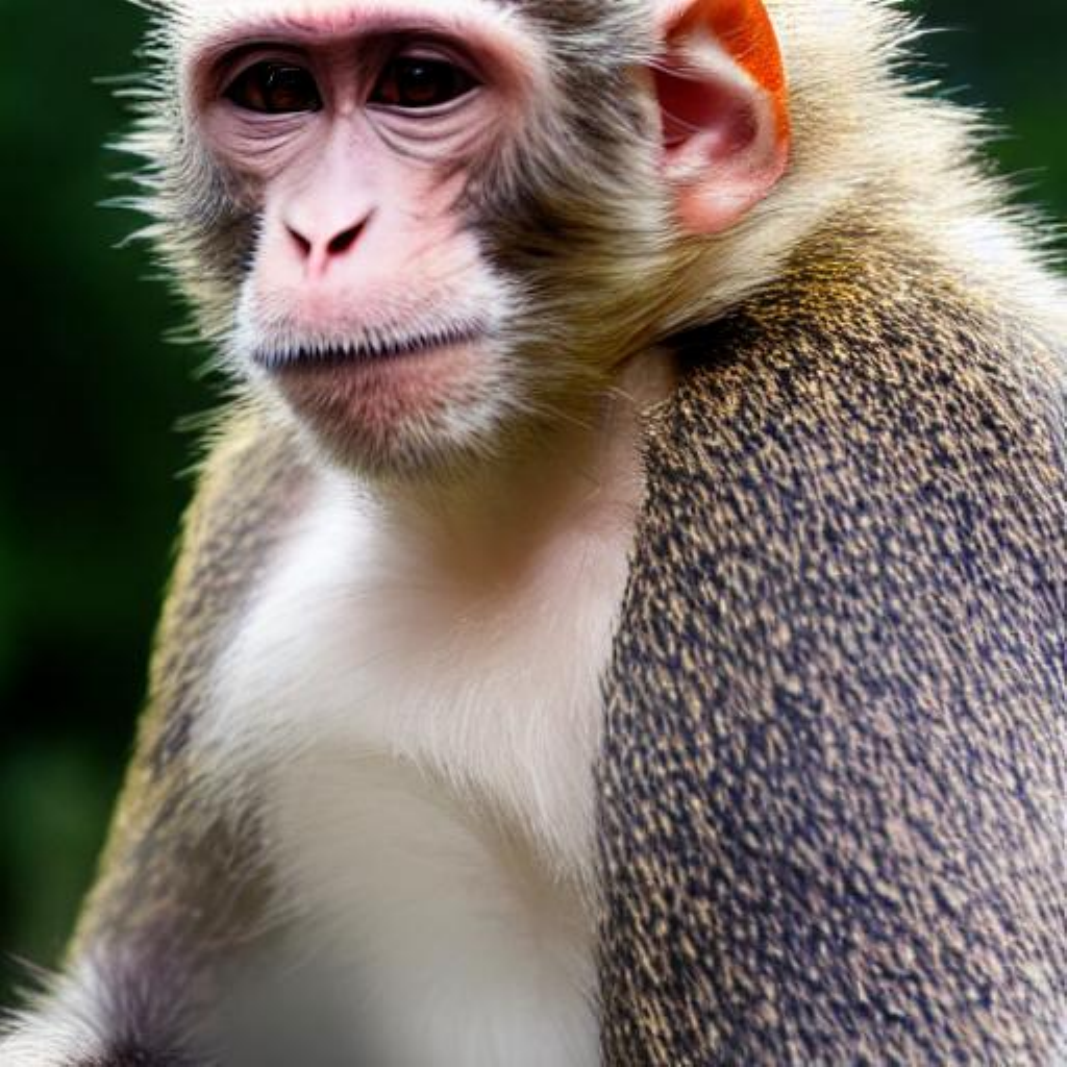} \end{subfigure}
    \begin{subfigure}[t]{0.19\textwidth} \includegraphics[width=\textwidth]{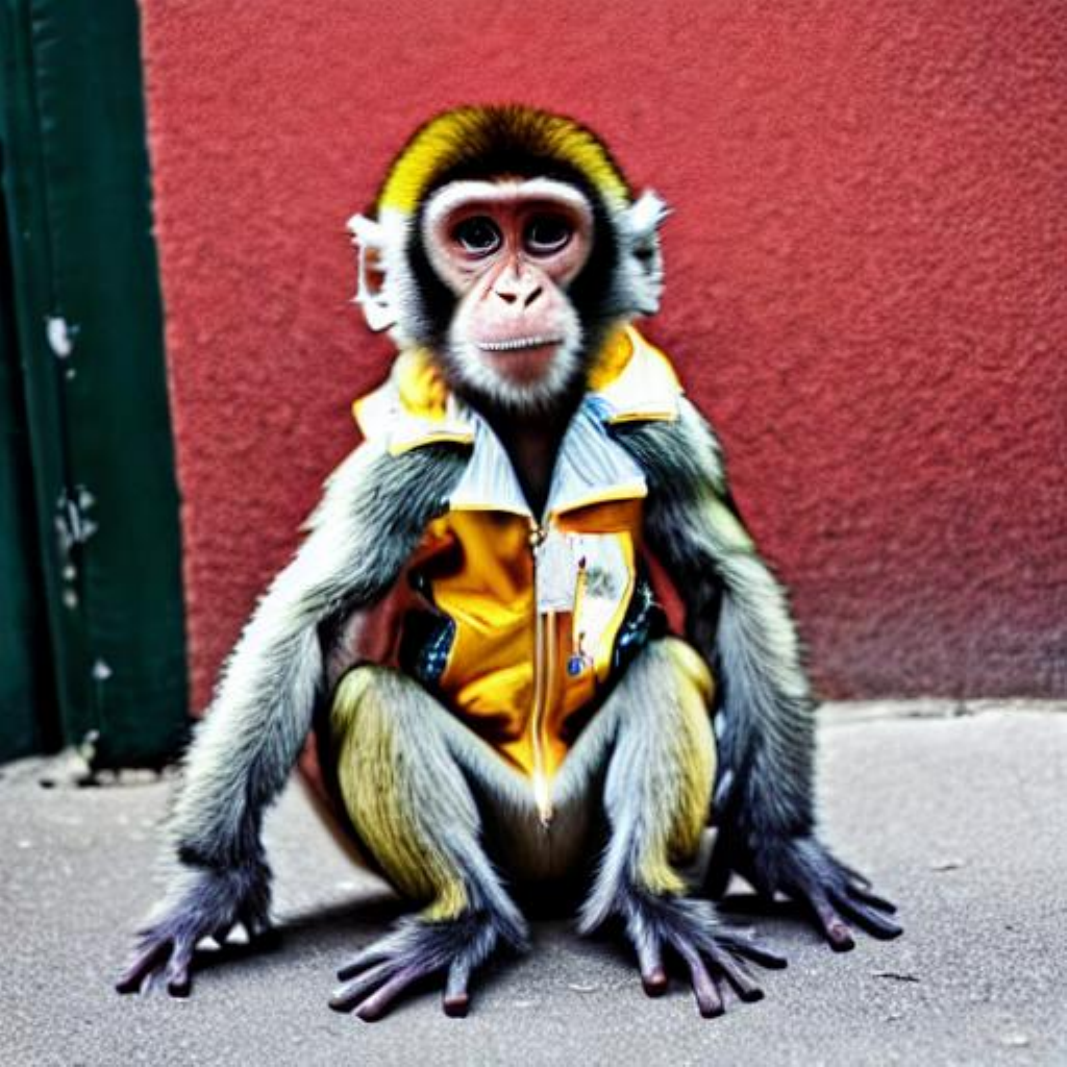} \end{subfigure}
    \begin{subfigure}[t]{0.19\textwidth} \includegraphics[width=\textwidth]{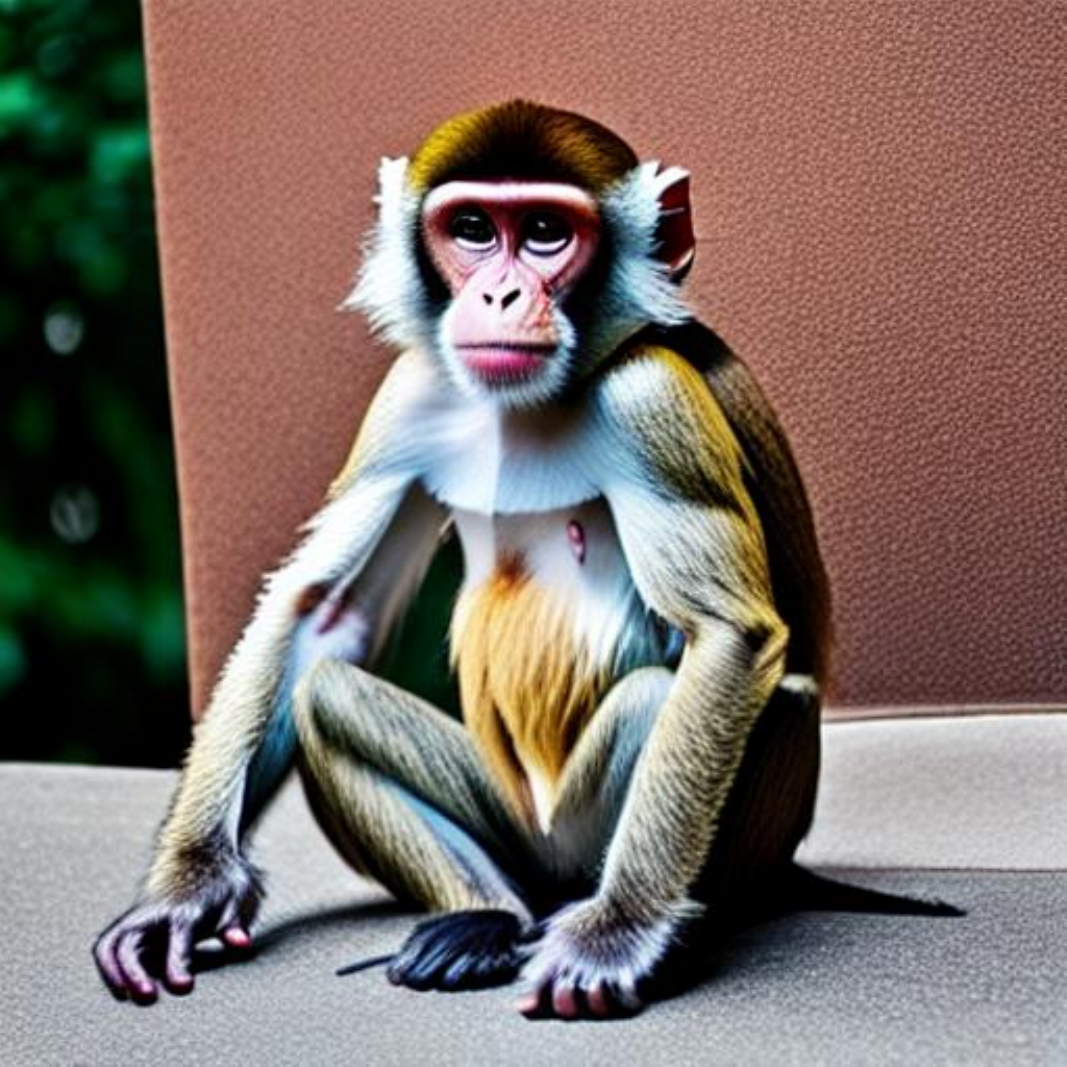} \end{subfigure}
    \begin{subfigure}[t]{0.19\textwidth} \includegraphics[width=\textwidth]{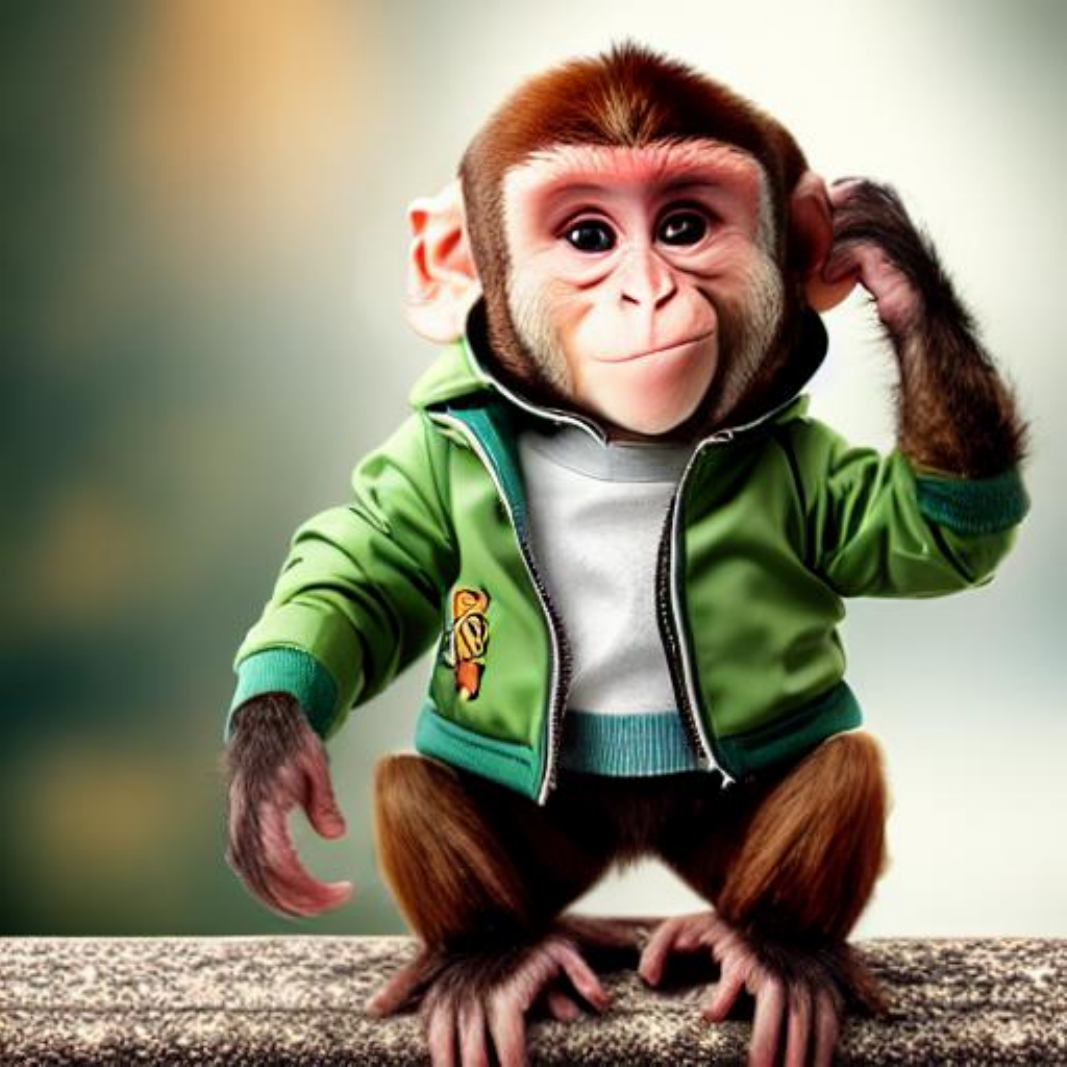} \end{subfigure}

    \parbox{1.02 \textwidth}{
        \centering
        An anime-style advertisement featuring a pizza and an explosion.

    }
    \begin{subfigure}[t]{0.19\textwidth} \includegraphics[width=\textwidth]{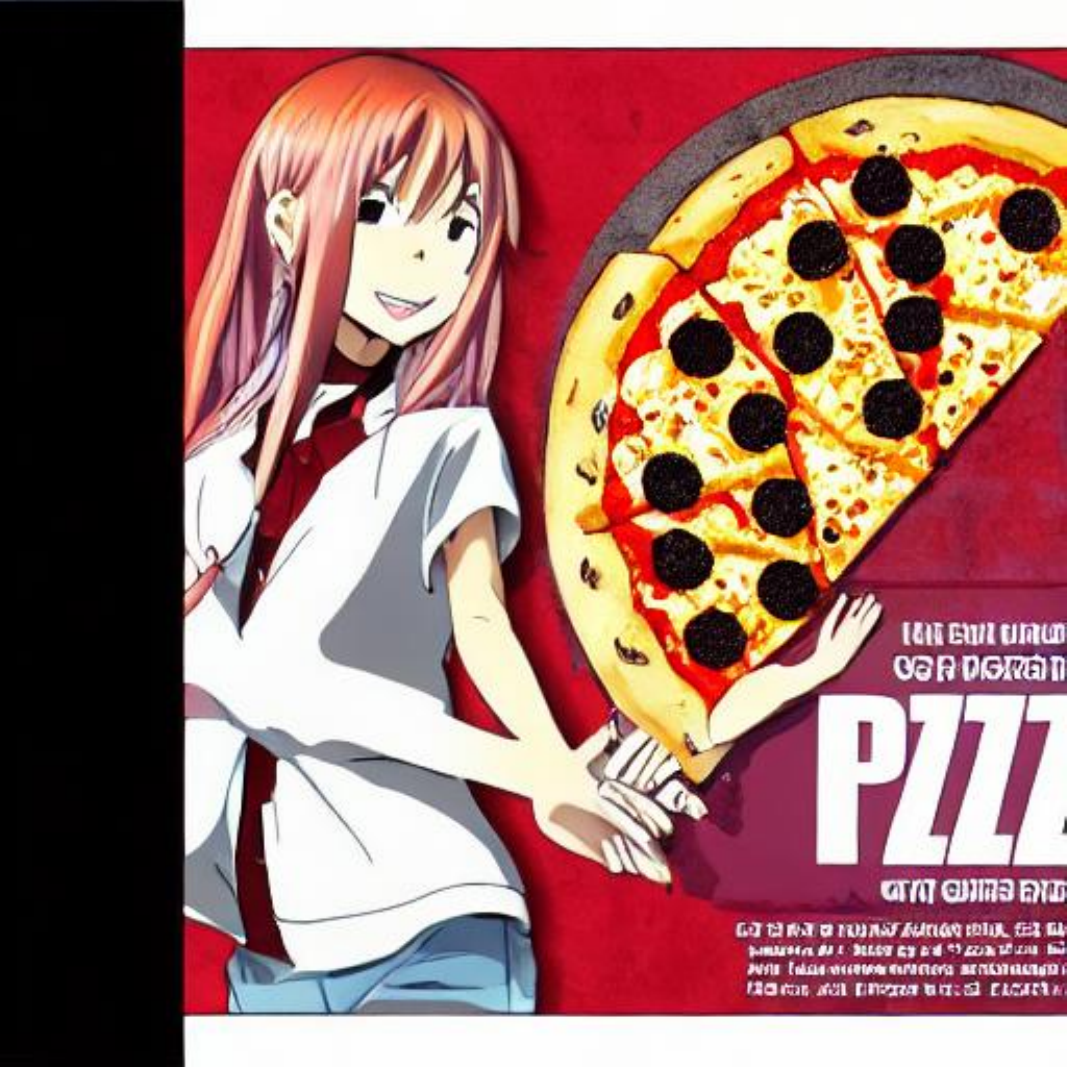} 
            \captionsetup{justification=centering}\caption{SD1.5~\cite{rombach2022high}}    \end{subfigure}
    \begin{subfigure}[t]{0.19\textwidth} \includegraphics[width=\textwidth]{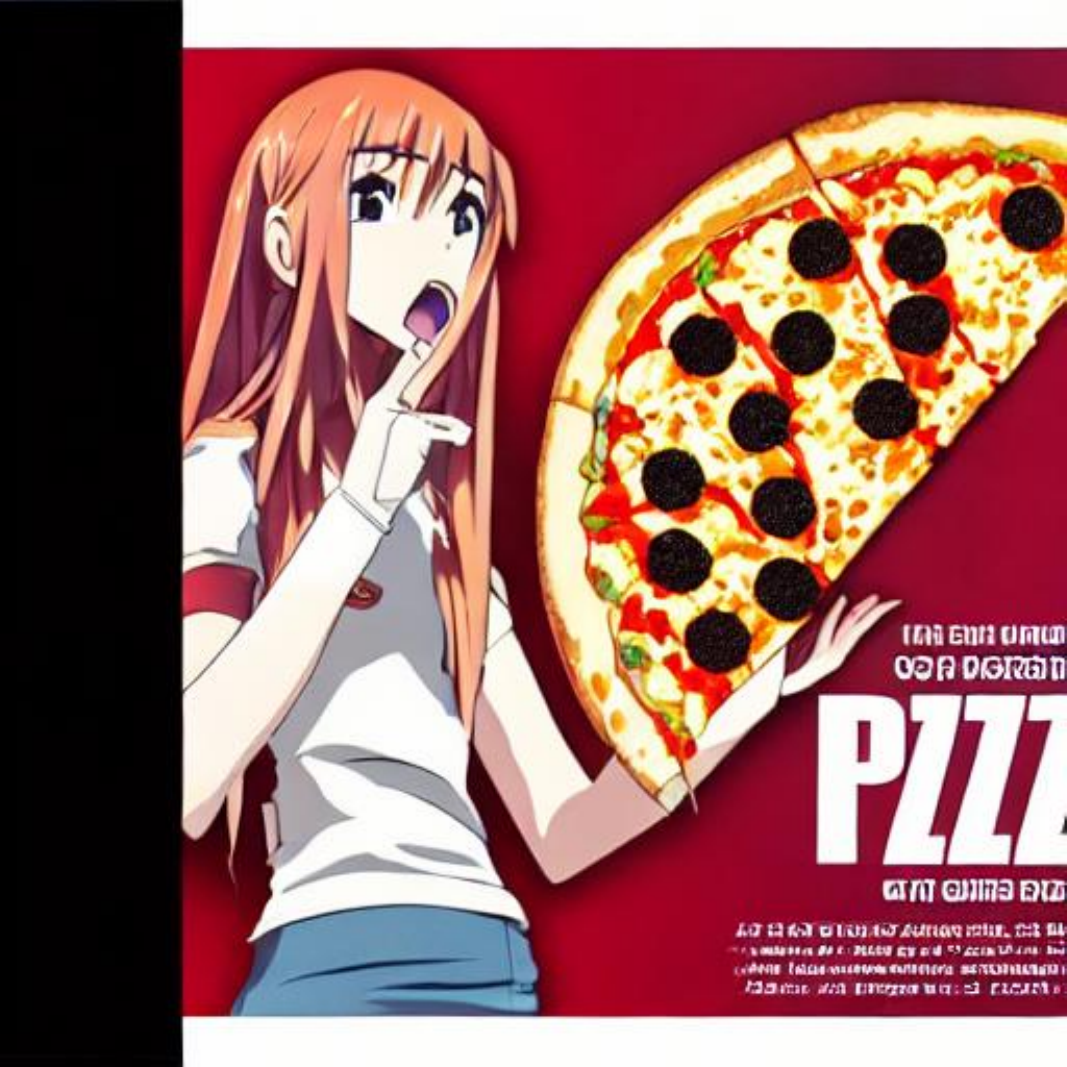} 
            \captionsetup{justification=centering}\caption{Diff-DPO~\cite{wallace2024diffusion}} \end{subfigure}
    \begin{subfigure}[t]{0.19\textwidth} \includegraphics[width=\textwidth]{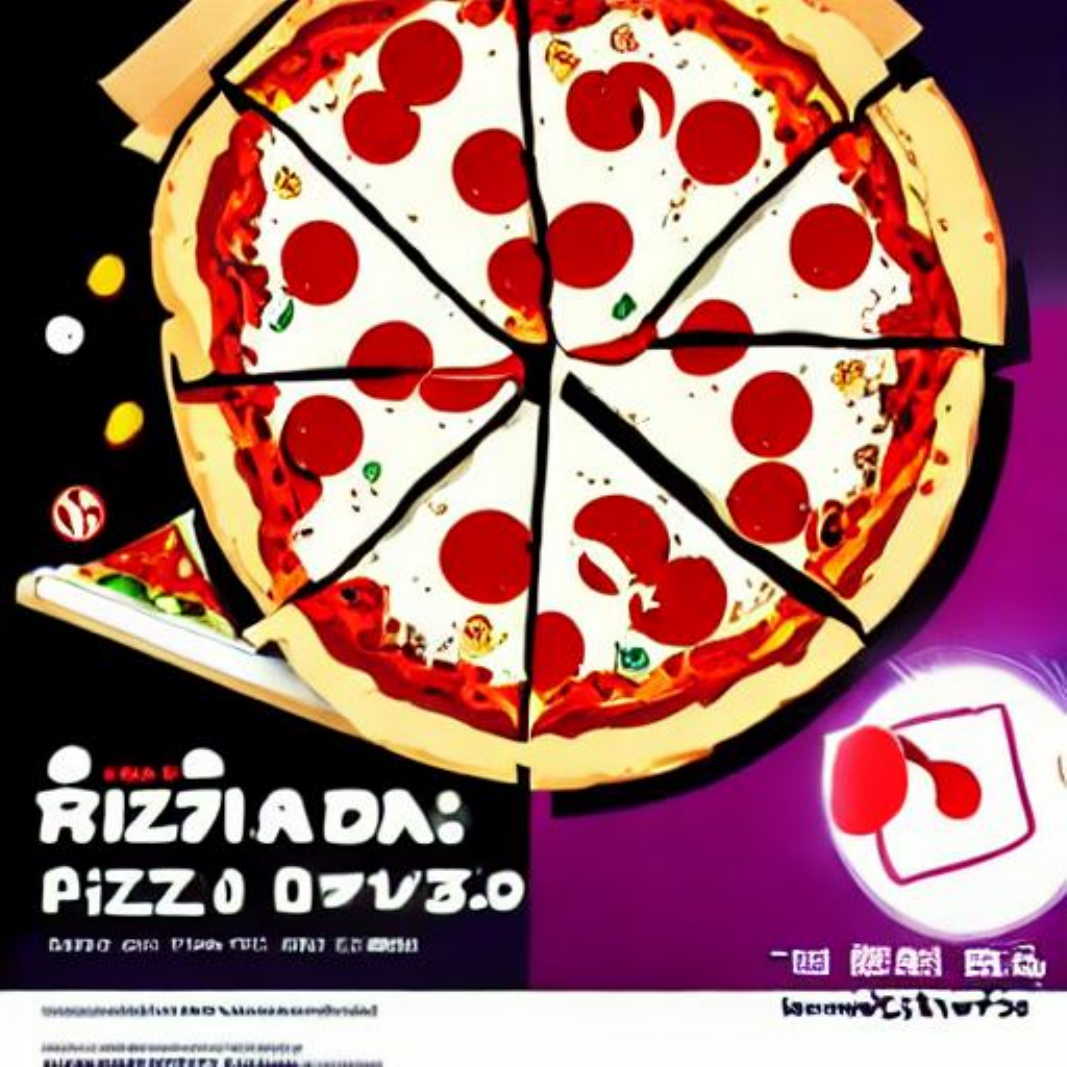}         \captionsetup{justification=centering}\caption{Diff-KTO~\cite{lialigning}} \end{subfigure}
    \begin{subfigure}[t]{0.19\textwidth} \includegraphics[width=\textwidth]{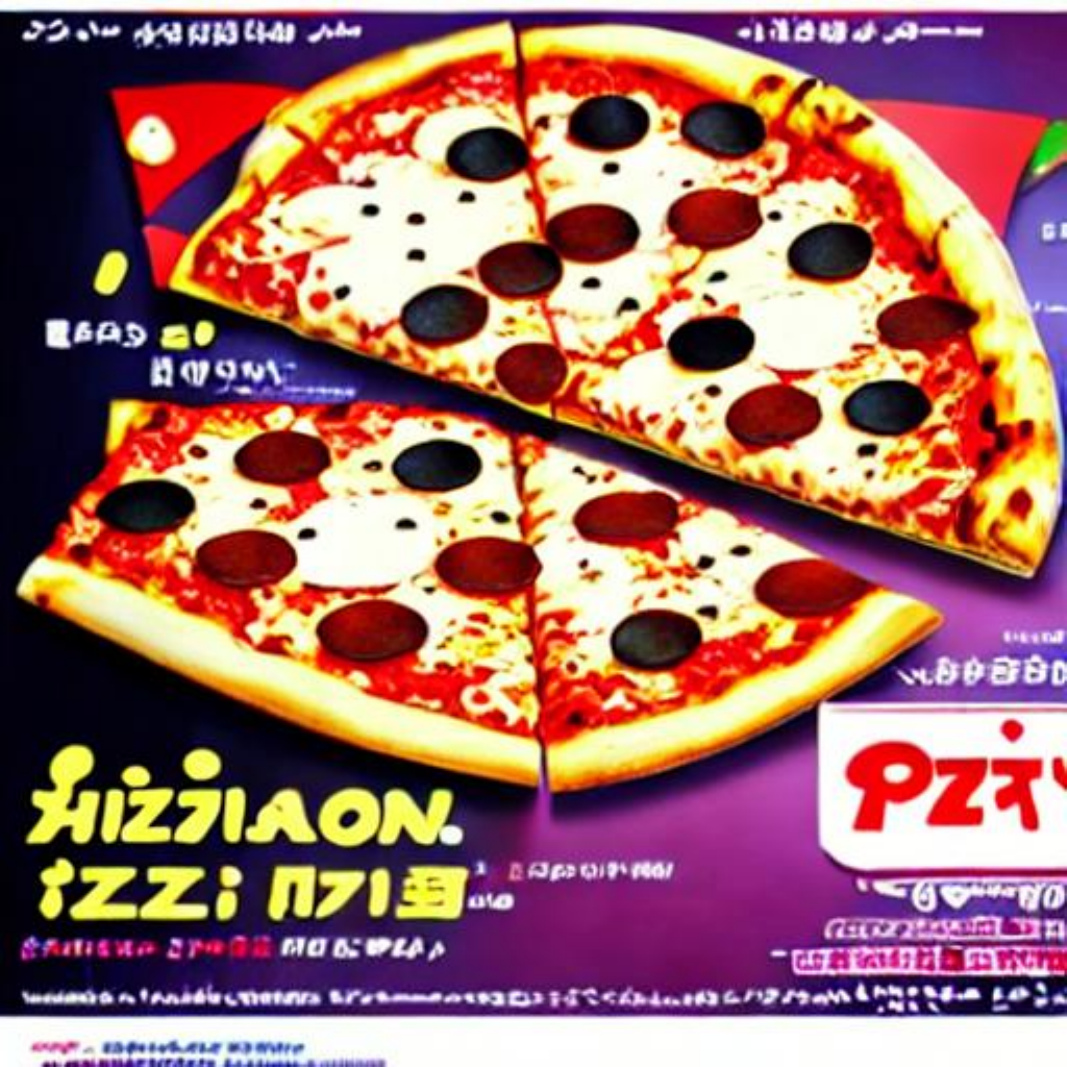}         \captionsetup{justification=centering}\caption{DSPO~\cite{zhu2025dspo}} \end{subfigure}
    \begin{subfigure}[t]{0.19\textwidth} \includegraphics[width=\textwidth]{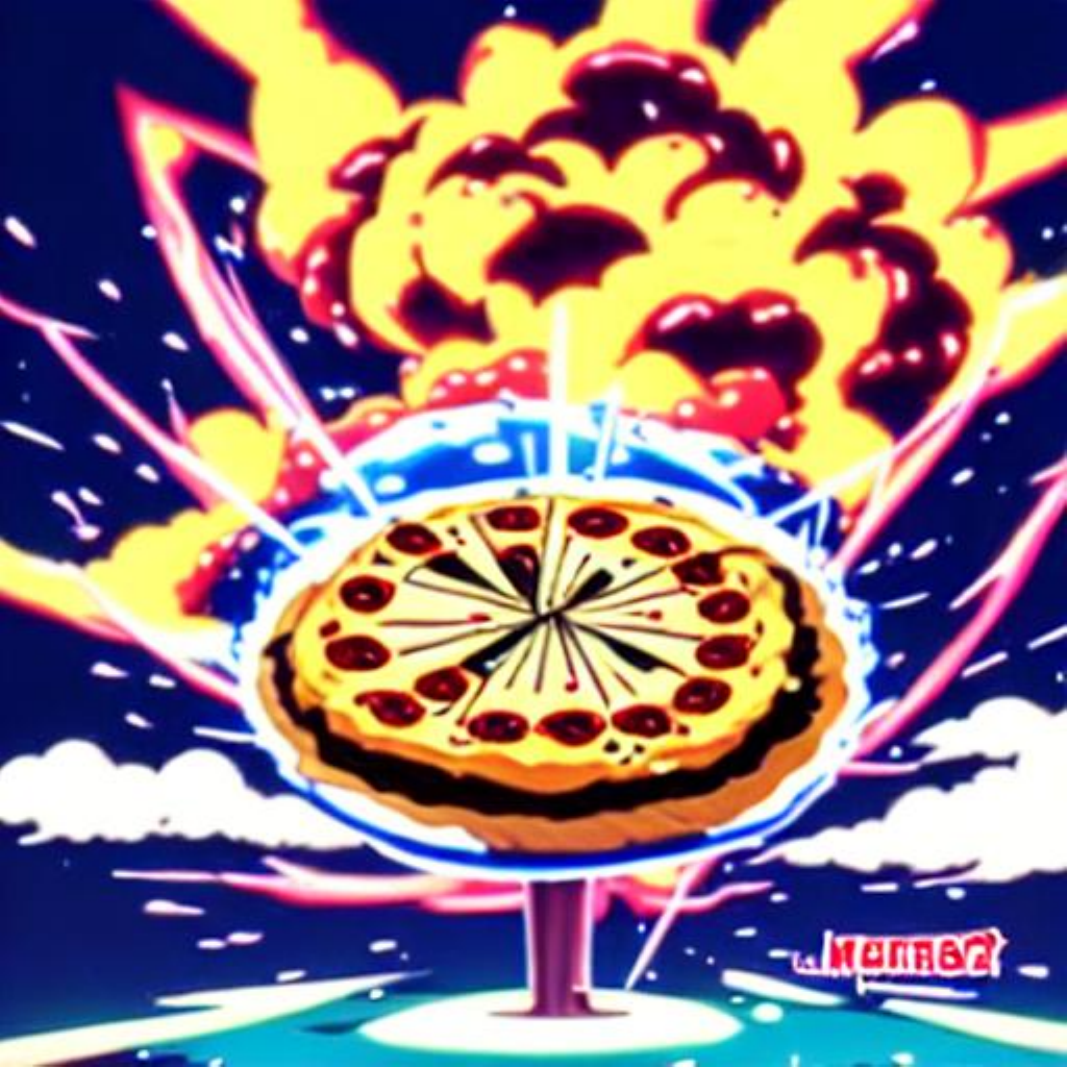}         \captionsetup{justification=centering}\caption{Ours} \end{subfigure}
    
    \caption{
        Qualitative comparisons on HPDv2 test set prompts.
    }
    \label{fig:appendix_qual_hpd}
\end{figure*}

\begin{figure*}[hb!]
    \centering
    \parbox{1.02 \textwidth}{
        \centering
        \large
           A woman with a pearl earring, blue eyes,in the style of blue and khaki, \\
           smiling and happy, meticulous, solapunk, li - core
    }
    \begin{subfigure}[b]{1.02\textwidth}
        \centering
        \begin{subfigure}[b]{0.45\textwidth}
            \centering
            \begin{subfigure}[b]{0.48\textwidth}
                \includegraphics[width=\linewidth]{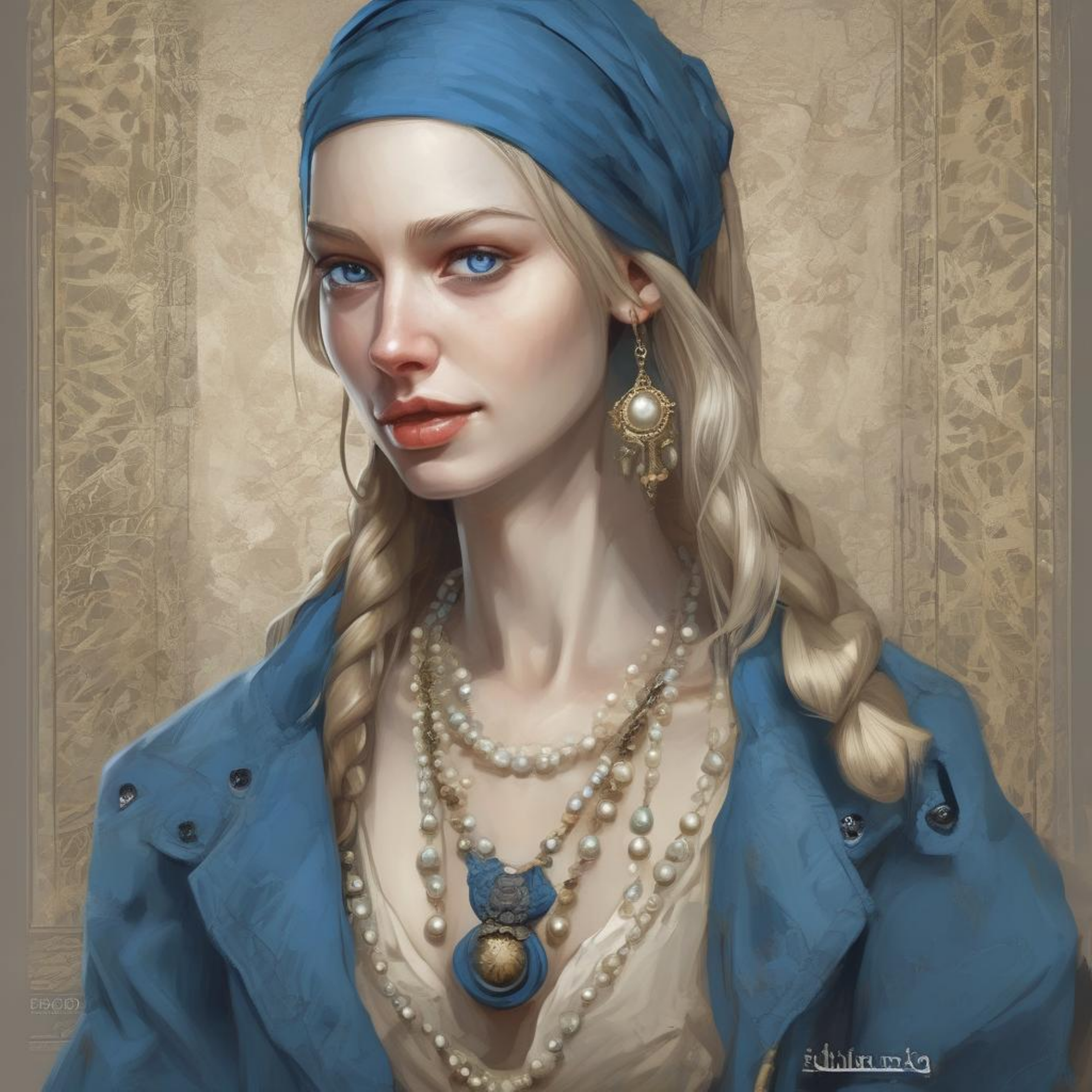}
                \caption{SDXL}
            \end{subfigure}
            \hfill
            \begin{subfigure}[b]{0.48\textwidth}
                \includegraphics[width=\linewidth]{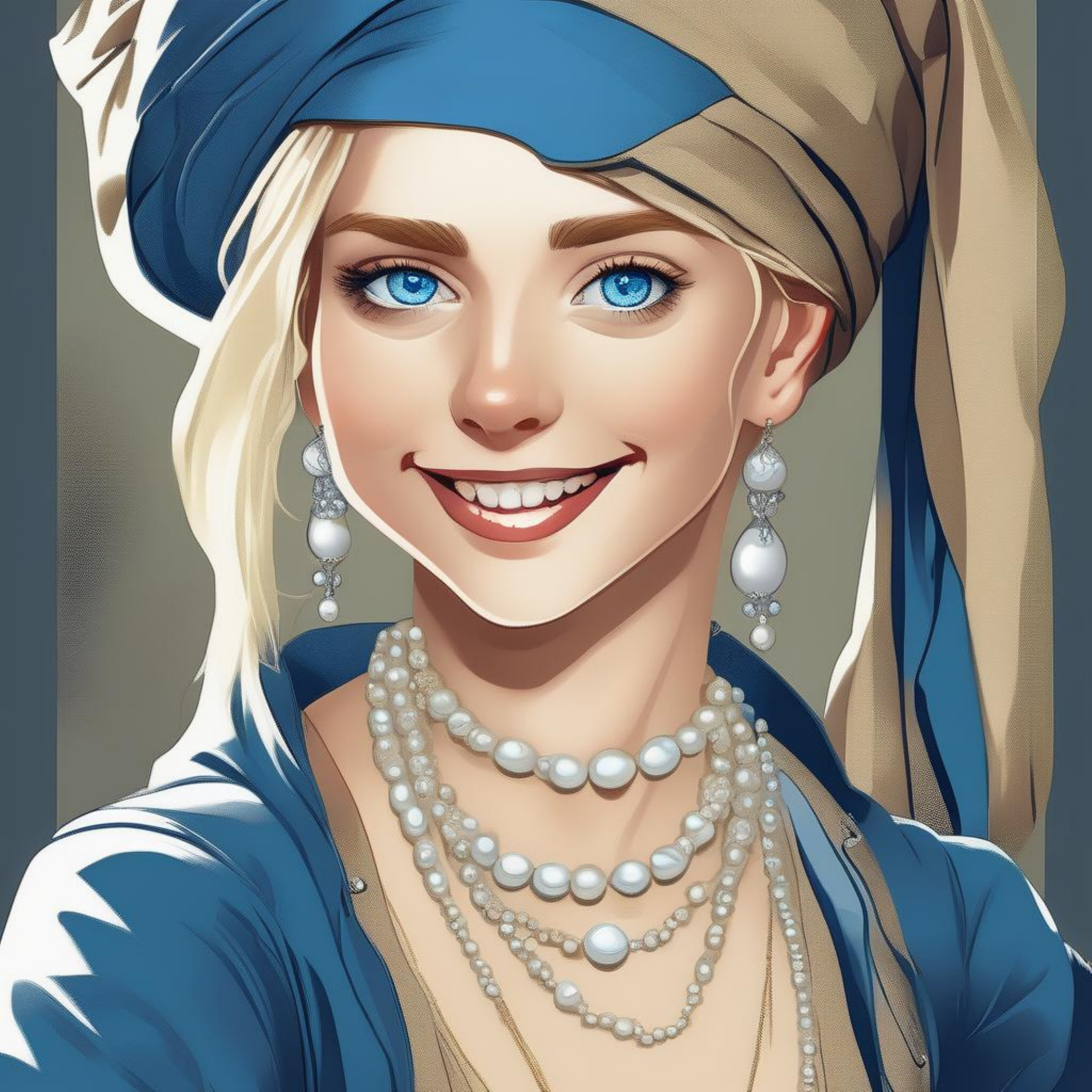}
                \caption{Diff-DPO}
            \end{subfigure}

            \begin{subfigure}[b]{0.48\textwidth}
                \includegraphics[width=\linewidth]{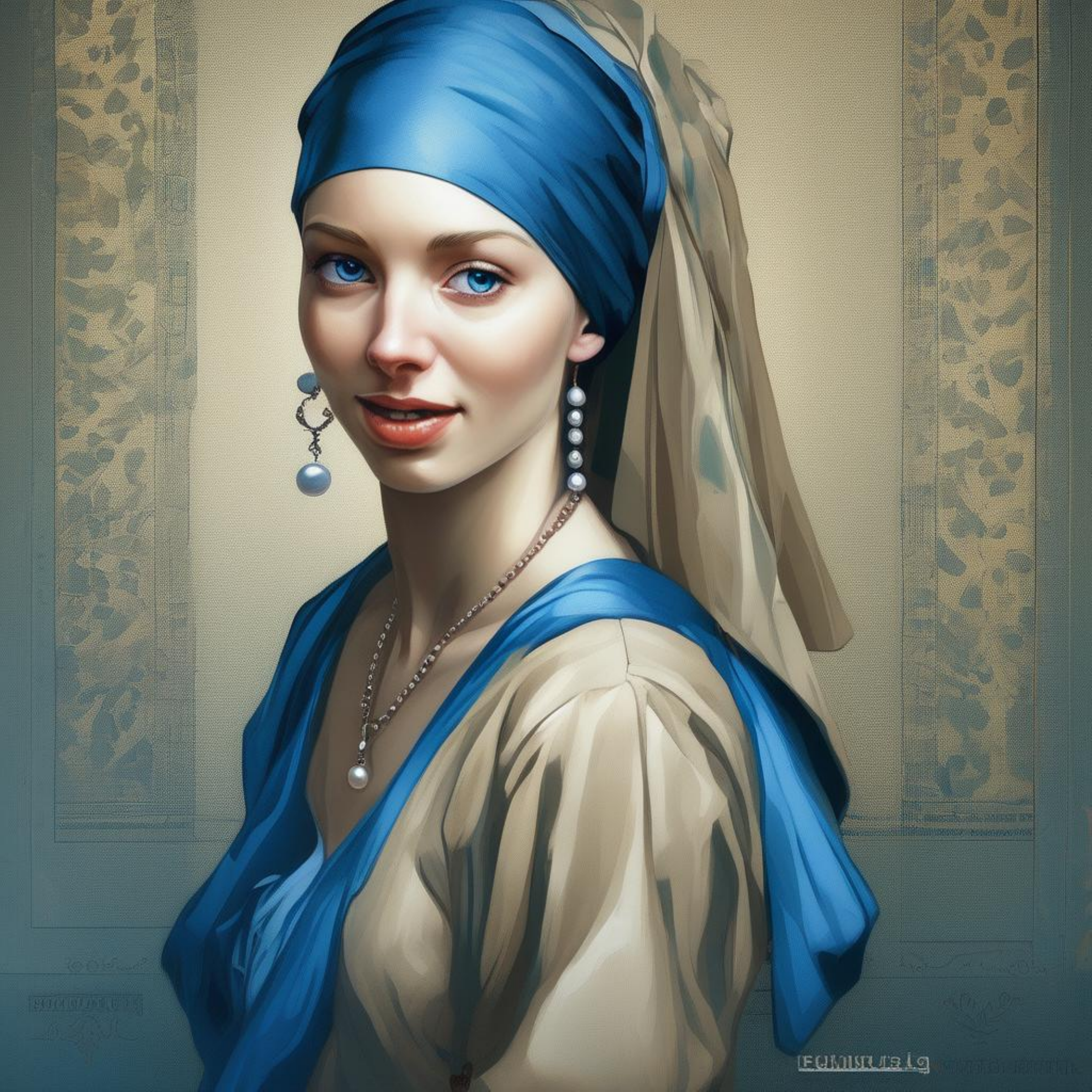}
                \caption{MAPO}
            \end{subfigure}
            \hfill
            \begin{subfigure}[b]{0.48\textwidth}
                \includegraphics[width=\linewidth]{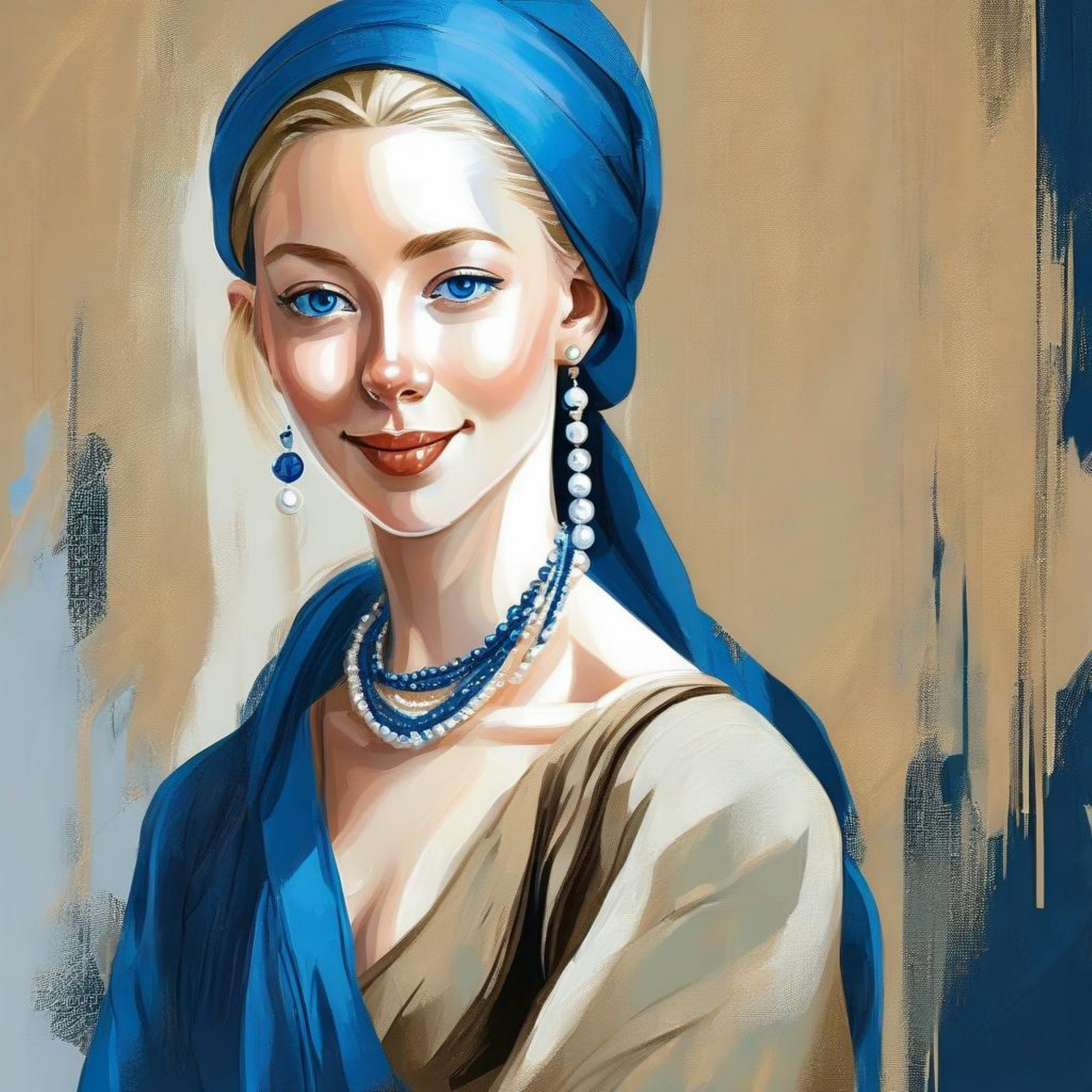}
                \caption{DSPO}
            \end{subfigure}
        \end{subfigure}
        \hfill
        \begin{subfigure}[b]{0.465\textwidth}
            \centering
            \includegraphics[width=\linewidth]{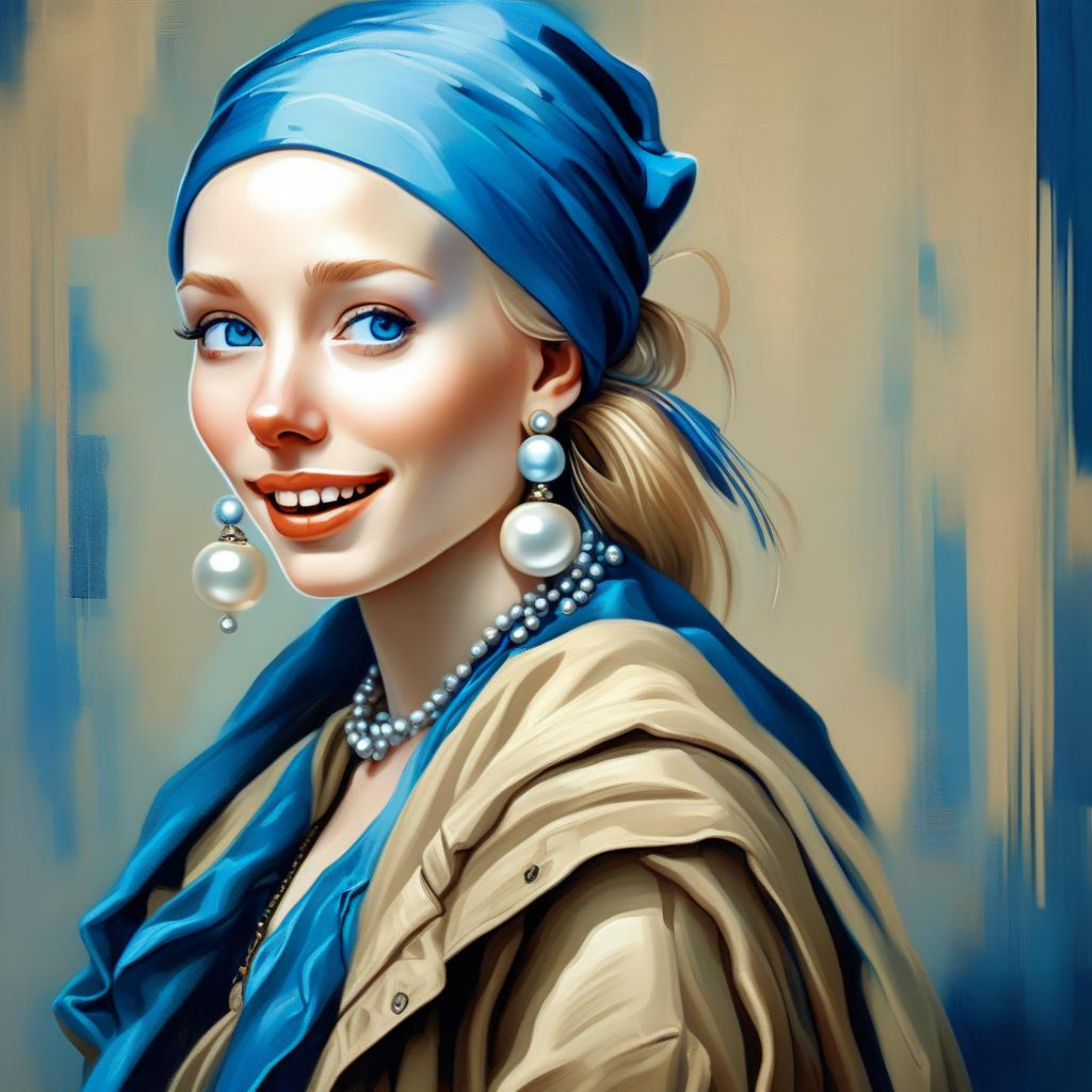}
            \caption{Ours}
        \end{subfigure}
    \end{subfigure}

    \setcounter{subfigure}{0}

    \parbox{1.02 \textwidth}{
        \centering
        \large
           smily french fries
    }
    \begin{subfigure}[b]{1.02\textwidth}
        \centering
        \begin{subfigure}[b]{0.45\textwidth}
            \centering
            \begin{subfigure}[b]{0.48\textwidth}
                \includegraphics[width=\linewidth]{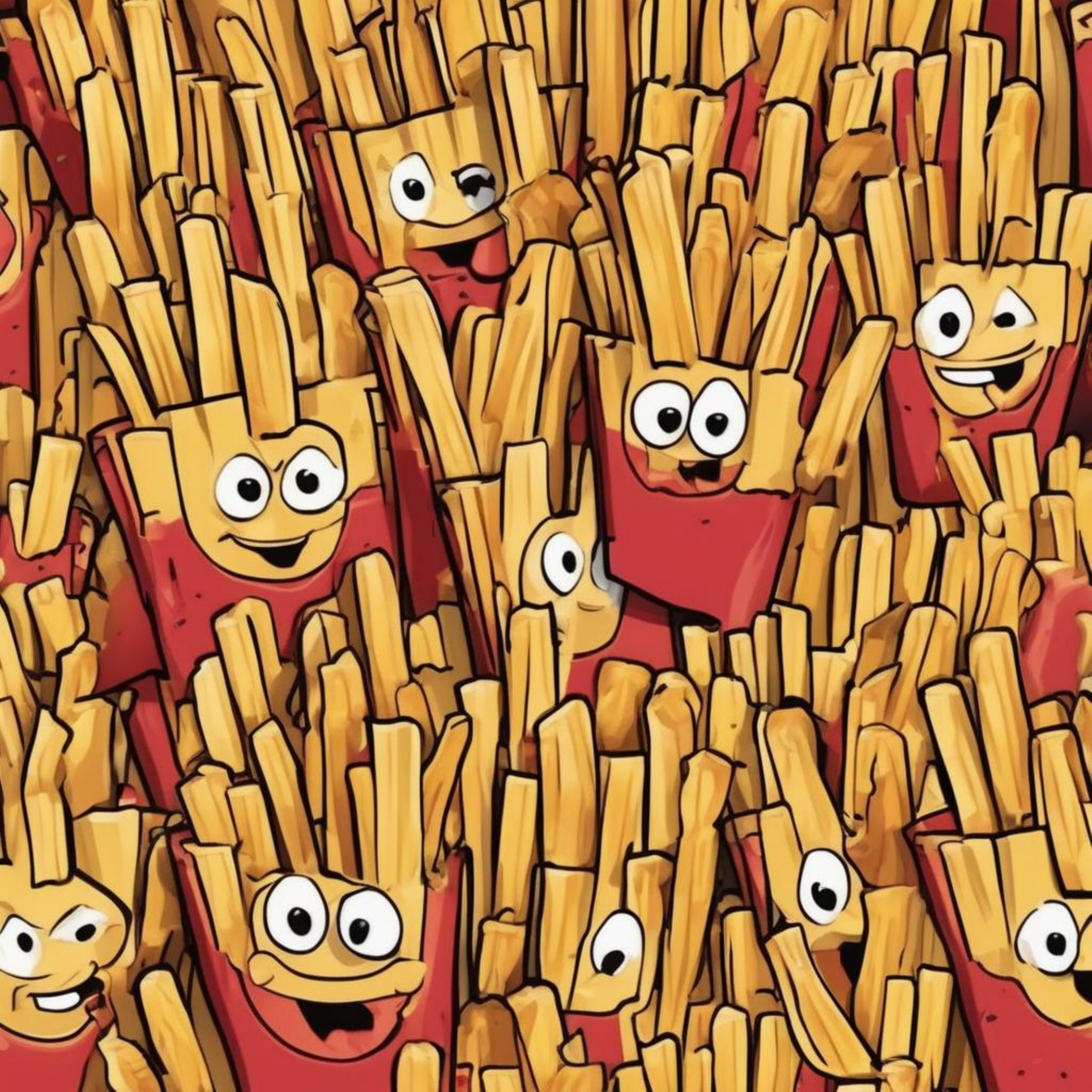}
                \caption{SDXL}
            \end{subfigure}
            \hfill
            \begin{subfigure}[b]{0.48\textwidth}
                \includegraphics[width=\linewidth]{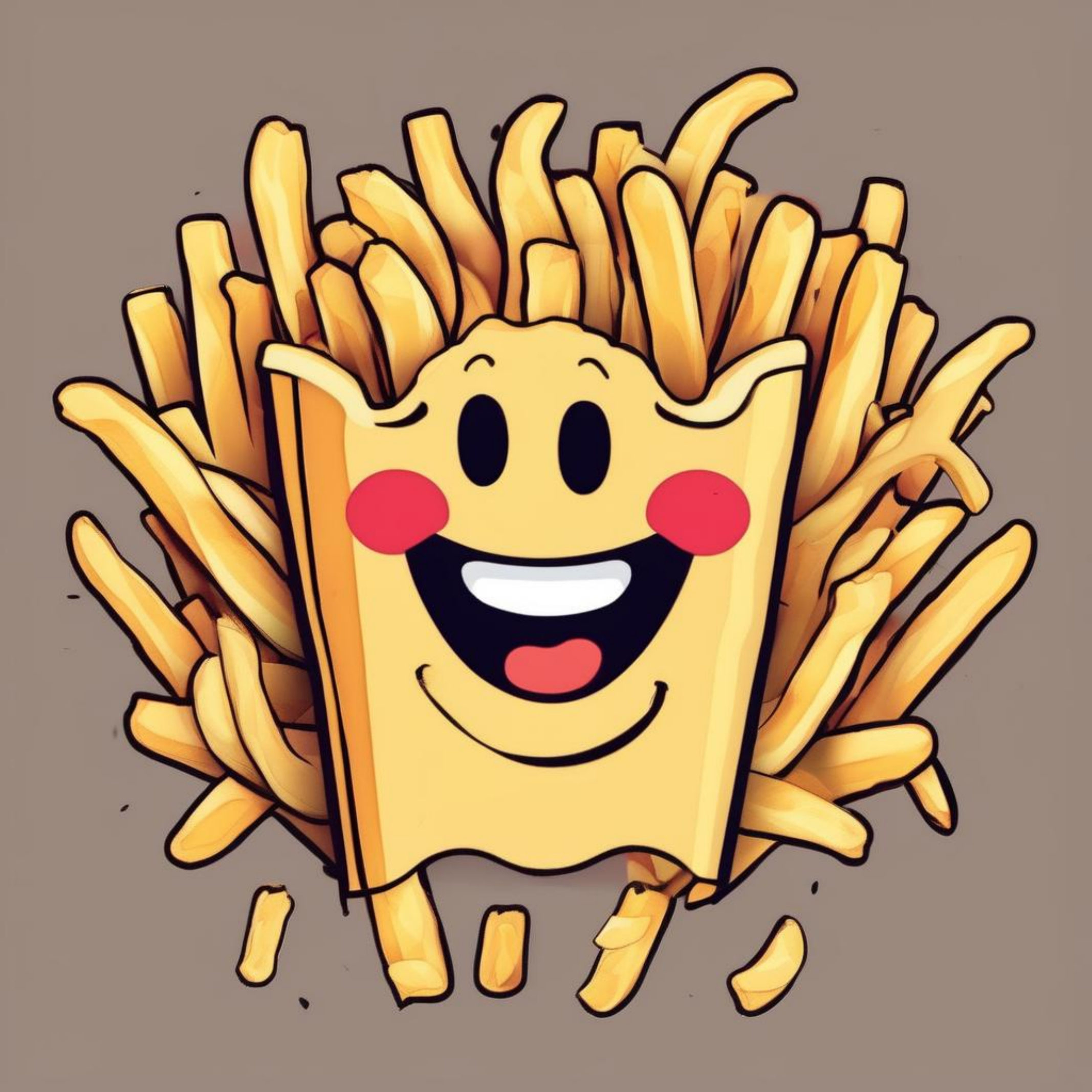}
                \caption{Diff-DPO}
            \end{subfigure}

            \begin{subfigure}[b]{0.48\textwidth}
                \includegraphics[width=\linewidth]{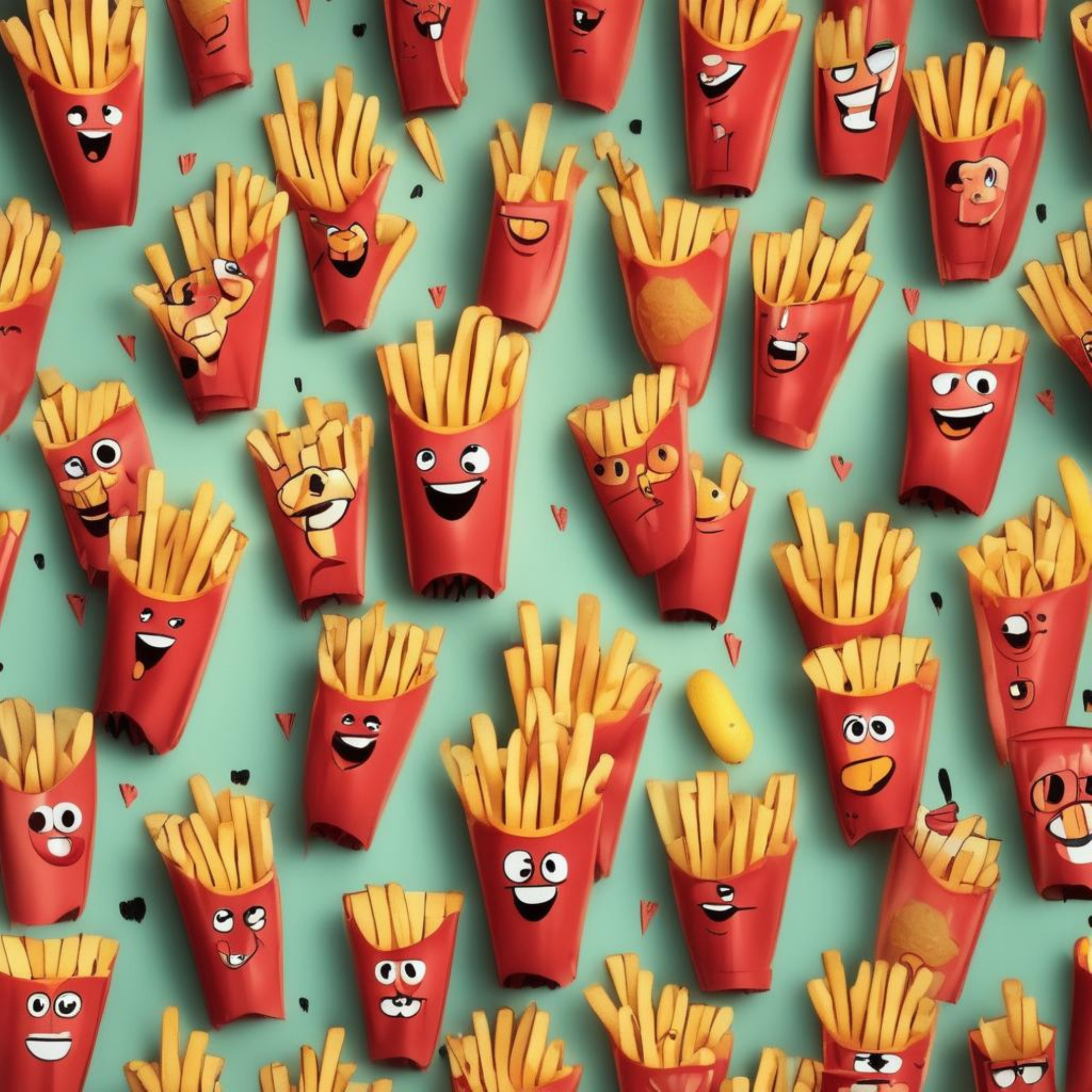}
                \caption{MAPO}
            \end{subfigure}
            \hfill
            \begin{subfigure}[b]{0.48\textwidth}
                \includegraphics[width=\linewidth]{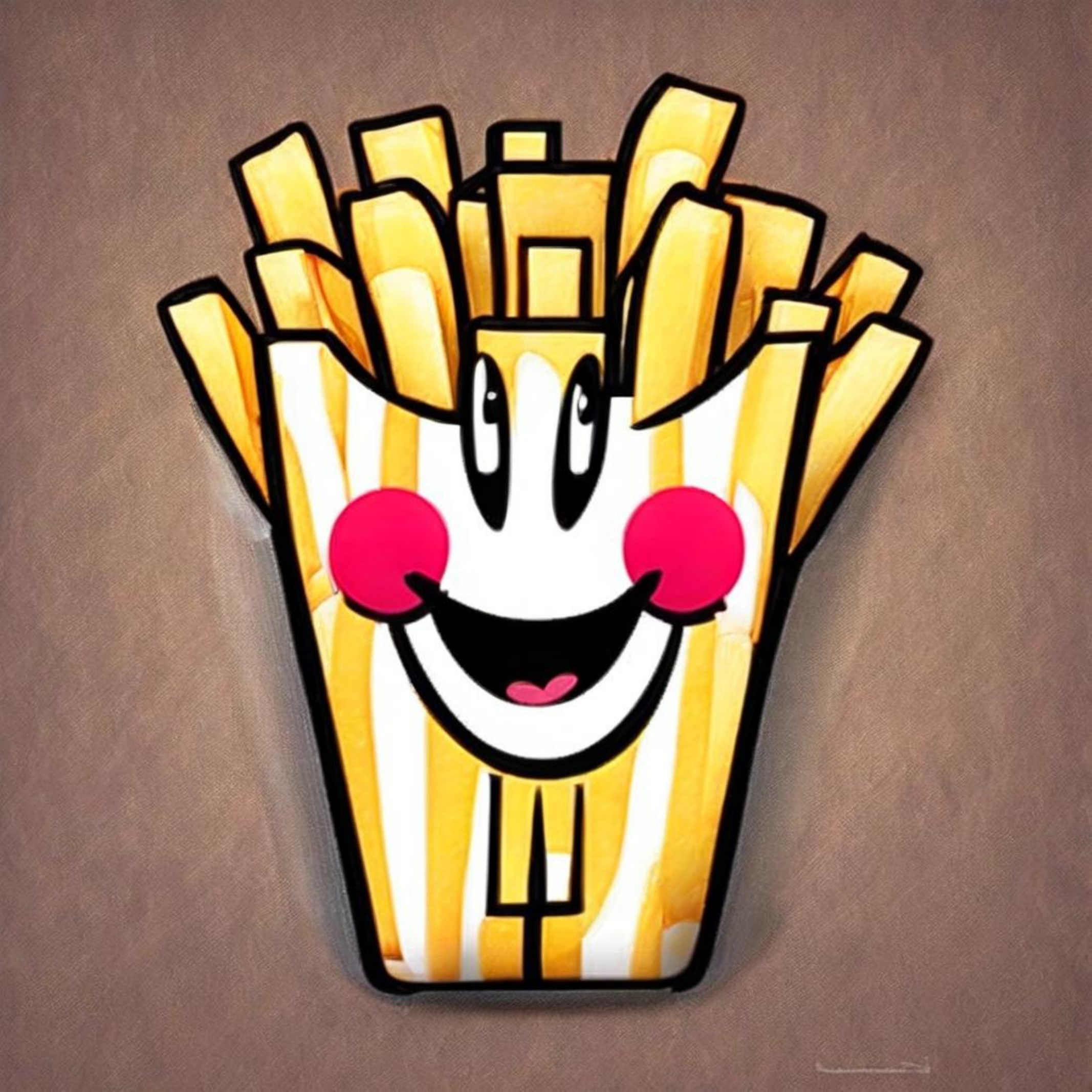}
                \caption{DSPO}
            \end{subfigure}
        \end{subfigure}
        \hfill
        \begin{subfigure}[b]{0.465\textwidth}
            \centering
            \includegraphics[width=\linewidth]{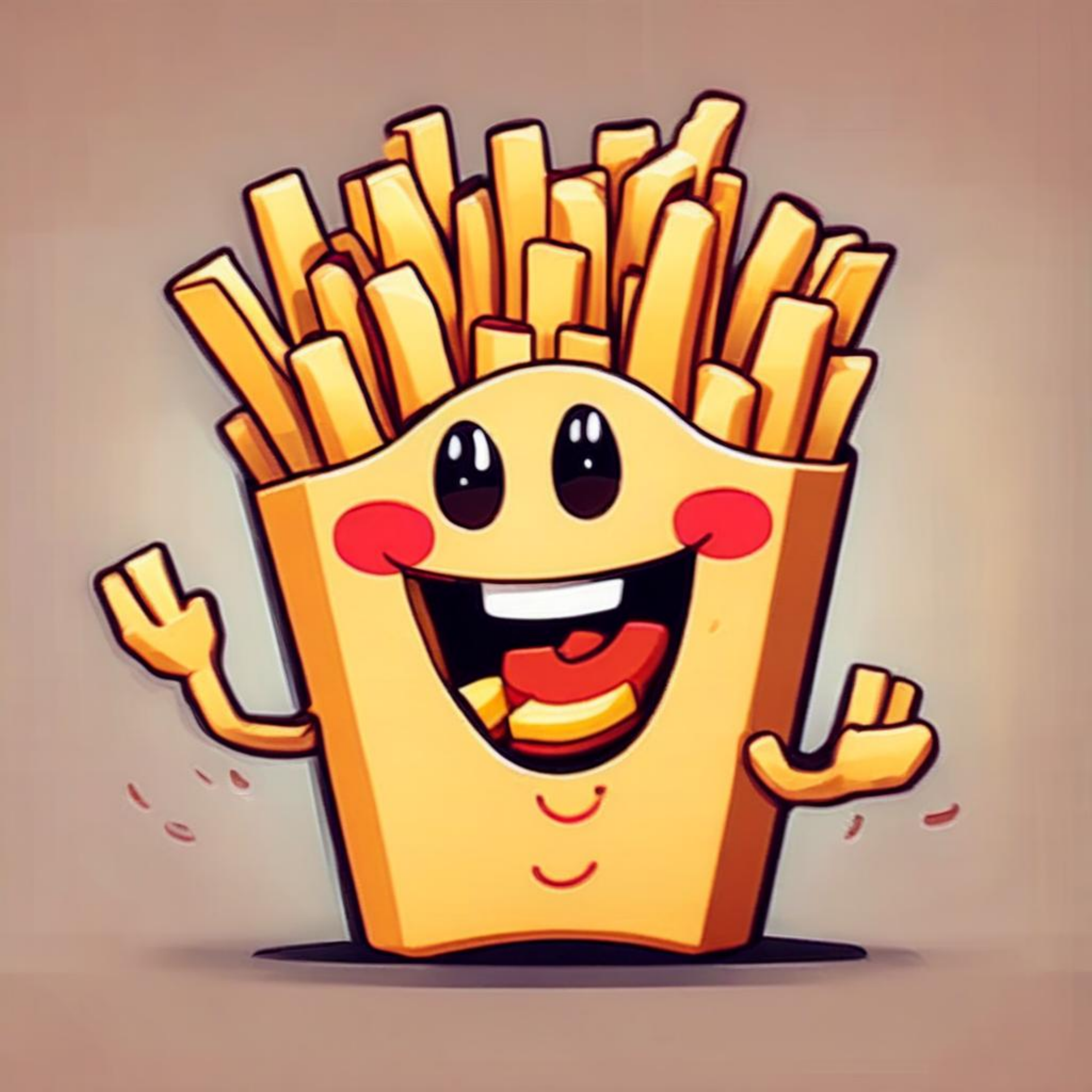}
            \caption{Ours}
        \end{subfigure}
    \end{subfigure}

    \caption{ Qualitative comparisons with the SDXL model.} 
    \label{fig:appendix_sdxl_first}
\end{figure*}

\begin{figure*}[htbp]
    \centering
    \parbox{1.02 \textwidth}{
        \centering
        \large
 a cute tiny bird wondering around water        
    }
    \begin{subfigure}[b]{1.02\textwidth}
        \centering
        \begin{subfigure}[b]{0.45\textwidth}
            \centering
            \begin{subfigure}[b]{0.48\textwidth}
                \includegraphics[width=\linewidth]{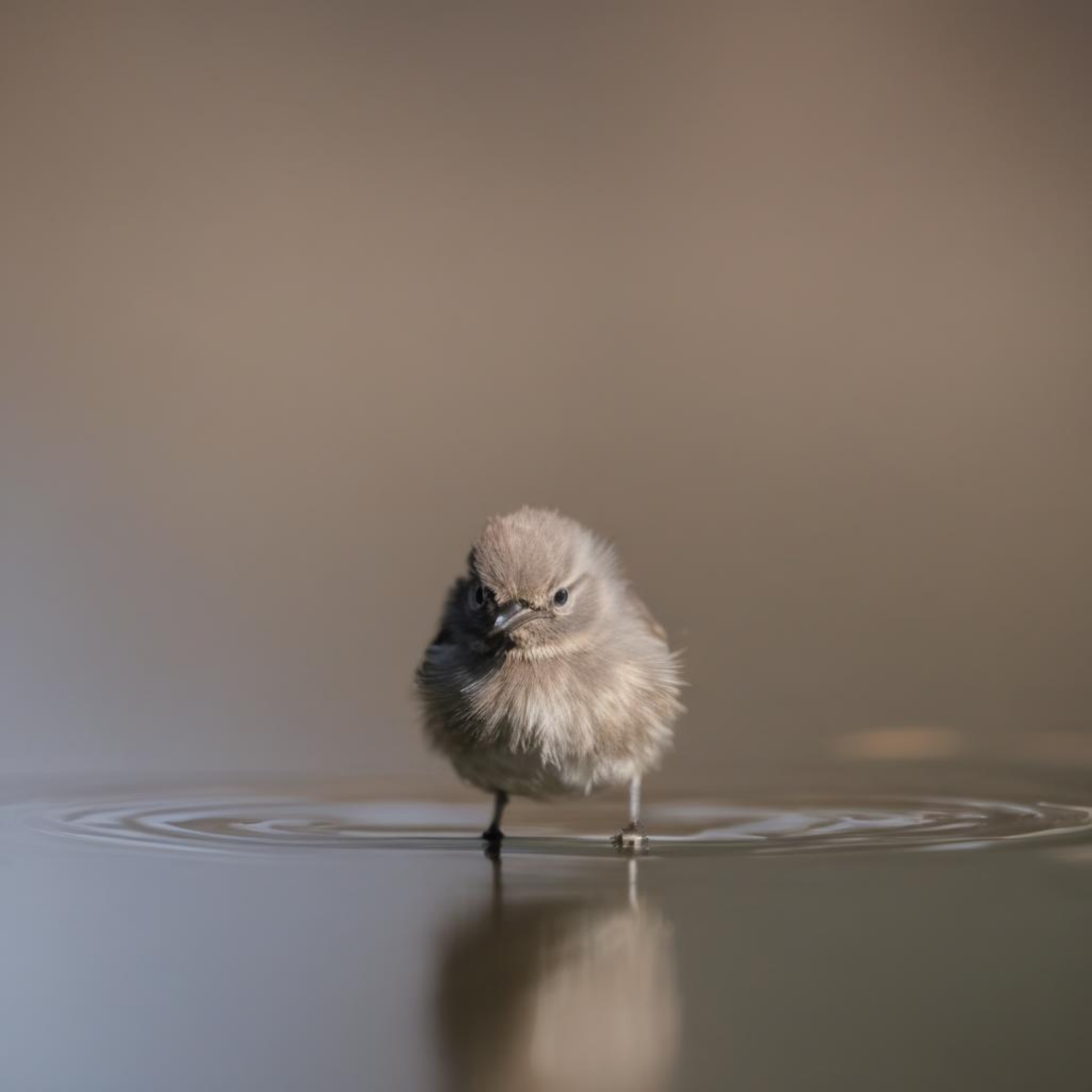}
                \caption{SDXL}
            \end{subfigure}
            \hfill
            \begin{subfigure}[b]{0.48\textwidth}
                \includegraphics[width=\linewidth]{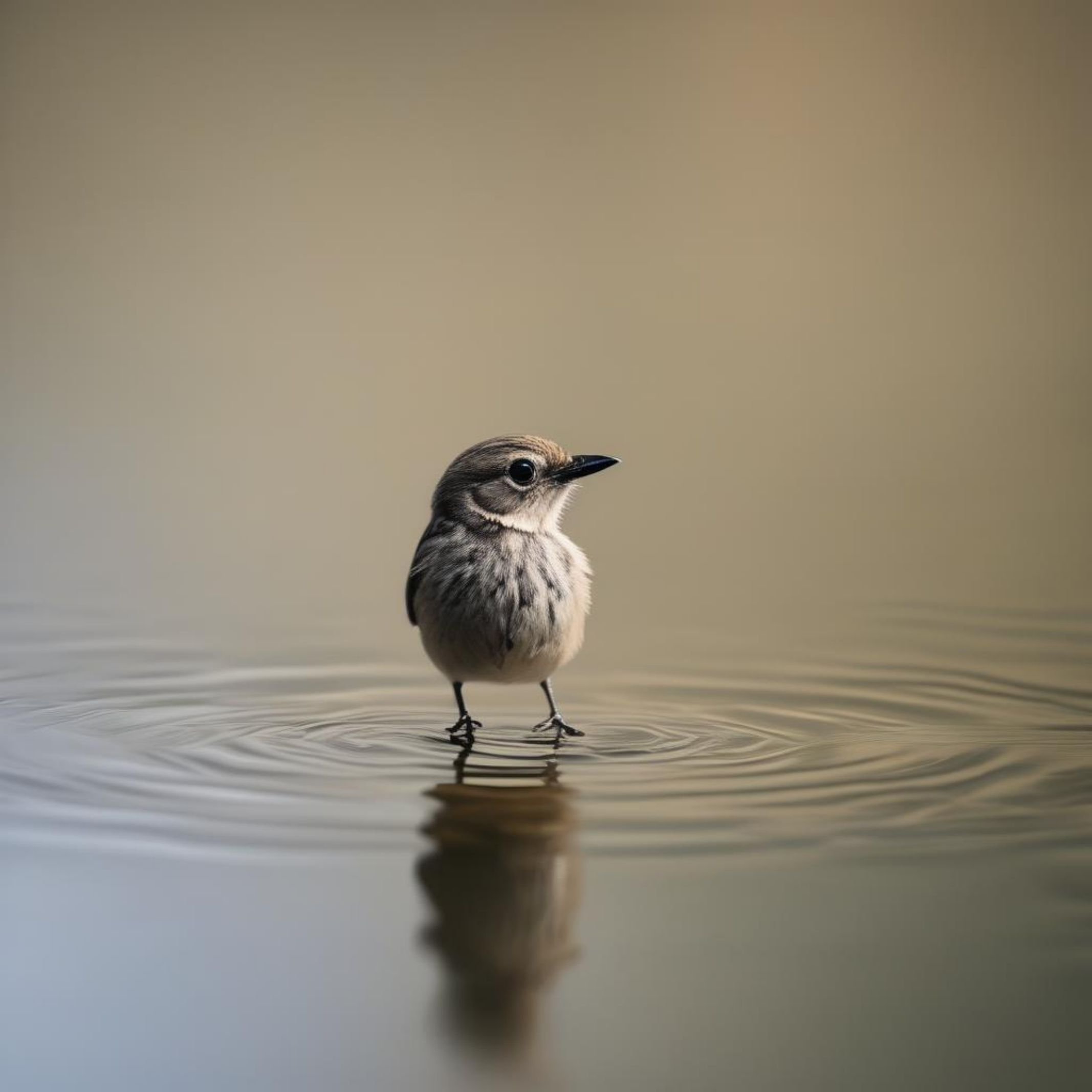}
                \caption{Diff-DPO}
            \end{subfigure}

            \begin{subfigure}[b]{0.48\textwidth}
                \includegraphics[width=\linewidth]{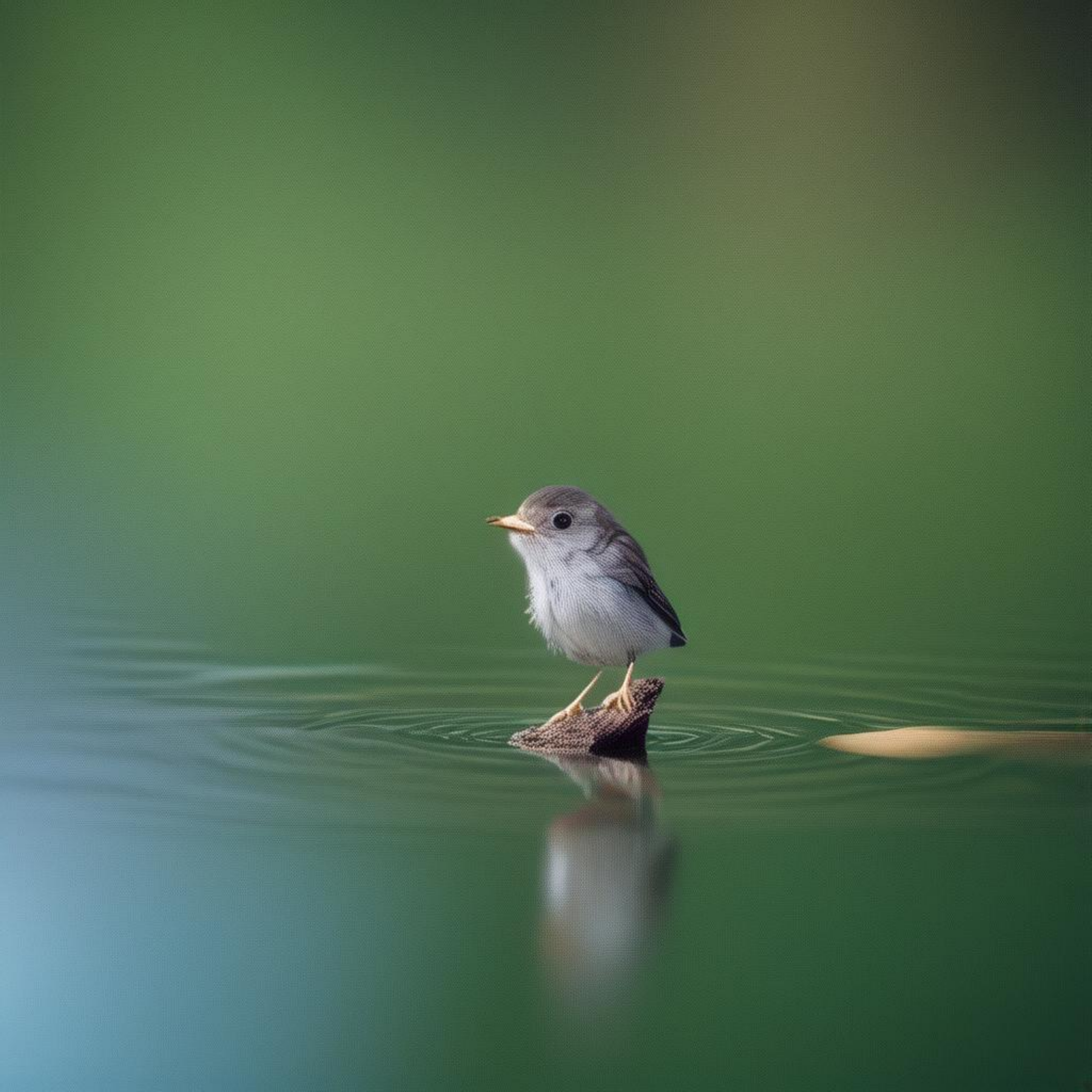}
                \caption{MAPO}
            \end{subfigure}
            \hfill
            \begin{subfigure}[b]{0.48\textwidth}
                \includegraphics[width=\linewidth]{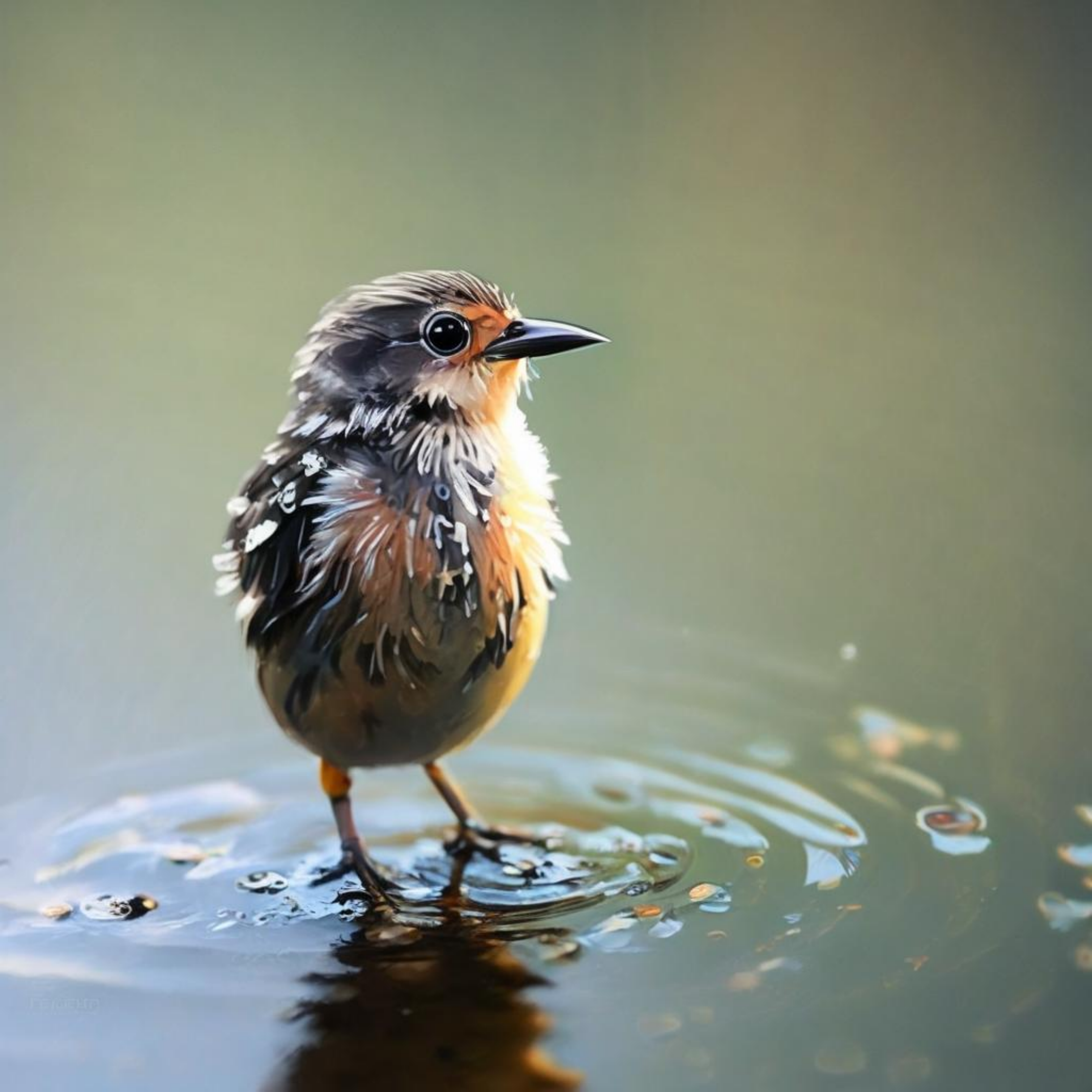}
                \caption{DSPO}
            \end{subfigure}
        \end{subfigure}
        \hfill
        \begin{subfigure}[b]{0.465\textwidth}
            \centering
            \includegraphics[width=\linewidth]{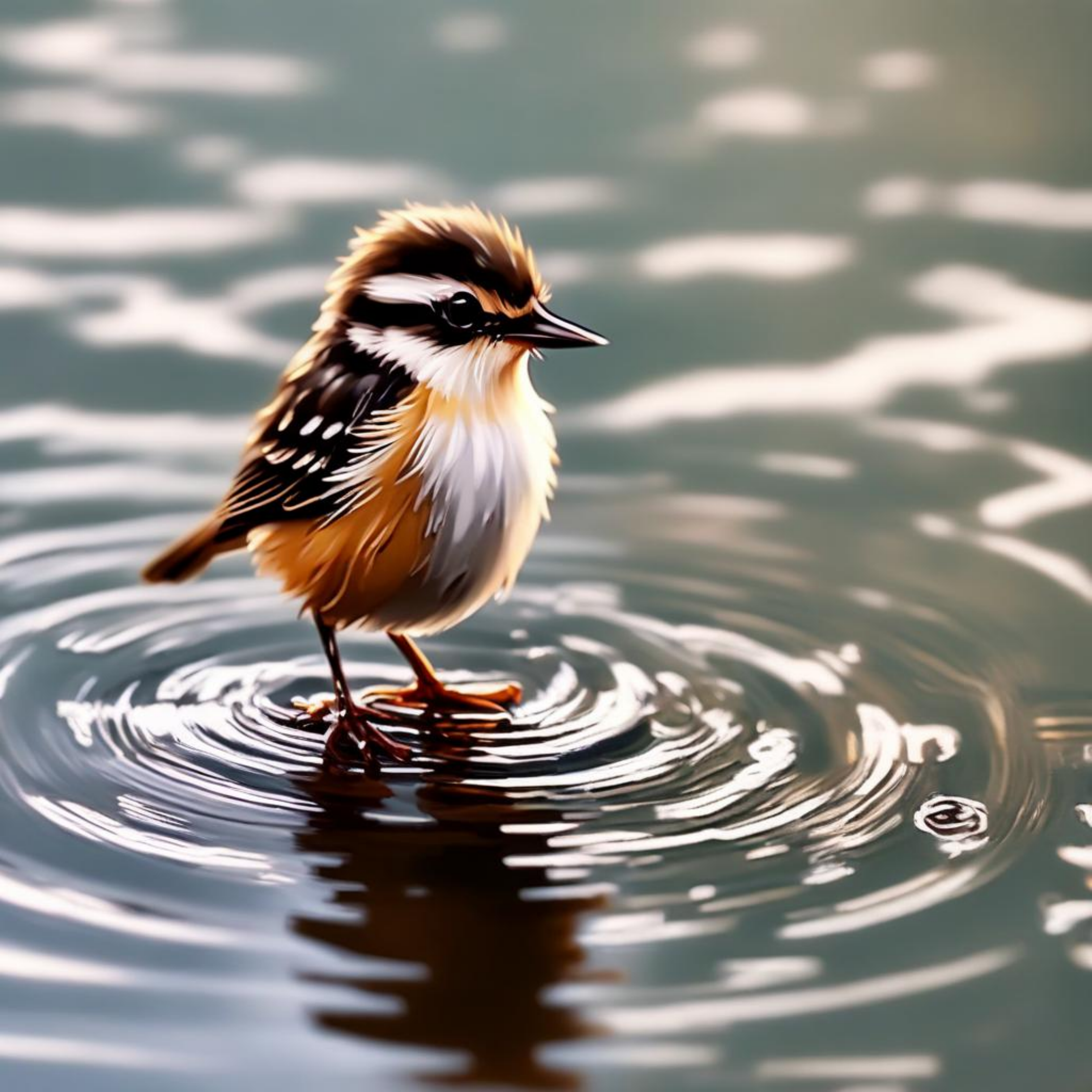}
            \caption{Ours}
        \end{subfigure}
    \end{subfigure}

    \setcounter{subfigure}{0}

    \parbox{1.02 \textwidth}{
        \centering
        \large
A cute hedgehog holding flowers       
    }
    \begin{subfigure}[b]{1.02\textwidth}
        \centering
        \begin{subfigure}[b]{0.45\textwidth}
            \centering
            \begin{subfigure}[b]{0.48\textwidth}
                \includegraphics[width=\linewidth]{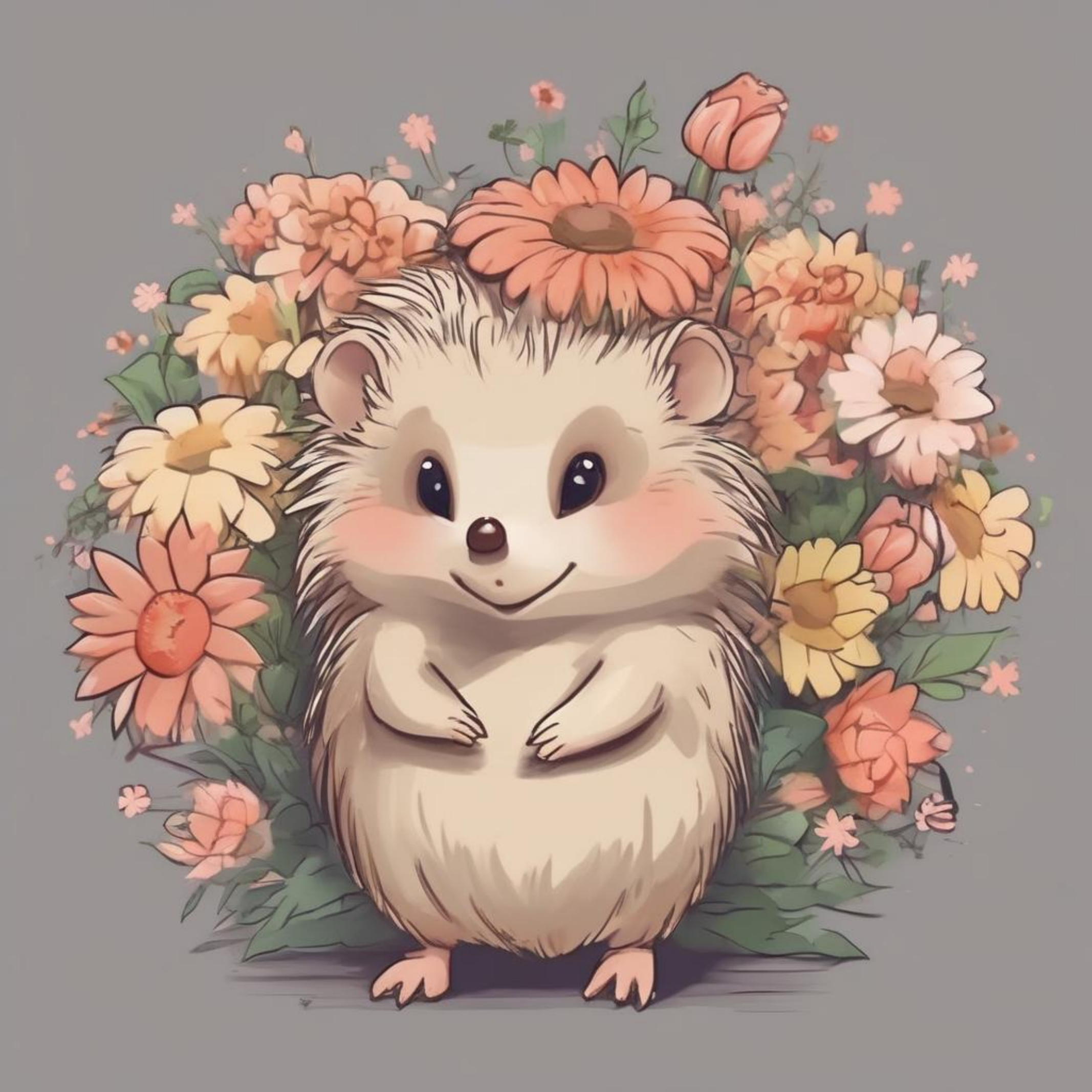}
                \caption{SDXL}
            \end{subfigure}
            \hfill
            \begin{subfigure}[b]{0.48\textwidth}
                \includegraphics[width=\linewidth]{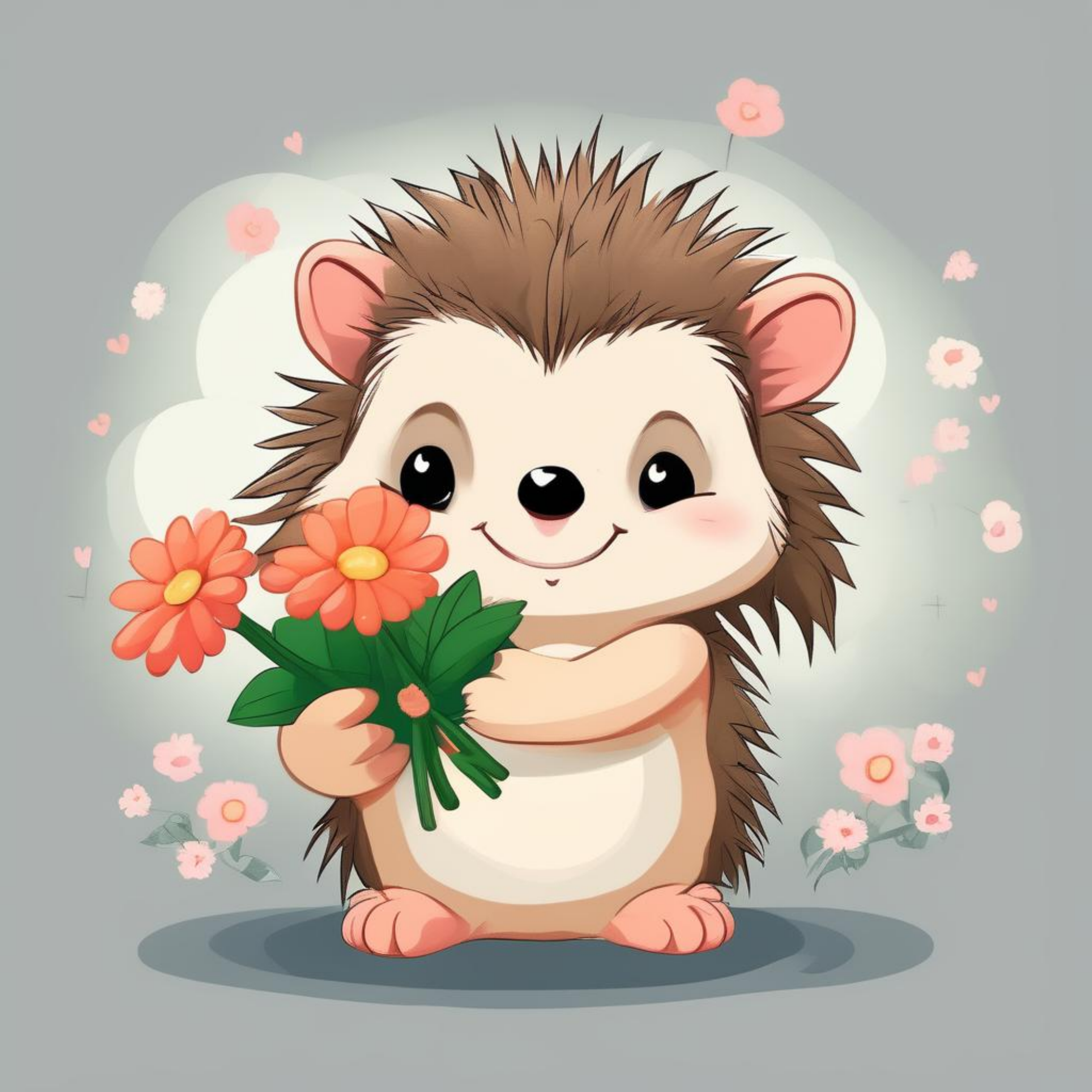}
                \caption{Diff-DPO}
            \end{subfigure}
            
            \begin{subfigure}[b]{0.48\textwidth}
                \includegraphics[width=\linewidth]{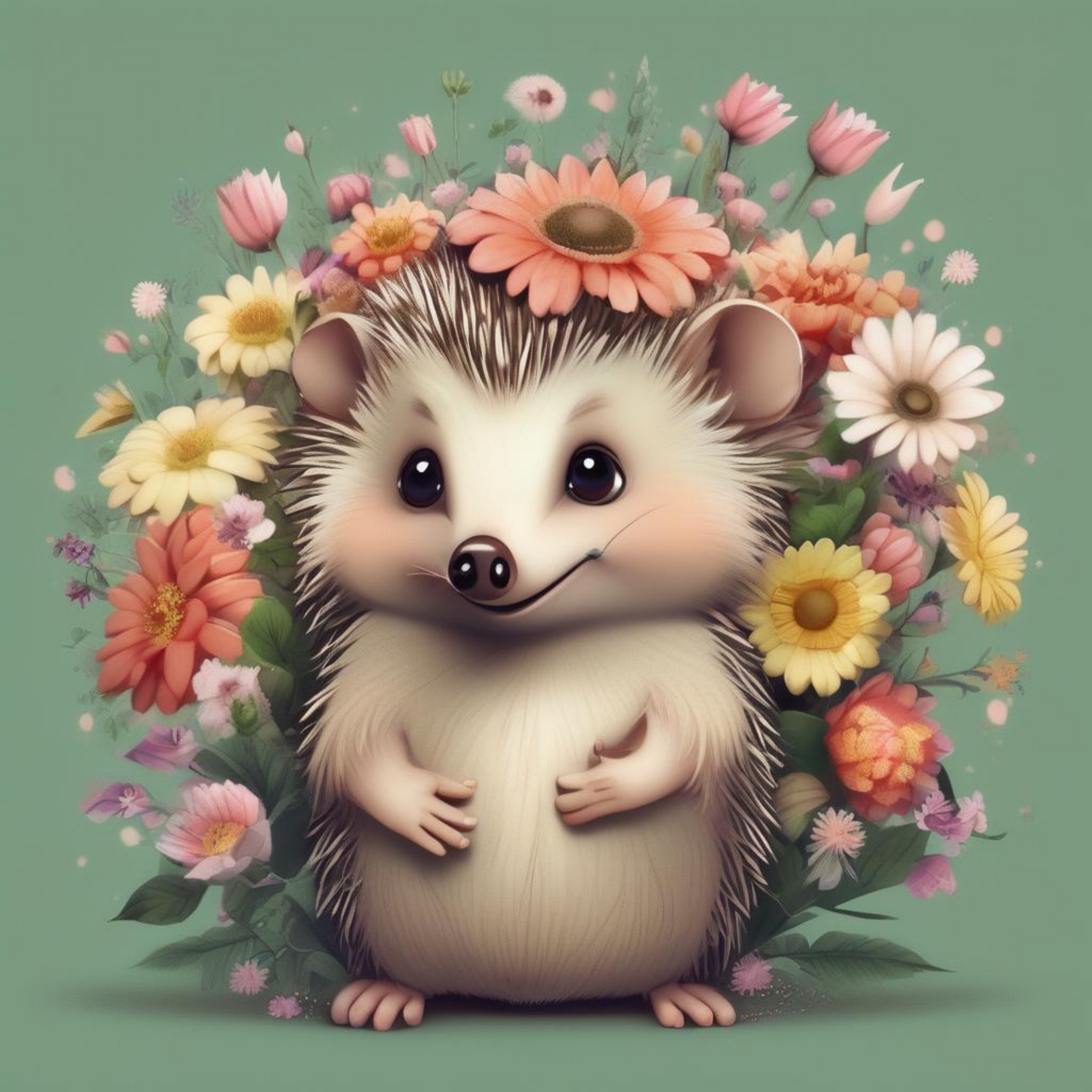}
                \caption{MAPO}
            \end{subfigure}
            \hfill
            \begin{subfigure}[b]{0.48\textwidth}
                \includegraphics[width=\linewidth]{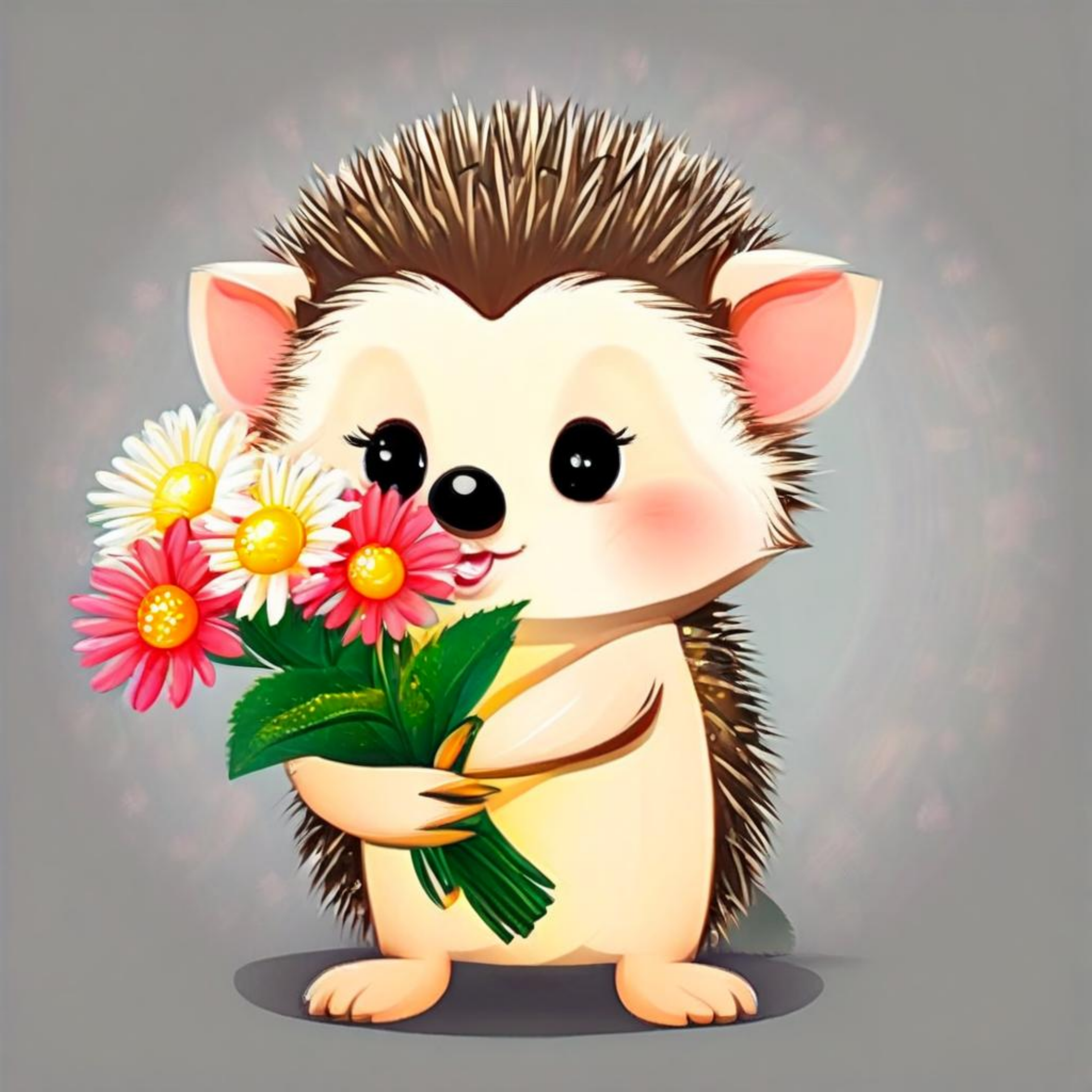}
                \caption{DSPO}
            \end{subfigure}
        \end{subfigure}
        \hfill
        \begin{subfigure}[b]{0.465\textwidth}
            \centering
            \includegraphics[width=\linewidth]{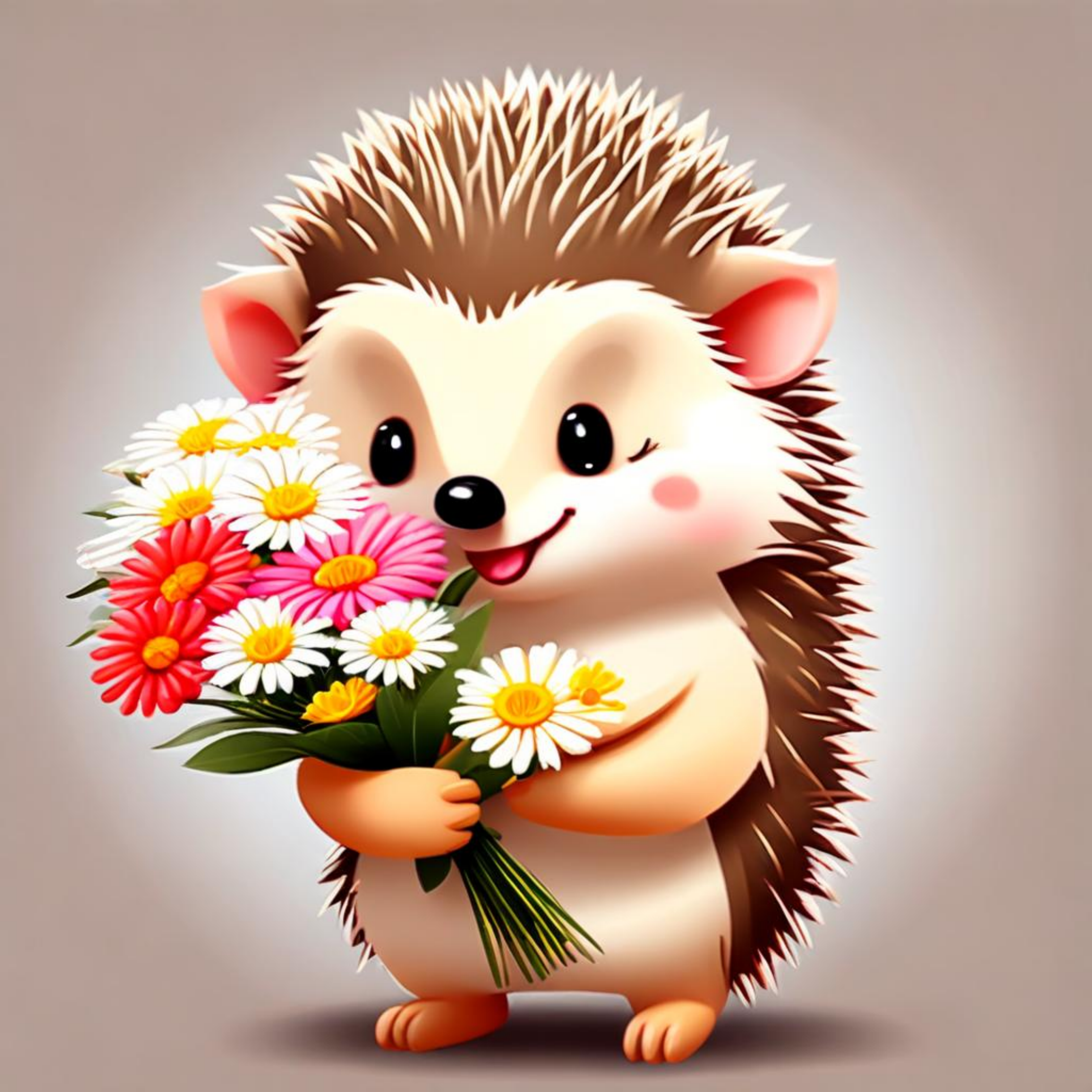}
            \caption{Ours}
        \end{subfigure}
    \end{subfigure}

    \caption{ Qualitative comparisons with the SDXL model.} 
    \label{fig:appendix_sdxl_last}
\end{figure*}

We provide more qualitative results in Figures~\ref{fig:appendix_qual_pick} --~\ref{fig:appendix_sdxl_last}.
In Figures~\ref{fig:appendix_qual_pick} --~\ref{fig:appendix_qual_hpd}, we show images generated by SD1.5 using evaluation prompts (Pick-a-Pic v2, PartiPrompts, HPDv2).
In particular, we use prompts that involve multiple objects or complex compositional relationships.
For example, the prompt \textit{a real flamingo...} describes a complex relationship between objects: the \textit{flamingo} is reading a large open book, and \textit{a stack of books} is placed next to it. 
While existing methods fail to accurately depict this relationship, our method captures the intended scene described in the prompt.                   
Finally, in Figures~\ref{fig:appendix_sdxl_first} --~\ref{fig:appendix_sdxl_last}, we present results for SDXL using prompts from the Pick-a-Pic v2 test set.

\begin{figure*}[htbp]
    \centering
    \parbox{1.02 \textwidth}{
        \centering
        \large
 pencil sketch of An old man looking outside through the first floor window at home
    }
    \begin{subfigure}[b]{1.02\textwidth}
        \centering
        \begin{subfigure}[b]{0.45\textwidth}
            \centering
            \begin{subfigure}[b]{0.48\textwidth}
                \includegraphics[width=\linewidth]{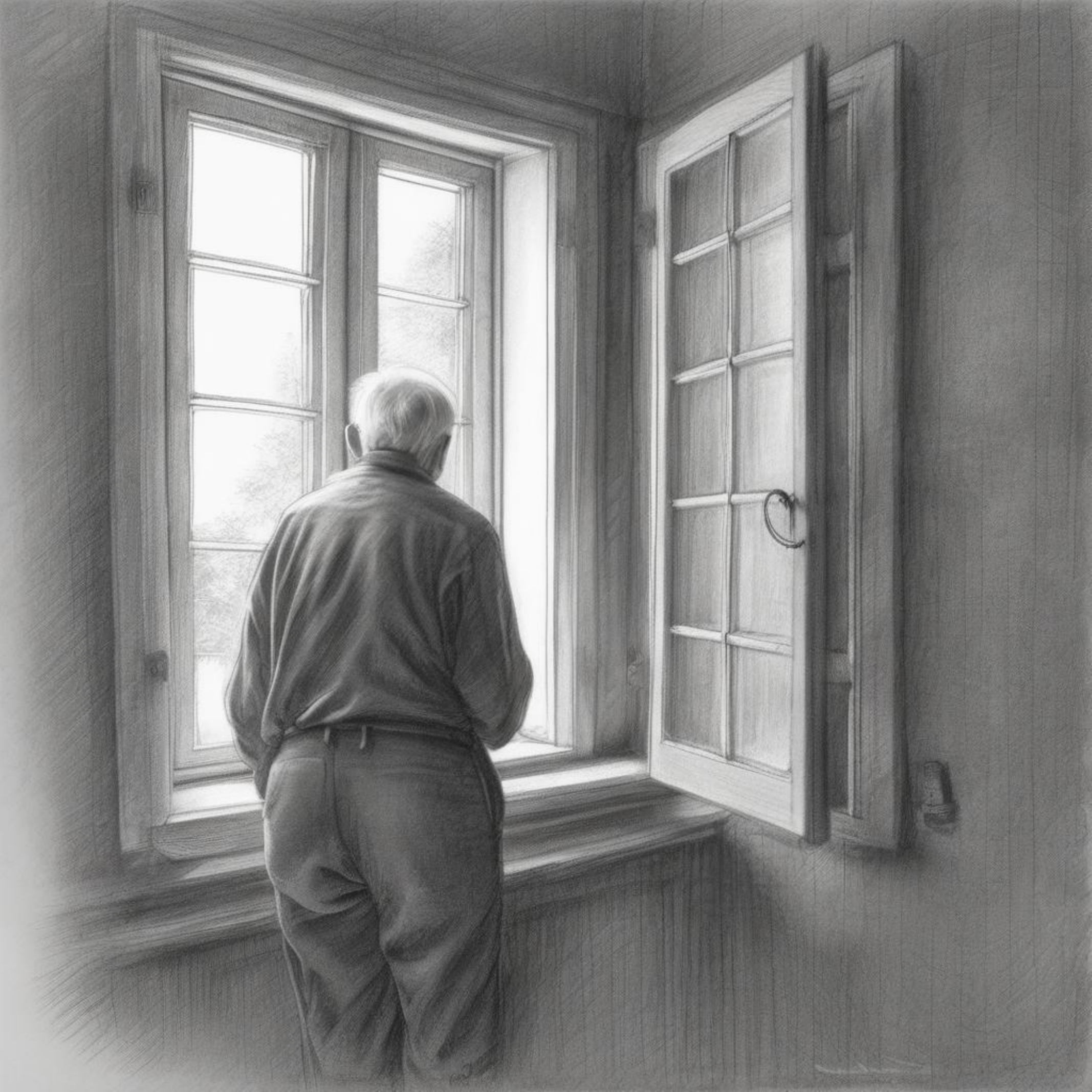}
                \caption{SDXL}
            \end{subfigure}
            \hfill
            \begin{subfigure}[b]{0.48\textwidth}
                \includegraphics[width=\linewidth]{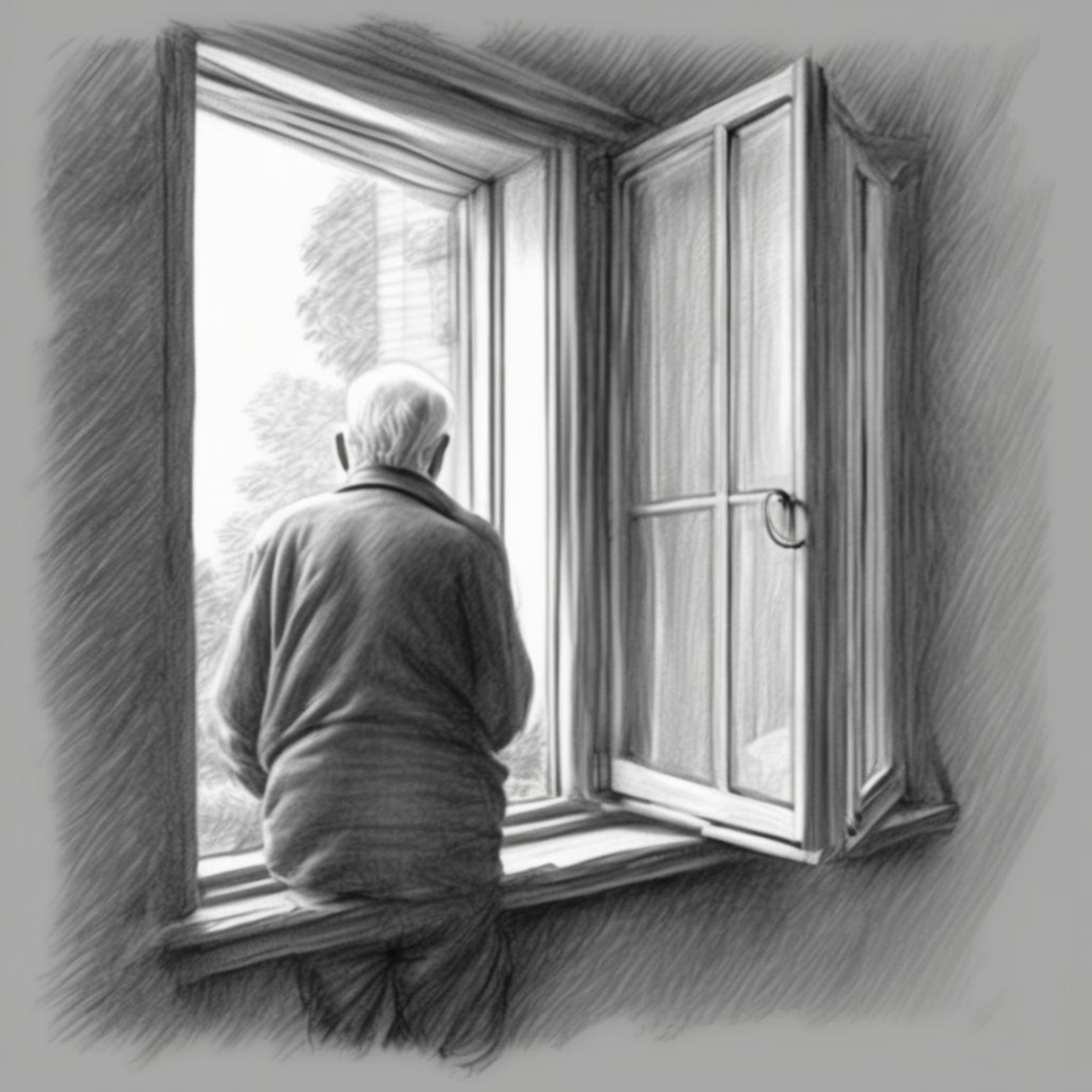}
                \caption{Diff-DPO}
            \end{subfigure}

            \begin{subfigure}[b]{0.48\textwidth}
                \includegraphics[width=\linewidth]{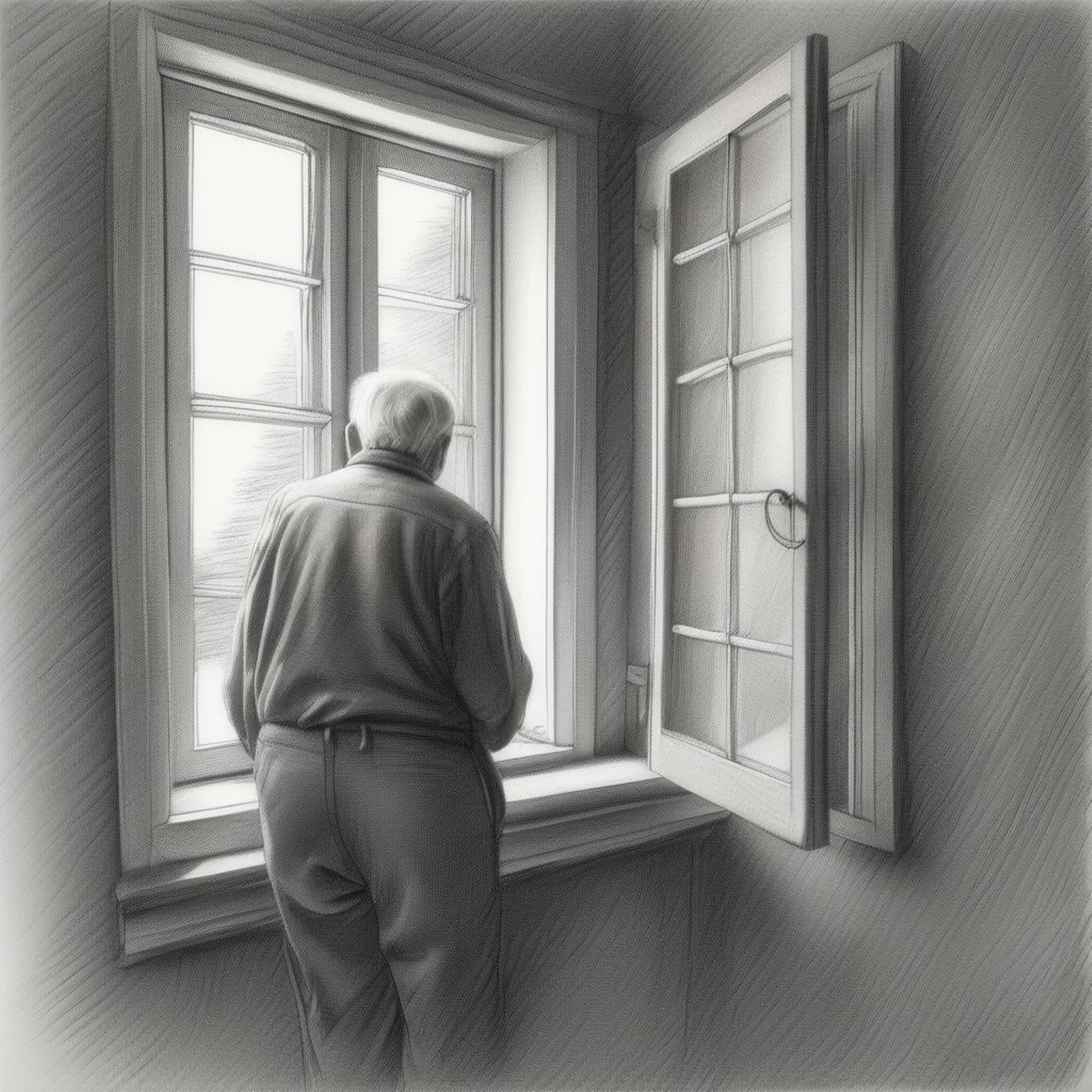}
                \caption{MAPO}
            \end{subfigure}
            \hfill
            \begin{subfigure}[b]{0.48\textwidth}
                \includegraphics[width=\linewidth]{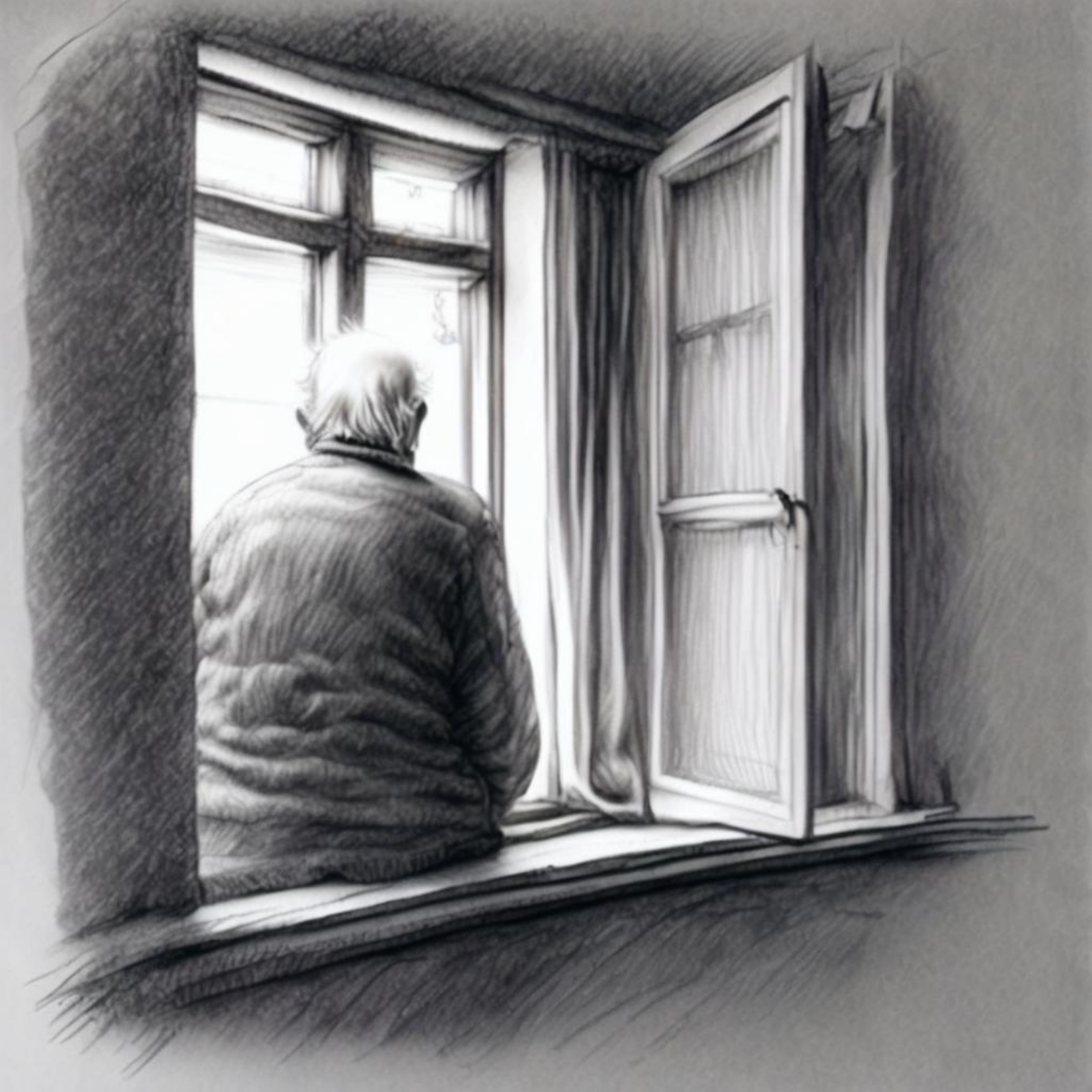}
                \caption{DSPO}
            \end{subfigure}
        \end{subfigure}
        \hfill
        \begin{subfigure}[b]{0.465\textwidth}
            \centering
            \includegraphics[width=\linewidth]{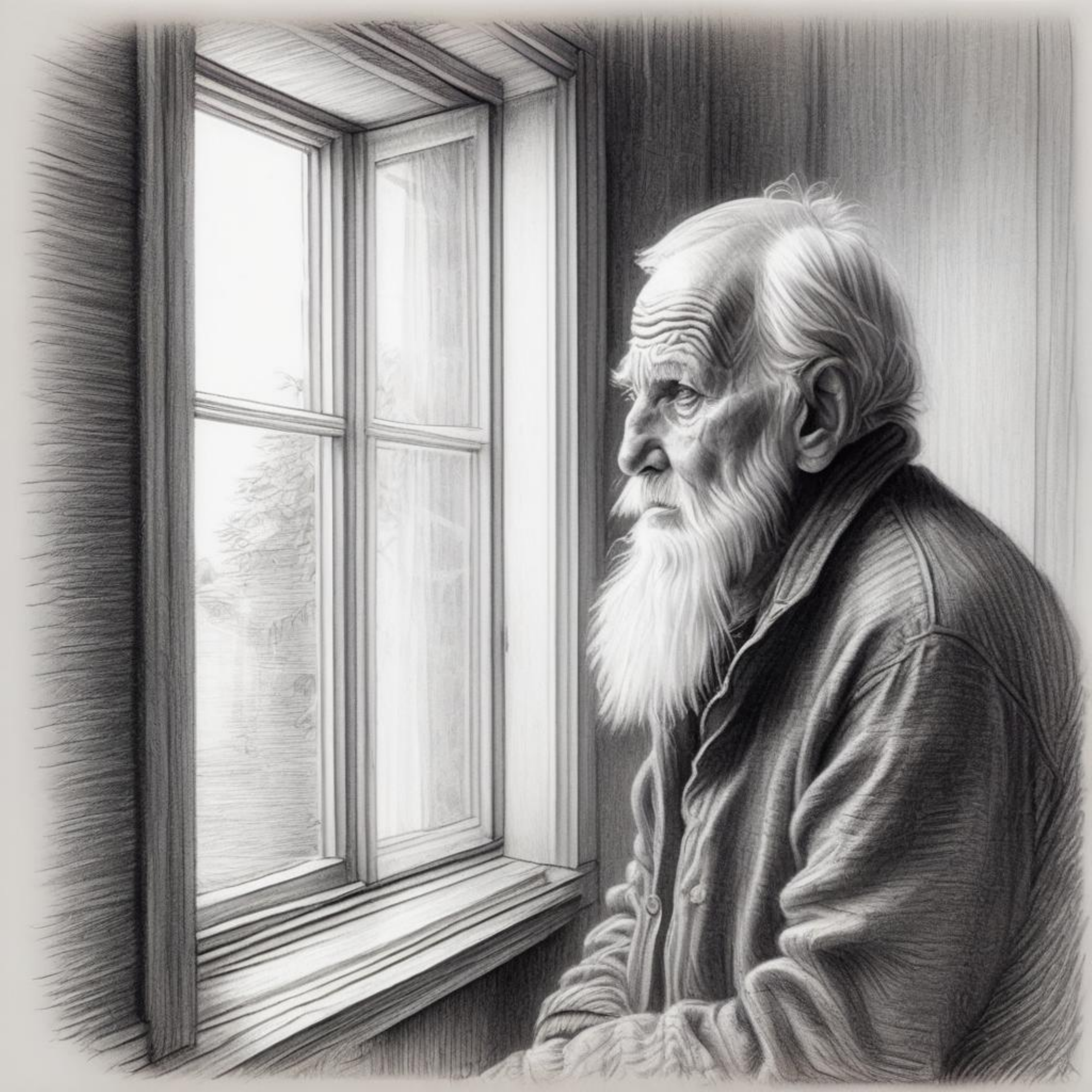}
            \caption{Ours}
        \end{subfigure}
    \end{subfigure}

    \setcounter{subfigure}{0}

    \parbox{1.02 \textwidth}{
        \centering
        \large
An evil villain holding a mini Earth, pixelart
    }
    \begin{subfigure}[b]{1.02\textwidth}
        \centering
        \begin{subfigure}[b]{0.45\textwidth}
            \centering
            \begin{subfigure}[b]{0.48\textwidth}
                \includegraphics[width=\linewidth]{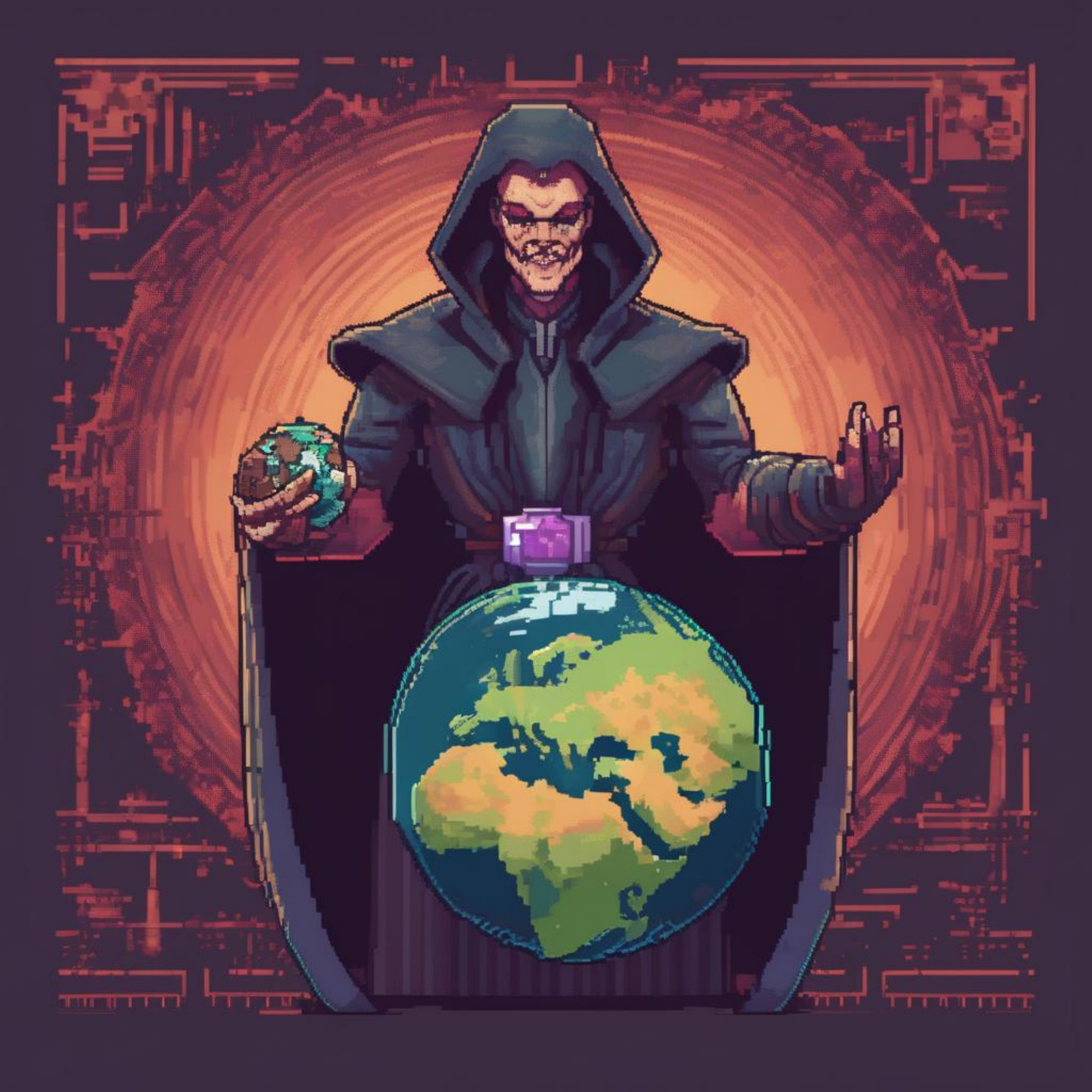}
                \caption{SDXL}
            \end{subfigure}
            \hfill
            \begin{subfigure}[b]{0.48\textwidth}
                \includegraphics[width=\linewidth]{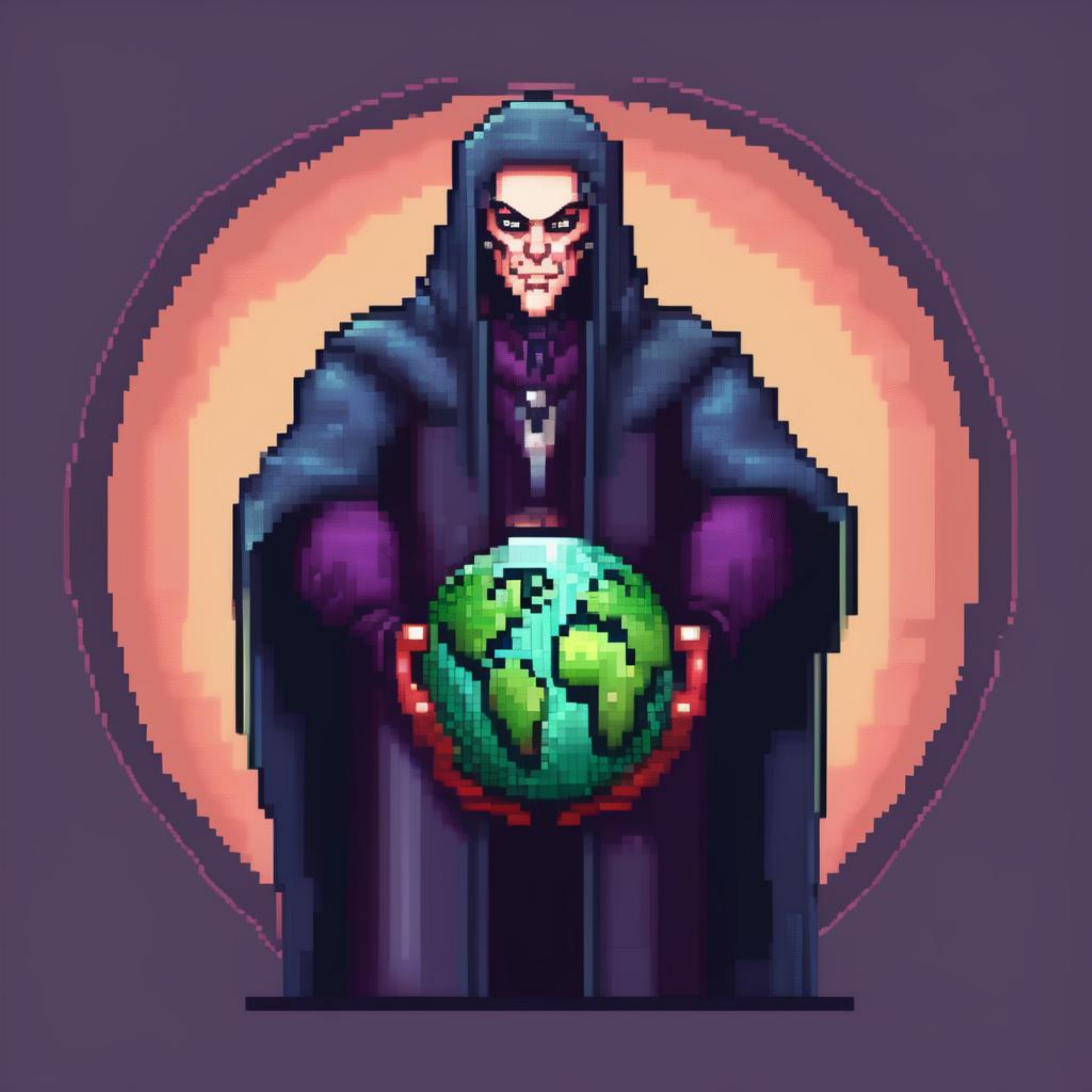}
                \caption{Diff-DPO}
            \end{subfigure}

            \begin{subfigure}[b]{0.48\textwidth}
                \includegraphics[width=\linewidth]{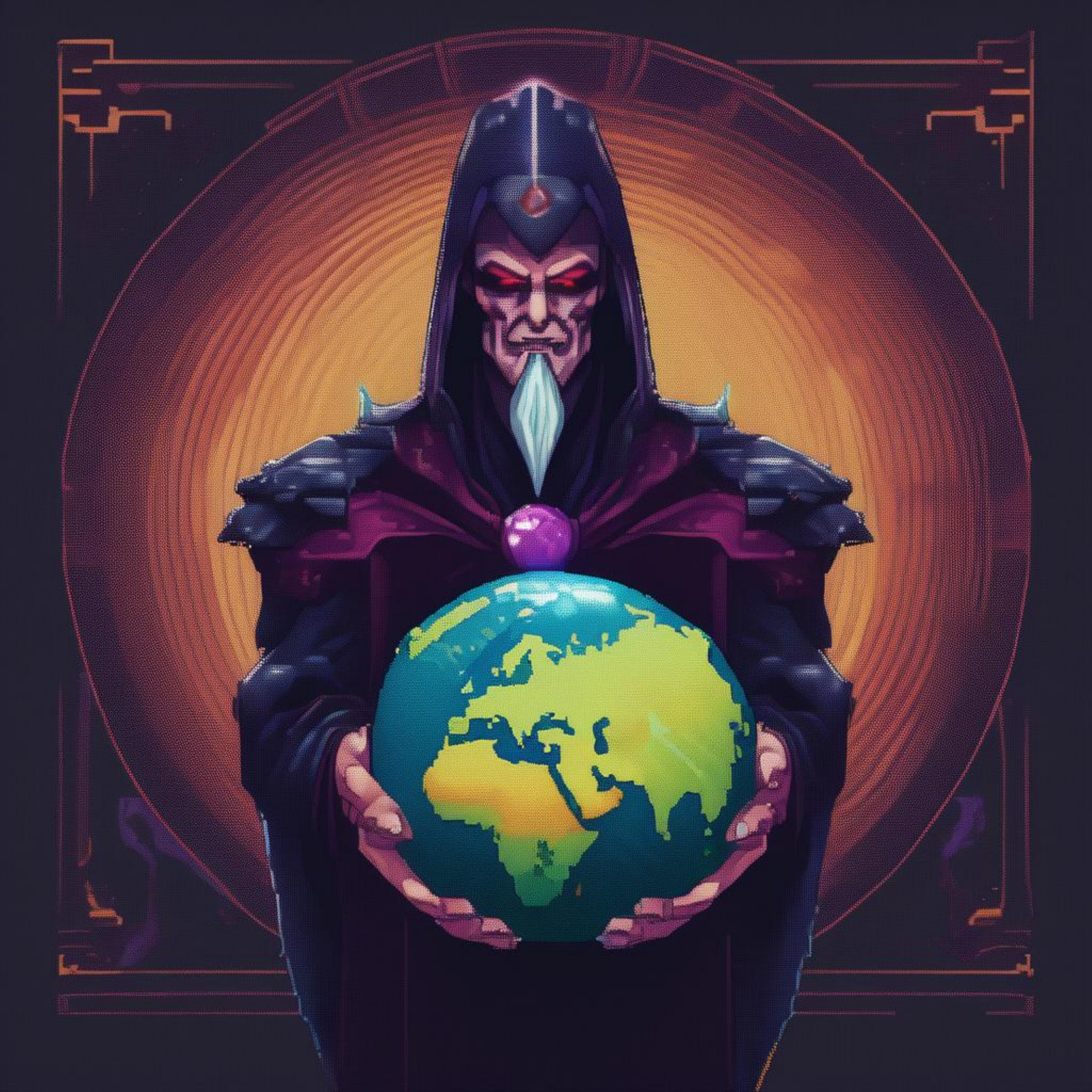}
                \caption{MAPO}
            \end{subfigure}
            \hfill
            \begin{subfigure}[b]{0.48\textwidth}
                \includegraphics[width=\linewidth]{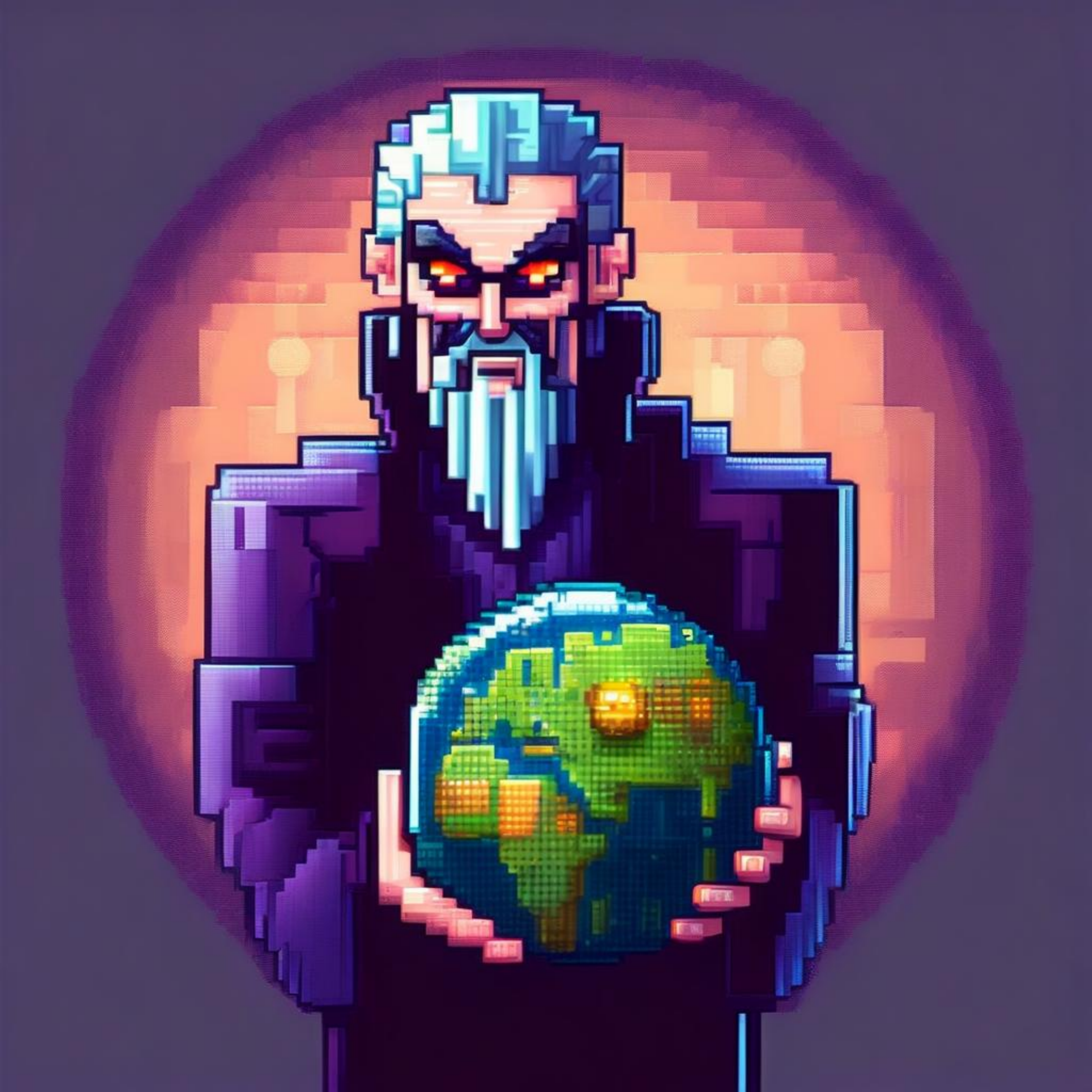}
                \caption{DSPO}
            \end{subfigure}
        \end{subfigure}
        \hfill
        \begin{subfigure}[b]{0.465\textwidth}
            \centering
            \includegraphics[width=\linewidth]{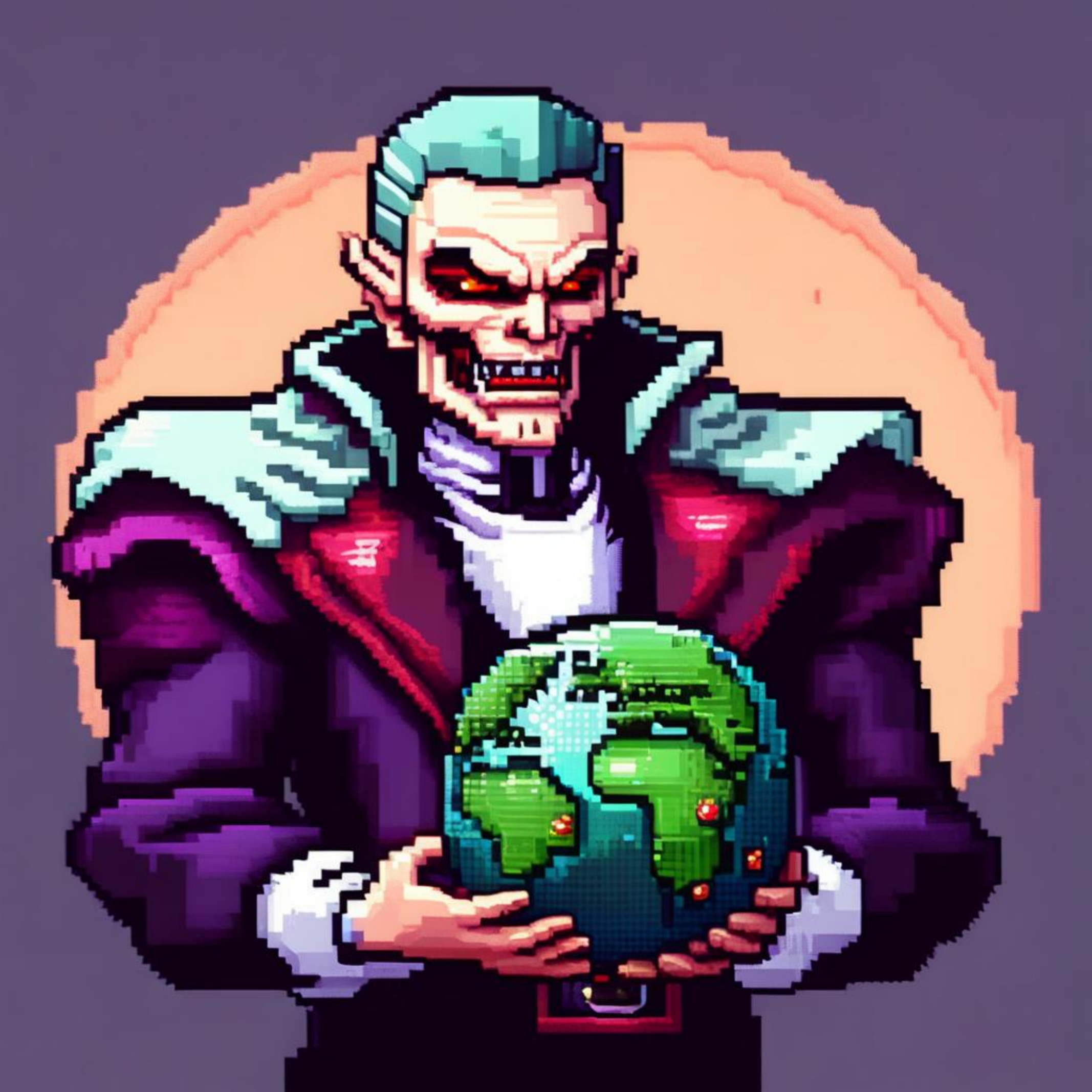}
            \caption{Ours}
        \end{subfigure}
    \end{subfigure}

    \caption{ Qualitative comparisons with the SDXL model.} 
\end{figure*}

\end{document}